\documentclass[a4paper, twoside]{article}

\usepackage[top=2.5cm, bottom=2.5cm, left=2.5cm, right=2.5cm]{geometry}
\usepackage[font=footnotesize,labelfont=bf, skip=4pt]{caption}

\usepackage{times}
\usepackage{url}
\usepackage{latexsym}
\usepackage{amsmath}
\usepackage{amssymb}
\usepackage[]{graphicx}
\usepackage[T1]{fontenc}
\usepackage{textcomp}
\usepackage{arabtex}
\usepackage{utf8}
\usepackage{twemojis}

\usepackage[ruled,vlined]{algorithm2e}
\usepackage{subcaption}
\usepackage{multirow}

\usepackage{titlesec}
\usepackage{hhline}

\titlespacing*{\section}{0pt}{2ex plus 1ex minus .2ex}{1ex plus .2ex}
\titlespacing*{\subsection}{0pt}{1.5ex plus 1ex minus .2ex}{0.8ex plus .2ex}
\titlespacing*{\subsubsection}{0pt}{1ex plus .5ex minus .2ex}{0.5ex plus .2ex}
\titlespacing*{\paragraph}{0pt}{1.5ex plus .5ex minus .2ex}{1.5ex plus .2ex}

\usepackage{natbib}
\let\cite\citep

\definecolor{DarkGreen}{RGB}{0,100,0}
\usepackage[unicode]{hyperref}%lineno
\usepackage{orcidlink}

\usepackage[normalem]{ulem}  % prevents ulem from overriding \emph

\usepackage{relsize} % for \smaller
\makeatletter
\DeclareRobustCommand{\texttt}[1]{%
	{\ttfamily\smaller[1]#1}%
}
\makeatother

\usepackage{fancyhdr}

\def\mytitle{{chDzDT}: Word-level morphology-aware language model for Algerian social media text}
\title{\mytitle}

\author{Abdelkrime Aries\,\orcidlink{0000-0002-3042-1477}}

\date{\itshape\small
	Laboratoire de la Communication dans les Systèmes Informatiques, \\Ecole Nationale Supérieure d'Informatique, Oued Smar, Algiers 16309, Algeria \\[.2cm]
	\texttt{ab\_aries@esi.dz}
}

\hypersetup{
	pdftex,
	colorlinks=true,
	linkcolor=red,
	citecolor=blue,
	urlcolor=DarkGreen,
	bookmarks=true,
	pdfauthor={A. Aries},
	pdftitle={\mytitle},
	pdfsubject={AI and Language},
	pdfkeywords={Algerian dialect, morphology, character-level language model, pre-trained language models, Arabizi, natural language processing},
}

\begin{document}
	
\setcode{utf8}

\maketitle

\begin{abstract}
	Pre-trained language models (PLMs) have substantially advanced natural language processing by providing context-sensitive text representations. 
	However, the Algerian dialect remains under-represented, with few dedicated models available. 
	Processing this dialect is challenging due to its complex morphology, frequent code-switching, multiple scripts, and strong lexical influences from other languages. 
	These characteristics complicate tokenization and reduce the effectiveness of conventional word- or subword-level approaches.
	
	To address this gap, we introduce \textbf{chDzDT}, a character-level pre-trained language model tailored for Algerian morphology.
	Unlike conventional PLMs that rely on token sequences, chDzDT is trained on isolated words.
	This design allows the model to encode morphological patterns robustly, without depending on token boundaries or standardized orthography.
	The training corpus draws from diverse sources, including YouTube comments, French, English, and Berber Wikipedia, as well as the Tatoeba project. 
	It covers multiple scripts and linguistic varieties, resulting in a substantial pre-training workload.
	
	Our contributions are threefold:
	(i) a detailed morphological analysis of Algerian dialect using YouTube comments;
	(ii) the construction of a multilingual Algerian lexicon dataset; and
	(iii) the development and extensive evaluation of a character-level PLM as a morphology-focused encoder for downstream tasks.
	The proposed approach demonstrates the potential of character-level modeling for morphologically rich, low-resource dialects and lays a foundation for more inclusive and adaptable NLP systems.
\end{abstract}


\section{Introduction}
\label{sec:intro}

% Pretrained models and their utility
% ===========================================
Pre-trained language models (PLMs) have transformed the encoding of textual data for machine learning systems. 
Traditional encoding methods, such as one-hot encoding and term frequency (TF) \cite{1972-sparckjones}, fail to capture semantic relationships between words, representing them as isolated entities without considering context. 
Word embedding techniques, including Word2Vec \cite{2013-mikolov-al} and GloVe \cite{2014-pennington-al}, marked a significant advance by embedding words in dense vector spaces that reflect their meanings based on co-occurrence patterns. 
Nevertheless, polysemy, where a word has multiple meanings depending on its context, continued to pose a challenge.
Contextual embedding methods, such as ELMo \cite{2018-peters-al}, address this limitation by representing each word dynamically, taking into account its surrounding context and usage. 
The advent of Transformers \cite{2017-vaswani-al} further revolutionized the field. 
This led to the development of numerous PLMs, including BERT \cite{2019-devlin-al}, which established new benchmarks in natural language understanding tasks.
Since then, many PLMs \cite{2023-Wang-al} have been developed across a variety of languages and dialects, including Arabic \cite{2024-mashaabi-al}.

% Algerian dialect and the need for more attention
% ================================================
Only a few pre-trained language models are available for Algerian dialectal text. 
DziriBERT \cite{2021-abdaoui-al}, the first attempt, is based on the BERT architecture and was trained on 150 MB of Algerian dialect tweets (approximately 20 million tokens). 
AraRoBERTa-DZ \cite{2022-alyami-al-zaidy}, built upon the RoBERTa framework \cite{2019-liu-al}, relied on an even smaller dataset of 103K tweets (1.9 million tokens). 
CAMeLBERT-DA \cite{2021-inoue-al}, although designed for multiple Arabic dialects including Algerian, was trained on a much larger 54 GB corpus (5.8 billion words) covering diverse sources. 
Despite these efforts, the Algerian dialect remains under-represented in large-scale pre-trained models compared with other languages and dialects.

% Problem of morphology and contribution listing
% ===========================================
The Algerian dialect poses unique difficulties for NLP systems due to its linguistic and orthographic variability. 
It integrates vocabulary from Berber, French, Turkish, Spanish, Italian, and, more recently, English, alongside Arabic. 
This reflects historical and cultural contact. 
Code-switching is pervasive, and words are often written in multiple scripts, including Arabic, Latin, Tifinagh, and Arabizi, with frequent mixing of French words in Arabic script and vice versa. 
These practices are largely unstandardized, leading to highly inconsistent spelling and morphology. 
The informal nature of social media further amplifies this variability through non-standard tokens such as emojis. 
Existing word- or subword-based models struggle to capture such irregular and fragmented morphology, particularly in the absence of annotated resources, resulting in limited effectiveness on downstream tasks.

% Our solution and contributions
% ===============================
To address these challenges, we introduce \textbf{chDzDT}, a character-level PLM specifically designed to model Algerian word-level morphology. 
Unlike PLMs trained on sentence-level corpora with subword or wordpiece tokenization, \textbf{chDzDT} is trained exclusively on isolated words, treating characters as atomic units.
This approach allows the model to capture morphological patterns without relying on token boundaries or standardized orthography.  
The training data spans a diverse vocabulary drawn from Algerian YouTube comments, French, English, and Berber Wikipedia, as well as the Tatoeba project. 
It covers multiple scripts, including Arabic, Latin, and Arabizi. 
This dataset reflects the multilingual and morphologically rich nature of Algerian text, incorporating Standard Arabic, French, Berber, English, and local slang.
In summary, the main contributions of this work are as follows:
\begin{itemize}
	\item an analysis of Algerian dialect morphology based on YouTube comments;
	\item the construction of a multilingual Algerian lexicon dataset compiled from YouTube, Wikipedia (French, English, Berber), and the Tatoeba project, preprocessed according to the morphological study;
	\item the development of a character-level PLM designed to encode isolated words solely from their character sequences, without relying on context or tokenization. This model functions as a purely morphological encoder and can be integrated into higher-level systems to mitigate issues such as out-of-vocabulary (OOV) words;
	\item a set of evaluation tasks and fine-tuned models to assess the effectiveness and generalizability of the proposed approach.
\end{itemize}

\section{Related works}
\label{sec:related}

Character-level and morphology-focused models have mostly relied on recurrent networks, convolutional architectures, or character n-gram features. 
These models are often trained in supervised or task-specific settings to capture subword patterns. 
Some pre-trained language models use character-level inputs, but they produce outputs at higher linguistic levels, such as words or sentences. 
They are optimized for general language understanding rather than morphology. 
Models targeting the Algerian dialect are scarce. 
Existing ones have limited coverage and do not adapt well to its complex morphology, code-switching, and non-standard orthography. 
To our knowledge, no transformer-based model has been trained exclusively on morphology-specific objectives using character-level word representations. 
This gap motivates our work.

\subsection{Character-based models}

Character-level models have proven useful for languages with rich morphology or limited resources.
\textit{Morfessor} \cite{2007-creutz-lagus} segments words into subword units, often morphemes, by optimizing a trade-off between lexicon compactness and data likelihood. 
Its effectiveness in low-resource scenarios makes it well-suited for under-resourced linguistic varieties. 
An improved variant, \textit{Morfessor EM+Prune} \cite{2020-gronroos-al}, replaces the original recursive training with Expectation-Maximization and lexicon pruning, yielding better segmentation across typologically diverse languages such as Finnish, Turkish, and North Sami. 
Such models are potentially valuable for dialects like Algerian Arabic, where orthographic standards are weak and corpora are limited.

Beyond segmentation, \citet{2018-godin-al} investigate the extent to which character-level convolutional and recurrent neural networks encode morphological structure. 
By extending the contextual decomposition framework, the authors demonstrate that these models focus on linguistically salient features, including prefixes, suffixes, and other morphologically meaningful subword patterns. 
This analysis shows how character-level models internalize morphological information. 
It maintains interpretability while delivering competitive performance, which is especially important in low-resource contexts.

Further evidence for subword-based representations comes from \citet{2024-ginn-palmer}, who examine fusional morphemes in morphologically complex, low-resource languages. 
They train static vector embeddings on morpheme sequences in Uspanteko and Tsez and find that the representations reflect known grammatical categories. 
These results highlight the viability of distributional approaches for modeling morphology without extensive linguistic resources.

Transformer models have also been applied to morphological transduction tasks, which typically involve learning direct mappings between input and output sequences at the character level.
\citet{2021-wu-al} evaluate transformer architectures on tasks such as morphological inflection and grapheme-to-phoneme conversion. 
Unlike pre-trained language models, these systems are trained from scratch for specific tasks and do not rely on large-scale unsupervised pre-training. 
While transformers outperform LSTM-based models in accuracy, they require more extensive hyperparameter tuning and longer training times. 
This highlights the importance of architecture-aware design choices for morphologically complex or low-resource languages.

In the Arabic context, \citet{2025-mohamed-alazani} evaluate various character-level models on privacy policy classification. 
The study compares CharCNNs, BiLSTMs, and the CANINE transformer, the latter of which will be discussed in the subsequent subsection. 
The use of data augmentation techniques, including vowel removal and style transformation, further enhances model performance. 
Their best system achieves a micro-F1 score of 93.8\% on a Saudi Arabic dataset. 
This demonstrates the efficacy of character-level methods in handling the morphological richness and orthographic variability characteristic of Arabic.

\subsection{Character-based PLMs}

Character-level modeling has emerged as a promising approach for handling morphologically rich and low-resource languages. 
While most widely adopted PLMs are trained on general language objectives, few focus exclusively on morphological learning using character-level input. 
Such models remain scarce, particularly in low-resource settings.
Traditional architectures, including RNNs, CNNs, and n-grams, have shown promise for morphologically unstable languages. 
With the advent of Transformers \cite{2017-vaswani-al}, several character-level PLMs have been developed.

\textit{CharacterBERT} \cite{2020-el-boukkouri-al} is a BERT-like \cite{2019-devlin-al} model that represents each word via its constituent characters. 
Character embeddings are processed through a convolutional layer and a highway network to generate word-level embeddings, which replace subword embeddings from WordPiece tokenization. 
CharacterBERT outputs both word-level and sentence-level embeddings and is particularly suited for sequence labeling tasks, such as part-of-speech tagging.
It is trained with the same masked language modeling (MLM) objective as BERT, predicting whole words from a fixed vocabulary. 
This dependence on a fixed lexicon limits adaptation to noisy or out-of-vocabulary words.

\textit{CharBERT} \cite{2020-ma-al} enhances BERT by integrating a character encoder alongside subword embeddings.
A dual-channel heterogeneous interaction module fuses and then redistributes character- and subword-level information, enabling complementary representations.
The model also introduces a Noisy Language Modeling (NLM) objective to improve robustness to misspellings.
Experiments on question answering, sequence labeling, and text classification show consistent gains, especially under adversarial spelling variations.
Compared to \textit{CharacterBERT}, \textit{CharBERT} is better suited to handling noisy text, yet its reliance on high-level downstream tasks biases character-based representations towards task-specific signals rather than morphological features.

\textit{ByT5} \cite{2021-xue-al} is a T5-like \cite{2020-raffel-al} model that operates on raw UTF-8 bytes, removing the need for subword tokenization. 
The encoder produces byte-level contextual embeddings, trained with a span corruption objective. 
ByT5 reduces the vocabulary size to only 255 tokens, simplifying multilingual processing and improving robustness to rare forms.
However, representing sentences as byte sequences requires substantially longer inputs, making the model less efficient for full-length texts and better suited to shorter inputs such as social media content.

\textit{Charformer} \cite{2021-tay-al} is another T5-like model that learns subword segmentation dynamically from characters or bytes using gradient-based tokenization. 
The resulting subwords are fed into a standard transformer for contextual representation learning, combining the benefits of character-level input with efficient subword processing.
Its main innovation lies in optimizing subword segmentation beyond the WordPiece algorithm \cite{2016-wu-al}.
Nonetheless, by ultimately reverting to subword embeddings, challenges related to noisy input and out-of-vocabulary words remain unresolved.

\textit{CANINE} \cite{2022-clark-al} is a BERT-like model that processes text at the character level, including whitespace, using hash-based representations. 
A shallow transformer encoder followed by convolutional downsampling produces compressed embeddings, which are then upsampled and processed through a second transformer to generate contextual embeddings. 
CANINE adopts a character-level autoregressive masked prediction objective and demonstrates strong performance in multilingual and low-resource scenarios.
Although it captures morphological information at the character level, its reliance on sentence-level context risks diluting word-level structural cues.
We argue that words should be encoded from characters, and sentences from words, in order to preserve the integrity of linguistic levels without interference.

Overall, these models illustrate a spectrum of strategies: some treat characters or bytes as the fundamental input units, while others integrate them with subword representations. 
While they collectively reduce reliance on fixed tokenizers, each approach faces trade-offs between efficiency, robustness, and linguistic adequacy. 
In particular, existing models either retain subword dependencies or risk inefficiencies in processing longer sequences, leaving open the challenge of developing character-level PLMs that capture morphology directly while remaining computationally tractable.
Table~\ref{tab:char_plm} summarizes the main characteristics of these representative models, including tokenization strategy, input/output formats, and parameter counts. 
These approaches highlight the advantages of tokenization-free or tokenization-light modeling, particularly in morphologically complex or low-resource languages.

\begin{table}[ht]
	\centering\small
	
	\caption{Comparative overview of representative character- and byte-level pre-trained language models. 
		Each model employs a transformer-based architecture (BERT- or T5-like), while differing in tokenization (Tok.), architecture (Arch.), input and output representations, and number of parameters (Params). 
		Abbreviations: W = words, SW = subwords, Ch = characters, B = bytes, T = trained, Emb = embeddings, Enc = encoding, Sent. = sentence.}

	\begin{tabular}{@{}p{0.12\textwidth}p{0.08\textwidth}p{0.08\textwidth}p{0.19\textwidth}p{0.27\textwidth}p{0.11\textwidth}@{}}%{llllll}
		\hline\hline
		\textbf{Model} & \textbf{Tok.} & \textbf{Arch.} & \textbf{Input} & \textbf{Output} & \textbf{Params} \\
		\hline
		CharacterBERT  & W & BERT & Ch-based W Emb & Sent. Emb + W Emb & 104.6M \\
		CharBERT       & SW + W & BERT &  SW + W & Sent. Emb + SW Emb + W Emb & 115M \\
		ByT5           & / & T5 & B & Next-byte Emb & 668M--1.23B \\
		Charformer     & T(SW) & T5 & SW & Next-subword Emb & 134M--206M \\
		CANINE         & / & BERT & Hash-based Ch Enc & Sent. Emb + Ch Emb & 127M \\
		\hline\hline
	\end{tabular}
	
	\label{tab:char_plm}
\end{table}

\subsection{Algerian dialect PLMs}

%Recent years have seen notable progress in developing PLMs tailored to the Algerian dialect. 
%As mentioned in the introduction, \textit{DziriBERT} \cite{2021-abdaoui-al} represents the first dedicated PLM for Algerian Arabic. 
%Trained on a Twitter corpus containing 20 million tokens, it employs a WordPiece vocabulary of 50,000 tokens---similar to the original BERT. 
%However, this vocabulary size is insufficient to capture the full range of lexical variation and morphological richness characteristic of Algerian dialects.
%
%As elaborated in the next section, the vocabulary fails to account for the high variability and lack of orthographic standardization in Algerian writing. 
%Inspection of the model's vocabulary on HuggingFace\footnote{\url{https://huggingface.co/alger-ia/dziribert}} reveals entries such as ``\textit{hhhhhhhhhhh}'' and ``\textit{\#\#hhhh}'', reflecting a lack of normalization. 
%This contributes to an inflated number of unique tokens and highlights issues with vocabulary design.
%
%Moreover, the model does not adequately consider the influence of standard languages such as Modern Standard Arabic (MSA) or French, nor does it address the challenge of Arabizi (Arabic written in Latin script), which dominates Algerian social media content. 
%As with other subword-based models, DziriBERT encodes words through token sequences, requiring downstream models to reconstruct or infer full-word representations from their constituent subword tokens.

Recent years have seen notable progress in PLMs for the Algerian dialect. 
\textit{DziriBERT} \cite{2021-abdaoui-al} is the first dedicated model, trained on a 20-million-token Twitter corpus with a WordPiece vocabulary of 50,000 tokens. 
This vocabulary is too small to capture the dialect's lexical variation and morphological richness. 
Inspection of the model's vocabulary on HuggingFace\footnote{\url{https://huggingface.co/alger-ia/dziribert}} reveals entries such as ``\textit{hhhhhhhhhhh}'' and ``\textit{\#\#hhhh}'', reflecting a lack of normalization and an inflated number of unique tokens.
The model also ignores the influence of Modern Standard Arabic and French. 
Arabizi, common on Algerian social media, is not properly addressed. 
Like other subword-based models, DziriBERT encodes words as token sequences, requiring downstream models to reconstruct or infer full-word representations from their constituent subword tokens.

A smaller distilled version, \textit{TinyDziriBERT} \cite{2025-laggoun-al}, was introduced to preserve performance while reducing model size. 
Although it maintains competitive accuracy across tasks such as sentiment analysis, dialect detection, and emotion recognition, it inherits the same limitations regarding tokenization, vocabulary coverage, and dialectal ambiguity.

Other initiatives have treated Algerian Arabic as part of broader dialect modeling efforts. 
For example, \textit{CAMeLBERT-DA} \cite{2021-inoue-al} is a BERT-based model covering multiple Arabic dialects, including Algerian. 
Pre-trained on 54GB of dialectal data (5.8 billion words), it also uses WordPiece tokenization with objectives similar to BERT, although not explicitly documented. 
A major limitation of this multi-dialect approach is the homogenization of dialect-specific vocabularies, which risks obscuring meaningful linguistic variation. 
One possible mitigation is the inclusion of an auxiliary classification objective to predict the dialect of each input sentence, encouraging dialect-sensitive representations. 
In our work, we adapt this idea at the word level using a multi-label classification objective, enabling each word to be associated with multiple dialects or languages, as appropriate.

A further contribution is the \textit{AraRoBERTa} project \cite{2022-alyami-al-zaidy}, which builds on the RoBERTa framework \cite{2019-liu-al} to pre-train a family of dialectal models, including \textit{AraRoBERTaDz} for Algerian Arabic. 
This model, trained on a comparatively small dataset of 103,000 tweets (1.9 million tokens), is far less comprehensive than DziriBERT. 
It employs Byte Pair Encoding (BPE) \cite{2016-sennrich-al} for tokenization, which, while different from WordPiece, still results in subword units and faces similar challenges regarding representation and generalization in low-resource, dialect-rich settings.

Overall, these models represent the earliest efforts to provide transformer-based PLMs for Algerian dialect, either in isolation or as part of broader dialectal coverage. 
Table~\ref{tab:dz_plm} summarizes their main characteristics, highlighting differences in training data, vocabulary, tokenization, architecture, and parameter size.  

\begin{table}[ht]
	\centering\small
	\caption{Comparative overview of representative Algerian dialect pre-trained language models. 
		All models employ transformer-based architectures (BERT- or RoBERTa-like). 
		Models labeled \textit{Dz} are specific to Algerian dialect, whereas \textit{Dz+} indicates inclusion of Algerian dialect alongside other dialects. 
		Abbreviations: WP = WordPiece, BPE = Byte-Pair Encoding.}
	\begin{tabular}{@{}lllllll@{}}
		\hline\hline
		\textbf{Model} & \textbf{Dialect} & \textbf{Training data} & \textbf{Vocab.} & \textbf{Tok.} & \textbf{Arch.} & \textbf{Params} \\
		\hline
		DziriBERT      & Dz  & $\sim$20M tokens (Twitter)  & 50k  & WP  & BERT    & 124M \\
		TinyDziriBERT  & Dz  & Same as above               & 50k  & WP  & BERT    & 18.6M \\
		CAMeLBERT-DA   & Dz+ & 5.8B words (LDC + others)   & 30k  & WP  & BERT    & 110M \\
		AraRoBERTaDz   & Dz  & 1.9M tokens (Twitter)       & 52k  & BPE & RoBERTa & 126M \\
		\hline\hline
	\end{tabular}
	\label{tab:dz_plm}
\end{table}

\section{Morphology in Algerian YouTube comments}
\label{sec:dzyoutube}

%Daily communication in Algeria is marked by considerable linguistic diversity, with extensive borrowing from multiple standard languages. 
%As noted by \cite{2023-sadouki}, speakers frequently alternate between Arabic, French, Berber (Tamazight), and, increasingly, English and Turkish. 
%Such code-switching behavior varies according to social context, interlocutor, and domain of conversation.
%
%On social media platforms, this diversity is reflected in distinct usage patterns. 
%Algerians select between standard languages and local dialects depending on the topic: standard Arabic predominates in religious content, Algerian dialect is prevalent in everyday discussions such as cooking or politics, while French is commonly used in conversations related to French–Algerian media. 
%Compared with Twitter, Arabizi (Latin-script Arabic with numerals) is less frequently observed on YouTube; however, this remains a preliminary finding that requires further validation across platforms and demographics.

Daily communication in Algeria is highly diverse, with extensive borrowing from multiple standard languages. 
Speakers often alternate between Arabic, French, Berber (Tamazight), and, increasingly, English and Turkish \cite{2023-sadouki}, with code-switching patterns shaped by social context, interlocutor, and topic. 
This linguistic diversity is also reflected on social media. 
Standard Arabic dominates religious content, Algerian dialect is common in everyday discussions such as cooking or politics, and French is used in conversations about French--Algerian media. 
Arabizi (Latin-script Arabic with numerals) appears less frequently on YouTube than on Twitter, though this requires further validation across platforms and demographics.

This section examines the morphological characteristics of Algerian YouTube comments. 
Whereas the morphology of standard languages is typically regarded as regular in the Chomsky hierarchy \cite{1956-chomsky}, Algerian social media presents a far more complex scenario. 
The use of multiple writing systems expands the character set and complicates tokenization. 
Even within standard languages, the coexistence of different morphological systems hinders unified processing. 
More commonly, users employ non-standard dialects shaped by regional variation in vocabulary, morphology, and pronunciation. 
The frequent borrowing and adaptation of foreign words adds another layer of complexity; for instance, French verbs are often assimilated into Arabic morphological patterns, generating hybrid forms. 
Although linguistically rich, such practices render traditional finite-state automata (FSA)-based morphological analysis inadequate.

\subsection{Writing systems in Algerian comments}

% Introduction
The written content in Algerian YouTube comments reflects a highly multilingual and multiscript environment. 
This diversity is shaped by historical, cultural, and technological factors that continue to influence online communication practices. 
Algerian users employ several writing systems, most prominently Arabic and Latin scripts, but also Tifinagh and, occasionally, scripts such as Japanese or Korean. 
Choice of script depends on factors such as topic, platform, keyboard availability, and stylistic preference. 
This multiplicity of scripts introduces challenges for morphological analysis, including script-dependent ambiguity, inconsistent spelling, and script mixing.

\paragraph{Arabic script.}
Arabic script is commonly used for both Standard Arabic and Algerian dialect (Table~\ref{tab:arabic-script}). 
It is particularly frequent in religious, cultural, and political discussions. 
In informal contexts, however, usage often departs from standard orthography, incorporating phonetic spellings, colloquial vocabulary, and inconsistent diacritics. 
Dialectal words are frequently represented in non-standard ways, increasing morphological variability. 
In addition, users sometimes transcribe French or English terms in Arabic script, such as ``\RL{مارشي}'' (from Fr: ``\textit{marché}'', ``\textit{market}'') or ``\RL{نايس}'' (from En: ``\textit{nice}'').

\begin{table}[!htp]
	\centering\small
	
	\caption{Illustration of Arabic script in Algerian YouTube comments, shown with ALA-LC transliteration (/.../) and English translation.}
	\begin{tabular}{p{0.9\textwidth}}
		\hline\hline
		\textbf{Standard Arabic:}\\
		\twemoji{blush} \RL{الإرادة القوية والإصرار والصدق والإخلاص للمبادئ كل هذا يصنع المعجزات}\\
		\textit{/al-\textquoteright ir\={a}dah al-qaw\={\i}yah wa-al-\textquoteright i\d{s}r\={a}r wa-al-\d{s}idq wa-al-\textquoteright ikhl\={a}\d{s} li-al-mab\={a}di\textquoteleft\ \ kullu h\={a}dh\={a} ya\d{s}na\textquoteleft u al-mu\textquoteleft jiz\={a}t/} \twemoji{blush}\\
		\textit{Strong will, determination, honesty, and loyalty to principles; all of this creates miracles \twemoji{blush}}\\
		\hline
		\textbf{Dialect:}\\
		\RL{لي تحب ماء الزهر تشمخ بيه العجين مع الماء تجي فيه بنة زمان}\\
		\textit{/l\={\i} t\d{h}abb m\=a az-zahr, tshammakh b\={\i}h la\textquoteleft j\={\i}n m\textquoteleft a al-m\=a, tj\={\i} f\={\i}h bannat zm\=an/}\\
		\textit{Whoever loves orange blossom water soaks the dough with it and water; it gives it the old-time flavor.}\\
		\hline\hline
	\end{tabular}

	\label{tab:arabic-script}
\end{table}

\paragraph{Latin script.}
Latin script is also widespread in Algerian YouTube comments (Table~\ref{tab:latin-script}). 
Its prominence is linked to historical factors such as French colonization, the role of French and English in education, and the ubiquity of Latin-script keyboards. 
It is used natively for French and English, as well as for Berber, which may alternatively be written in Tifinagh or Arabic script. 
Latin script is also commonly used for Arabizi, in which Arabic words are transcribed phonetically using Latin characters and numerals (e.g., ``3'' for ``\RL{ع}'', ``7'' for ``\RL{ح}''). 
Because no orthographic standard exists, Arabizi exhibits extreme spelling variability, complicating tasks such as normalization and morphological analysis.

\begin{table}[!htp]
	\centering\small
	
	\caption{Illustration of Latin script in Algerian YouTube comments, including English, French, Kabyle, and Arabizi examples.}
	\begin{tabular}{p{0.9\textwidth}}
		\hline\hline
		\textbf{English:}\\
		Wish all the best for the two contries\twemoji{tada}\twemoji{tada}\\
		\hline
		\textbf{French:}\\
		Malheureusement ce sont toujours les gens les plus pauvres qui souffrent le plus.\\ 
		\textit{Unfortunately, it is always the poorest people who suffer the most.} \\
		\hline
		\textbf{Kabyle [Variant of Berber]:}\\
		Adhi ssahlu yillu ladjruh ifkad sber ithassa. athnirham yillu.\\
		\textit{May God heal your wounds and give you patience. May God have mercy on you.} \\
		\hline
		\textbf{Arabizi [Dialect in Latin and numerals]:}\\
		\twemoji{2764} darto lyoum jani raw3a ya3tik saha\twemoji{2764}\twemoji{2764}\\
		\textit{\twemoji{2764} I made it today, it was amazing. Thank you! \twemoji{2764}\twemoji{2764}} \\
		\hline\hline
	\end{tabular}

	\label{tab:latin-script}
\end{table}

\paragraph{Tifinagh.}
Tifinagh, the indigenous script of Tamazight (Berber), is only occasionally observed in Algerian YouTube comments. 
Its use is typically associated with the expression of Amazigh identity or cultural pride. 
Although Tamazight is primarily spoken, its online representation serves to reinforce visibility in digital spaces. 
Due to limited keyboard and font support, however, users more often resort to Latin script when writing Tamazight. 
In our dataset, no clear examples of Tifinagh were detected, although a systematic search was not undertaken. 
Consequently, its impact on morphological processing remains limited.

\paragraph{Other scripts.}
Scripts from non-native languages such as Japanese, Korean, or Turkish appear occasionally. 
They are usually restricted to references to anime, K-dramas, Turkish television series, or related popular culture. 
Such instances are largely stylistic, consisting mainly of names, titles, or short phrases, and rarely extend to full sentences. 
Their frequency is low, and their impact on morphological analysis is negligible.

\paragraph{Emojis and non-alphabetic symbols.}
Emojis, repeated punctuation (e.g., ``!!!''), and non-alphabetic symbols are frequently used to convey tone, emotion, or emphasis. 
Emojis may substitute for words, reinforce meaning, or indicate sentiment. 
Although they are not linguistic units in the strict sense, they affect tokenization and segmentation in natural language processing. 
Our dataset contains a wide range of such elements, which must be treated explicitly during preprocessing.

\paragraph{Code-switching and script mixing.}
Although not a morphological process in itself, code-switching significantly shapes the morphological environment of Algerian comments. 
As observed by \citet{2023-sadouki}, users often alternate between Arabic, French, and occasionally English, sometimes within a single sentence. 
This is frequently accompanied by script mixing, such as French words written in Arabic script or Arabic words transcribed in Latin characters. 
These hybrid constructions produce non-standard forms that complicate tokenization, segmentation, and morphological analysis. 
In the absence of a standardized orthography, morphology must therefore be analyzed in context-sensitive and script-aware ways.

%\vspace{6pt}
%\newpage
\subsection{Vocabulary in Algerian comments}

%Introduction
The vocabulary in Algerian YouTube comments reflects a highly multilingual and dynamic linguistic environment. 
It blends Standard Arabic, Algerian dialect, French, Arabizi, Berber (Tamazight), and, increasingly, English. 
Usage varies with region, education, topic, and the informality of digital communication. 
Consequently, lexical choice is fluid, with shifts even within a single thread, which poses particular challenges for automatic processing.

\paragraph{Standard languages.}
Although Algerian dialect is pervasive in everyday discourse, standard languages occupy a substantial proportion of YouTube interactions. 
Standard Arabic is common in religious, political, and formal content, typically using classical lexical items (e.g., ``\RL{عدل}'' /\textit{\textquoteleft adl}/ ``\textit{justice}'', ``\RL{حقيقة}'' /\textit{\d{h}aq\={\i}qah}/ ``\textit{truth}'', ``\RL{حرية}'' /\textit{\d{h}urr\={\i}yah}/ ``\textit{freedom}''). 
French appears frequently in entertainment, education, and technology (e.g., ``\textit{paroles}'' ``\textit{lyrics}'', ``\textit{télécharger}'' ``\textit{download}''), supported by widespread familiarity. 
Tamazight is less common but visible, often written in Latin script for accessibility (e.g., ``\textit{azul}'' ``\textit{hello}'', ``\textit{aselway}'' ``\textit{prayer}'', ``\textit{thanemirth}'' ``\textit{thank you}''). 
English is increasingly present among younger users, especially for education, science, and technology (e.g., ``\textit{compete}'', ``\textit{beneficial}'', ``\textit{classroom}''). 
Other languages (e.g., Spanish, Korean, Japanese) occur only marginally and may be omitted from most models without materially affecting coverage. 
While each standard language has a finite lexicon governed by clear derivational and inflectional rules, their combination substantially enlarges the effective vocabulary space, making character-level representations attractive relative to word-level encoding.

\paragraph{Algerian dialect and Arabizi.}
Algerian dialect dominates informal discourse and shows pronounced regional variation (e.g., ``\textit{what}'' may surface as ``\textit{shawala}'' in Oran and ``\textit{wash}'' in Algiers). 
It is written in both Arabic script (e.g., ``\RL{علاش}'' /\textit{\textquoteleft l\={a}sh}/ ``\textit{Why}'') and Latin script (e.g., ``\textit{3lech}'' ``\textit{Why}''). 
Spelling often mirrors local pronunciation and personal preference, yielding multiple surface forms for the same word. 
In Arabic script, conventions are somewhat more recognizable, though informal elongation persists (e.g., ``\RL{علاااااش}'' /\textit{\textquoteleft l\={a}aaaash}/ ``\textit{Whyyyyy}'', for emphasis). 
Arabizi (Latin-script Algerian dialect) is considerably less standardized: the phrase ``\textit{you frightened me}'' can appear as ``\textit{Khla3touni}'', ``\textit{Khla3toni}'', ``\textit{5la3touni}'', ``\textit{5l3touni}'', ``\textit{5la3tony}'', and others. 
This variability complicates tokenization, normalization, and downstream NLP.

\paragraph{Hybridization and obfuscation.}
Code-switching often involves borrowing French or English words and writing them in Arabic script, usually to avoid switching keyboard layouts.
Examples include ``\RL{الأكسس}'' /\textit{al-\textquoteright aksas}/ (from En: ``\textit{access}'') and ``\RL{الكابل}'' /\textit{al-k\={a}bl}/ (from FR: \textit{cable}'').
These borrowed forms are further adapted with Arabic morphology, such as adding the definite article ``\RL{ال}''.
Conversely, foreign roots in Latin script are inflected with Arabic/Algerian dialect affixes (e.g., ``\textit{nvisitih}'' ``\textit{I visit him}'', ``\textit{tsupprimiouah}'' ``\textit{you delete him/it}''). 
To evade moderation, users also obfuscate offensive terms via symbol substitution (e.g., ``\textit{idi*t}'' for ``\textit{idiot}'') or script mixing (e.g., blending Arabic and Latin characters within ``\RL{حمار}'' /\textit{\d{h}m\={a}r}/ ``\textit{donkey/idiot}'' to get ``\RL{ار}m\RL{ح}''). 
Such irregular forms introduce additional noise, challenging tokenization, morphological analysis, and sentiment classification, and motivate script-aware, context-sensitive processing.

\section{Training dataset design}
\label{sec:dataset}

% Introduction
This section describes the creation of a rich, multilingual dataset for training a character-based language model on Algerian dialectal text. 
The dataset integrates multiple sources, including social media (YouTube), Wikipedia in several languages, and the Tatoeba corpus. 
Given the multilingual nature of Algerian communication, it is designed to capture vocabulary across different languages and scripts. 
%We first describe the data collection process, then present the preprocessing steps, and finally outline the data merging and labeling procedure.

\subsection{Data collection}

\paragraph{YouTube.}
YouTube was selected as the primary data source due to its abundance of diverse, user-generated content. 
Compared to platforms such as Twitter, YouTube comments are generally cleaner and contain fewer offensive expressions, making them particularly suitable for morphological modeling. 
The platform provides a free API with daily access quotas, which enabled large-scale data collection at minimal cost. 
Beyond comments, YouTube offers valuable metadata, including video titles and descriptions, allowing comments to be interpreted as responses to both the visual content and accompanying text, which provides a foundation for future multimodal studies. 
We targeted 347 Algerian YouTube channels and collected data from 45,842 videos, yielding a corpus of 15,368,543 user comments written in multiple languages and scripts.

\paragraph{Wikipedia.}
To expand vocabulary coverage and introduce standardized language input, we incorporated Wikipedia articles in four languages commonly used by Algerians: Arabic, French, English, and Kabyle.
These languages are widely present in online communication, either individually or through code-switching, and their inclusion provides formal vocabulary that enhances model generalization.
The number of collected articles per language was 2,243,581 for Arabic, 489,847 for French, 384,714 for English, and 8,607 for Kabyle. 
During preprocessing (see the next subsection), most corpora exceeded 1.5 GB, except for Kabyle, which was considerably smaller at 3.2 MB. 
To address this imbalance and strengthen Berber vocabulary coverage, we supplemented the dataset with material from the Tatoeba project\footnote{Tatoeba is a collaborative database of example sentences across many languages.}.

\paragraph{Tatoeba.}
From Tatoeba, we extracted all sentences labeled as Kabyle and Berber. 
While Kabyle is a distinct variety, it is often regarded as part of the wider Berber family, so both sources were merged. 
The final counts are 677,555 Kabyle sentences and 643,875 Berber sentences, providing additional coverage of low-resource linguistic material for training.

\subsection{Data preprocessing}

Following data collection, each dataset underwent preprocessing tailored to its structure and linguistic characteristics. 
Wikipedia articles, being well-structured and written in standardized language, required minimal normalization. 
In contrast, YouTube comments, which are informal and often ungrammatical, necessitated more extensive processing to address inconsistencies and noise. 
Some operations, such as deduplication (removal of repeated comments within individual video files), were considered less critical for morphological analysis, as the focus lies on word-level patterns rather than full sentence structure.

\begin{figure}[!htbp]
	\centering
	\includegraphics[width=0.5\textwidth]{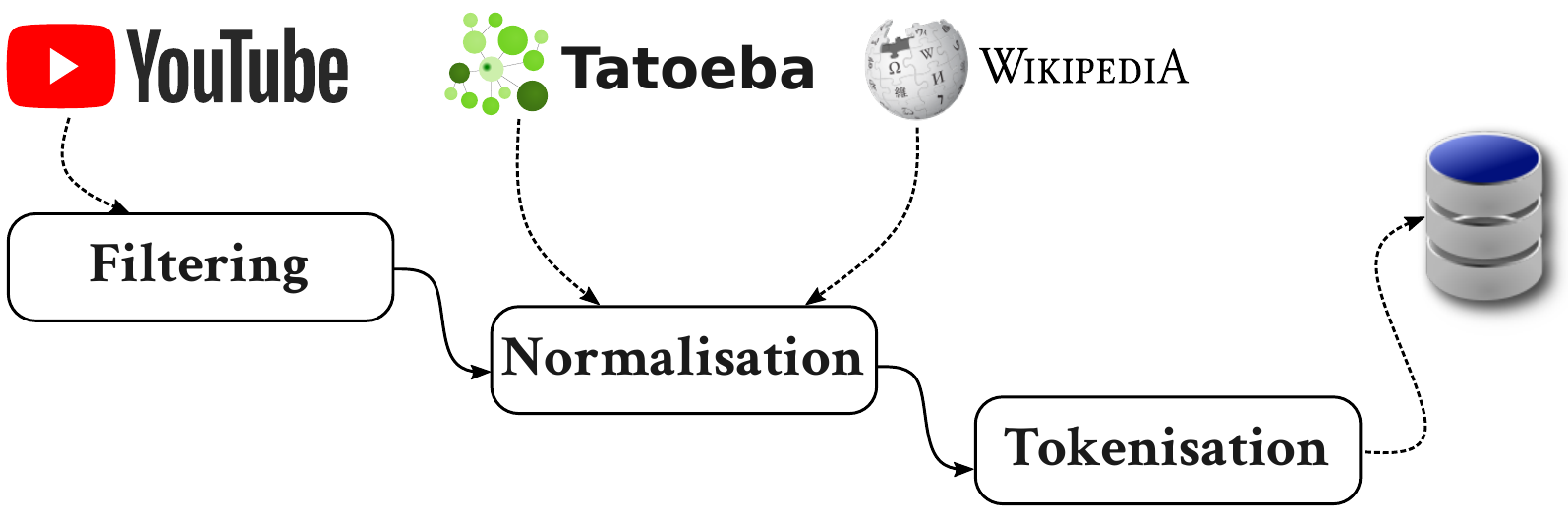}
	\caption{Preprocessing pipeline across data sources. YouTube comments required an additional filtering stage due to higher noise compared with Tatoeba and Wikipedia.}
	\label{fig:prep}
\end{figure}

Figure~\ref{fig:prep} outlines the preprocessing pipeline.  
For YouTube comments, an initial filtering step excluded region-specific vocabulary using regular expressions, removing terms distinctive to neighboring dialects such as Moroccan, Tunisian, or Gulf Arabic. 
Although North African dialects share a large proportion of vocabulary, each also preserves unique expressions.  
For example, ``\RL{دابا}'' (/d\={a}b\={a}/, ``\textit{Now}'') is primarily Moroccan, ``\RL{برشا}'' (/barsh\={a}/, ``\textit{A lot}'') is common in Tunisia, and ``\RL{كفو}'' (/kaf\={u}/, ``\textit{Well done}'') is typical in Gulf Arabic.  
YouTube data required heavy normalization to address a wide range of irregularities:
\begin{itemize}
	
	\item \textbf{Emoji normalization}: Emojis are frequent in Algerian YouTube comments, often in exaggerated quantities. 
	Consecutive identical emojis were reduced to a maximum of two, balancing expressive intent with data cleanliness.  
	For example, ``\textit{\twemoji{2764}\twemoji{2764}\twemoji{2764}\twemoji{2764} Macha Allah merci beaucoup}'' becomes ``\textit{\twemoji{2764} \twemoji{2764} Macha Allah merci beaucoup}''.  
	Textual emoji forms were replaced by their graphical counterparts to ensure uniform representation.  
	For instance, the string ``:-)'' was converted into ``\twemoji{1f642}''.
	
	\item \textbf{Character normalization}: Quotation marks, hyphens, dashes, bars, and various space symbols were unified into consistent forms to reduce token fragmentation.  
	For example, different forms of double quotation marks (\textquotedblleft, \textquotedblright, \guillemotleft, \guillemotright) were normalized to the standard ASCII double quote (\textquotedbl).
	
	\item \textbf{Character deduplication}: Repeated letters used for elongation were limited to a fixed maximum.  
	For example, ``\RL{برافووووووو}'' (/br\={a}f\={u}uuuuuu/, ``\textit{bravo}'') was normalized to ``\RL{برافوو}''.
	
	\item \textbf{Spacing corrections}: Punctuation attached directly to words was separated by spaces.  
	Similarly, consecutive emojis were separated to avoid unnecessary vocabulary inflation, as each emoji is treated as a distinct token.  
	When emojis were attached to words, they were detached for clarity.  
	For example, ``\textit{\twemoji{2705}\twemoji{2705}yes. No!\twemoji{2705}}'' became ``\textit{\twemoji{2705} \twemoji{2705} yes . No ! \twemoji{2705}}''.
	
	\item \textbf{Character filtering}: Non-essential symbols such as Arabic diacritics (Tashkīl) were removed to simplify downstream processing.  
	For example, ``\RL{مَجَلَّــــــــــــة}'' (/\textit{majallah}/, ``\textit{Journal}'') was normalized to ``\RL{مجلة}''.
\end{itemize}
Finally, all text was tokenized into words using space delimiters.  
Wikipedia and Tatoeba corpora, being more standardized, required far less intervention; however, for consistency, the same preprocessing pipeline was applied across all sources.

\subsection{Data merging and labeling}

Following preprocessing, words from all sources were consolidated into a unified, word-level vocabulary.  
Each word was treated as an independent instance, since the model operates at the character level.  
In line with Zipf's brevity law \cite{1950-Zipf}, which suggests that less frequent words tend to be longer, we excluded unusually long forms.  
Such items are likely to fall outside the genuine vocabularies of the target languages, often representing unrecognized tokens or scrambled strings.  
A maximum word length of 30 characters was therefore imposed.

\begin{figure}[!htbp]
	\centering
	\includegraphics[width=0.4\textwidth]{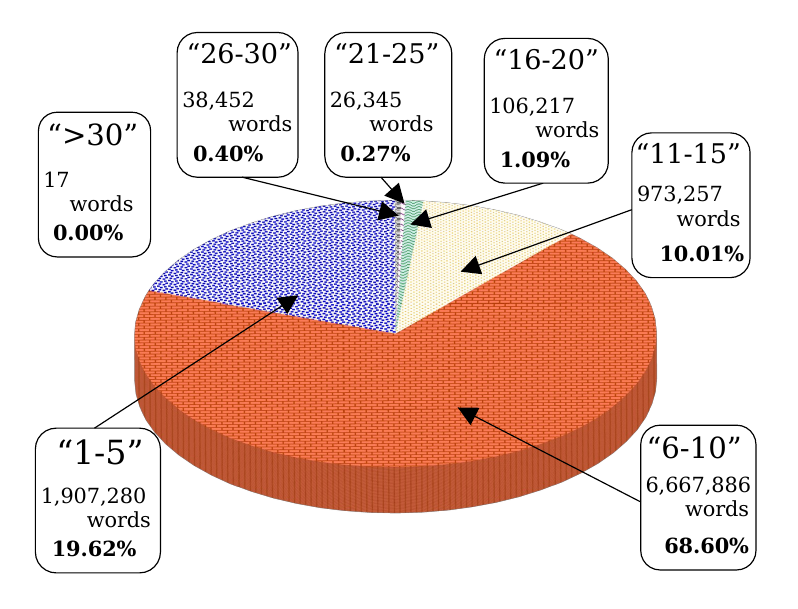}
	\caption{Distribution of word lengths in the training dataset. Lengths are grouped into five-character intervals, with both frequencies and percentages reported.}
	\label{fig:words-size}
\end{figure}

We examined the distribution of word lengths in the final dataset.  
Character counts per word vary considerably, reflecting the morphological diversity and multiple writing systems across languages. 
Figure~\ref{fig:words-size} illustrates the frequency and percentage distribution of word lengths, grouped into five-character intervals.  
As shown, over 99\% of words contain 1--20 characters.  
Consequently, a transformer with a maximum context window of 20 characters is sufficient to encode the majority of the vocabulary for single-word pretraining.  
However, for tasks involving comparisons between words, such as next-word prediction, a context window of roughly double that size is required.

To address the multilingual composition of the dataset, we applied a multi-label annotation scheme indicating the language(s) of origin for each word.  
Words from standard corpora such as Wikipedia and Tatoeba were annotated with language codes AR (Arabic), BER (Berber), EN (English), and FR (French).  
Words from Algerian YouTube comments were labeled with DZ, corresponding to Algeria's ISO 3166-1 alpha-2 country code.  
A single word could receive one to five labels depending on its distribution across corpora.  
For example, a word might occur in both Arabic and Berber sources, or across all five labels (AR, BER, EN, FR, DZ).  
This approach allows for fine-grained tracking of linguistic overlap, particularly relevant in code-switched and diglossic\footnote{Diglossia: the use of two distinct forms of the same language within a community, each suited to different social contexts [Cambridge Dictionary].} contexts such as Algeria.

Some words appeared in corpora attributed to a given language even though they did not originate in that language.  
For instance, a French Wikipedia article on an Algerian locality may include place names or terms in Arabic or Berber.  
We chose not to exclude such items or apply frequency thresholds, thereby preserving the long-tail distribution of rare words, which can be informative for morphology and named entity recognition.
This multi-label framework yields 31 possible label combinations: five monolingual, ten bilingual, ten trilingual, five quadrilingual, and one spanning all five labels.  
Table~\ref{tab:words-labels} presents their distribution.  
Several insights emerge:
\begin{itemize}
	\item The DZ-only category (3,808,202 words) is by far the most populous.  
	This indicates that a substantial portion of the lexicon drawn from Algerian YouTube comments is distinctive to the local dialect or informal usage, with limited overlap with standard Arabic or French vocabularies.  
	This underscores the need for dedicated resources and models for Algerian dialect and similar under-resourced varieties.
	
	\item The second most frequent category is AR-only, with 2,304,728 words.  
	This reflects the morphological richness and lexical depth of Arabic.  
	Although many Arabic terms are used by Algerian speakers online, a considerable portion of the lexicon remains unused in informal contexts, highlighting the breadth of Arabic beyond everyday digital communication.
	
	\item Significant counts in hybrid categories such as AR-DZ (529,010), DZ-FR (64,710), and DZ-EN (29,139) confirm the strong influence of Arabic, French, and English on Algerian online discourse.  
	These figures provide quantitative evidence of code-switching and linguistic hybridization in a multilingual society shaped by both colonial and global language exposure.
	
	\item Cross-script overlaps such as AR-EN (97,626) and AR-FR (29,494) demonstrate the integration of Arabic in English- or French-language articles and vice versa.  
	Although such combinations may appear inconsistent with strict linguistic boundaries, they arise from multilingual articles, loanwords, and the presence of named entities and cultural terms in foreign-language texts.  
	These cases warrant closer inspection to distinguish genuine multilingual usage from artifacts of content mixing or annotation.
\end{itemize}

\begin{table}[!htp]
	\centering\footnotesize
	\caption{Distribution of words across language label combinations. 
		The first two blocks list bilingual and trilingual label counts, while the last block (``Other'') covers monolingual cases and higher-order combinations.}
	\begin{tabular}{@{}l@{}rl@{}r@{}p{5mm}@{}l@{}rl@{}r@{}p{5mm}@{}l@{}rl@{}r@{}}
		\hline\hline
		\multicolumn{4}{c}{\textbf{Two labels}} && \multicolumn{4}{c}{\textbf{Three labels}} && \multicolumn{4}{c}{\textbf{Other}} \\
		\cline{1-4}\cline{6-9}\cline{11-14}
		AR-DZ  &\ \ 529,010 & DZ-EN  & 29,139 && AR-EN-FR  & 133,527 & BER-DZ-FR &\ \ 1,293 && DZ  & 3,808,202 & AR-DZ-EN-FR  & 55,052  \\
		EN-FR  & 352,139 & BER-DZ & 5,816  && DZ-EN-FR  & 42,079  & BER-DZ-EN & 694   && AR  & 2,304,728 & AR-BER-EN-FR & 2,766 \\
		AR-EN  & 97,626  & BER-FR & 1,834  && AR-DZ-EN  & 10,427  & AR-BER-EN & 441   && EN  & 1,104,555 & BER-DZ-EN-FR & 2,461 \\
		DZ-FR  & 64,710  & BER-EN & 1,008  && AR-DZ-FR  & 6,475   & AR-BER-DZ & 309   && FR  & 975,697   & AR-BER-DZ-FR & 308 \\
		AR-FR  & 29,494  & AR-BER & 541    && BER-EN-FR & 1,988   & AR-BER-FR & 199   && BER & 147,471   & AR-BER-DZ-EN & 220 \\
		&         &        &        &&           &         &           &       &&     &           & (All)        & 9,244 \\
		\hline\hline
	\end{tabular}
	\label{tab:words-labels}
\end{table}

%\begin{table}[!htp]
%	\centering\small
%	\begin{tabular}{p{0.1\textwidth}rlp{0.15\textwidth}rlp{0.2\textwidth}r}
%		\hline
%		\multicolumn{2}{c}{Two labels} && \multicolumn{2}{c}{Three labels} && \multicolumn{2}{c}{The rest} \\
%		\cline{1-2}\cline{4-5}\cline{7-8}
%		&&&&&&&\\
%		AR-BER & 541     && AR-BER-DZ & 309     && AR  & 2,304,728 \\
%		AR-DZ  & 529,010 && AR-BER-EN & 441     && BER & 147,471 \\
%		AR-EN  & 97,626  && AR-BER-FR & 199     && DZ  & 3,808,202 \\
%		AR-FR  & 29,494  && AR-DZ-EN  & 10,427  && EN  & 1,104,555 \\
%		BER-DZ & 5,816   && AR-DZ-FR  & 6,475   && FR  & 975,697 \\
%		BER-EN & 1,008   && AR-EN-FR  & 133,527 && AR-BER-DZ-EN & 220 \\
%		BER-FR & 1,834   && BER-DZ-EN & 694     && AR-BER-DZ-FR & 308 \\
%		DZ-EN  & 29,139  && BER-DZ-FR & 1,293   && AR-BER-EN-FR & 2,766 \\
%		DZ-FR  & 64,710  && BER-EN-FR & 1,988   && AR-DZ-EN-FR  & 55,052  \\
%		EN-FR  & 352,139 && DZ-EN-FR  & 42,079  && BER-DZ-EN-FR & 2,461 \\
%		&  &&   &   && (All) & 9,244 \\
%		\hline
%	\end{tabular}
%	\caption{Frequency of words across language label combinations.}
%	\label{tab:words-labels}
%\end{table}

\section{chDzDT design}
\label{sec:chdzdt}

We introduce \textbf{chDzDT} (Algerian Dialect Transformer), a lightweight transformer model operating at the character level.  
Symbolically denoted as $c \cdot h \cdot \frac{\partial z}{\partial t}$, it reflects its design: character-level processing ($c$), deep representations ($h$), and transformation over time ($\frac{\partial z}{\partial t}$).  
The model is specifically tailored to address the challenges of morphologically rich, noisy, multilingual, and multi-script dialects such as Algerian Arabic.
It is built to be robust against orthographic variation and morphological obfuscation prevalent in informal online text.  
By encoding words at the character level and relying exclusively on morphological features, chDzDT is particularly suitable for tasks involving word morphology, such as inflection and derivation.
It can serve as a preprocessing module to enhance conventional language models, especially for handling out-of-vocabulary (OOV) tokens.

\begin{figure}[!htp]
	\centering
	\includegraphics[width=0.5\textwidth]{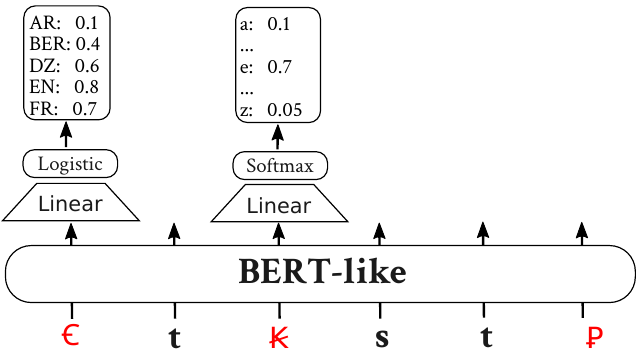}
	\caption{Pre-training architecture of chDzDT. 
		Each word is tokenized into characters and processed by a BERT encoder. 
		Special tokens are used for padding (``\texttt{\sout{P}}''), masking (``\texttt{\sout{K}}''), and word-level representation (``\texttt{\sout{C}}''). 
		The model jointly optimizes a character-level masked language modeling objective and a word-level multi-label classification objective.}
	\label{fig:arch}
\end{figure}

As illustrated in Figure~\ref{fig:arch}, each input word is decomposed into a sequence of characters and processed by a BERT encoder.  
The sequence is padded with a dedicated special token (``\texttt{\sout{P}}'') to maintain a fixed length, while a subset of characters is randomly masked (``\texttt{\sout{K}}'') and predicted via a character-level softmax classifier, implementing the masked language modeling (MLM) objective.  
In parallel, the model encodes the full word into a contextual representation extracted from the final hidden state of the special token (``\texttt{\sout{C}}'') and passes it through a sigmoid-activated logistic regression layer for multi-label classification.  
This dual-head architecture enables the model to capture both orthographic regularities and higher-level linguistic signals.  
Such capacity is essential for handling dialectal, multilingual, and noisy web-based text effectively.

\subsection{Character tokenizer}

Since chDzDT processes one word at a time, the tokenizer is designed to function at the word level while still allowing efficient batch processing. 
To ensure flexibility, the vocabulary can be expanded either by explicitly adding individual characters or by specifying entire character sets using UTF-8 decimal code ranges.  
Tokenization is performed by mapping each character in a word to its corresponding index in the vocabulary.  
As noted earlier, Algerian users employ a wide range of scripts in online communication (e.g., YouTube comments).  
To accommodate this diversity, the vocabulary includes multiple character sets, even for characters that are rarely used.  
This inclusion improves robustness by allowing the model to generalize to unseen characters that may appear during inference.  
The following non-exhaustive list outlines the incorporated character sets:  
\begin{itemize}
	\item \textbf{Non-alphabetic characters}: dingbats, emoticons, symbols, pictographs, punctuation, and related symbols.  
	Algerian social media users often employ these to express emotion or even convey complete messages.  
	For example, the sequence ``\twemoji{1f441} \twemoji{2764} \twemoji{1f370}'' may be interpreted as ``\textit{I love cake}''.
	
	\item \textbf{Latin characters}: the Basic Latin set, Latin-1 Supplement, and several extended Latin sets.  
	French is commonly used in Algerian online discourse, English is increasingly adopted, and the Latin script is widely used for both Arabizi and Berber writing.
	
	\item \textbf{IPA extensions}: selected International Phonetic Alphabet characters occasionally used in informal Berber transcriptions.
	
	\item \textbf{Arabic characters}: used in both Modern Standard Arabic and Darija (Algerian Arabic) written in Arabic script.
	
	\item \textbf{Tifinagh characters}: although less widely adopted than Latin script for Berber, Tifinagh is included to support potential or future use.
\end{itemize}

Similar to BERT, the proposed tokenizer employs a fixed maximum context length, here set to 20 characters, covering more than 99\% of word lengths observed in our dataset.  
Each input word is encoded as a character sequence, prefixed with a special classification token ``\texttt{\sout{C}}'' (functionally analogous to BERT's ``\texttt{[CLS]}'').  
Overlong words are truncated, while shorter ones are padded with a dedicated padding token ``\texttt{\sout{P}}'' (equivalent to ``\texttt{[PAD]}'') to ensure consistent input length.  
Characters not present in the vocabulary are replaced by an unknown token ``\texttt{\sout{U}}'' (corresponding to ``\texttt{[UNK]}'').  
During masked language modeling, characters are replaced with a masking token ``\texttt{\sout{K}}'' (analogous to ``\texttt{[MASK]}'').  
When multiple words are processed, a separator token ``\texttt{\sout{S}}'' (serving the role of BERT's ``\texttt{[SEP]}'') is used.  
Unlike conventional BERT implementations, which reserve multi-character string tokens, chDzDT represents all special tokens using single-character placeholders.  
This design simplifies token handling and enables users to insert special tokens directly into character sequences without unintentional fragmentation.

\subsection{Pre-training setup}

We pre-trained chDzDT using two complementary objectives: one at the character-sequence level and the other at the word level via the BERT encoder.  
For the character-level objective, we adopted the masked language modeling (MLM) strategy from BERT, replacing subword units with individual characters.  
This enables the model to acquire context-sensitive character representations informed by both local and positional information.

A natural analogue to BERT's next sentence prediction (NSP) would be a next word prediction (NWP) task, in which word pairs $(w_1, w_2)$ are drawn from real-world text, with consecutive pairs labeled ``\textit{Next}'' and randomly shuffled pairs labeled ``\textit{Not Next}''.  
While this could encourage learning of syntactic relationships and sequential dependencies, it shifts focus toward higher-level linguistic structures such as word order and syntax.  
Given the noisy, multilingual, and multi-script nature of Algerian social media text, we instead employ a multi-label classification objective.  
Here, the model predicts whether a given word is likely to have been used in each of the following language contexts: Arabic (AR), Berber (BER), Algerian dialect (DZ), French (FR), or English (EN).  
This objective promotes the encoding of linguistic cues that reflect multilingual usage across scripts.

One might question why the two objectives are not extended with NWP.  
In practice, the assumption underlying next word prediction, that adjacent words typically belong to the same language, is often violated in code-switched text.  
This introduces noise and reduces task alignment.  
Furthermore, combining multiple classification tasks on the same special token embedding (e.g., ``\texttt{[CLS]}'') may cause gradient interference, ultimately degrading the learned representations.
  
\paragraph{Forward pass.}
Let $x = [x_1, x_2, \dots, x_T]$ denote a sequence of $T$ characters representing a single word, where each $x_i$ belongs to the character vocabulary $\mathcal{V}_c$.  
The sequence is tokenized at the character level and mapped to dense vectors $E = [e_1, \dots, e_T]$.  
Positional encodings $P$ are added element-wise to capture character order.  
The combined sequence $E \oplus P$ is processed by a stack of $N$ BERT blocks, yielding contextualized representations $H = [h_1, \dots, h_T]$, with each $h_i \in \mathbb{R}^d$ encoding a context-aware representation of $x_i$.  
This forward pass is formalized in Equation~\ref{eq:forward}.  
Segment embeddings are not used in this configuration, as classification is performed over individual words, though they are reserved for potential extensions.  

\begin{equation}
	H = \mathrm{BERT}^{(N)}(E \oplus P)
	\label{eq:forward}
\end{equation}

\paragraph{Masked language modeling.}
The first objective is a character-level masked language modeling (MLM) task.  
Given an input word $x$, a random subset of its characters is selected for masking, producing a corrupted input $\tilde{x}$.  
Each masked character $x_i$ is replaced by a special mask token ``\texttt{\sout{K}}'', and the model is trained to recover the original characters using surrounding context.  
Let $\mathcal{M}$ denote the set of masked positions.  
The loss is computed only over these positions via a softmax layer over the character vocabulary, as in Equation~\ref{eq:L-MLM}, where $\theta$ denotes the model parameters.  

\begin{equation}
	\mathcal{L}_{\text{MLM}} = - \sum_{i \in \mathcal{M}} \log P_{\theta}(x_i \mid \tilde{x})
	\label{eq:L-MLM}
\end{equation}

\paragraph{Multi-label classification.}
In parallel, the model is trained for multi-label classification over language tags.  
For a word $x$, its character sequence is encoded by the BERT encoder into a holistic word representation.  
Specifically, the final hidden state of the classification token ``\texttt{\sout{C}}'' is extracted and passed through a sigmoid-activated output layer with $L$ units, each corresponding to a language label.  
Each unit independently estimates the probability $\hat{y}_l$ that the word belongs to class $l$.  
Binary cross-entropy is used for training with ground-truth labels $y_l \in \{0,1\}$, as formalized in Equation~\ref{eq:L-multilabel}.  

\begin{equation}
	\mathcal{L}_{\text{multi-label}} = - \sum_{l=1}^{L} \left[ y_l \log \hat{y}_l + (1 - y_l) \log (1 - \hat{y}_l) \right]
	\label{eq:L-multilabel}
\end{equation}

\paragraph{Overall training objective.}
The two objectives are optimized jointly to train the character-level BERT model.  
The MLM task captures context-sensitive character representations, while the multi-label classification task encourages language-specific encoding within the holistic word representation.  
The total training loss is defined as the sum of the two components, as shown in Equation~\ref{eq:L-total}.  

\begin{equation}
	\mathcal{L}_{\text{total}} = \mathcal{L}_{\text{MLM}} + \mathcal{L}_{\text{multi-label}}
	\label{eq:L-total}
\end{equation}

%\begin{algorithm}[!ht]
%	\caption{Joint Training of Character-Level BERT with MLM and Multi-Label Classification}
%	\label{algo:pretrain}
%	\KwIn{Corpus of words $\mathcal{D}$, label vectors $Y$, masking ratio $r$}
%	\KwOut{Trained model parameters $\theta$}
%	
%	\ForEach{epoch}{
%		\ForEach{batch $x \in \mathcal{D}$}{
%			Randomly mask $r\%$ of the characters in $x$ to obtain $\tilde{x}$ and record the masked positions $\mathcal{M}$\;
%			Encode $\tilde{x}$ using Equation~\ref{eq:forward}\;
%			
%			\tcp{Masked Language Modelling Loss}
%			Compute $\mathcal{L}_{\text{MLM}}$ using Equation~\ref{eq:L-MLM}\;
%			
%			\tcp{Multi-Label Classification Loss}
%			Predict label probabilities $\hat{y}^{(i)}$ and compute $\mathcal{L}_{\text{multi-label}}$ using Equation~\ref{eq:L-multilabel}\;
%			
%			\tcp{Total Loss and Parameter Update}
%			Calculate $\mathcal{L}_{\text{total}}$ using Equation~\ref{eq:L-total} and normalise by batch size\;
%			Update model parameters $\theta$ via gradient descent\;
%		}
%	}
%\end{algorithm}

\subsection{Implementation details}

We implemented multiple variants of chDzDT, differing in the number of Transformer blocks $N$, attention heads $H$, and hidden size $d$.  
To ensure computational efficiency and maintain a lightweight design, the input context was limited to a maximum of 20 characters per word.  
This configuration reflects the typical morphological structure of the corpus while allowing the model to capture subword-level regularities relevant to both derivational and inflectional morphology.  
Although all variants share the same core architecture, they differ in scale and training dynamics.  
Each configuration was trained for 20 epochs on the pre-training tasks described in Section~\ref{sec:chdzdt}.  
Table~\ref{tab:imp-models} summarizes the configurations, reporting the number of parameters $|\theta|$ for the uncased encoder (excluding task-specific heads), batch size $B$, training time $t$, and throughput in samples per second ($s/s$).

\begin{table}[!htp]
	\centering\small
	
	\caption{Architectural configurations of chDzDT variants, varying in the number of Transformer blocks ($N$), attention heads ($H$), and hidden size ($d$).  
		Parameter counts ($|\theta|$), batch size ($B$), training time ($t$), and average throughput (samples per second, $s/s$) are also reported.}

	\begin{tabular}{lrrrrrrr}
		\hline\hline
		\textbf{Model} & $N$ & $H$ & $d$ & $|\theta|$ & $B$ & $t$ & $s/s$ \\
		\hline
		 \texttt{chDzDT\_5x4x128} & 5 & 4 & 128 & 4,664,192 & 1,500 & 10h58 & 4,921.430 \\
		 \texttt{chDzDT\_4x4x64}  & 4 & 4 & 64  & 1,840,512 & 1,800 & 5h44  & 9,405.017 \\
		 \texttt{chDzDT\_4x4x32}  & 4 & 4 & 32  & 908,992   & 2,000 & 4h36  & 11,736.553 \\
		\hline\hline
	\end{tabular}
	
	\label{tab:imp-models}
\end{table}

\begin{figure}[!htp]
	\centering
	
	\begin{subfigure}[b]{0.47\textwidth}
		\includegraphics[width=\linewidth]{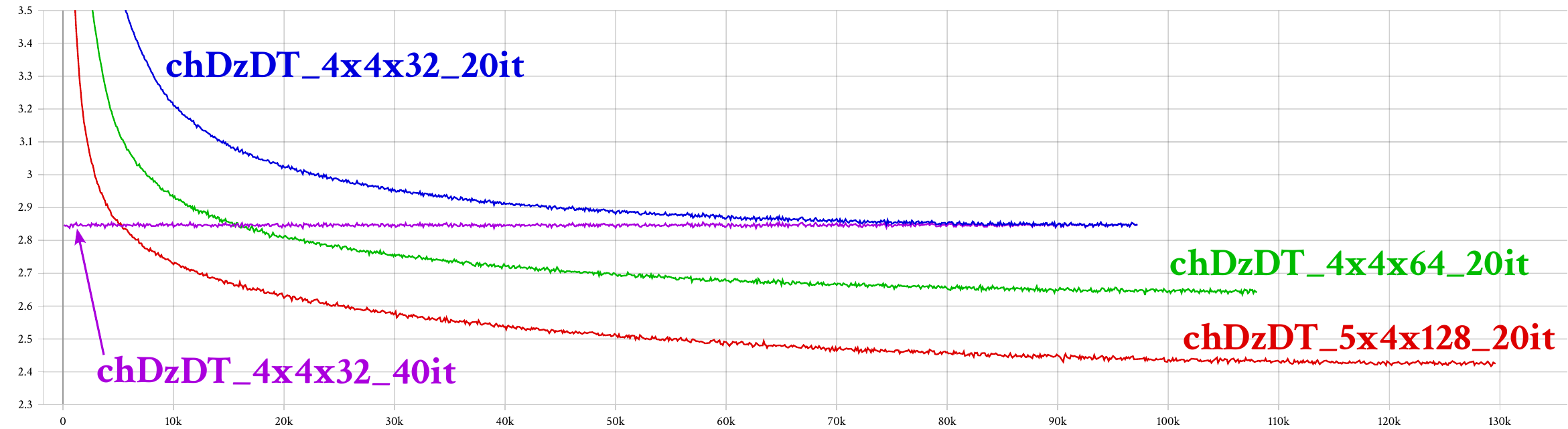}
		\caption{Total loss}
	\end{subfigure}
	
	\begin{subfigure}[b]{0.47\textwidth}
		\includegraphics[width=\linewidth]{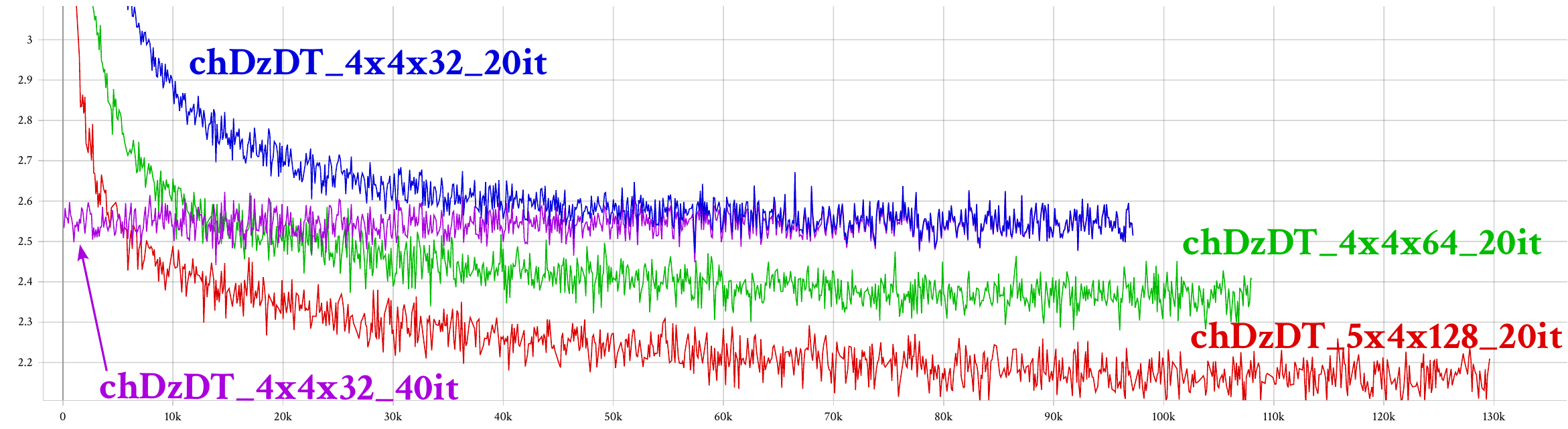}
		\caption{MLM loss}
	\end{subfigure}
	\hfill
	\begin{subfigure}[b]{0.47\textwidth}
		\includegraphics[width=\linewidth]{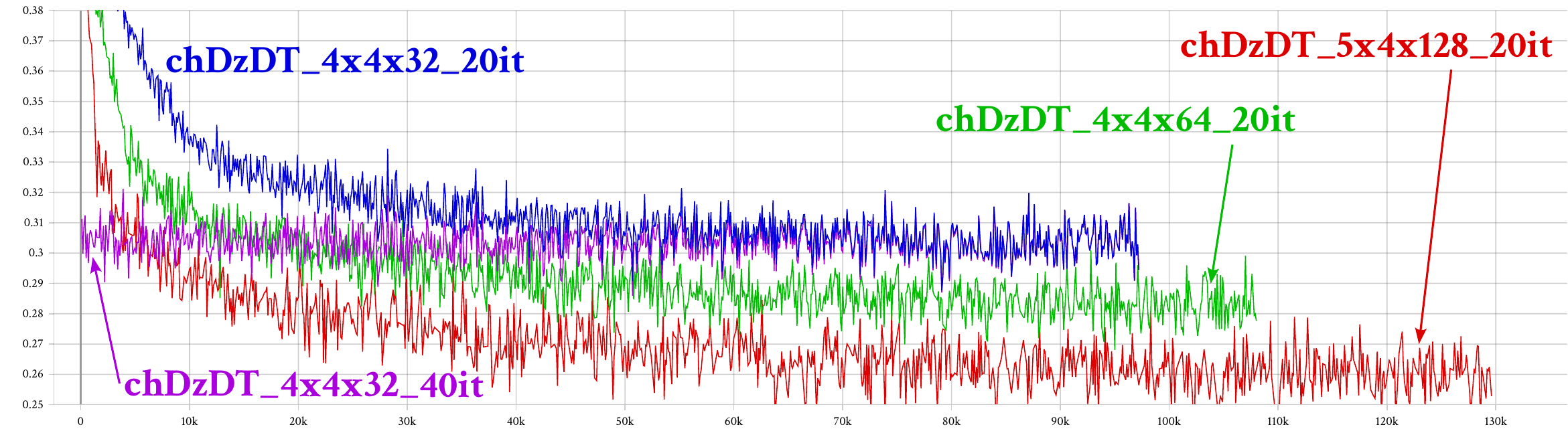}
		\caption{Multi-label loss}
	\end{subfigure}

	\caption{Pre-training loss curves for the main models.  
		The extended model \texttt{chDzDT\_4x4x32\_40it} was initialized from \texttt{chDzDT\_4x4x32\_20it} and trained for an additional 20 epochs.  
		Its curve restarts from zero because training was resumed with a new scheduler.  
		No further reduction in loss is observed beyond the initial 20 epochs.}
	
	\label{fig:pretrain-history}
\end{figure}

All experiments were conducted on a laptop equipped with an Intel Core i7-12650H CPU, 32\,GB of RAM, and an NVIDIA GeForce RTX 3060 GPU (6\,GB VRAM).  
The software environment included KDE Neon, Python 3.10, PyTorch v2.3.1 (CUDA 12.1), and HuggingFace Transformers v4.42.4.  
Figure~\ref{fig:pretrain-history} shows the pre-training loss curves for the three main models.  
Across all variants, 20 epochs proved sufficient, with the loss plateauing well before the final epoch, particularly for smaller models with reduced hidden size $d$.  
To validate this observation, the model \texttt{chDzDT\_4x4x32\_20it} was further trained for an additional 20 epochs using a higher learning rate to accelerate convergence.  
However, the extended model \texttt{chDzDT\_4x4x32\_40it} did not surpass the lowest loss achieved by its 20-epoch counterpart, confirming that additional training yields negligible benefit at this capacity.  
This result suggests that meaningful performance gains are unlikely without increasing model complexity, principally the number of layers $N$ and hidden size $d$.  
Such expansions, however, are constrained by hardware limitations and the design objective of combining this model with syntax-based counterparts at the sentence level.  
Moreover, increasing the hidden size would reduce compatibility with resource-limited downstream applications.  
The loss curves in Figure~\ref{fig:pretrain-history} also show that both the masked language modeling and multi-label losses decrease steadily, indicating no imbalance in the joint objective function and no need for re-weighting.  
In addition, seven smaller ablation models were trained, and their results will be presented in the subsequent experimentation section.

%\newpage
\section{Experiments}
\label{sec:exp}

We conducted a series of experiments to evaluate our model.  
Each subsection addresses one of the following research questions:
\begin{itemize}
	\item \textbf{Q1. Intrinsic evaluation of word embeddings:} To what extent are the character-based word embeddings representative in the absence of external task supervision?
	
	\item \textbf{Q2. Impact on downstream tasks:} How effectively do the unsupervised embeddings support downstream tasks without fine-tuning the full model?
	
	\item \textbf{Q3. Effect of fine-tuning on downstream tasks:} Does fine-tuning the model enhance performance compared to using frozen (static) embeddings?
	
	\item \textbf{Q4. Ablation study:} Which architectural features and hyperparameters most significantly affect embedding quality and task performance?
\end{itemize}

Because standard languages are highly prevalent in Algerian YouTube content, the evaluations cover Arabic, English, and French, which also benefit from abundant publicly available resources.  
Although Arabizi shares structural similarities with Arabic and uses a Latin-based script like French and English, its orthography is highly variable, making reliable annotation more difficult.  
In addition, resources for Arabizi and the Algerian dialect in general are limited.  
Therefore, the Algerian dialect (particularly Arabizi) is evaluated only when resources are available or when annotation is not prohibitively time-consuming.  
Nevertheless, insights gained from standard languages are expected to generalize to the Algerian dialect due to shared morphological patterns and its hybrid script influences.

Since the same models are used across multiple experiments, we describe them here and refer only to the relevant ones in each subsection.  
For chDzDT, we evaluate three variants: \texttt{chDzDT\_5x4x128}, \texttt{chDzDT\_4x4x64}, and \texttt{chDzDT\_4x4x32}, as described earlier.  
As a point of comparison, we include DziriBERT\footnote{\url{https://huggingface.co/alger-ia/dziribert}} \cite{2021-abdaoui-al}, a contextual model pre-trained on Algerian tweets.  
Contextual models pose two challenges: they require contextual information, which is absent in some tasks, and they produce subword embeddings that must be aggregated.  
To address this, we encode each word in isolation and extract both the \texttt{[CLS]} token representation (\texttt{DziriBERT\_cls}) and the mean of subword token embeddings (\texttt{DziriBERT\_tok}).  
For Arabic, we also evaluate AraBERT \cite{2020-antoun-al}, using the \texttt{bert-base-arabertv02-twitter}\footnote{\url{https://huggingface.co/aubmindlab/bert-base-arabertv02-twitter}} variant.  
For English, we use BERT \cite{2019-devlin-al} in its \texttt{bert-base-uncased}\footnote{\url{https://huggingface.co/google-bert/bert-base-uncased}} form.  
For French, we evaluate FlauBERT \cite{2020-le-al}, using the \texttt{flaubert\_base\_uncased}\footnote{\url{https://huggingface.co/flaubert/flaubert_base_uncased}} version.  
All contextual models are processed using both \texttt{[CLS]} and token centroid representations, as with DziriBERT.  
We also include CANINE \cite{2022-clark-al}, using the \texttt{CANINE-C}\footnote{\url{https://huggingface.co/google/canine-c}} variant.  
Since this model outputs both character-level and sentence-level encodings, we use only the \texttt{[CLS]} output.

\subsection{Intrinsic evaluation of word embeddings}

We evaluate the ability of character-based word embeddings to represent morphology, orthography, and compositional structure without external supervision.
Our evaluation includes five intrinsic tests: morphological consistency, robustness to orthographic variation, interpretability through morphemic probing, compositional vector arithmetic, and semantic similarity alignment.

\subsubsection{Morphological consistency}

This experiment addresses the research question: \textit{Do word embeddings cluster morphologically related forms in close proximity within the embedding space?}  
Specifically, we test whether derivational and inflectional variants of a word appear near their root form.  
For example, the verb ``\textit{to know}'' (Ar: ``\RL{علم}'' /\textit{\textquoteleft alima}/, Fr: ``\textit{savoir/connaître}'') has numerous morphological variants.  
Derivational forms include ``\textit{known}'' (Ar: ``\RL{معلوم}'' /\textit{ma\textquoteleft l\={u}m}/, Fr: ``\textit{connu}''),  
``\textit{knower}'' (Ar: ``\RL{عليم}'' /\textit{\textquoteleft al\={\i}m}/, Ar: ``\textit{connaisseur}''),  
and ``\textit{knowledge}'' (Ar: ``\RL{معلومات}'' /\textit{ma\textquoteleft l\={u}m\={a}t}/, Fr: ``\textit{savoir/connai-ssance}'').  
Inflectional forms include conjugations such as  
``\textit{I know}'' (Ar: ``\RL{أنا أعلم}'' /\textit{an\={a} a\textquoteleft lam}/, Fr: ``\textit{je sais/connais}''),  
``\textit{he knows}'' (Ar: ``\RL{هو يعلم}'' /\textit{howa ya\textquoteleft lam}/, Fr: ``\textit{il sait/connaît}''), and  
``\textit{she knew}'' (Ar: ``\RL{هي علمت}'' /\textit{hiya \textquoteleft alimat}/, Fr: ``\textit{elle savait/connaissait}'').  
A character-level encoder is expected to capture these morphological relationships in its representations.  

\paragraph{Task.}  
We focus on verbs as base forms because of their extensive derivational and inflectional potential across the three target languages.  
The hypothesis is that embeddings of base verbs should lie close to those of their derived and inflected forms.  
We frame this as a clustering task with predefined clusters, addressing two questions:  
(1) To what extent do inflectional and derivational variants cluster appropriately in the embedding space?  
(2) How similar are derived forms to their roots?  

\paragraph{Datasets.}  
For Arabic derivational morphology, we used the Arramooz resource\footnote{\url{https://github.com/linuxscout/arramooz-pysqlite}}, which links roots to derived verbs and nouns.  
Clusters were formed with root verbs as centers and derivations as members, yielding 3,191 clusters of at least three words.  
We created three datasets: ``Full'' ($\geq 3$ words), ``Avg'' ($\geq 15$), and ``Min'' ($\geq 30$).  
Nominal derivations were prioritized due to their greater diversity.  
For inflectional morphology, conjugations were generated with Qutrub\footnote{\url{https://github.com/linuxscout/qutrub}}, applying thresholds of 3, 75, and 100.  

For English and French, we used MorphyNet\footnote{\url{https://github.com/kbatsuren/MorphyNet}} \cite{2021-batsuren-al}, which provides both derivational and inflectional forms.  
Datasets followed the same three-tier scheme, with thresholds adjusted for morphological richness:  
English derivations (3, 30, 78); French derivations (3, 15, 30); French inflections (30, 40, 42); and English inflections (4, 6, 7).  

Table~\ref{tab:morph-data-stat} reports word and cluster counts.  
Datasets with many clusters and few words are suited for root--derivation similarity evaluation, while larger clusters are preferable for clustering and visualization.  
We denote datasets as Full/Avg/Min, but exclude the Full version from experiments due to noise and excessive memory usage.

\begin{table}[!htp]
	\centering\small
	
	\caption{Word and cluster counts for morphological datasets across three variants: 
		``\textit{Full}'', ``\textit{Avg}'', and ``\textit{Min}''. 
		The Full sets provide broad coverage but are noisy and memory-intensive, while the Avg and Min sets are used in experiments.}
	
	\begin{tabular}{lrrrrrrrr}
		\hline\hline
		\multirow{2}{*}{\textbf{Dataset}} & \multicolumn{2}{c}{\textbf{Full}} && \multicolumn{2}{c}{\textbf{Avg}} && \multicolumn{2}{c}{\textbf{Min}}  \\
		\cline{2-3}\cline{5-6}\cline{8-9}
		& \textbf{Words} & \textbf{Clusters} && \textbf{Words} & \textbf{Clusters} && \textbf{Words} & \textbf{Clusters}\\
		\hline
%		Arabic word/root deriv. & 27,670  & 3,191 && 9,808  & 496  && 545  &  12  \\
		Arabic noun/root deriv. & 24,165  & 3,019 && 6,953  &  371  && 191 & 6 \\
		English word/root deriv. & 221,044  & 34,294 && 18,384 & 458  && 1,091  &  12 \\
		French word/root deriv. & 70,448  & 15,571 && 4,489 & 205 && 869 &  18 \\
		\hline
		Arabic word/root infl. & 328,440 & 6,826  && 22,228  & 251 && 488  & 4 \\
		English word/root infl. & 98,204 & 24,005  && 1,607  & 262 && 215  & 30 \\
		French word/root infl. & 252,119 & 6,569 && 9,308 & 231 && 493 & 11\\
		\hline\hline
	\end{tabular}
	
	\label{tab:morph-data-stat}
\end{table}

\paragraph{Models.} 
We evaluated our three character-based models alongside DziriBERT and CANINE for all languages. 
AraBERT was included for Arabic, BERT for English, and FlauBERT for French. 
For subword-based models, both the CLS embedding (``\_cls'') and the centroid of token embeddings (``\_tok'') were extracted.

\paragraph{Metrics.}
To assess similarity to roots, we computed the average cosine similarity $ACS \in [0,1]$ (Equation~\ref{eq:avg-cos-sim}).  
For clusters $\mathcal{C} = \{ (r, c)\}$ of root $r$ and members $c$, similarity was calculated using embeddings $emb(w)$ and $emb(r)$.  
Larger values indicate stronger alignment. 
\begin{equation}
	ACS = \frac{1}{\sum_{c \in \mathcal{C}} |c|} \sum_{(r, c) \in \mathcal{C}} \sum_{w \in c} \frac{emb(w) \cdot emb(r)}{||emb(w)|| \, ||emb(r)||}
	\label{eq:avg-cos-sim}
\end{equation}

We also measured the corrected average Euclidean distance $AED \in [0, +\infty)$ (Equation~\ref{eq:avg-eucl-dist}), normalized by $\sqrt{d}$ to allow comparison across dimensions.  
Lower values are preferable, indicating that variants lie close to their roots.  
\begin{equation}
	AED = \frac{1}{\sqrt{d}} \cdot \frac{1}{\sum_{c \in \mathcal{C}} |c|} \sum_{(r, c) \in \mathcal{C}} \sum_{w \in c} ||emb(w) - emb(r)||
	\label{eq:avg-eucl-dist}
\end{equation}

Cluster quality was further assessed with the Silhouette Score \cite{1987-rousseeuw}, defined for each word $i$ by intra-cluster distance $a(i)$ and nearest-cluster distance $b(i)$ (Equation~\ref{eq:sil}).  
Values approaching +1 indicate well-separated clusters, values near 0 suggest cluster overlap, and negative values imply potential misassignments.  
\begin{equation}
	sil = \frac{1}{m} \sum_{i=1}^{m} \frac{b(i) - a(i)}{\max \{ a(i), b(i) \}}
	\label{eq:sil}
\end{equation}

Finally, clustering accuracy was evaluated using the Adjusted Rand Index (ARI) \cite{1985-hubert-arabie}, which compares $k$-means results to gold-standard labels.  
ARI ranges from $-1$ (complete disagreement) to $1$ (perfect agreement), with higher values indicating more accurate cluster recovery.  

\paragraph{Results and discussion.}
Table~\ref{tab:test-morph-deriv} reports the results of the derivational morphology evaluation for the selected models. 
Clustering-based metrics, namely the Adjusted Rand Index (ARI) and Silhouette Score, consistently yield values close to zero, indicating that clusters are poorly separated and often overlapping.
One possible explanation is that $k$-means fails to identify optimal groupings from the embeddings, which would account for the low ARI. 
However, the near-zero Silhouette Score further suggests that, for a given word, the average distance to members of its own cluster is comparable to that of the nearest neighboring cluster.

\begin{table}[!htp]
	\centering\small
	
	\caption{Evaluation of derivational morphology across models. 
		For each model, only the best-performing variant is reported. 
		Bold values indicate the highest score within each column and language. 
		The Silhouette score (Sil) is not bolded, as all models yielded weak clustering results.}
	
	\begin{tabular}{p{1cm}lrrrrrrrrr}
		\hline\hline
		\multirow{2}{*}{\textbf{Lang.}} & \multirow{2}{*}{\textbf{Model}} & \multicolumn{4}{c}{\textbf{Avg}} && \multicolumn{4}{c}{\textbf{Min}}  \\
		\cline{3-6}\cline{8-11}
		&& \textbf{ARI} & \textbf{Sil} & \textbf{ACS} & \textbf{AED} && \textbf{ARI} & \textbf{Sil} & \textbf{ACS} & \textbf{AED} \\
		\hline
		\multirow{4}{*}{\shortstack{Arabic\\ noun/root\\ deriv.}} 
		&  \texttt{AraBERT\_cls} & 0.0133 & -0.1046 & 0.8929 & 0.5547 && 0.0493 & -0.0082 & 0.9051 & 0.5229 \\
		%		&  \texttt{AraBERT\_tok} & 0.0239 & -0.0795 & 0.7385 & 0.6038  && 0.0538 & -0.0074 & 0.7759 & 0.5586 \\
		%		&  \texttt{DziriBERT\_cls} & 0.0219 & -0.0556 & 0.7367 & 0.6193 && 0.0377 & 0.0058 &  0.7124 & 0.6510 \\
		&  \texttt{DziriBERT\_tok} & 0.0211 & -0.0536 & 0.7851 & 0.3652  && 0.0387 & 0.0029 &  0.7789 & 0.3768 \\
		&  \texttt{CANINE-C\_cls} &  \textbf{0.1716} & 0.0313 & 0.7541 & \textbf{0.2580} &&  \textbf{0.7445} & 0.1057 & 0.7488 & \textbf{0.2613} \\
		%		&  \texttt{chDzDT\_5x4x128} & 0.0581 & -0.0371 & 0.8600 & 0.5119  && 0.1516 & 0.0640 & 0.8322 & 0.5606 \\
		%		&  \texttt{chDzDT\_4x4x64} & 0.0224 & -0.1081 & 0.8607 & 0.5375 && 0.0948 & 0.0220 &  0.8523 & 0.5525 \\
		&  \texttt{chDzDT\_4x4x32} & 0.0295 & -0.1200 & \textbf{0.9215} & 0.4240  && 0.0304 & 0.0475 &  \textbf{0.9105} & 0.4505 \\
		\hline
		\multirow{4}{*}{\shortstack{English\\ word/root\\ deriv.}} 
		&  \texttt{BERT\_cls} & 0.1200 & -0.0480 & \textbf{0.8864} & \textbf{0.2411} && 0.4608 & 0.0896 & \textbf{0.8488} & 0.2804 \\
		%		&  \texttt{BERT\_tok} & 0.1634 & 0.0074 &  0.6950 & 0.2584  && 0.4591 & 0.0863 &   0.6395 & 0.2858 \\
		&  \texttt{DziriBERT\_cls} & \textbf{0.2674} & 0.0332 & 0.7249 & 0.6095  && 0.5489 & 0.1582 &   0.7296 & 0.6016  \\
		%		&  \texttt{DziriBERT\_tok} & 0.2553 & 0.0228 & 0.8277 & 0.2826 && 0.6125 & 0.1294 &   0.8181 & 0.2869  \\
		&  \texttt{CANINE-C\_cls} & 0.0913 & -0.0083 & 0.7397 & 0.2649 && 0.6168 & 0.0629 & 0.7537 & \textbf{0.2606} \\
		&  \texttt{chDzDT\_5x4x128} & 0.1586 & -0.0177 & 0.7509 & 0.7073 && \textbf{0.6248} & 0.1440 &  0.7391 & 0.7286 \\
		%		&  \texttt{chDzDT\_4x4x64} & 0.0434 & -0.0766 & 0.7097 & 0.7963  && 0.2882 & 0.0467 &  0.6641 & 0.8663  \\
		%		&  \texttt{chDzDT\_4x4x32} & 0.0395 & -0.1554 & 0.8204 & 0.6643  && 0.1537 & -0.0096 &  0.7899 & 0.7261  \\
		\hline
		\multirow{4}{*}{\shortstack{French\\ word/root\\ deriv.}} 
		%		&  \texttt{FlauBERT\_cls} & 0.0210 & -0.1369 & 0.4979 & 1.4475  && 0.0383 & -0.0572 &  0.4555 & 1.5222  \\
		&  \texttt{FlauBERT\_tok} & 0.0220 & -0.1575 & 0.4995 & 0.5060  && 0.0456 & -0.0844 &  0.4585 & 0.9675  \\
		%		&  \texttt{DziriBERT\_cls} & \textbf{0.2943} & 0.0444 &   0.7385 & 0.5897  && 0.3944 & 0.1206 &  0.7245 & 0.6027  \\
		&  \texttt{DziriBERT\_tok} & \textbf{0.2679} & 0.0238 &  \textbf{0.8319} & 0.2800  && \textbf{0.4635} & 0.1119 &  \textbf{0.8005} & 0.2923  \\
		&  \texttt{CANINE-C\_cls} & 0.1329 & 0.0021 & 0.7260 & \textbf{0.2674} && 0.2567 & 0.0534 & 0.7196 & \textbf{0.2732}  \\
		&  \texttt{chDzDT\_5x4x128} & 0.1849 & -0.0478 &  0.7478 & 0.7011 && 0.2730 & 0.0638 &  0.7713 & 0.6737  \\
		%		&  \texttt{chDzDT\_4x4x64} & 0.0975 & -0.1006 &  0.7239 & 0.7638 && 0.2220 & -0.0168 &  0.7191 & 0.7825  \\
		%		&  \texttt{chDzDT\_4x4x32} & 0.0914 & -0.1758 &  0.7919 & 0.7067 && 0.1906 & -0.0647 &  0.7648 & 0.7686  \\
		\hline\hline
		
	\end{tabular}
	
	\label{tab:test-morph-deriv}
\end{table}

Although overall scores remain modest in the average dataset, ARI values are relatively higher for English and French, with DziriBERT performing best, closely followed by our model, chDzDT. 
Arabic derivational morphology is considerably more complex than English and French: root verbs can undergo substantial transformations through infixes and extended affixes, sometimes exceeding the root in length (e.g., ``\RL{بكى}'' /\textit{bak\={a}}/ ``to cry'' $\rightarrow$ ``\RL{متباكي}'' /\textit{mutab\={a}k\={\i}}/ ``\textit{one who pretends to cry}''). 
In such cases, subword-based models may struggle because token similarity diminishes when the root form is obscured. 
Character-based models may also be affected, as affix characters frequently outnumber the root's three-character base. 
Among the evaluated systems, CANINE is the only model to achieve consistently strong results under these conditions.

In the minimal dataset, clustering performance improves substantially, particularly for English (BERT) and French (DziriBERT). 
This gain likely stems from the simpler cluster structure, characterized by fewer clusters with more items each. 
Interestingly, for French, token-centroid embeddings outperform CLS-based embeddings, suggesting that tokenizers in these models successfully isolate root and suffix components. 
Our model chDzDT performs competitively with DziriBERT, which is noteworthy given that BERT and FlauBERT are language-specific, whereas chDzDT is multilingual.

A likely reason for the poor clustering results overall is that models tend to emphasize affixes rather than root morphemes, making words with similar affixes appear closer even when their roots differ. 
By contrast, average cosine similarity (ACS) indicates strong alignment between derived forms and their roots across all models and datasets. 
Specifically, chDzDT achieves the highest ACS for Arabic, BERT for English, and DziriBERT for French, with chDzDT ranking second in the latter two languages. 
This suggests that while all models capture root--derivation similarity, they do so via different mechanisms: subword models leverage semantic co-occurrence and shared subwords, while character-based models rely primarily on morphological patterns.

For Adjusted Euclidean Distance (AED), chDzDT produces smaller values, though this may partly reflect its lower dimensionality. 
When normalized, the distances of the three other models become more competitive. 
Nonetheless, because chDzDT is designed for downstream word-level applications, its compact size and competitive performance position it as a practical alternative to larger, language-specific models.

Table~\ref{tab:test-morph-infl} reports results of the inflectional morphology evaluation for selected models.
As in the previous table, only the best-performing variant of each model is shown to conserve space.
The inflectional forms considered here are primarily verb conjugations, which are generally more morphologically diverse and complex than nominal forms.
The overall trends mirror those observed in the derivational morphology results: chDzDT consistently ranks first or second based on ARI and ACS scores.
Notably, \texttt{chDzDT\_4x4x32} ($d = 32$) outperformed its larger counterparts ($d = 64$ and $d = 128$), highlighting the effectiveness of the proposed architecture in capturing morphological regularities, supporting higher-level tasks, and simultaneously reducing vocabulary sparsity and memory usage.

\begin{table}[!htp]
	\centering\small
	
	\caption{Evaluation of inflectional morphology across models. 
		Only the best-performing variant of each model is shown. 
		Bold values indicate the highest score within each column and language. 
		The Silhouette score (Sil) is not bolded, as all models yielded weak results.}
	\begin{tabular}{p{1cm}lrrrrrrrrr}
		\hline\hline
		\multirow{2}{*}{\textbf{Lang.}} & \multirow{2}{*}{\textbf{Model}} & \multicolumn{4}{c}{\textbf{Avg}} && \multicolumn{4}{c}{\textbf{Min}}  \\
		\cline{3-6}\cline{8-11}
		&& \textbf{ARI} & \textbf{Sil} & \textbf{ACS} & \textbf{AED} && \textbf{ARI} & \textbf{Sil} & \textbf{ACS} & \textbf{AED} \\
		\hline
		\multirow{4}{*}{\shortstack{Arabic\\ word/root\\ infl.}} 
		&  \texttt{AraBERT\_cls} & 0.0178 & -0.0769 & \textbf{0.8988} & 0.5384 && 0.0250 & 0.0148 & \textbf{0.9109} & 0.5128 \\
%		&  \texttt{AraBERT\_tok} & 0.0430 & -0.0097 & 0.7527 & 0.5976 &&  0.0644 & 0.0568 & 0.7162 & 0.6225 \\
%		&  \texttt{DziriBERT\_cls} & 0.0317 & -0.0260 & 0.7383 & 0.6164 && 0.0567 & 0.0150 & 0.7246 & 0.6359 \\
		&  \texttt{DziriBERT\_tok} & 0.0367 & -0.0145 & 0.8135 & 0.3320 && 0.0634 & 0.0188 & 0.7636 & 0.3764 \\
		&  \texttt{CANINE-C\_cls} & 0.0534 & 0.0038 & 0.6047 & \textbf{0.3284}  && 0.0589 & 0.0489 & 0.5949 & \textbf{0.3414}  \\
%		&  \texttt{chDzDT\_5x4x128} & 0.0140 & -0.0544 & 0.6076 & 0.9107 && 0.0873 & 0.0485 & 0.6343 & 0.8771 \\
%		&  \texttt{chDzDT\_4x4x64} & 0.0142 & -0.0884 & 0.6262 & 0.9188 && 0.1006 & 0.0236 & 0.6978 & 0.8188 \\
		&  \texttt{chDzDT\_4x4x32} & 0.0276 & -0.1245 & 0.8222 & 0.6837 && 0.0725 & 0.0036 & 0.8297 & 0.6678 \\
		\hline
		\multirow{4}{*}{\shortstack{English\\ word/root\\ infl.}} 
		&  \texttt{BERT\_cls} & 0.0300 & -0.1038 & 0.9193 & \textbf{0.1996} && 0.0265 & -0.0913 & \textbf{0.9182} & \textbf{0.2003} \\
%		&  \texttt{BERT\_tok} & 0.1581 & 0.0325 & 0.7609 & 0.2216 && 0.2138 & 0.0188 & 0.7475 & 0.2252 \\
%		&  \texttt{DziriBERT\_cls} & 0.1288 & 0.0496 & 0.8180 & 0.5045 && 0.2375 & 0.0372 & 0.7790 & 0.5475 \\
		&  \texttt{DziriBERT\_tok} & 0.2498 & 0.0933 & 0.8709 & 0.2616 && 0.3919 & 0.0732 & 0.8392 & 0.2880 \\
		&  \texttt{CANINE-C\_cls} &  \textbf{0.3334} & 0.0785 & 0.7506 & 0.2537 && \textbf{0.4298} & 0.0858 & 0.7107 & 0.2761   \\
%		&  \texttt{chDzDT\_5x4x128} & 0.1347 & 0.0025 & 0.8262 & 0.5791 && 0.1852 & 0.0303 & 0.7874 & 0.6357 \\
%		&  \texttt{chDzDT\_4x4x64} & 0.1129 & -0.0367 & 0.8468 & 0.5638 && 0.3141 & 0.0333 & 0.8160 & 0.6125 \\
		&  \texttt{chDzDT\_4x4x32} & 0.2490 & -0.0103 & \textbf{0.9287} & 0.3975 && 0.3413 & 0.0723 & 0.8948 & 0.4682 \\
		\hline
		\multirow{4}{*}{\shortstack{French\\ word/root\\ infl.}} 
%		&  \texttt{FlauBERT\_cls} & 0.0376 & -0.1335 & 0.4753 & 1.3876 && 0.0972 & -0.0662 & 0.5185 & 1.4147\\
		&  \texttt{FlauBERT\_tok} & 0.0424 & -0.1368 & 0.4814 & 0.8528 && 0.1030 & -0.0462 & 0.5177 & 0.7950\\
%		&  \texttt{DziriBERT\_cls} & 0.1247 & 0.0544 & 0.7898 & 0.5262 && 0.4463 & 0.0713 & 0.7451 & 0.5825\\
		&  \texttt{DziriBERT\_tok} & 0.2869 & 0.0974 & 0.8700 & 0.2425 && 0.5005 & 0.1151 & 0.8165 & 0.2981\\
		&  \texttt{CANINE-C\_cls} & \textbf{0.4327} & 0.0958 & 0.8009 & \textbf{0.2241}  && \textbf{0.5430} & 0.1462 & 0.7448 & \textbf{0.2512 } \\
%		&  \texttt{chDzDT\_5x4x128} & 0.0762 & -0.0098 & 0.8283 & 0.5721 && 0.2256 & 0.1015 & 0.8046 & 0.6207\\
%		&  \texttt{chDzDT\_4x4x64} & 0.1636 & -0.0042 & 0.8679 & 0.5125 && 0.3779 & 0.1304 & 0.8266 & 0.5950\\
		&  \texttt{chDzDT\_4x4x32} & 0.2293 & -0.0059 & \textbf{0.9260} & 0.3922 && 0.4693 & \textbf{0.2066} & \textbf{0.9214} & 0.4170\\
		\hline\hline
		
	\end{tabular}
	
	\label{tab:test-morph-infl}
\end{table}

\subsubsection{Robustness to orthographic noise}

This experiment addresses the research question: \textit{Are word embeddings resilient to spelling obfuscation and other forms of non-standard orthography?}
As in other online communities, Algerian users often employ \emph{grawlix}, in which characters are replaced with symbols such as ``*'' or ``\#'' to obscure taboo or sensitive terms.
Obfuscation can involve replacing a character with a visually similar symbol (e.g., ``o'' $\rightarrow$ ``0'') or with a phonetically similar character from another script (e.g., Latin ``d'' $\rightarrow$ Arabic ``\RL{د}'').
Additionally, Algerian Arabizi lacks standardized orthography, so a single word may appear in multiple variant forms.

\paragraph{Task.}
Given a list of standard words, we evaluate embedding similarity to noisy counterparts created via obfuscation or orthographic variation.
Two evaluation modes are considered:
(1) computing similarity between clean and noisy forms, and
(2) clustering multiple variants of the same word.
Taboo words are selected as the primary test case, given their frequent obfuscation in online communication.

\paragraph{Datasets.}
For English\footnote{\url{https://github.com/MauriceButler/badwords}} and French\footnote{\url{https://github.com/darwiin/french-badwords-list}}, we used publicly available lists of taboo words.
These resources also provide obfuscated variants generated by character substitutions, e.g.:
``a'' $\rightarrow$ ``@'' or ``4'', 
``s'' $\rightarrow$ ``5'' or ``\$'',
``i'' $\rightarrow$ ``!'' or ``1'', 
``o'' $\rightarrow$ ``0'', 
``e'' $\rightarrow$ ``3'', 
``t'' $\rightarrow$ ``+'' or ``7''.
Multiple obfuscated forms may exist per word.
For the English set, some inflected forms were removed and additional obfuscations were programmatically added to enlarge clusters.
Each cluster is represented by its canonical, normalized form.

For Algerian Arabizi, no pre-existing resource is available.
We compiled a list of taboo terms with spelling variants that are not strictly obfuscations but orthographic alternatives.
For instance, the phoneme ``\RL{خ}'' /\textit{kh}/ may appear as ``kh'', ``5'', or ``7''.
Other variations include differences in vowel representation (``ou'' vs.\ ``o''), optional insertion of ``e'' to separate consonants, and alternate Latin encodings of Arabic digraphs (``ch'' vs.\ ``sh''). 
For example, the word ``\textit{fatso}'' can appear as ``\textit{bouche7ma}'' or ``\textit{bouch7ma}''.
Further substitutions, such as ``ou'' $\rightarrow$ ``oo/o'', ``ch'' $\rightarrow$ ``sh'', and ``7'' $\rightarrow$ ``h'', can produce up to 24 variants.
Although some forms may not occur naturally, they provide a useful stress test for embeddings.
All obfuscations and non-standard spellings are treated as orthographic noise.

Standard Arabic, being highly codified, exhibits negligible intra-word variation; non-standard forms typically occur only in dialectal contexts.
For each language, we constructed a compact dataset of condensed clusters for visualization.

We also created a supplementary dataset for targeted obfuscation tests, including grawlix substitutions (replacing one or two characters with * or \#), visually similar Unicode replacements, and phonetic substitutions from other scripts.
For English and French, these replacements follow the patterns used in the clustering dataset (one variant retained per word).
For Arabizi, Arabic-character substitutions are replaced with phonetically similar alternatives, and for Arabic, taboo terms are modified with Latin-script equivalents.
Table~\ref{tab:noise-data-stat} summarizes statistics for the noisy-word datasets.

\begin{table}[!htp]
	\centering\small
	
	\caption{Statistics of the orthographic-noise datasets. 
		``Noisy clusters'' refer to the complete sets of obfuscated or variant forms, while ``Compact noisy clusters'' are reduced subsets used for visualization. 
		``Obfuscation tuples'' denote word--variant pairs for targeted tests.}

	\begin{tabular}{lrrrrrrr}
		\hline\hline
		\multirow{2}{*}{\textbf{Lang.}} & \multicolumn{2}{c}{\textbf{Noisy clusters}} && \multicolumn{2}{c}{\textbf{Compact Noisy clusters}} && \multirow{2}{*}{\textbf{\shortstack{Obfuscation\\ Tuples}}}  \\
		\cline{2-3}\cline{5-6}
		& \textbf{variants} & \textbf{words} && \textbf{variants} & \textbf{words} &&  \\
		\hline
		Arabic & /  & / && /  & /  && 262  \\
		Arabizi & 2,548  & 249 && 526  & 16  && 284  \\
		English & 10,235  & 308 && 4,381  & 16  && 308  \\
		French & 2,438  & 234 && 676  & 15  &&  233 \\
		\hline\hline
	\end{tabular}
	
	\label{tab:noise-data-stat}
\end{table}

\paragraph{Models.} 
In this experiment, we evaluated our three character-based models alongside DziriBERT and CANINE for all languages. 
AraBERT was included for Arabic, BERT for English, and FlauBERT for French. 
For each subword-based model, we extracted both the CLS embedding (``\_cls'') and the centroid of token embeddings (``\_tok'').

\paragraph{Metrics.} 
Evaluation used two metrics: (1) average cosine similarity (ACS) between clean and noisy forms, and (2) the Adjusted Rand Index (ARI) with $k$-means clustering for grouping variants. 
t-SNE \cite{2008-vandermaaten-hinton} was applied for qualitative visualization.

\paragraph{Results and discussion.}
Table~\ref{tab:test-noisy} presents the results of the orthographic-noise tasks, where \textit{ACS1*} refers to ACS with one letter replaced by ``*'', \textit{ACS1\#} with replacement by ``\#'', and \textit{ACS1c} with replacement by a visually or phonetically similar character. 
Clustering performance (ARI) shows that English embeddings struggle to group noisy variants, except for character-based models (chDzDT and CANINE). 
The dataset is synthetic in nature and often contains unattested word forms. 
Thus, these results highlight the models' ability to integrate novel forms into the embedding space while preserving proximity to their canonical variants.
DziriBERT performs strongly on French and Arabizi, likely due to exposure to Algerian Twitter data containing such variation.

Based on ACS, chDzDT consistently yields higher similarity scores between noisy forms and their standard counterparts. 
This is especially true for single-character obfuscations, confirming its robustness. 
Among the three chDzDT configurations, the 32-dimensional variant performs best, which is encouraging given its suitability for larger-scale applications. 
Across all models, ``*'' obfuscations outperform ``\#'', likely reflecting the greater web frequency of ``*'' patterns during training.

\begin{table}[!htp]
	\centering\small
	
	\caption{Performance of models on orthographic-noise tasks across languages. 
		For each model, only the best-performing variant is shown. 
		Bold values mark the highest score within each column and language.}
	
	\begin{tabular}{p{1cm}lrrrrrr}
		\hline\hline
		\multirow{2}{*}{\textbf{Lang.}} & \multirow{2}{*}{\textbf{Model}} & \multicolumn{2}{c}{\textbf{Noisy clusters}} && \multicolumn{3}{c}{\textbf{Tuple (word, noise)}}  \\
		\cline{3-4}\cline{6-8}
		&& \textbf{ARI} & \textbf{ACS} && \textbf{ACS1*} & \textbf{ACS1\#} & \textbf{ACS1c} \\
		\hline
		\multirow{4}{*}{Arabic} 
		&  \texttt{AraBERT\_cls} & / & / && 0.8611 & 0.8262 & 0.8667 \\
%		&  \texttt{AraBERT\_tok} & & && 0.6585 & 0.6087 & 0.7135  \\
%		&  \texttt{DziriBERT\_cls} & & && 0.7338 & 0.7183 & 0.7683 \\
		&  \texttt{DziriBERT\_tok} & / & / && 0.7840 & 0.7775 & 0.8092  \\
		&  \texttt{CANINE-C\_cls} & / & / && 0.7158 & 0.6940 & 0.8547   \\
%		&  \texttt{chDzDT\_5x4x128} & & && 0.9501 & 0.7356 & 0.9137 \\
%		&  \texttt{chDzDT\_4x4x64} & & && 0.9646 & 0.8002 & 0.9338 \\
		&  \texttt{chDzDT\_4x4x32} & / & / && \textbf{0.9832} & \textbf{0.8539} & \textbf{0.8980} \\
		\hline
		\multirow{3}{*}{Arabizi} 
		%		&  \texttt{DziriBERT\_cls} & 0.2035 & 0.7859 && 0.7674 & 0.7428 & 0.7991 \\
		&  \texttt{DziriBERT\_tok} & 0.2156 & 0.8559 && 0.8264 & 0.8014 & 0.8468 \\
		&  \texttt{CANINE-C\_cls} & \textbf{0.3484} & 0.7554 && 0.6725 & 0.6398 & 0.8249 \\
		%		&  \texttt{chDzDT\_5x4x128} & 0.2593 & 0.8347 && 0.9416 & 0.8065 & 0.8523 \\
		&  \texttt{chDzDT\_4x4x64} & 0.2050 & \textbf{0.9009} && \textbf{0.9640} & \textbf{0.8539} & \textbf{0.8843} \\
		%		&  \texttt{chDzDT\_4x4x32} & 0.1442 & 0.9465 && 0.9794 & 0.9574 & 0.9030 \\
		\hline
		\multirow{4}{*}{English} 
		&  \texttt{BERT\_cls} & 0.0360 & \textbf{0.8388} && 0.9014 & 0.8903 & 0.8795 \\
%		&  \texttt{BERT\_tok} & 0.0458 & 0.5931 && 0.6389 & 0.6557 & 0.6829  \\
%		&  \texttt{DziriBERT\_cls} & 0.0536 & 0.6716 && 0.7837 & 0.7516 & 0.8100 \\
		&  \texttt{DziriBERT\_tok} & 0.0544 & 0.7578 && 0.8512 & 0.8208 & 0.8724 \\
		&  \texttt{CANINE-C\_cls} & 0.1031 & 0.6103 && 0.6900 & 0.6548 & 0.7618 \\
%		&  \texttt{chDzDT\_5x4x128} & 0.1701 & 0.5902 && 0.9486 & 0.7404 & 0.7150 \\
%		&  \texttt{chDzDT\_4x4x64} & 0.1597 & 0.6907 && 0.9727 & 0.7624 & 0.8039 \\
		&  \texttt{chDzDT\_4x4x32} & \textbf{0.1267} & 0.8086 && \textbf{0.9882} & \textbf{0.9241} & \textbf{0.8960} \\
		\hline
		\multirow{4}{*}{French} 
%		&  \texttt{FlauBERT\_cls} & 0.0268 & 0.5783 && 0.6353 & 0.6434 & 0.6408\\
		&  \texttt{FlauBERT\_tok} & 0.0206 & 0.5875 && 0.6391 & 0.6472 & 0.6439 \\
%		&  \texttt{DziriBERT\_cls} & 0.1130 & 0.6745 && 0.7373 & 0.7037 & 0.7746 \\
		&  \texttt{DziriBERT\_tok} & 0.1248 & 0.7597 && 0.8153 & 0.7816 & 0.8438 \\
		&  \texttt{CANINE-C\_cls} & \textbf{0.2204} & 0.6660 && 0.7023 & 0.6618 & 0.7572\\
%		&  \texttt{chDzDT\_5x4x128} & 0.2461 & 0.6071 && 0.9616 & 0.7870 & 0.7397 \\
%		&  \texttt{chDzDT\_4x4x64} & 0.2314 & 0.6882 && 0.9766 & 0.7826 & 0.7775 \\
		&  \texttt{chDzDT\_4x4x32} & 0.2169 & \textbf{0.7754} && \textbf{0.9897} & \textbf{0.9355} & \textbf{0.8587}\\
		\hline\hline
		
	\end{tabular}
	
	\label{tab:test-noisy}
\end{table}

\begin{figure}[!htp]
	\centering\small
	
	\begin{tabular}{c@{}c@{}c@{}c@{}c@{}c}
		\includegraphics[width=0.15\textwidth]{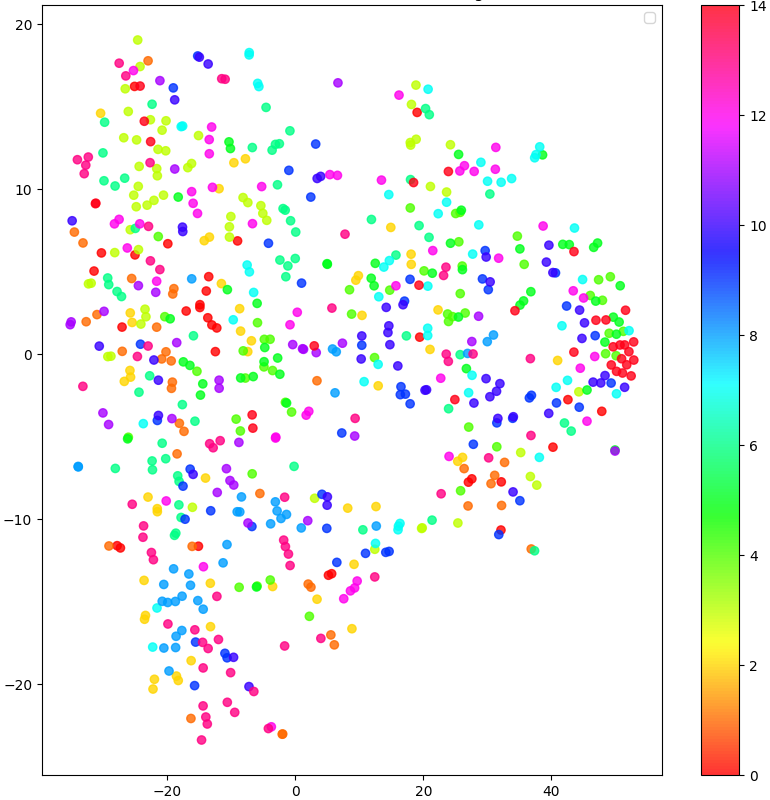} &
		\includegraphics[width=0.15\textwidth]{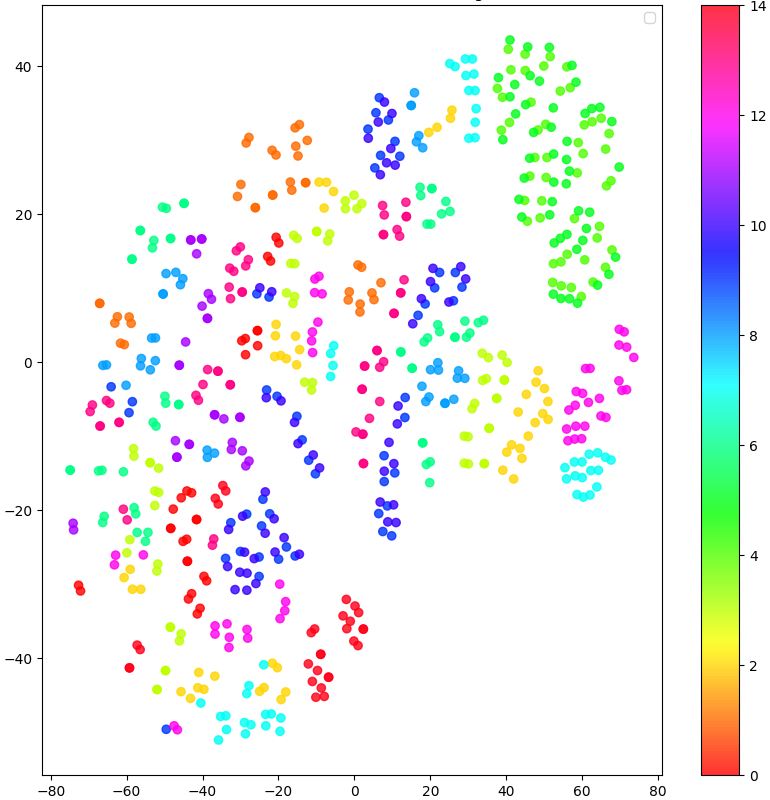} &
		\includegraphics[width=0.15\textwidth]{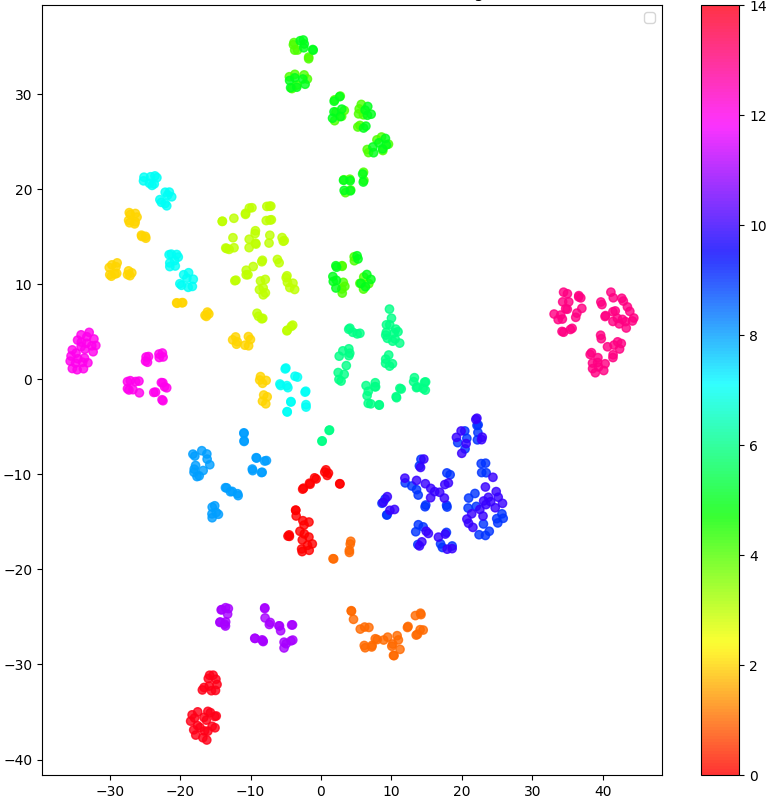} & 
		\includegraphics[width=0.15\textwidth]{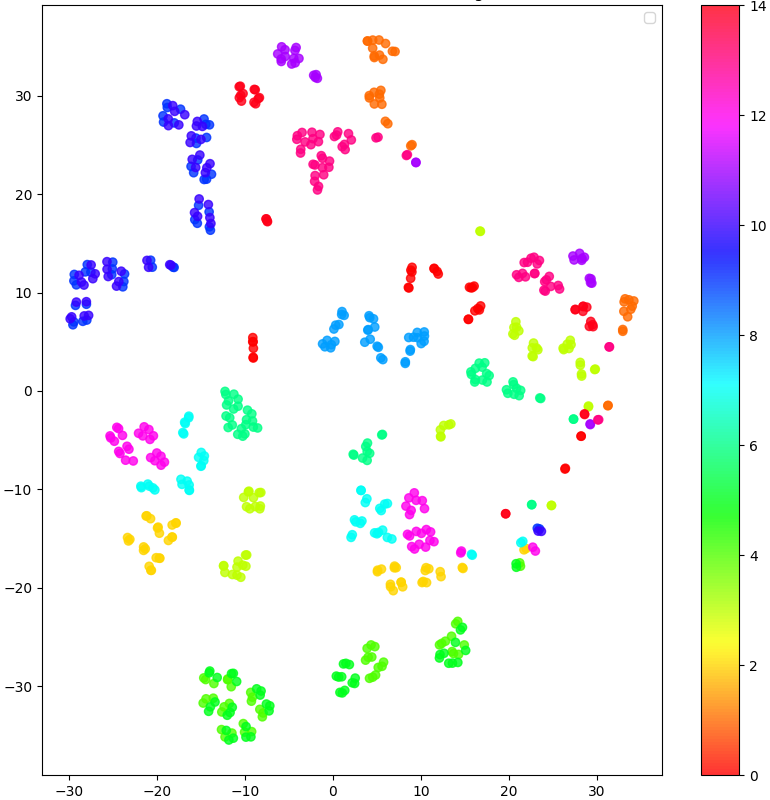} &
		\includegraphics[width=0.15\textwidth]{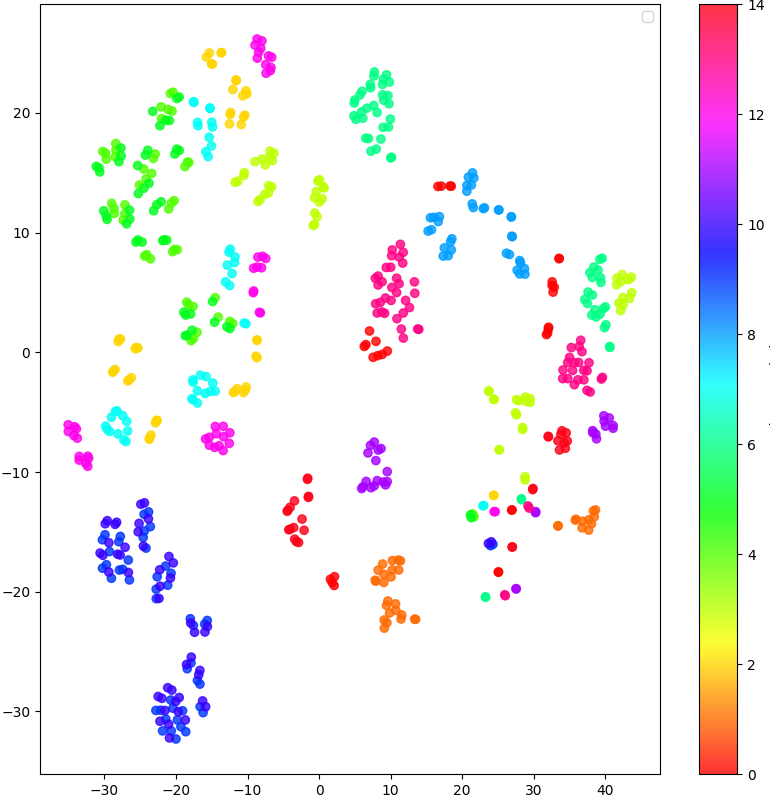} &
		\includegraphics[width=0.15\textwidth]{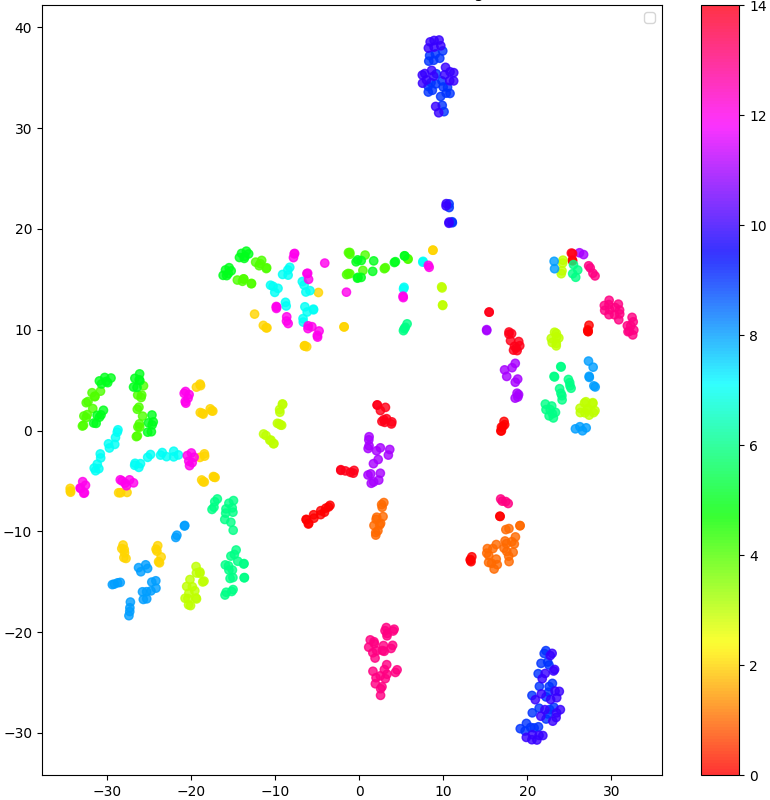} \\
		\texttt{FlauBERT\_cls} &  \texttt{DziriBERT\_cls} & \texttt{CANINE-C\_cls} & \texttt{chDzDT\_5x4x128} & \texttt{chDzDT\_4x4x64} & \texttt{chDzDT\_4x4x32} \\
	\end{tabular}
	
	\caption{t-SNE visualization of embeddings from the compact French noisy-clusters dataset. 
		The figure displays 15 word-variant clusters, each shown in a different color (best viewed in color). 
		Ideally, embeddings from the same cluster should be compact and clearly separated from those of other clusters.}
	
	\label{fig:test-noisy-tsne}
\end{figure}

To further assess clustering capacity, we visualized embeddings from the compact French noisy-clusters dataset using t-SNE.  
This dataset was selected for its lower synthetic bias and to conserve space.  
Figure~\ref{fig:test-noisy-tsne} shows that CANINE achieves the clearest separation of word variants, consistent with the results in Table~\ref{tab:test-noisy}.  
The next-best performance is obtained by the three chDzDT models, followed by DziriBERT, with FlauBERT showing the least separation.

\subsubsection{Interpretability through morphemic probing}

This experiment addresses the following research question: \textit{Do word embeddings encode morphemic structure such that affixes can be linearly separated in the representation space?}  
Specifically, we test whether embeddings capture affixes (e.g., \textit{un-}, \textit{-ness} in \textit{unhappiness}) in a way that allows them to be isolated by a linear decision boundary.  
To this end, a linear probe (logistic regression or linear SVM) is trained on the embeddings to predict the presence of specific morphemes.  
If the probe achieves strong performance, this suggests that the embedding space retains sufficient morphological information to make morphemes linearly separable. 
This, in turn, provides an interpretable link between the representation and internal word structure.

\paragraph{Task.} 
For each dataset and encoding model, we train a multi-label logistic regression classifier to detect designated affixes.  
A multi-label setup is necessary because words may contain multiple affixes simultaneously.
This property is common to Arabic, English, and French, and naturally extended to Arabizi.

\paragraph{Datasets.} 
Experiments are limited to French and English, for which morpheme--affix lexicons are readily available.  
Arabic was excluded due to the required additional annotation effort, particularly for infixes.  
Although morphologically richer, Arabic patterns can to some extent be generalized from these two languages.  
For English, we used MorphoLex-En\footnote{\url{https://github.com/hugomailhot/MorphoLex-en}} \cite{2018-Sanchez-Gutierrez-al}, removing words without affixes to reduce size and retaining only affixes with at least 200 instances.  
For French, we relied on MorphoLex-Fr\footnote{\url{https://github.com/hugomailhot/morpholex-fr}} \cite{2020-Mailhot-al} without word removal, as the dataset is already moderate in scale.  
Both datasets were split into training and test sets (60/40) using stratified sampling via \texttt{iterative-stratification}\footnote{\url{https://github.com/trent-b/iterative-stratification}} to preserve affix distribution.  
Table~\ref{tab:prob-data-stat} summarizes the dataset statistics.

\begin{table}[!htp]
	\centering\small
	
	\caption{Dataset statistics for the morphemic probing experiments. 
		Each instance represents a word, with affixes encoded as binary features (present or absent). 
		Only affixes with at least 200 occurrences are included.}
	
	\begin{tabular}{lrrr}
		\hline\hline
		\textbf{Lang.} & \textbf{Train words} & \textbf{Test words} & \textbf{Affixes}  \\
		\hline
		English & 17,410 & 11,607 & 58\\
		French & 9,376 & 6,252  & 23\\
		\hline\hline
	\end{tabular}
	
	\label{tab:prob-data-stat}
\end{table}

\paragraph{Models.} 
We evaluated our three character-based models along with DziriBERT and CANINE.  
For English, we included BERT; for French, FlauBERT.  
For both BERT and FlauBERT, we assessed the \texttt{[CLS]} embedding (``\_cls'') and the centroid of token embeddings (``\_tok'').

\paragraph{Metrics.} 
Performance is reported using macro-averaged precision, recall, and F1-score, which give equal weight to both rare and frequent labels.  
For a label $l$, let $TP_l$, $FP_l$, and $FN_l$ denote true positives, false positives, and false negatives, respectively.  
Precision $P_l$, recall $R_l$, and F1-score $F1_l$ are defined in Equation~\ref{eq:prf1}.
\begin{equation}
	P_l = \frac{TP_l}{TP_l + FP_l}, \quad
	R_l = \frac{TP_l}{TP_l + FN_l}, \quad
	F1_l = \frac{2 P_l R_l}{P_l + R_l}.
	\label{eq:prf1}
\end{equation}

\begin{figure}[!htp]
	\centering
	
	\begin{subfigure}{0.49\textwidth}
		\includegraphics[width=\textwidth]{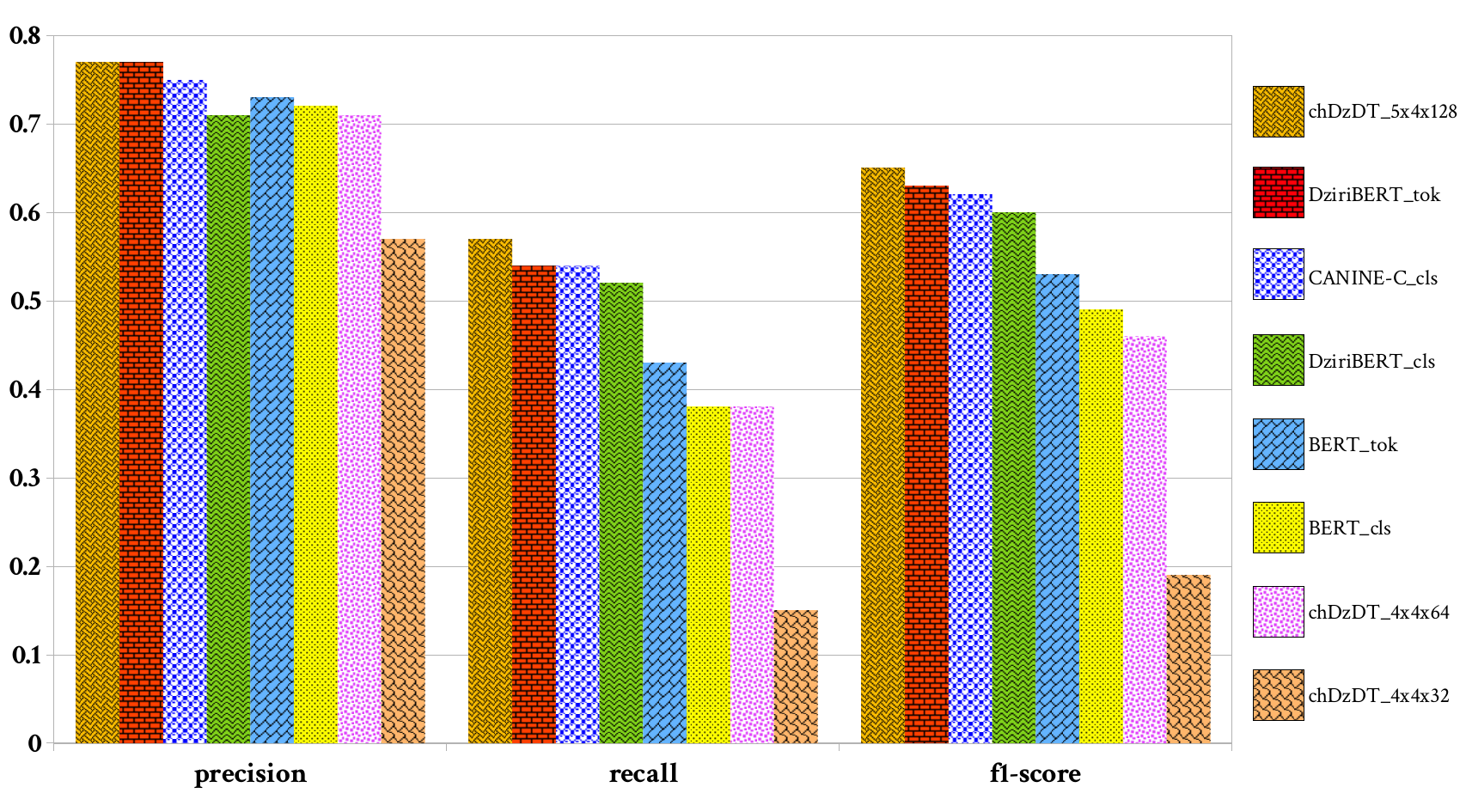}
		\subcaption{English}
	\end{subfigure}
	\begin{subfigure}{0.49\textwidth}
		\includegraphics[width=\textwidth]{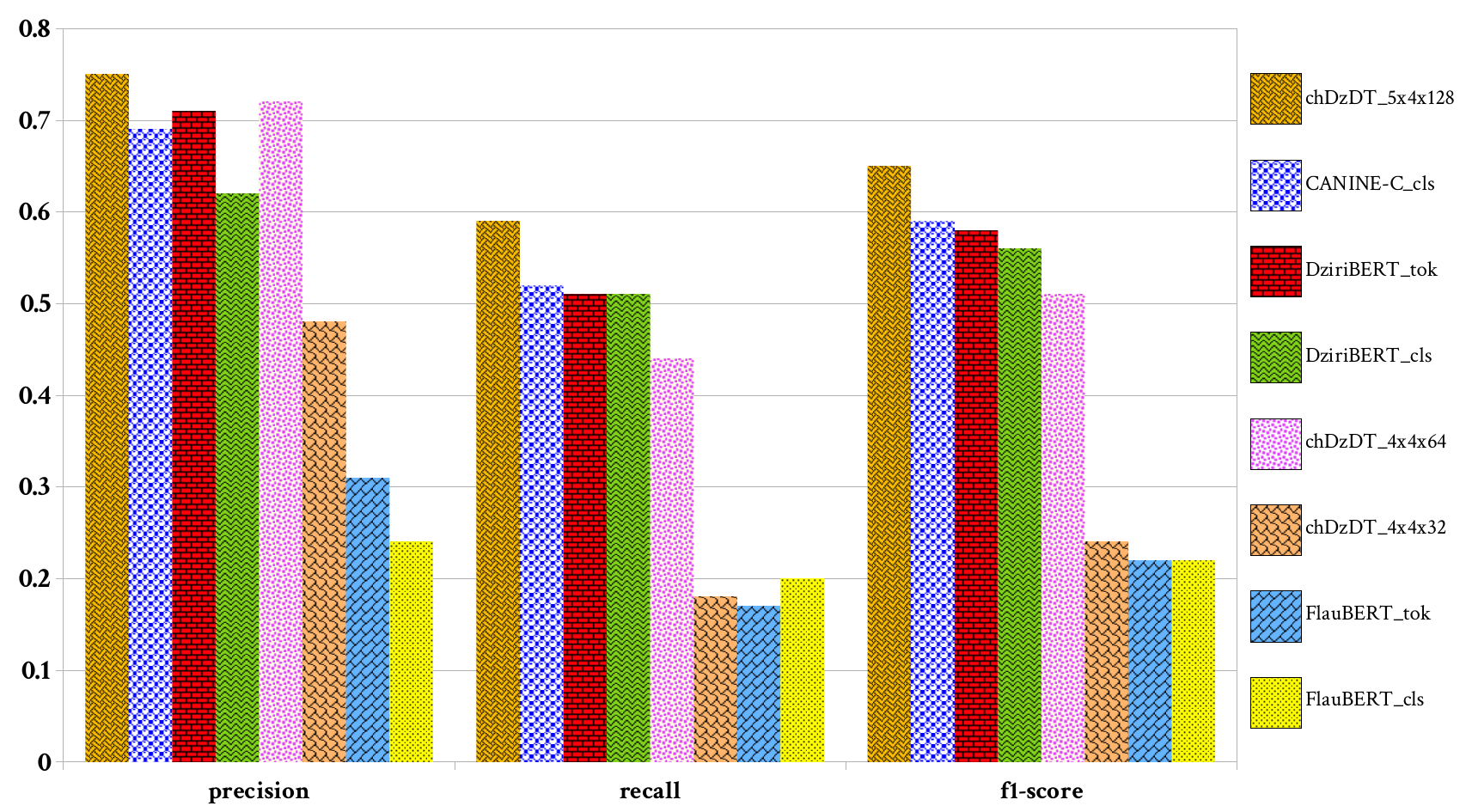}
		\subcaption{French}
	\end{subfigure}
	
	\caption{Macro-averaged precision, recall, and F1-scores for morphemic probing in English and French. 
		Models are ordered by descending F1-score.}

	\label{fig:test-prob}
\end{figure}

\paragraph{Results and discussion.}
Figure~\ref{fig:test-prob} reports macro-averaged precision, recall, and F1-scores.  
The 128-dimensional model achieved the best performance in both languages, suggesting that larger embeddings better capture morphological information.  
The other two character-based models perform worst in English and second-worst in French, surpassing only FlauBERT.  
In both French and English, higher-performing models show consistently higher precision (slightly above 0.7) than recall (slightly above 0.5).  
This pattern indicates a conservative prediction strategy: the models are generally correct when predicting an affix, but they often fail to identify all relevant instances.  
The result reflects a bias toward minimizing false positives at the expense of increased false negatives.  
This behavior likely arises from the high lexical variability and morphological ambiguity in both languages, where affixes occur across diverse grammatical and semantic contexts.  
While the models capture strong signals for prototypical affix forms, their ability to generalize to rarer or noisier forms remains limited, constraining recall.  

\subsubsection{Compositional vector arithmetic}

This experiment addresses the following research question: \textit{Is it possible to approximate a word embedding (e.g., ``unhappiness'') by composing the embeddings of its morphological components (e.g., ``un'', ``happy'', ``ness'')?}  
A simple function is additive composition; for example: \textit{vec(unhappiness) = vec(un) + vec(happy) + vec(ness)?}.  
In this experiment, we test several compositional functions to investigate how our model represents such relationships compared with other models.  
We restrict the study to words with exactly one prefix, one root, and one suffix, since words can have more complex structures, and analyzing those would warrant a separate paper.

\paragraph{Task.} 
Given a word vector $\vec{w}$, we test whether there exists an arithmetic composition function $f$ that combines the vectors of the prefix $\vec{p}$, root $\vec{r}$, and suffix $\vec{s}$ into a reconstructed vector $\hat{\vec{w}} = f(\vec{p}, \vec{r}, \vec{s}) \approx \vec{w}$.  
We then measure the similarity between the original word vector $\vec{w}$ and the reconstructed vector $\hat{\vec{w}}$.  
The following composition functions are evaluated:
\begin{itemize}
	\item Additive composition: $\hat{\vec{w}} = \vec{p} + \vec{r} + \vec{s}$;
	\item Multiplicative composition: $\hat{\vec{w}} = \vec{p} \odot \vec{r} \odot \vec{s}$;
	\item Weighted additive composition: $\hat{\vec{w}} = \alpha \vec{p} + \beta \vec{r} + \gamma \vec{s}$, where $\alpha, \beta, \gamma$ are trainable scalar parameters;
	\item Weighted multiplicative composition: $\hat{\vec{w}} = \vec{p}^{\alpha} \odot \vec{r}^{\beta} \odot \vec{s}^{\gamma}$, where $\alpha, \beta, \gamma$ are trainable scalar parameters;
	\item Linear mapping on concatenation: $\hat{\vec{w}} = W [\vec{p}; \vec{r}; \vec{s}]$, where $W$ is a trainable matrix and $[\,;\,;\,]$ denotes concatenation;
	\item Linear mapping on addition: $\hat{\vec{w}} = W (\vec{p} + \vec{r} + \vec{s})$, where $W$ is a trainable matrix.
\end{itemize}

\paragraph{Datasets.} 
We used MorphoLex-En and MorphoLex-Fr, as described in the previous experiment.  
For this study, we retained only words with exactly one prefix, one root, and one suffix.  
The datasets were split into training and test sets (60\%/40\%) using stratified sampling over the affixes.  
The test set was used to evaluate all compositional functions, while the training set was used only for functions requiring parameter learning.  
For English, this yielded 2,169 training samples and 1,453 test samples.  
For French, the split produced 822 training samples and 545 test samples.  
Notably, the French dataset contains a suffix labeled ``\textit{[VB]}'' to indicate verbal inflection, which we retained as a source of noise.  

\paragraph{Models.} 
We evaluated our three character-based models along with DziriBERT and CANINE for all languages.  
BERT was included for English and FlauBERT for French.  
For each subword-based model, we extracted both the CLS embedding (``\_cls'') and the centroid of token embeddings (``\_tok'').  

\paragraph{Metrics.} 
Evaluation employed two previously described metrics:  
(1) average cosine similarity (ACS) between composed and original word embeddings, and  
(2) adjusted average Euclidean distance (AED).  
We also used visualizations to inspect the learned mapping weights.

\paragraph{Results and discussion.}
Table~\ref{tab:test-comp} reports the average cosine similarity (ACS) between composed and original word embeddings across six composition strategies: additive (\textsc{Add}), multiplicative (\textsc{Mul}), weighted additive (\textsc{WAdd}), weighted multiplicative (\textsc{WMul}), linear mapping on concatenation (\textsc{MpCnc}), and linear mapping on addition (\textsc{MpAdd}). 
For each model, only the best-performing variant is reported for brevity. 
In English, BERT achieves the highest ACS in most additive-style settings (\textsc{Add}, \textsc{Mul}, \textsc{WAdd}), whereas CANINE-C performs best under weighted multiplicative composition. 
Our character-based chDzDT model, however, obtains the highest ACS in the more flexible linear mapping strategies (\textsc{MpCnc} and \textsc{MpAdd}), surpassing the strongest Transformer baseline by $0.0439$ in \textsc{MpCnc}. 
For French, CANINE-C again leads in additive-style compositions, while chDzDT outperforms all baselines in multiplicative and both linear mapping strategies, with a particularly large margin of $0.0615$ in \textsc{MpCnc}. 
These findings suggest that while subword- and token-based Transformers retain an advantage in simple compositional arithmetic, our fully character-based approach benefits more from learnable projection over concatenated morpheme representations, potentially offering a better alignment with morphological structure. 
Across both languages, CANINE-C also achieved the lowest corrected AED in most settings, indicating stronger preservation of absolute distances (full AED results omitted for brevity).

\begin{table}[!htp]
	\centering\small
	
	\caption{Average cosine similarity (ACS) between composed and original embeddings for six composition strategies in English (EN) and French (FR). 
		For each model, only the best-performing variant is reported. 
		Bold values mark the highest ACS within each column and language.}
	
	\begin{tabular}{p{1cm}lrrrrrr}
		\hline\hline
		\textbf{Lang.} & \textbf{Model} & \textbf{\textsc{Add}} & \textbf{\textsc{Mul}} & \textbf{\textsc{WAdd}} & \textbf{\textsc{WMul}} & \textbf{\textsc{MpCnc}} & \textbf{\textsc{MpAdd}} \\
		\hline
		\multirow{4}{*}{English}  
		&  \texttt{BERT\_cls} & \textbf{0.8789} & \textbf{0.6425} & \textbf{0.8797} & 0.0263 & 0.9285 & 0.9274 \\
%		&  \texttt{BERT\_tok} & 0.6334 & 0.4105 & 0.6468 & 0.0095 & 0.8053 & 0.8014 \\
%		&  \texttt{DziriBERT\_cls} & 0.6572 & 0.5002 & 0.6647 & 0.1240 & 0.8459 & 0.8404 \\
		&  \texttt{DziriBERT\_tok} & 0.7191 & 0.5500 & 0.7296 & 0.0796 & 0.8934 & 0.8895 \\
		&  \texttt{CANINE-C\_cls} & 0.7397 & 0.4623 & 0.7440 & \textbf{0.2642} & 0.9095 & 0.9045 \\
%		&  \texttt{chDzDT\_5x4x128} & 0.4855 & 0.3533 & 0.5704 & 0.1403 & 0.9346 & 0.9148 \\
%		&  \texttt{chDzDT\_4x4x64} & 0.4398 & 0.4677 & 0.5254 & 0.1145 & 0.9404 & 0.9178 \\
		&  \texttt{chDzDT\_4x4x32} & 0.6011 & 0.5192 & 0.6880 & 0.2314 & \textbf{0.9724} & \textbf{0.9602} \\
		\hline
		\multirow{4}{*}{French}  
%		&  \texttt{FlauBERT\_cls} & 0.5524 & 0.2749 & 0.5618 & 0.1289 & 0.6268 & 0.6297 \\
		&  \texttt{FlauBERT\_tok} & 0.5433 & 0.2653 & 0.5612 & 0.1194 & 0.6306 & 0.6347 \\
%		&  \texttt{DziriBERT\_cls} & 0.6858 & 0.5159 & 0.6908 & 0.1186 & 0.8319 & 0.8307 \\
		&  \texttt{DziriBERT\_tok} & 0.7218 & 0.5375 & 0.7347 & 0.0781 & 0.8526 & 0.8506 \\
		&  \texttt{CANINE-C\_cls} & \textbf{0.7246} & 0.4753 & \textbf{0.7424} & \textbf{0.2871} & 0.8935 & 0.8908 \\
%		&  \texttt{chDzDT\_5x4x128} & 0.5243 & 0.3196 & 0.6039 & 0.1470 & 0.9191 & 0.9099 \\
%		&  \texttt{chDzDT\_4x4x64} & 0.5164 & 0.4586 & 0.5980 & 0.1487 & 0.9336 & 0.9203 \\
		&  \texttt{chDzDT\_4x4x32} & 0.6479 & \textbf{0.5713} & 0.7029 & 0.2328 & \textbf{0.9550} & \textbf{0.9451} \\
		\hline\hline
		
	\end{tabular}

	\label{tab:test-comp}
\end{table}

\begin{figure}[!htp]
	\centering
	
	\begin{subfigure}{0.49\textwidth}
		\includegraphics[width=\textwidth]{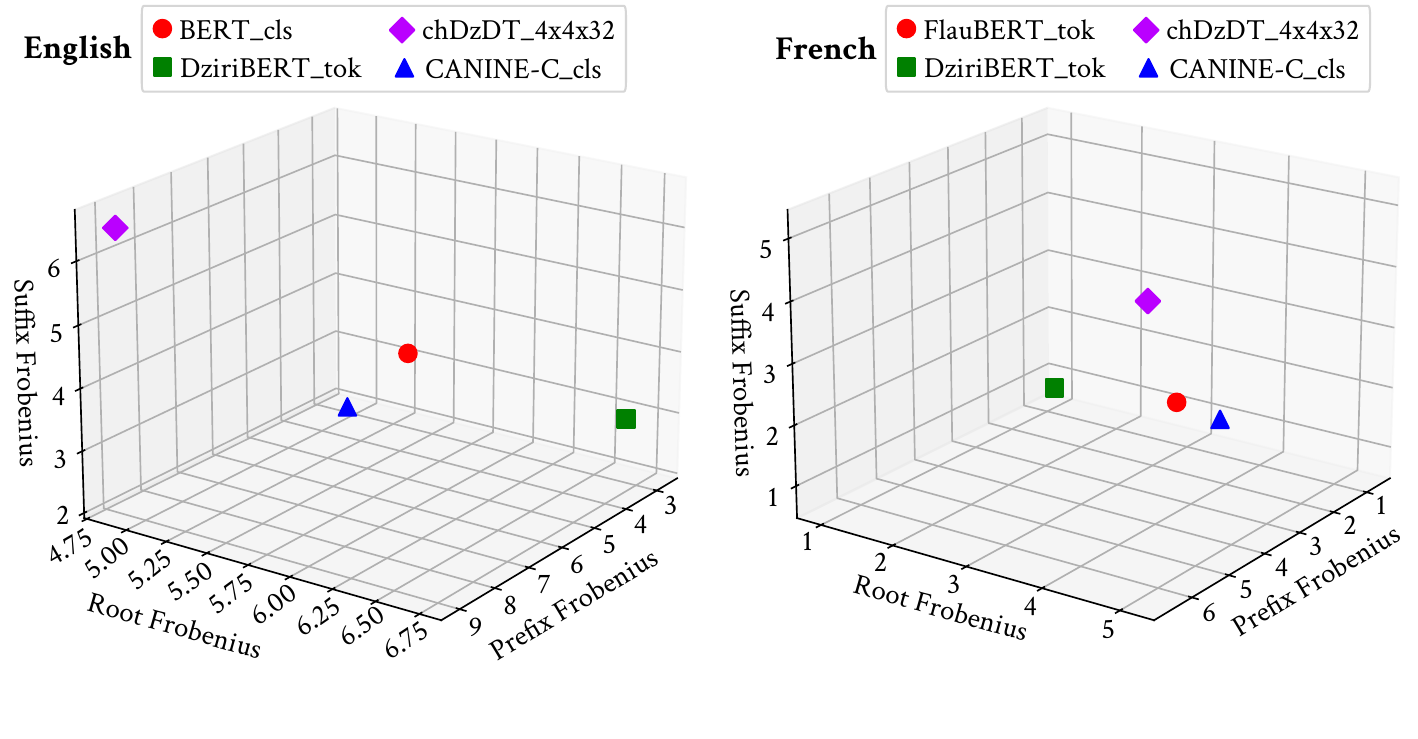}
		\subcaption{\textsc{MpCnc}}
	\end{subfigure}
	\begin{subfigure}{0.49\textwidth}
		\includegraphics[width=\textwidth]{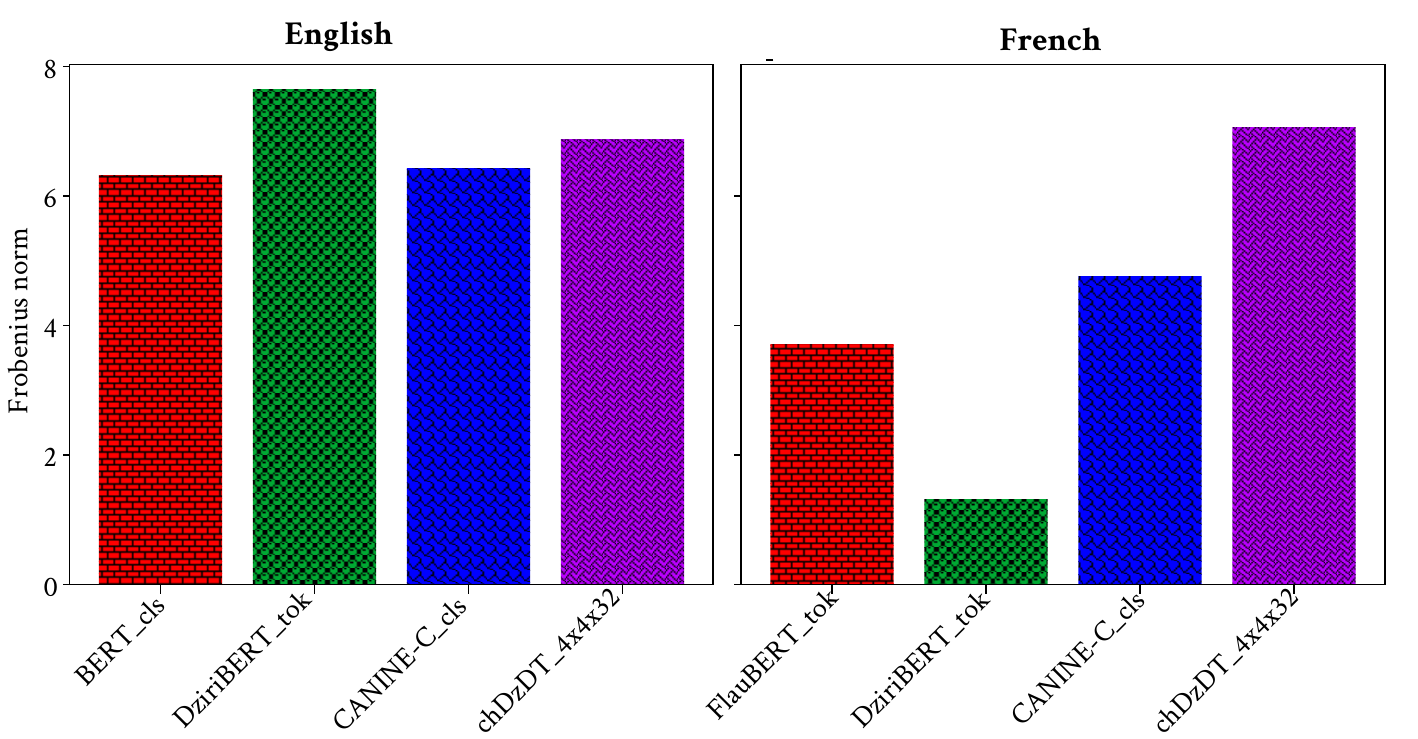}
		\subcaption{\textsc{MpAdd}}
	\end{subfigure}
	
	\caption{Frobenius norms of projection matrices for English and French models.  
		(a) Concatenation-based mapping (\textsc{MpCnc}) reports separate norms for prefix, root, and suffix components.  
		(b) Additive mapping (\textsc{MpAdd}) applies a single projection matrix to the sum of all morpheme components, yielding one norm value per model.}  
	
	\label{fig:test-frobenius_proj}
\end{figure}

Figure~\ref{fig:test-frobenius_proj} compares the Frobenius norms of projection matrices for \textsc{MpCnc} (prefix, root, suffix; 3D plot) and \textsc{MpAdd} (prefix + root + suffix; bar plot) in English and French.  
These norms quantify the extent to which each mapping reshapes morpheme embeddings.  

In English, \textsc{MpCnc} values are highest for chDzDT, particularly in the prefix and suffix components, indicating stronger, more localized transformations.  
When considered alongside the table of ACS results, models that draw heavily on both prefix and suffix information (e.g., chDzDT) perform best, followed by those emphasizing the root and one affix (CANINE and BERT).  
By contrast, models such as DziriBERT, which distribute emphasis more evenly and prioritize the root, tend to perform less well.  
This pattern suggests that exploiting both prefix and suffix information in conjunction with the root is especially beneficial in English.  

In French, chDzDT again yields the largest \textsc{MpCnc} norms, though they are more evenly distributed across morphemes.  
Token-based models such as BERT and DziriBERT exhibit lower norms, especially for suffixes, suggesting weaker transformations.  
However, no consistent relationship emerges between individual component norms and ACS performance, implying that the correspondence may be arbitrary.  
One plausible explanation is the inclusion of the ``\textit{[VRB]}'' label in the French dataset: not a true suffix but an indicator of verbal inflection, which we retained intentionally and which may introduce noise.  

For \textsc{MpAdd}, only one dimension is plotted, since prefix, root, and suffix are first combined by addition before projection.  
In English, all models show similarly high Frobenius norms, suggesting that additive composition produces a uniform degree of transformation regardless of model type.  
In French, the values vary more, though again without a clear connection to ACS performance.  
Overall, chDzDT with concatenation-based mapping appears to rely on stronger, morpheme-specific transformations, whereas syntax-oriented models favor more uniform adjustments across components.  

\begin{figure}[!htp]
	\centering
	
	\begin{subfigure}{0.49\textwidth}
		\includegraphics[width=\textwidth]{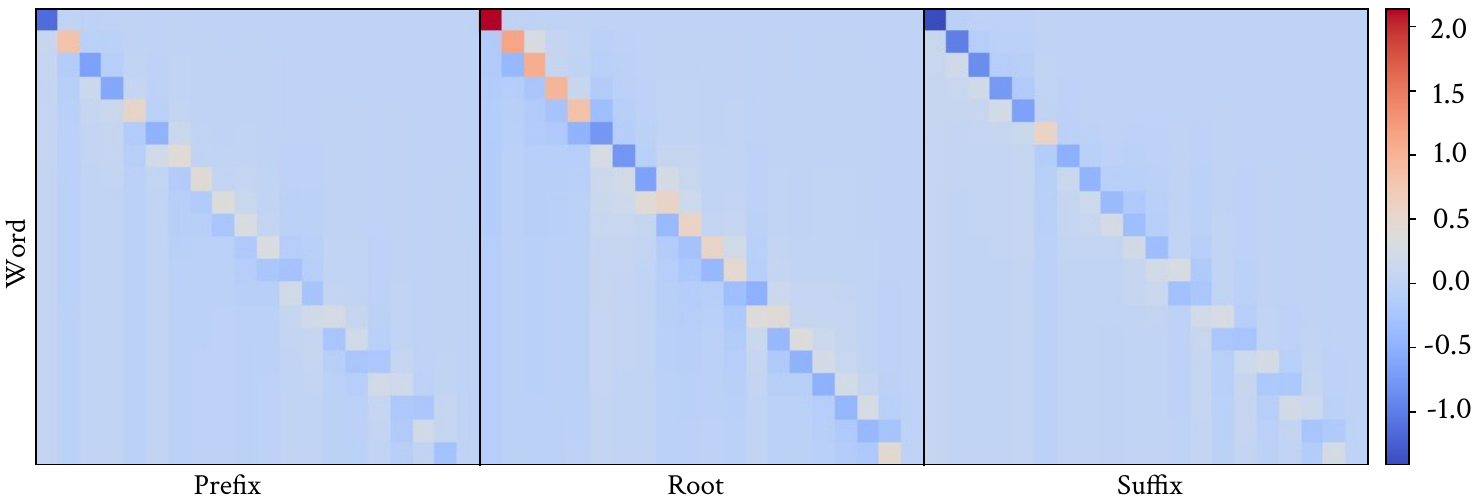}
		\subcaption{BERT\_cls mapping reduced from 768 to 20 dimensions.}
	\end{subfigure}
	\begin{subfigure}{0.49\textwidth}
		\includegraphics[width=\textwidth]{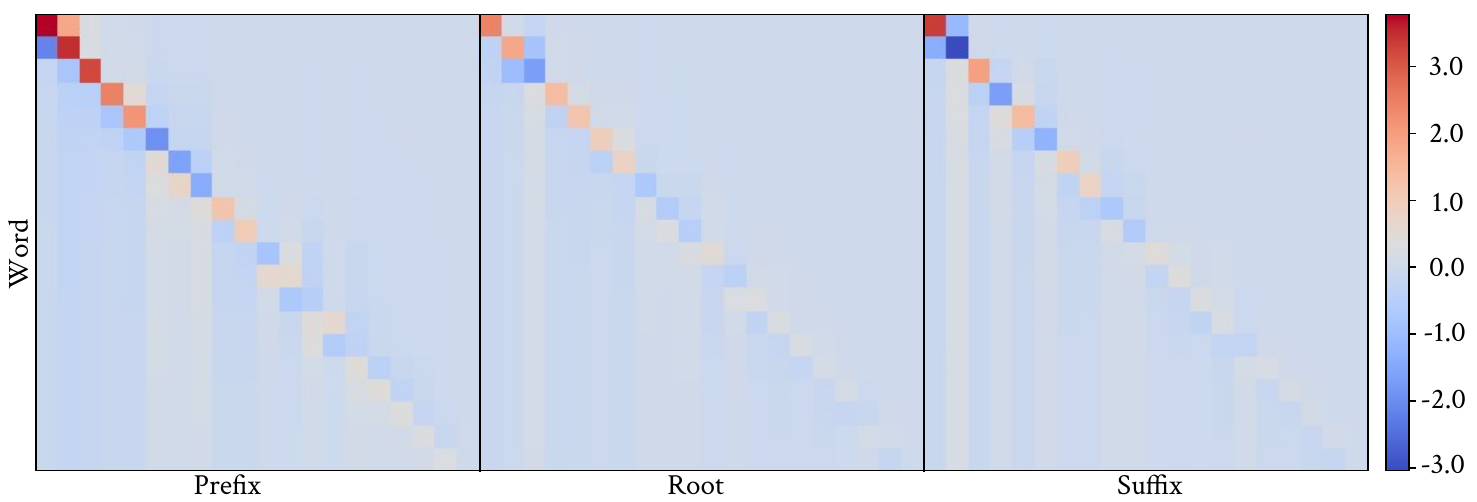}
		\subcaption{chDzDT mapping reduced from 32 to 20 dimensions.}
	\end{subfigure}
	
	\caption{PCA-compressed heatmaps of the learned mapping $W = [W_p ; W_r ; W_s]$ for English, shown for the two best-performing models. 
		Each block corresponds to prefix, root, and suffix mappings, respectively.}
	
	\label{fig:test-map_proj}
\end{figure}

Figure~\ref{fig:test-map_proj} presents a PCA-reduced visualization of the learned linear mapping $W = [W_p ; W_r ; W_s]$ for chDzDT and BERT on English.
Each block $W_p,\ W_r,\ W_s$ is reduced to a $20 \times 20$ summary via two-sided PCA (applied first to rows, then to columns) and plotted side by side, yielding a compact representation of the mapping structure.
The heatmaps reveal coarse structural patterns, such as energy concentration and diagonal alignment.
Because PCA compresses the original dimensionality, each cell represents a linear combination of original coordinates rather than a one-to-one mapping, and patterns should therefore be interpreted qualitatively.

A visual comparison of the PCA-compressed mappings highlights distinct differences between the two models.
For BERT, each block retains a diagonal structure, though with relatively mild magnitudes, typically ranging from $[-1, 2]$.
The prefix block ($W_p$) shows moderate activation, while the suffix block ($W_s$) is even less pronounced, suggesting a smoother, more uniform transformation.
By contrast, chDzDT exhibits much stronger diagonals across all three blocks, with values reaching up to $\pm 3$ and concentrated energy in the upper-left regions of the matrices.
This pattern indicates greater reliance on the leading reduced dimensions and aligns with the larger Frobenius norms observed earlier, pointing to more specialized and morpheme-sensitive mappings.

\subsubsection{Semantic similarity alignment}

This experiment addresses the following research question: \textit{To what extent do word embeddings reflect human-perceived semantic similarity?}
That is, how closely do vector representations of words align with human intuitions about meaning?
A model that effectively captures semantic similarity should assign higher similarity scores to semantically related word pairs, as judged by humans.

\paragraph{Task.}
The goal of this task is to evaluate semantic similarity between pairs of words.
Embeddings are generated for each word, and cosine similarity is computed between them.
These scores are then compared to human-annotated similarity ratings using correlation measures.
Because chDzDT is character-based, it primarily encodes morphology rather than semantics, and thus is not expected to perform strongly on meaning-focused tasks.
Nevertheless, we investigate whether character-level patterns in Arabic, English, and French provide any capacity to capture semantic information.

\paragraph{Datasets.}
We used the Multilingual WordSimilarity-353 dataset \cite{2016-freitas-al}, an extension of the original WordSimilarity-353 benchmark \cite{2002-finkelstein-al}.
The dataset contains 353 word pairs, each annotated with an average human-assigned similarity score on a scale from 1 to 10.
Because Algerian dialect incorporates many terms from Modern Standard Arabic, French, and increasingly English, we evaluated the models on all three languages to assess their capacity for cross-lingual semantic representation.

\paragraph{Models.}  
We evaluated the character-based model \texttt{chdzdt\_5x4x128}, selected for its larger embedding dimensionality.
The other two chDzDT variants yielded comparable results, so only this representative model is reported.
AraBERT was used for Arabic, BERT for English, and FlauBERT for French.

\paragraph{Metrics.}  
We computed the correlation between model-based similarity scores $\hat{Y}$ and human-annotated scores $Y$ using Pearson, Spearman, and Kendall correlation coefficients.
Kendall's $\tau$ is particularly appropriate here due to its robustness to nonlinearity and sensitivity to the ordinal structure of the data, which aligns with subjective human ratings.
Given $|Y| = |\hat{Y}| = n$, Kendall's $\tau \in [-1, 1]$ is calculated using Equation~\ref{eq:kendall}.
\begin{equation}
	\tau = \frac{2}{n (n-1)} \sum_{i<j} sign(\hat{Y}_i - \hat{Y}_j) sign(Y_i - Y_j)
	\label{eq:kendall}
\end{equation}

\paragraph{Results and discussion.}  
%As shown in Table~\ref{tab:test-sim-corr}, the cosine similarity scores $\hat{Y}$ produced by chDzDT show virtually no correlation with human similarity ratings $Y$ across all three languages.  
Cosine similarity scores $\hat{Y}$ from chDzDT show virtually no correlation with human similarity ratings $Y$ across all three languages.
This outcome is expected, as character-based models lack access to semantic cues in alphabetic languages such as Arabic, English, and French.
In contrast, logographic systems such as Chinese or Japanese inherently encode meaning at the character level.
Contextual models are trained on full sentences and are expected to capture word semantics through distributional information.
However, they also exhibit low correlation when evaluated on isolated words.
This suggests that such models rely heavily on context and may not encode strong stand-alone semantic representations for individual words.
Only the English BERT model with token centroid aggregation (\texttt{BERT\_tok}) shows a modest correlation (0.2226), highlighting its relative strength in encoding some semantic structure even outside of context.

%\begin{table}[!htp]
%	\centering
%	\begin{tabular}{lrrrrrrrr}
%		\hline\hline
%		\multirow{2}{*}{\textbf{Model}} & \multicolumn{2}{c}{\textbf{Arabic}} && \multicolumn{2}{c}{\textbf{English}} && \multicolumn{2}{c}{\textbf{French}} \\
%		\cline{2-3}\cline{5-6}\cline{8-9}
%		& \textbf{CLS} & \textbf{centroid} && \textbf{CLS} & \textbf{centroid} && \textbf{CLS} & \textbf{centroid}\\
%		\hline
%		chDzDT\_5x4x128 & 0.0439 & / && 0.0271 & / && 0.0393 & / \\
%		DziriBERT       & 0.0009 & -0.0066 && 0.0694 & 0.0273 && 0.0270 & 0.0112 \\
%		AraBERT         & -0.0136 & -0.0185 && / & / && / & /\\
%		BERT            & / & / && 0.0898 & 0.2226 && / & / \\
%		FlauBERT        & / & / && / & / && 0.1064 & 0.0929 \\
%		\hline\hline
%	\end{tabular}
%	\caption{Kendall correlation between model-based similarity scores and human ratings from WordSimilarity-353.}
%	\label{tab:test-sim-corr}
%\end{table}

\subsection{Impact of embeddings on downstream tasks}

We examine the effectiveness of unsupervised embeddings when applied to downstream tasks without end-to-end fine-tuning.
Specifically, we evaluate three subtasks using shallow decoders trained on frozen embeddings: morphological tagging, part-of-speech (PoS) tagging, and sentiment analysis.
These tasks span different levels of linguistic processing: morphological tagging at the morphological level, PoS tagging at the morpho-syntactic level, and sentiment analysis at the semantic--pragmatic level.

\subsubsection{Morphological tagging}

This experiment addresses the question: \textit{How effectively do morphology-based word embeddings capture the grammatical attributes of words?}
Given an inflected verb, we test whether decoders can recover features such as tense, person, number, gender, or case.
For example, in Arabic, the word ``\RL{تذهبون}'' (/\textit{tadhhab\={u}n}/, ``\textit{you go}'') derives from the verb ``\RL{ذهب}'' (/\textit{dhahaba}/, ``\textit{to go}''), conjugated in the present tense, second person, plural, masculine.
In French, the equivalent form comes from ``\textit{aller}'' as ``\textit{allez}'', which applies to both masculine and feminine.
Unlike the earlier probing task, which tested linear relations between embeddings and affixes, this experiment examines more complex relations to evaluate whether embeddings encode morphological information, rather than whether they are easily interpretable.

\paragraph{Task.} 
Grammatical features include tense, case, voice, aspect, person, number, gender, mood, animacy, and definiteness.
Each language encodes a subset of these features through inflection: verbs via conjugation and nouns, adjectives, or adverbs via declension.
Here, we focus on verb conjugation, which exhibits the richest variation.
The task consists of training a model that takes verb embeddings as input and predicts the corresponding feature values.

\paragraph{Datasets.} 
We used UniMorph\footnote{\url{https://unimorph.github.io/}} resources for Arabic, English, and French to extract verb conjugations and their associated grammatical features.
When a feature was not applicable, we assigned the label ``\textit{NA}'' (e.g., tense for non-finite verbs or gender for first person).
Each dataset was split into training (60\%) and test (40\%) sets.
Table~\ref{tab:morph-tag-stat} summarizes dataset sizes and their corresponding features.

\begin{table}[!htp]
	\centering\small
	
	\caption{Statistics of morphological tagging datasets for verb conjugation. 
		For each language, the table reports training and test sizes along with the set of output features.\newline
		\textit{Abbreviations:} Num = Number, Mood = Mood, Asp = Aspect, Pers = Person, Gen = Gender, ActV = Active Voice,
		Tns = Tense, NonFin = Nonfinite, Sg = Singular, 3rdP = Third Person, ImpSubj = Imperative--Subjunctive, Part = Participle.}
	
	\begin{tabular}{lrrl}
		\hline\hline
		\textbf{Lang.}  & \textbf{Train} & \textbf{Test} & \textbf{Features [Number of tags]} \\ 
		\hline
		Arabic & 81,139  & 54,093  & Num [3], Mood [5], Asp [3], Pers [3], Gen [3], ActV [2] \\
		English & 76,508  & 51,006  & Tns [3], NonFin [2], Sg [2], 3rdP [2], ImpSubj [2], Part [2] \\
		French & 207,082 & 138,055 & Pers [4], Asp [3], Mood [5], NonFin [2], Tns [4], Num [3]\\
		\hline\hline
	\end{tabular}
	
	\label{tab:morph-tag-stat}
\end{table}

\paragraph{Models.} 
We evaluated our three character-based encoders alongside subword-based PLMs used as frozen feature extractors: DziriBERT and CANINE-C (all languages), AraBERT (Arabic), BERT (English), and FlauBERT (French).
For each subword-based model, we derived two representations: the CLS embedding (``\_cls'') and the centroid of the token embeddings (``\_tok'').
Each word embedding was fed into a shared multi-task layer followed by parallel classification heads, one per grammatical feature.
Depending on the feature's cardinality, heads used either sigmoid (binary) or softmax (multiclass) activations.
The total loss was computed as the sum of binary and categorical cross-entropies, and models were optimized with Adam for up to 100 epochs, with early stopping once the overall error dropped below 0.1.

\paragraph{Metrics.} 
Since most features are multiclass and binary features are equally important, we used accuracy as the evaluation metric.
Accuracy is defined as the proportion of correctly predicted labels over the total number of samples.

\paragraph{Results and discussion.}
Table~\ref{tab:test-morph-tag-ara} reports the accuracy scores for each grammatical feature in Arabic.
Our 128-dimensional model achieved the highest accuracy across all features, while the other two variants performed comparably.
CANINE-C, also a character-based model, yielded competitive results, reinforcing the view that character-level approaches are well suited to morphology-oriented tasks, at least in Arabic.

\begin{table}[!htp]
	\centering\small

	\caption{Accuracy of different models on the Arabic morphological tagging task. 
		For each model, only the best-performing variant is reported. Bold values indicate the highest accuracy for each grammatical feature.\newline
		\textit{Abbreviations:} Num = Number, Mood = Mood, Asp = Aspect, Pers = Person, Gen = Gender, ActV = Active voice.}
		
	\begin{tabular}{lrrrrrrr}
		\hline\hline
		\textbf{Model} & \textbf{Num} & \textbf{Mood} & \textbf{Asp} & \textbf{Pers} & \textbf{Gen} & \textbf{ActV} & \textbf{Overall} \\
		\hline 
		 \texttt{AraBERT\_cls} & 0.9797 & 0.7142 & 0.9725 & 0.7880 & 0.7910 & 0.8391 &  0.8474\\
%		 \texttt{AraBERT\_tok} & 0.9818 & 0.7031 & 0.9748 & 0.7745 & 0.7786 & 0.8323 & 0.841 \\
%		 \texttt{DziriBERT\_cls} & 0.9726 & 0.6970 & 0.9609 & 0.7621 & 0.7678 & 0.8366 & 0.833 \\
		 \texttt{DziriBERT\_tok} & 0.9719 & 0.7022 & 0.9642 & 0.7703 & 0.7676 & 0.8378 & 0.8357\\
		 \texttt{CANINE-C\_cls} & 0.9934 & 0.9595 & 0.9948 & 0.8174 & 0.8166 & 0.9938 & 0.9292\\
		 \texttt{chDzDT\_5x4x128} & \textbf{0.9956} & \textbf{0.9674} & \textbf{0.9949} & \textbf{0.8602} & \textbf{0.8593} & \textbf{0.9946} & \textbf{0.9453} \\
%		 \texttt{chDzDT\_4x4x64} & 0.9873 & 0.9592 & 0.9937 & 0.8605 & 0.8576 & 0.9916 & 0.942 \\
%		 \texttt{chDzDT\_4x4x32} & 0.9071 & 0.8854 & 0.9704 & 0.8283 & 0.8036 & 0.9337 & 0.888  \\
		\hline\hline
		
	\end{tabular}
	
	\label{tab:test-morph-tag-ara}
\end{table}

For English, all model variants achieved broadly similar accuracies across grammatical features, with a slight advantage for chDzDT.
Table~\ref{tab:test-morph-tag-eng} reports the results.
Overall, the models perform nearly identically, reflecting the relative simplicity of English verb conjugation.
The features ``\textit{Sg}'' and ``\textit{3rdP}'' yield almost identical accuracies, as expected, since they co-occur in the present tense (where the ``s'' suffix is added).
From a technical perspective, they can be considered the same feature because their dataset distributions are identical, although semantically they remain distinct.

\begin{table}[!htp]
	\centering\small
	
	\caption{Accuracy of different models on the English morphological tagging task. 
		For each model, the best-performing variant is reported. 
		Bold values indicate the highest accuracy for each grammatical feature.\newline
		\textit{Abbreviations:} Tns = Tense, NonFin = Nonfinite, Sg = Singular, 3rdP = Third person, ImpSubj = Imperative--subjunctive, Part = Participle.}
	
	\begin{tabular}{lrrrrrrr}
		\hline\hline
		\textbf{Model} & \textbf{Tns} & \textbf{NonFin} & \textbf{Sg} & \textbf{3rdP} & \textbf{ImpSubj} & \textbf{Part} & \textbf{Overall} \\
		\hline 
		 \texttt{BERT\_cls}       & 0.9786  & 0.9789  & 0.9960  & 0.9960  & 0.9790 & 0.7419 & 0.9451 \\
%		 \texttt{BERT\_tok}       & 0.9771  & 0.9788  & 0.9948  & 0.9948  & 0.9787 & 0.7126 & \\
%		 \texttt{DziriBERT\_cls}  & 0.9739  & 0.9757  & 0.9931  & 0.9931  & 0.9757 & 0.7421 & \\
		 \texttt{DziriBERT\_tok}  & 0.9750  & 0.9766  & 0.9939  & 0.9939  & 0.9767 & 0.7485 & 0.9441 \\
		 \texttt{CANINE-C\_cls}   & 0.9766  & 0.9760  & 0.9963  & 0.9962  & 0.9760 & 0.7291 & 0.9417 \\
		 \texttt{chDzDT\_5x4x128} & \textbf{0.9801}  & \textbf{0.9814}  & \textbf{0.9974}  & \textbf{0.9975}  & \textbf{0.9814} & \textbf{0.7894} & \textbf{0.9545}\\
%		 \texttt{chDzDT\_4x4x64}  & 0.9699  & 0.9715  & 0.9908  & 0.9907  & 0.9714 & 0.7974 & \\
%		 \texttt{chDzDT\_4x4x32}  & 0.8821  & 0.9035  & 0.9566  & 0.9567  & 0.9034 & 0.7927 & \\
		\hline\hline
		
	\end{tabular}
		
	\label{tab:test-morph-tag-eng}
\end{table}

Table~\ref{tab:test-morph-tag-fra} reports accuracy scores for each grammatical feature in French.
The 128-dimensional model consistently achieved the highest accuracy across all features, outperforming the other models.
This suggests that higher-dimensional embeddings are better able to encode feature-related information.
Nonetheless, the other models are not far behind, and the overall results indicate that French affixes are strongly predictive of grammatical features.
The feature ``\textit{NonFin}'' achieved equally high accuracy across all models, showing that it represents a well-separated and easily identifiable category.

\begin{table}[!htp]
	\centering\small
	
	\caption{Accuracy of different models on the French morphological tagging task. 
		For each model, the best-performing variant is reported. Bold values indicate the highest accuracy for each grammatical feature.\newline
		\textit{Abbreviations:} Pers = Person, Asp = Aspect, Mood = Mood, NonFin = Nonfinite, Tns = Tense, Num = Number.}
		
	\begin{tabular}{lrrrrrrr}
		\hline\hline 
		\textbf{Model} & \textbf{Pers} & \textbf{Asp} & \textbf{Mood} & \textbf{NonFin} & \textbf{Tns} & \textbf{Num} & \textbf{Overall} \\
		\hline 
		 \texttt{FlauBERT\_cls}   & 0.7964 & 0.8626 & 0.7060 & 0.9907 & 0.7400 & 0.9331 & 0.8381 \\
%		 \texttt{FlauBERT\_tok}   & 0.7997 & 0.8598 & 0.6992 & 0.9917 & 0.7414 & 0.9354 & 0.8379 \\
%		 \texttt{DziriBERT\_cls}  & 0.8727 & 0.9236 & 0.7571 & 0.9983 & 0.8326 & 0.9919 & 0.8960 \\
		 \texttt{DziriBERT\_tok}  & 0.8733 & 0.9268 & 0.7576 & 0.9983 & 0.8372 & 0.9929 & 0.8977 \\
		 \texttt{CANINE-C\_cls}   & 0.8680 & 0.9167 & 0.7440 & 0.9989 & 0.8188 & 0.9922 & 0.8898 \\
		 \texttt{chDzDT\_5x4x128} & \textbf{0.8847} & \textbf{0.9425} & \textbf{0.7872} & \textbf{0.9994} & \textbf{0.8593} & \textbf{0.9970} & \textbf{0.9117} \\
%		 \texttt{chDzDT\_4x4x64}  & 0.8804 & 0.9033 & 0.7624 & 0.9980 & 0.8050 & 0.9895 & 0.8898 \\
%		 \texttt{chDzDT\_4x4x32}  & 0.7679 & 0.8350 & 0.6734 & 0.9862 & 0.6793 & 0.9023 & 0.8074 \\
		\hline\hline
		
	\end{tabular}
	
	\label{tab:test-morph-tag-fra}
\end{table}

\subsubsection{Part-of-speech tagging}

In this experiment, we address the following research question:
\textit{To what extent can morphology-based word embeddings, in a lightweight configuration, support a morpho-syntactic task such as PoS tagging?}
The assumption is that purely word-level models lack explicit syntactic information unless it is introduced during pre-training, which is not the case for our model.
Our goal, therefore, is to assess the extent to which morphological information alone contributes to performance in syntactically sensitive tasks.

\paragraph{Task.} 
The task is to assign grammatical categories to each word in a sentence.

\paragraph{Datasets.} 
We used data from Universal Dependencies\footnote{\url{https://universaldependencies.org/}} \cite{2021-de-marneffe-al} for PoS tagging.
For Arabic, we used the PADT treebank; for Arabizi, the Maghrebi Arabic–French–Arabizi treebank; for English, the GUM treebank; and for French, the GSD treebank.
The datasets are pre-divided into training and test sets.
Table~\ref{tab:pos-tag-stat} summarizes their statistics.

\begin{table}[!htp]
	\centering\small
	
	\caption{Statistics for PoS tagging datasets. 
		For each language, the table reports the number of sentences (\# Sent.) in the training and test sets, the average sentence length in words (AW), and the five most frequent classes with their frequencies.}
	
	\begin{tabular}{llrrl}
		\hline\hline
		\textbf{Lang.} & \textbf{Set}  & \textbf{\# Sent.} & \textbf{AW} & \textbf{Five most frequent tags and their frequencies} \\ 
		\hline
		\multirow{2}{*}{Arabic} 
		& Train & 6,075 & 42 & NOUN: 74,546; ADP: 33,617; \_: 30,485; ADJ: 23,498; PUNCT: 17,511\\
		& Test  &   680 & 47 & NOUN: 9,547; ADP: 4,528; \_: 3,865; ADJ: 2,937; VERB: 2,191 \\
		\hline
		\multirow{2}{*}{Arabizi} 
		& Train & 1,003 & 16 & NOUN: 2,808; VERB: 2,697; DET: 1,694; PRON: 1,526; PROPN: 1,431  \\
		& Test  &   145 & 16 & NOUN: 404; VERB: 400; DET: 222; PROPN: 211; PRON: 202\\
		\hline
		\multirow{2}{*}{English} 
		& Train & 10,224 & 18 & NOUN: 29,292; PUNCT: 24,563; VERB: 18,700; ADP: 16,673; PRON: 15,200\\
		& Test  &  1,464 & 20 & NOUN: 4,905; PUNCT: 3,550; VERB: 3,029; ADP: 2,867; DET: 2,521 \\
		\hline
		\multirow{2}{*}{French} 
		& Train & 14,450 & 25 & NOUN: 66,661; ADP: 56,356; DET: 54,112; PUNCT: 39,017; VERB: 28,189\\
		& Test  &    416 & 25 & NOUN: 1,870; ADP: 1,479; DET: 1,479; PUNCT: 1,186; VERB: 821\\
		\hline\hline
	\end{tabular}
	
	\label{tab:pos-tag-stat}
\end{table}

\paragraph{Models.}  
We evaluated our three character-based models alongside DziriBERT across all languages.
For comparison, AraBERT was used for Arabic, BERT for English, and FlauBERT for French.
 
Our word-level Transformer generated word embeddings, which were passed to a BiGRU encoder, yielding a 768-dimensional representation to ensure comparability with the subword-based encoders.
This design captures both past and future contextual information without feeding predicted tokens into subsequent time steps.
Each token representation was then passed through a shallow decoder consisting of a dense layer with 768 units, followed by a classification output layer.
The BiGRU encoder was trained jointly with the decoder, while padded tokens were ignored during training.

Subword-based models, by contrast, produce representations for both special tokens and subwords.
Special tokens, including padding, were ignored, and only the first subword of each word was considered for classification.
The same shallow decoder architecture was applied as in the word-level Transformer.

\begin{figure}[!htp]
	\centering
	
	\includegraphics[width=0.8\textwidth]{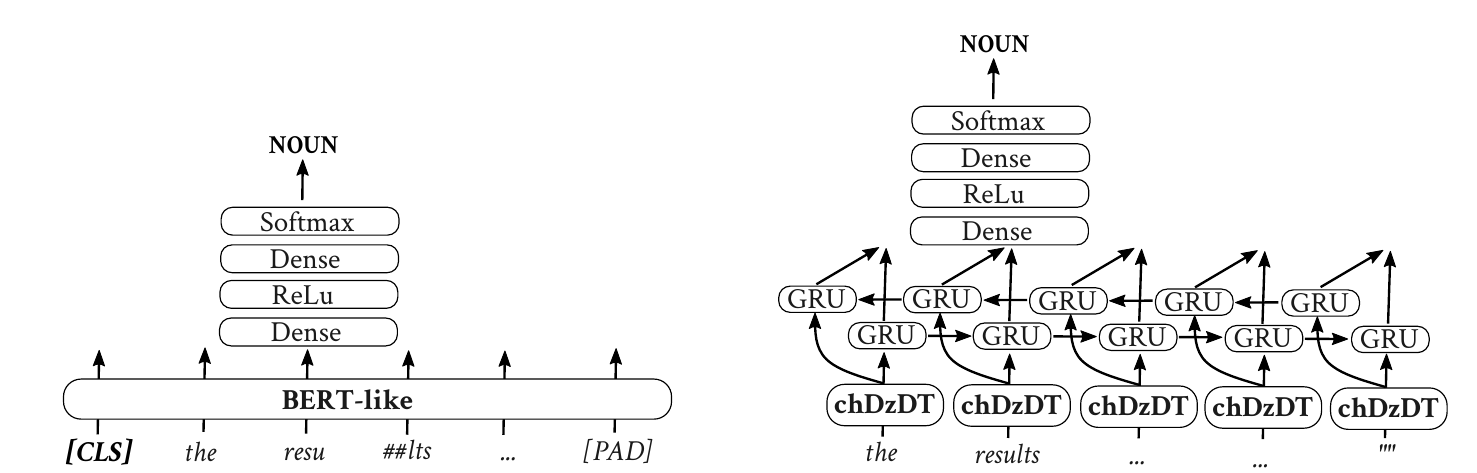} 
	
	\caption{Part-of-speech architectures: subword-based Transformers (left) and character-based models (right).}
	
	\label{fig:pos-arch}
\end{figure}

Figure~\ref{fig:pos-arch} illustrates the architectures: the subword-based Transformer (left) and the character-based model (right).
All sentences were truncated to 60 words to account for the limitations of RNNs in handling long-distance dependencies; the same restriction was applied to subword-based PLMs for fairness.
All models were trained with cross-entropy loss and the Adam optimizer, up to a maximum of 50 epochs.
Early stopping was applied once the error fell below 0.1.
The encoding models were used solely as a preprocessing stage and were not fine-tuned together with the decoder.

\paragraph{Metrics.} 
Since PoS tagging is a multiclass classification task, accuracy was used as the evaluation metric.
In this context, accuracy is equivalent to the micro-average of precision, recall, and F1 score.
It is defined as the proportion of correctly predicted labels over the total number of samples.

\paragraph{Results and discussion.}
Table~\ref{tab:test-pos} reports model accuracies across languages.
Our character-based models (chDzDT) consistently deliver competitive performance, though slightly below that of subword-based PLMs.
This outcome is expected: subword PLMs are pre-trained on large-scale sentence corpora and therefore encode morpho-syntactic and semantic information.
By contrast, our model is designed to capture morphological regularities from the character level upward, without direct exposure to syntactic or contextual training signals.

Despite this limitation, the performance gap is relatively small in Arabic, English, and French, indicating that morphological information alone provides a strong basis for morpho-syntactic tasks such as PoS tagging.
The largest gap occurs in Arabizi, where orthographic variability and limited training data reduce the reliability of morphology as a predictor.
Overall, the findings confirm that morphology-oriented embeddings are highly informative. 
They suggest that incorporating morpho-syntactic context during pre-training, while retaining character-level sensitivity, could further enhance performance.

\begin{table}[!htp]
	\centering\small

	\caption{Accuracy of the models on PoS tagging. 
		For *BERT, this refers to AraBERT for Arabic, BERT for English, and FlauBERT for French. 
		Bold values mark the highest accuracy in each column.}
	\begin{tabular}{lrrrr}
		\hline\hline
		\textbf{Model} & \textbf{Arabic} & \textbf{Arabizi} & \textbf{English} & \textbf{French}\\
		\hline
		 \texttt{*BERT}           & \textbf{0.9603} & / & \textbf{0.9542} & 0.9512\\
		 \texttt{DziriBERT}       & 0.9412 & \textbf{0.8050} & 0.9272 & \textbf{0.9566}\\
		
		 \texttt{chDzDT\_5x4x128} & 0.9279 & 0.5397 & 0.9169 & 0.9417 \\ 
		 \texttt{chDzDT\_4x4x64}  & 0.9132 & 0.5070 & 0.9008 & 0.9291 \\
		 \texttt{chDzDT\_4x4x32}  & 0.8815 & 0.4507 & 0.8664 & 0.9119 \\
		\hline\hline
		
	\end{tabular}	
	
	\label{tab:test-pos}
\end{table}

\subsubsection{Sentiment analysis}

In this experiment, we address the following research question:
\textit{To what extent can morphology-based word embeddings, in a minimal configuration, support a task that requires both semantic understanding and pragmatic reasoning, such as sentiment analysis?}
In principle, purely word-level models are not expected to outperform Transformer-based PLMs, which capture higher-order linguistic phenomena, including irony and sarcasm.
Our objective is to evaluate the extent to which morphological information contributes to sentiment classification. 
We also estimate the relative role of semantic and pragmatic information by comparing the performance of our character-based models against subword-based PLMs.

\paragraph{Task.} 
We selected Twitter sentiment analysis (polarity classification) because tweets are short, making them computationally tractable for Transformer-based models.
This task also benefits from the availability of numerous datasets, including those in the Algerian dialect.
The objective is straightforward: given a tweet, determine whether its polarity is positive, negative, or neutral.

\begin{figure}[!htp]
	\centering\small
	
	\begin{tabular}{c@{}cc@{}cc@{}cc@{}cc}
		\multicolumn{2}{c}{\textbf{Arabic (Ar)}}  & \multicolumn{2}{c}{\textbf{Algerian (Dz)}} & \multicolumn{2}{c}{\textbf{English (En)}} & \multicolumn{2}{c}{\textbf{French (Fr)}} & \\ 
		\textbf{Train} & \textbf{Test} & \textbf{Train} & \textbf{Test} & \textbf{Train} & \textbf{Test} & \textbf{Train} & \textbf{Test} & \\
		\includegraphics[width=0.1\textwidth]{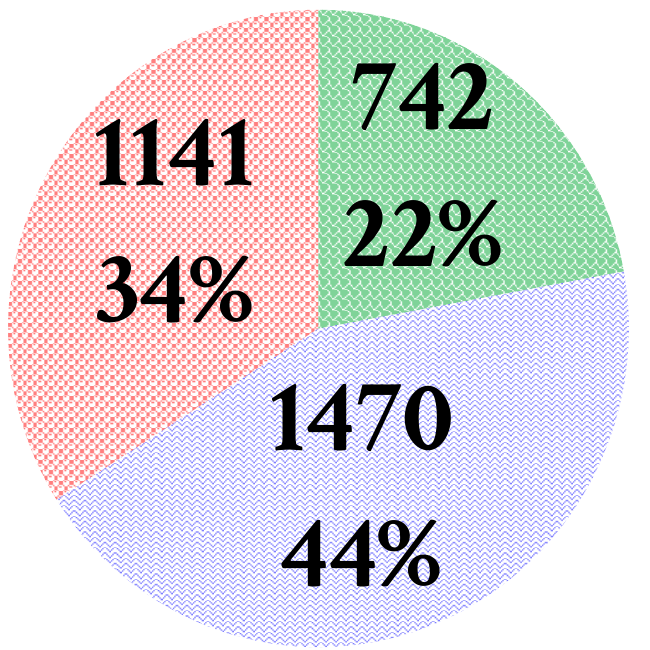} &
		\includegraphics[width=0.1\textwidth]{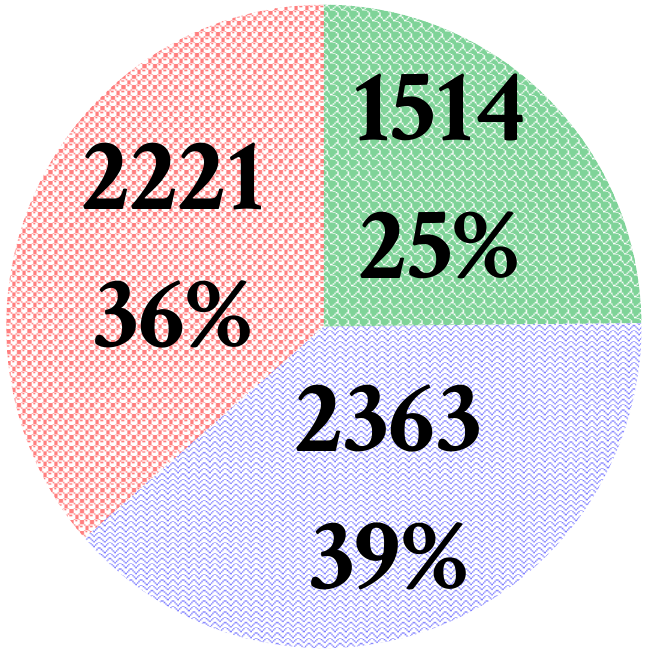} &
		
		\includegraphics[width=0.1\textwidth]{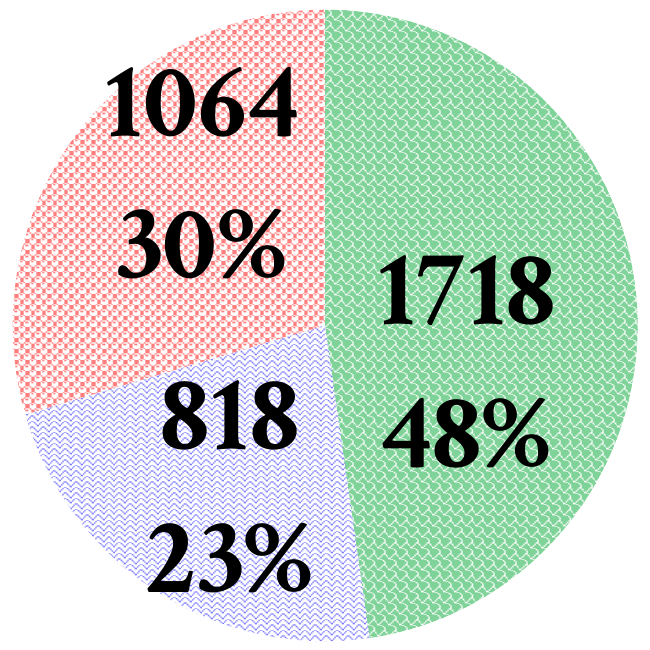} &
		\includegraphics[width=0.1\textwidth]{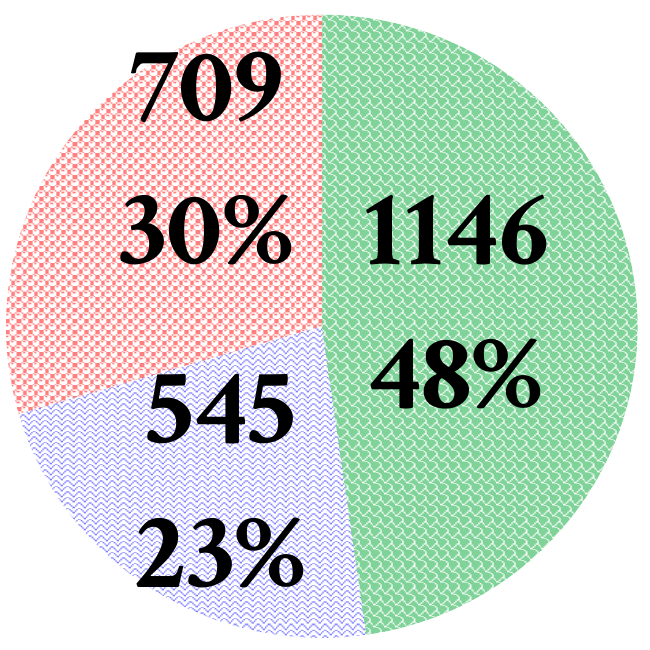} &
		
		\includegraphics[width=0.1\textwidth]{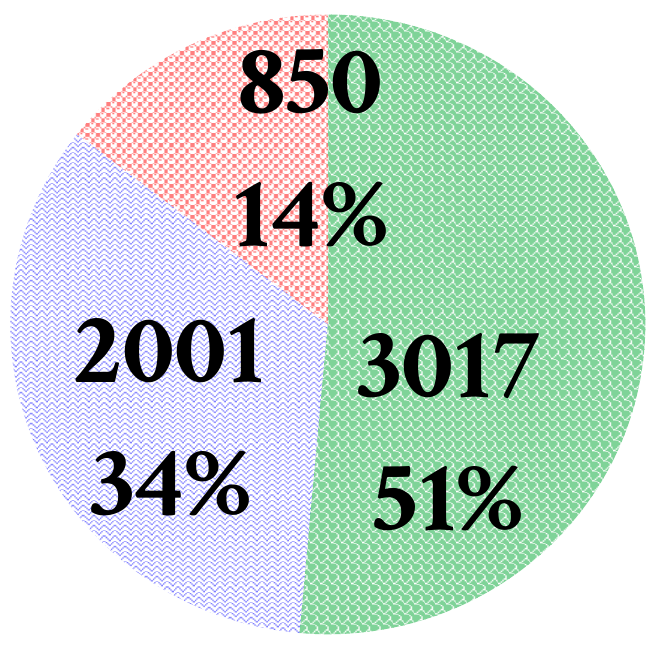} &
		\includegraphics[width=0.1\textwidth]{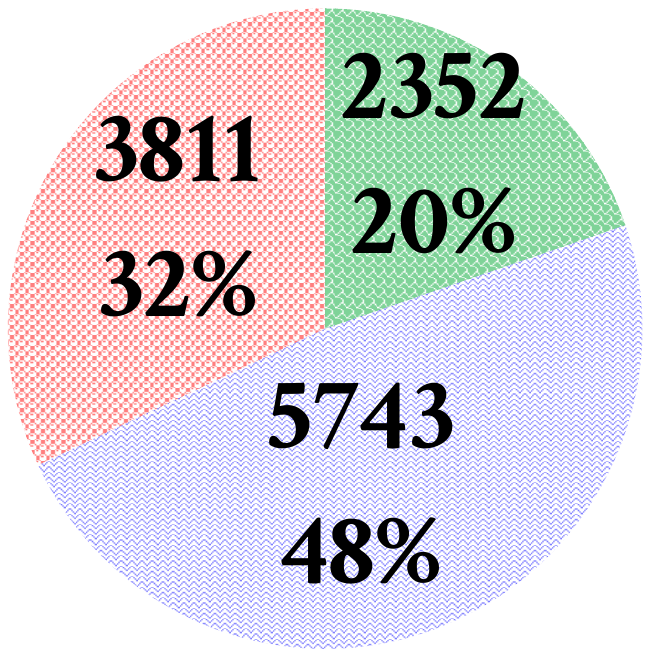} &
		
		\includegraphics[width=0.1\textwidth]{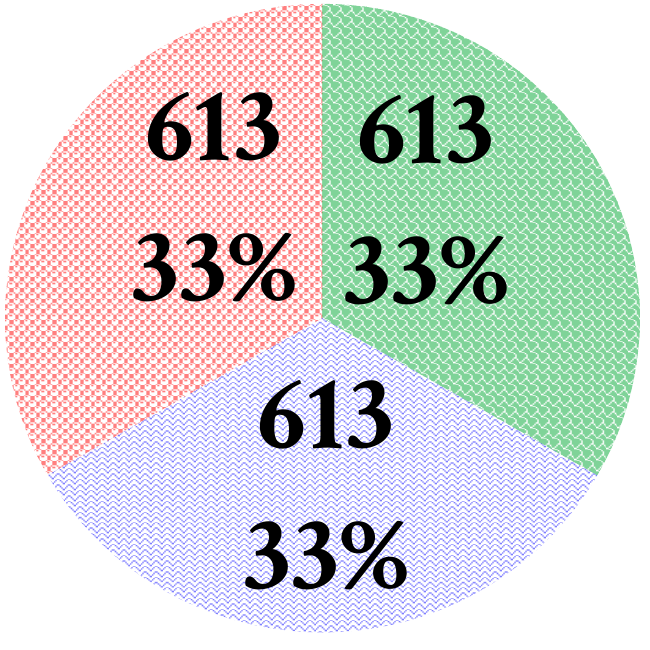} &
		\includegraphics[width=0.1\textwidth]{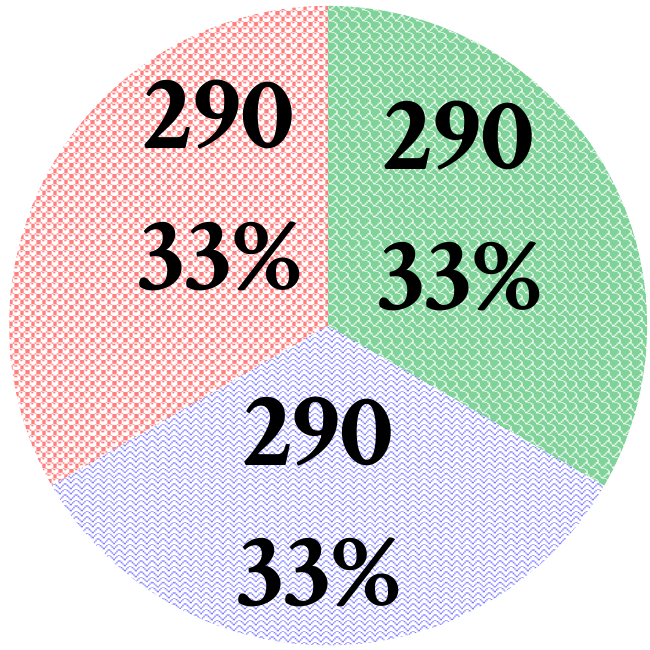} &
		
		\includegraphics[width=0.025\textwidth]{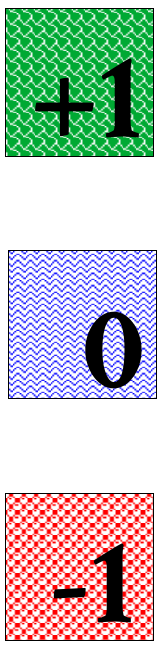}\\
	\end{tabular}
	
	\caption{Polarity distribution for each dataset split. 
		Positive (+1), neutral (0), and negative (-1) classes are reported with both percentages and absolute counts.}
	
	\label{fig:sa-data-stat}
\end{figure}

\paragraph{Datasets.} 
For Arabic and English, we employed the SemEval-2017 Task~4 \cite{2017-rosenthal-al} subtask ``A'' dataset for Twitter sentiment analysis.  
For French, we used the TweetNLP\footnote{\url{https://github.com/cardiffnlp/tweetnlp}} \cite{2022-camacho-collados-al} multilingual tweet sentiment dataset, which had already been filtered to remove URLs and normalized by replacing user mentions with ``\texttt{@user}''.
For the Algerian dialect, we used Twifil \cite{2020-moudjari-al}, which contains tweets from North African users, particularly Algerians.
It is written in a mixture of languages, not exclusively Arabizi.  
All datasets were pre-partitioned into training and test sets, except for Twifil, which we split into 60\% training and 40\% testing.  
Figure~\ref{fig:sa-data-stat} shows the polarity distribution for each dataset and split.  

\paragraph{Models.} 
We evaluated our three character-based models alongside DziriBERT across all languages.  
AraBERT was used for Arabic, BERT for English, and FlauBERT for French.  
For BERT, FlauBERT, and DziriBERT, the \texttt{[CLS]} token output was taken as the sentence representation, which was then passed through a dense layer with 768 units, followed by a 3-unit output layer for classification.  
Our word-level Transformer produced word embeddings that were fed into a BiGRU sentence encoder (maximum 30 words), yielding a 768-dimensional vector to ensure comparability with the other encoders.  
The BiGRU-based encoder was trained jointly with the decoder.  

\begin{figure}[!htp]
	\centering
	
	\includegraphics[width=0.8\textwidth]{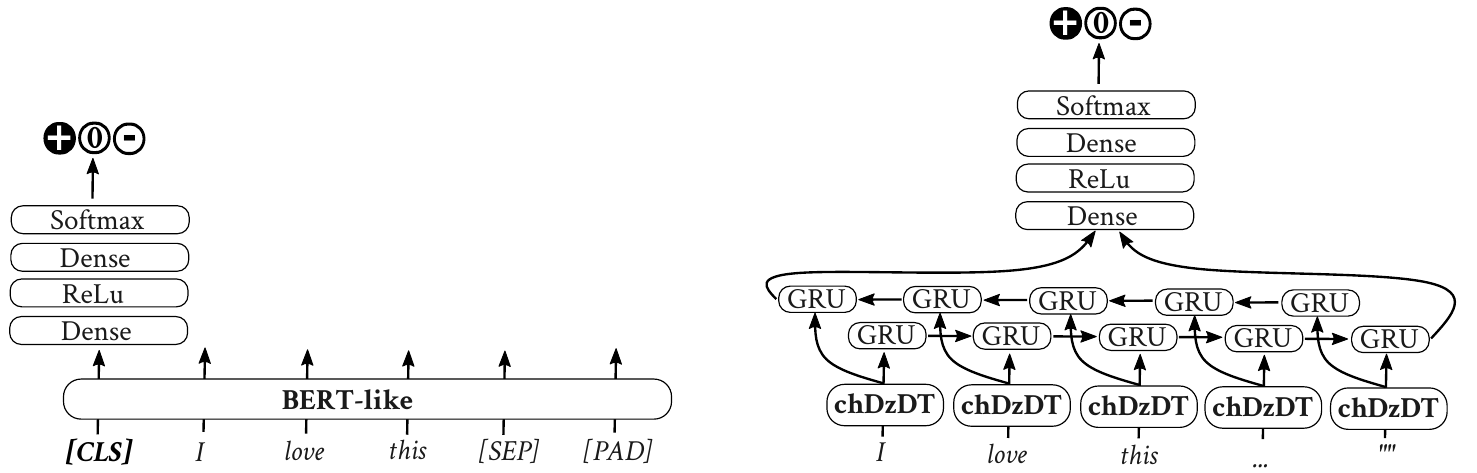} 
	
	\caption{Sentiment analysis architectures: subword-based Transformers (left) and character-based models (right).}

	\label{fig:sa-arch}
\end{figure}

Figure~\ref{fig:sa-arch} illustrates the architectures: the subword-based Transformer (left) and the character-based model (right).  
All models were trained with cross-entropy loss and the Adam optimizer.  
For fairness, training was limited to a maximum of 100 epochs, with early stopping triggered once the error dropped below 0.1.  
Importantly, the encoding models were used solely as a preprocessing stage and were not fine-tuned jointly with the decoder.  
%During the experiments, CANINE-C consistently failed to complete inference, even with small batch sizes, across multiple hardware configurations, and was therefore excluded from further evaluation.

\paragraph{Metrics.} 
Because polarity classification involves three classes, accuracy was used as the evaluation metric.  
In this setting, accuracy is equivalent to the micro-average of precision, recall, and F1 score.  
It is defined as the proportion of correctly predicted labels over the total number of samples.  

\paragraph{Results and discussion.}
Figure~\ref{tab:test-sentiment} presents the accuracy of the models for each language.  
As expected, our character-based models did not outperform language-specific PLMs or DziriBERT, except on the French dataset, where our model exceeded FlauBERT.  
The observed performance gap reflects the ability of PLMs to encode semantic and pragmatic knowledge during pre-training, whereas our embeddings primarily capture morphological and limited syntactic information.  
Nevertheless, the shortfall remains moderate ($\approx$33\%), indicating that morphology alone contributes meaningfully to sentiment classification.  
The estimated relative contribution of semantic and pragmatic information is approximately 33\% for Arabic, 14\% for the Algerian dialect, 33\% for English, and 23\% for French.  
The comparatively lower figure for Algerian may reflect the tendency of users to employ more direct, less figurative expressions, with reduced use of metaphor or irony.

Across languages, our models achieve broadly comparable performance, with consistently higher accuracy on the Algerian dialect.  
When normalized by Arabic performance, our model achieves a relative improvement of approximately 52\% on Algerian data, compared with around 20\% for DziriBERT, which was pre-trained exclusively on Algerian tweets.  
This difference is unlikely to result solely from random variation.  
We hypothesize that it reflects a model bias toward Algerian data, consistent with its design.  
However, label distribution across datasets may act as a confounding factor: positive tweets account for 48\% of the Algerian test set, compared with $\le$33\% in the other datasets, potentially inflating accuracy due to class imbalance.  
Finally, our model appears robust to preprocessing steps such as URL and user mention removal, maintaining performance comparable to other models, except DziriBERT.  
This robustness suggests an ability to disregard non-semantic artifacts without degrading classification accuracy.

\begin{table}[!htp]
	\centering\small
	
	\caption{Accuracy of the models on sentiment analysis for raw and filtered tweets. 
		For *BERT, this corresponds to AraBERT for Arabic, BERT for English, and FlauBERT for French. 
		The French dataset was already pre-processed. 
		Bold values indicate the highest accuracy in each column.}

	\begin{tabular}{llrrrrrrrrr}
		\hline\hline
		\multirow{2}{*}{\textbf{Model}} & \multicolumn{2}{c}{\textbf{Arabic}} && \multicolumn{2}{c}{\textbf{Algerian}} && \multicolumn{2}{c}{\textbf{English}} && \multirow{2}{*}{\textbf{French}}\\
		\cline{2-3}\cline{5-6}\cline{8-9}
		& \textbf{Raw} & \textbf{Filtered} && \textbf{Raw} & \textbf{Filtered} && \textbf{Raw} & \textbf{Filtered} && \\
		\hline
		 \texttt{*BERT} & \textbf{0.6322} & \textbf{0.6294} && / & /  && \textbf{0.5455} & \textbf{0.5351} && 0.3931\\
		 \texttt{DziriBERT} & 0.5786 & 0.5971 && \textbf{0.7042} & \textbf{0.7167} && 0.4308 & 0.4735 && \textbf{0.5690}\\
		
		 \texttt{chDzDT\_5x4x128} & 0.4182 & 0.4128 && 0.6071 & 0.6229 && 0.3503  & 0.3643 && 0.4425 \\
		 \texttt{chDzDT\_4x4x64} & 0.4282 & 0.4169  && 0.6012 & 0.6029  && 0.3466 & 0.3289 && 0.4287 \\
		 \texttt{chDzDT\_4x4x32} & 0.4154 & 0.4175  && 0.6067 & 0.5150  && 0.3536 & 0.3513 && 0.4322 \\
		\hline\hline
		
	\end{tabular}	
	
	\label{tab:test-sentiment}
\end{table}

\subsection{Effect of fine-tuning on downstream tasks}

We examine the impact of fine-tuning compared with using frozen embeddings on downstream tasks. 
Specifically, we evaluate three sub-tasks: 
morphological tagging, part-of-speech (PoS) tagging, and sentiment analysis.
We compare the performance of shallow classifiers trained on frozen embeddings with models fine-tuned end-to-end.

\subsubsection{Morphological tagging}

This experiment addresses the following research question: \textit{How effectively do fine-tuned morphology-based PLMs capture the grammatical attributes of words?}
We compare models trained on frozen embeddings with their fine-tuned counterparts.
The task, datasets, models, and evaluation metrics are identical to those described in the earlier ``Morphological Tagging'' experiments. 
The only difference is that the PLMs are fine-tuned rather than frozen.

\paragraph{Results and Discussion.}
Across all three languages, fine-tuning consistently improved the performance of our character-based models on morphological tagging.
The magnitude of improvement, however, varied by language and model size.
For Arabic (Table~\ref{tab:test-ft-morph-tag-ara}), fine-tuned models yielded notable gains in ``\textit{Pers}'' and ``\textit{Gen}'', which are among the most challenging features for frozen embeddings. 
Even the smallest model (\texttt{chDzDT\_4x4x32}) achieved competitive overall accuracy after fine-tuning, substantially narrowing the gap with larger configurations.

\begin{table}[!htp]
	\centering\small
	
	\caption{Accuracy of frozen and fine-tuned character-based models on the Arabic morphological tagging task. 
		Fine-tuning consistently improves overall performance, with notable gains for features such as person and gender.\newline
		\textit{Abbreviations:} Num = Number, Mood = Mood, Asp = Aspect, Pers = Person, Gen = Gender, ActV = Active voice}
	
	\begin{tabular}{llrrrrrrr}
		\hline\hline
		\textbf{Model} & \textbf{Type} & \textbf{Num} & \textbf{Mood} & \textbf{Asp} & \textbf{Pers} & \textbf{Gen} & \textbf{ActV} & \textbf{Overall} \\
		\hline 
		\multirow{2}{*}{\texttt{chDzDT\_5x4x128}}
		& Frozen     & 0.9956 & 0.9674 & 0.9949 & 0.8602 & 0.8593 & 0.9946 & 0.9453 \\
		& Fine-tuned & 0.9964 & 0.9701 & 0.9952 & 0.8705 & 0.8711 & 0.9956 & 0.9498 \\
		\hline
		\multirow{2}{*}{\texttt{chDzDT\_4x4x64}}  
		& Frozen      & 0.9873 & 0.9592 & 0.9937 & 0.8605 & 0.8576 & 0.9916 & 0.9417 \\
		& Fine-tuned & 0.9963 & 0.9710 & 0.9957 & 0.8704 & 0.8705 & 0.9945 & 0.9497 \\
		\hline
		\multirow{2}{*}{\texttt{chDzDT\_4x4x32}} 
		& Frozen       & 0.9071 & 0.8854 & 0.9704 & 0.8283 & 0.8036 & 0.9337 & 0.8881 \\
		& Fine-tuned & 0.9957 & 0.9691 & 0.9947 & 0.8722 & 0.8713 & 0.9936 & 0.9494 \\
		\hline\hline
		
	\end{tabular}
	
	\label{tab:test-ft-morph-tag-ara}
\end{table}

In English (Table~\ref{tab:test-ft-morph-tag-eng}), fine-tuning had a mixed effect.
The largest model showed a slight decrease in overall accuracy, which could reflect either mild overfitting due to limited data or under-training given its higher capacity.
In contrast, the mid-sized and smaller models exhibited substantial gains, particularly for the ``\textit{Part}'' feature, which was poorly captured in the frozen setting.
This suggests that fine-tuning can correct deficiencies in specific morphological categories, even if overall accuracy does not always improve uniformly.

\begin{table}[!htp]
	\centering\small
	
	\caption{Accuracy of frozen and fine-tuned character-based models on the English morphological tagging task. 
		Fine-tuning generally improves performance, particularly for participle and nonfinite forms.\newline
		\textit{Abbreviations:} Tns = Tense, NonFin = Nonfinite, Sg = Singular-number, 3rdP = Third-person, ImpSubj = Imperative--subjunctive, Part = Participle.}
	
	\begin{tabular}{llrrrrrrr}
		\hline\hline
		\textbf{Model} & \textbf{Type} & \textbf{Tns} & \textbf{NonFin} & \textbf{Sg} & \textbf{3rdP} & \textbf{ImpSubj} & \textbf{Part} & \textbf{Overall} \\
		\hline 
		\multirow{2}{*}{\texttt{chDzDT\_5x4x128}}  
		& Frozen     & 0.9801  & 0.9814  & 0.9974 & 0.9975   & 0.9814 & 0.7894 & 0.9545 \\
		& Fine-tuned & 0.9645  & 0.9668  & 0.9973 & 0.9973   & 0.9668 & 0.8017 & 0.9491 \\
		\hline
		\multirow{2}{*}{\texttt{chDzDT\_4x4x64}}   
		& Frozen     & 0.9699  & 0.9715  & 0.9908 & 0.9907   & 0.9714 & 0.7974 & 0.9486 \\
		& Fine-tuned & 0.9792  & 0.9808  & 0.9983 & 0.9983   & 0.9807 & 0.8106 & 0.9580 \\
		\hline
		\multirow{2}{*}{\texttt{chDzDT\_4x4x32}}   
		& Frozen     & 0.8821  & 0.9035  & 0.9566 & 0.9567   & 0.9034 & 0.7927 & 0.8992 \\
		& Fine-tuned & 0.9763  & 0.9777  & 0.9983 & 0.9983   & 0.9777 & 0.8105 & 0.9565 \\
		\hline\hline
		
	\end{tabular}
	
	\label{tab:test-ft-morph-tag-eng}
\end{table}
 
For French (Table~\ref{tab:test-ft-morph-tag-fra}), fine-tuning had a more pronounced effect.
All models showed consistent improvements, with notable gains in ``\textit{Asp}'', ``\textit{Tns}'', and ``\textit{Pers}''.
The smallest model benefited the most, indicating that fine-tuning can compensate for limited model capacity by better aligning representations with the task-specific signal.

\begin{table}[!htp]
	\centering\small
	
	\caption{Accuracy of frozen and fine-tuned character-based models on the French morphological tagging task. 
		Fine-tuning yields clear improvements across most features, particularly aspect, tense, and person, resulting in higher overall accuracy.\newline
		\textit{Abbreviations:} Pers = Person, Asp = Aspect, Mood = Mood, NonFin = Nonfinite, Tns = Tense, Num = Number}
		
	\begin{tabular}{llrrrrrrr}
		\hline\hline 
		\textbf{Model} & \textbf{Type} & \textbf{Pers} & \textbf{Asp} & \textbf{Mood} & \textbf{NonFin} & \textbf{Tns} & \textbf{Num} & \textbf{Overall} \\
		\hline 
		\multirow{2}{*}{\texttt{chDzDT\_5x4x128}}
		& Frozen      & 0.8847 & 0.9425 & 0.7872 & 0.9994 & 0.8593 & 0.9970 & 0.9117 \\
		& Fine-tuned  & 0.8956 & 0.9535 & 0.8056 & 0.9998 & 0.8816 & 0.9978 & 0.9223 \\
		\hline
		\multirow{2}{*}{\texttt{chDzDT\_4x4x64}}  
		& Frozen      & 0.8804 & 0.9033 & 0.7624 & 0.9980 & 0.8050 & 0.9895 & 0.8898 \\
		& Fine-tuned  & 0.8960 & 0.9503 & 0.8035 & 0.9993 & 0.8763 & 0.9975 & 0.9205 \\
		\hline
		\multirow{2}{*}{\texttt{chDzDT\_4x4x32}}  
		& Frozen      & 0.7679 & 0.8350 & 0.6734 & 0.9862 & 0.6793 & 0.9023 & 0.8074 \\
		& Fine-tuned  & 0.8947 & 0.9464 & 0.7992 & 0.9989 & 0.8671 & 0.9955 & 0.9170 \\
		\hline\hline
		
	\end{tabular}
	
	\label{tab:test-ft-morph-tag-fra}
\end{table}

Taken together, these results confirm that fine-tuning improves the ability of morphology-based PLMs to encode grammatical features. 
This effect is particularly noticeable in languages with richer morphology, such as Arabic and French.
However, the improvements are not uniform across features or languages. 
This indicates that the benefits of fine-tuning depend on both the morphological complexity of the language and the representational capacity of the model.

\subsubsection{Part-of-speech tagging}

In this experiment, we aim to address the following research question: 
\textit{How effectively do fine-tuned morphology-based PLMs perform on a morpho-syntactic task such as PoS tagging?} 
To this end, we compare models trained on frozen embeddings with their fine-tuned counterparts.
The task, dataset, models, and evaluation metrics are identical to those described in the earlier ``PoS tagging'' experiments.
The only difference being that the PLMs are fine-tuned instead of used in a frozen configuration.

\begin{figure}[!htp]
	\centering

	\includegraphics[width=0.7\textwidth]{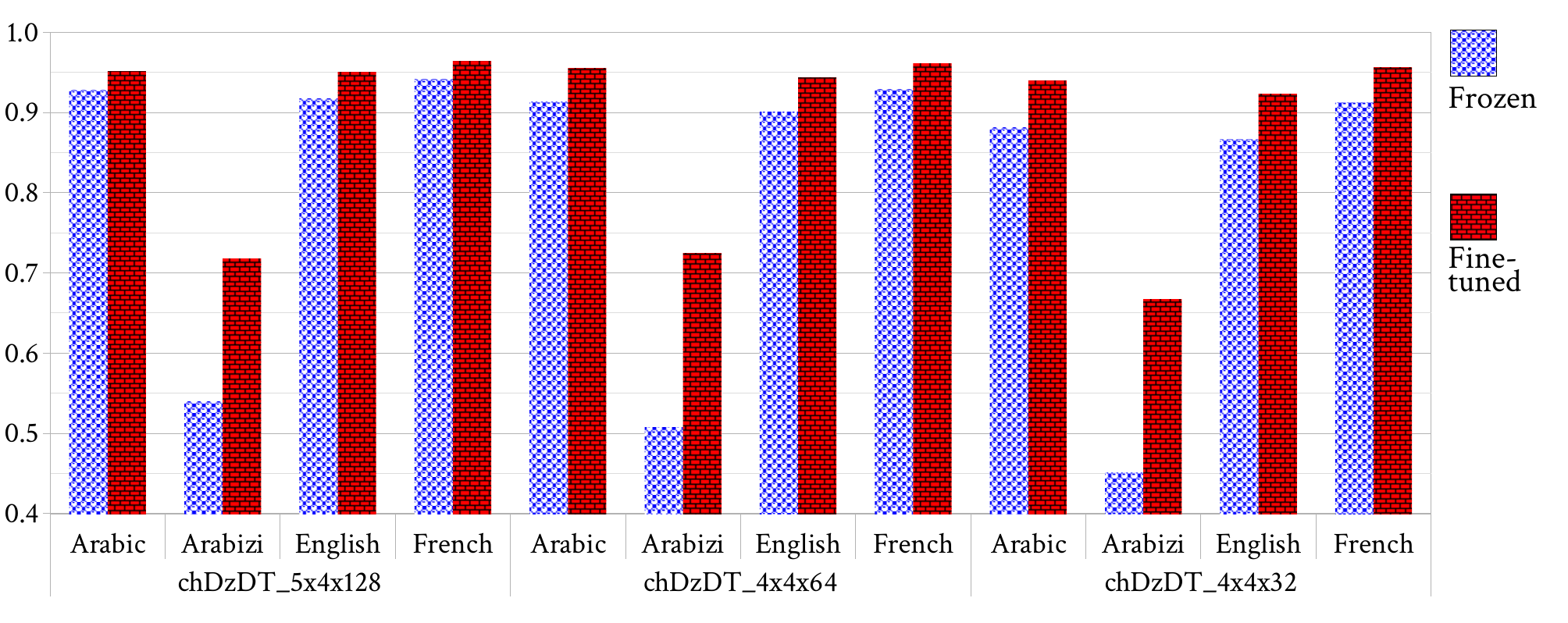}
	
	\caption{PoS tagging accuracy of frozen vs.~fine-tuned models across four languages. Fine-tuning yields consistent improvements, particularly for Arabizi.}

	\label{fig:test-ft-pos}
\end{figure}

\paragraph{Results and discussion.}
The results in Figure~\ref{fig:test-ft-pos} show consistent improvements when fine-tuning morphology-based PLMs compared to relying on frozen embeddings. 
Gains are observed across all four languages, though the magnitude varies. 
The most substantial improvement is seen in Arabizi (e.g., +0.21 for \texttt{chDzDT\_4x4x32}), reflecting the noisy and under-resourced nature of this variety. 
Fine-tuning allows the model to adapt more effectively to irregular orthography and code-switching phenomena. 
Arabic and French also benefit considerably, with accuracy improvements of roughly 3--4 percentage points.
This indicates that even morphologically rich and well-resourced languages gain from contextual adjustment of embeddings. 
English, while already strong in the frozen setting, still shows notable improvements (+0.03--0.04).
This suggests that fine-tuning provides complementary information beyond purely morphological cues.  

Taken together, these findings indicate that fine-tuning both enhances representational quality and improves transferability to morpho-syntactic tasks.  
The relative scale of the gains suggests that languages with higher orthographic variability or fewer resources (e.g., Arabizi) benefit most.  
This pattern aligns with prior studies on multilingual PLMs, which report that fine-tuning is often critical for achieving competitive task-level performance \cite{2018-howard-ruder,2019-peters-al,2020-conneau-al,2020-hu-al}.

%\begin{table}[!htp]
%	\centering\small
%	
%	\caption{Accuracy of frozen and fine-tuned character-based models on PoS tagging task.}
%	\begin{tabular}{lrrcrrcrrcrr}
%		\hline\hline
%		\multirow{2}{*}{\textbf{Model}} & \multicolumn{2}{c}{\textbf{Arabic}} && \multicolumn{2}{c}{\textbf{Arabizi}} && \multicolumn{2}{c}{\textbf{English}} && \multicolumn{2}{c}{\textbf{French}}\\
%		\cline{2-3}\cline{5-6}\cline{8-9}\cline{11-12}
%		& \textbf{Frozen} & \textbf{FTuned} && \textbf{Frozen} & \textbf{FTuned} && \textbf{Frozen} & \textbf{FTuned} && \textbf{Frozen} & \textbf{FTuned} \\
%		\hline
%		\texttt{chDzDT\_5x4x128} & 0.9279 & 0.9511 && 0.5397 & 0.7177 && 0.9169 & 0.9502 && 0.9417 & 0.9640 \\ 
%		\texttt{chDzDT\_4x4x64}  & 0.9132 & 0.9547 && 0.5070 & 0.7243 && 0.9008 & 0.9437 && 0.9291 & 0.9608 \\
%		\texttt{chDzDT\_4x4x32}  & 0.8815 & 0.9396 && 0.4507 & 0.6667 && 0.8664 & 0.9232 && 0.9119 & 0.9562 \\
%		\hline\hline
%		
%	\end{tabular}	
%	
%	\label{tab:test-ft-pos}
%\end{table}

\subsubsection{Sentiment analysis}

In this experiment, we aim to address the following research question: 
\textit{How effectively do fine-tuned morphology-based PLMs perform on a semantic--pragmatic task such as sentiment analysis?}
To this end, we compare models trained on frozen embeddings with their fine-tuned counterparts.
The task, dataset, models, and evaluation metrics are identical to those described in the earlier ``Sentiment analysis'' experiments.
The only difference being that the PLMs are fine-tuned rather than frozen.

\begin{figure}[!htp]
	\centering
	
	\includegraphics[width=0.7\textwidth]{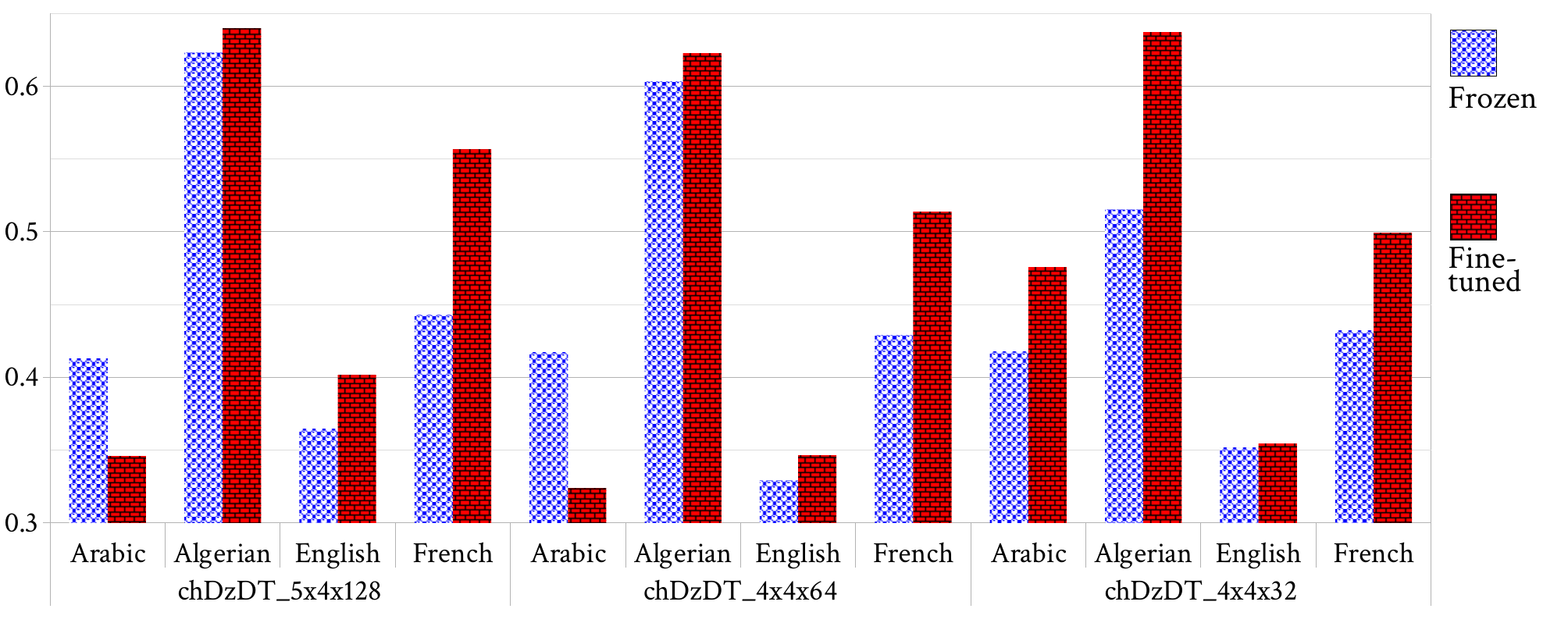}
	
	\caption{Sentiment analysis accuracy of frozen vs.~fine-tuned models across four languages.}

	\label{fig:test-ft-sa}
\end{figure}

\paragraph{Results and discussion.}
The results in Figure~\ref{fig:test-ft-sa} reveal a more nuanced picture than for morpho-syntactic tasks. 
Fine-tuning does not consistently improve performance: in Arabic, accuracy declines across two of the three models.
This suggests potential overfitting or instability when adapting to the sentiment classification signal.  
By contrast, French shows clear gains, particularly for the larger models (e.g., +0.11 for \texttt{chDzDT\_5x4x128}).
This indicates that fine-tuning can enhance semantic discriminability when sufficient training signal is available.  
For Algerian (Arabizi), fine-tuning yields moderate improvements across all models.
This is consistent with the language's noisy and heterogeneous nature, where contextual adaptation appears beneficial.  
English remains relatively stable, with only minor improvements (+0.02--0.04) in most settings.

Taken together, these findings suggest that the benefits of fine-tuning are less uniform for semantic--pragmatic tasks than for morpho-syntactic ones.  
While fine-tuning can improve sentiment classification in resource-rich settings (French) and noisy, under-resourced varieties (Arabizi), it may also amplify noise or dataset bias.
This effect is particularly noticeable in Arabic, underscoring the challenge of sentiment analysis.
It depends not only on morphological and syntactic cues but also on pragmatic and cultural nuances that may not be fully captured through fine-tuning alone.  
Overall, the evidence suggests that character-based models are best used in combination with subword- or word-level representations, rather than as standalone solutions for semantic tasks.

%\begin{table}[!htp]
%	\centering\small
%	
%	\caption{Sentiment analysis accuracy of frozen vs.~fine-tuned models across four languages.}
%	\begin{tabular}{lrrcrrcrrcrr}
%			\hline\hline
%			\multirow{2}{*}{\textbf{Model}} & \multicolumn{2}{c}{\textbf{Arabic}} && \multicolumn{2}{c}{\textbf{Algerian}} && \multicolumn{2}{c}{\textbf{English}} && \multicolumn{2}{c}{\textbf{French}}\\
%			\cline{2-3}\cline{5-6}\cline{8-9}\cline{11-12}
%			& \textbf{Frozen} & \textbf{FTuned} && \textbf{Frozen} & \textbf{FTuned} && \textbf{Frozen} & \textbf{FTuned} && \textbf{Frozen} & \textbf{FTuned} \\
%			\hline
%			\texttt{chDzDT\_5x4x128} & 0.4128 & 0.3455 && 0.6229 & 0.6396 && 0.3643 & 0.4014 && 0.4425 & 0.5563 \\ 
%			\texttt{chDzDT\_4x4x64}  & 0.4169 & 0.3236 && 0.6029 & 0.6225 && 0.3289 & 0.3461 && 0.4287 & 0.5138 \\
%			\texttt{chDzDT\_4x4x32}  & 0.4175 & 0.4754 && 0.5150 & 0.6367 && 0.3513 & 0.3544 && 0.4322 & 0.4989 \\
%			\hline\hline
%			
%	\end{tabular}	
%	
%	\label{tab:test-ft-sa}
%\end{table}

\subsection{Ablation study}

The purpose of this ablation study is to identify which architectural components most significantly affect embedding quality and downstream task performance. 
We systematically vary three key factors: the number of BERT blocks ($N$), the number of attention heads ($H$), and the embedding size ($d$). 
To isolate each factor, we fix the remaining parameters: 
when testing $N$, we fix $H=2$ and $d=16$ with $N \in \{1, 2, 3\}$; 
when testing $H$, we fix $N=2$ and $d=16$ with $H \in \{1, 2, 4\}$ (since $d$ is not divisible by 3, four heads are used instead); 
when testing $d$, we fix $N=2$ and $H=2$ with $d \in \{8, 16, 32\}$. 
This yields seven model variants, summarized in Table~\ref{tab:ablation-models}, where $|\theta|$ denotes the number of parameters for the encoder (excluding task-specific heads), $t$ the training time, and $s/s$ the average number of training samples processed per second.

\begin{table}[!htp]
	\centering\small
	
	\caption{Architectural configurations of chDzDT variants for the ablation study. 
		Each model differs in the number of Transformer blocks ($N$), attention heads ($H$), and hidden size ($d$). 
		The table also reports parameter counts ($|\theta|$), training time ($t$), and average throughput in samples per second ($s/s$).}
	
	\begin{tabular}{rrrrrrp{16pt}rrrrrr}
		\hhline{======~======}
		$N$ & $H$ & $d$ & $|\theta|$ & $t$ & $s/s$ && $N$ & $H$ & $d$ & $|\theta|$ & $t$ & $s/s$ \\
		\cline{1-6}\cline{8-13}
		2 & 2 & 8  & 128,096 & 3h04 &  27,844.677 && 2 & 4 & 16 & 251,200 & 3h06 &  17,330.112 \\
		1 & 2 & 16 & 148,656 & 1h56  &  16,352.247 && 3 & 2 & 16 & 353,744 & 3h20 &  16,180.028 \\
		2 & 1 & 16 & 251,200 & 3h18  &  17,573.737  && 2 & 2 & 32 & 500,864 & 4h20 &  12,431.466 \\
		\hhline{~~~~~~~======}
		2 & 2 & 16 & 251,200 &  3h03 & 17,666.323  && &&&&&\\
		\hhline{======~~~~~~~}
	\end{tabular}
	
	\label{tab:ablation-models}
\end{table}

For each factor ($N$, $H$, or $d$), we assess the impact across multiple tasks: model efficiency, morphological consistency, robustness to orthographic noise, morphological tagging, PoS tagging, and sentiment analysis, as detailed in the previous sections.
Regarding model efficiency, we note that wall-clock training time is an imperfect proxy for computational cost.
It can be influenced by system-level conditions such as background processes, I/O contention, and caching.
Therefore, the reported times should be considered indicative rather than exact.

\subsubsection{Transformer depth}

A key architectural choice in Transformer-based models is depth, defined as the number of Transformer blocks ($N$). 
\textit{To what extent does increasing $N$ enhance representational capacity, and at what computational cost?} 
Deeper models are generally expected to capture richer contextual patterns.
However, they introduce more parameters and longer training times, with the risk of diminishing returns.  
To examine this, we fixed $H=2$ and $d=16$ and varied $N \in \{1, 2, 3\}$.

\begin{figure}[!htp]
	\centering
	
	\includegraphics[width=0.4\textwidth]{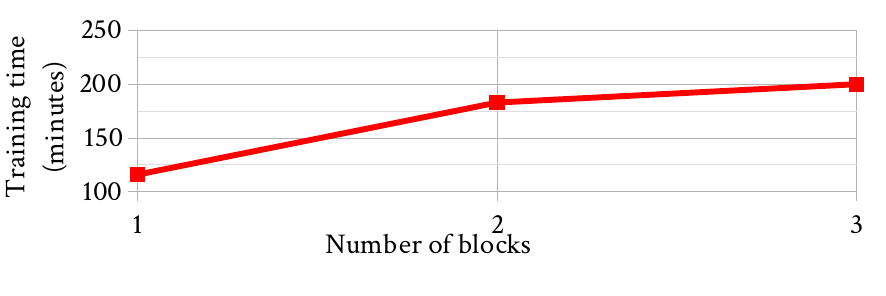}
	\hfill
	\includegraphics[width=0.4\textwidth]{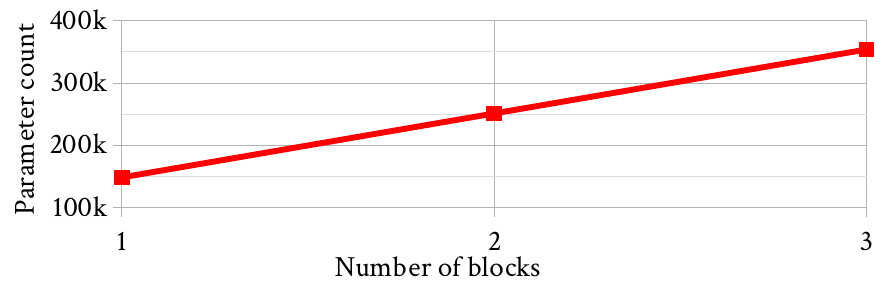}
	
	\caption{Effect of Transformer depth ($N$) on training time and parameter count. Both increase with $N$, though training time grows more steeply.}
	
	\label{fig:ablation-N-t_theta}
\end{figure}

\paragraph{Model efficiency.}
Figure~\ref{fig:ablation-N-t_theta} shows that both parameter count and training time increase as Transformer depth $N$ grows. 
Parameter growth is roughly linear, since each additional block contributes a fixed set of weights. 
Training time, however, rises more steeply, reflecting accumulated sequential computations and system-level overhead. 
This indicates that adding more blocks increases model capacity but reduces training efficiency, highlighting a trade-off between representational power and computational cost.

\begin{figure}[!htp]
	\centering\small

	\begin{subfigure}[t]{0.22\textwidth}
		\includegraphics[width=\textwidth]{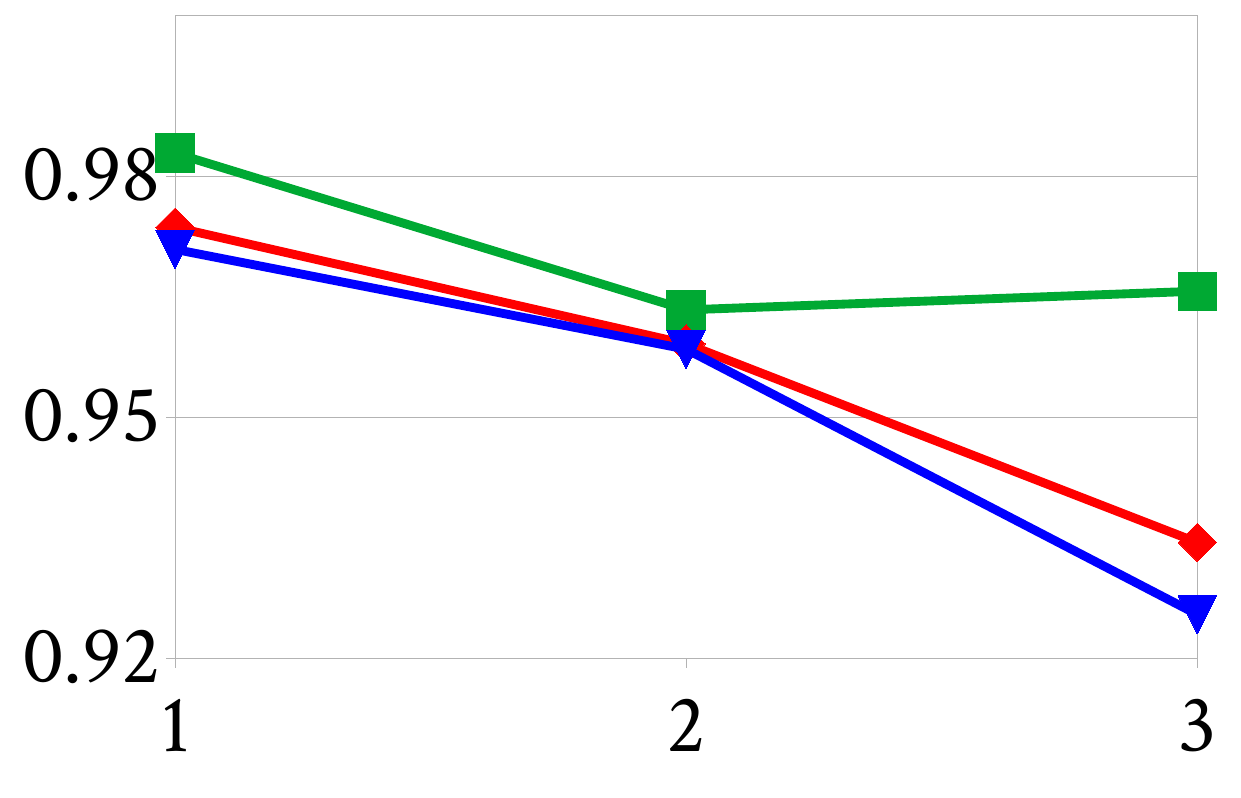}
		\subcaption{Deriv--Avg}
	\end{subfigure}
	\begin{subfigure}[t]{0.22\textwidth}
		\includegraphics[width=\textwidth]{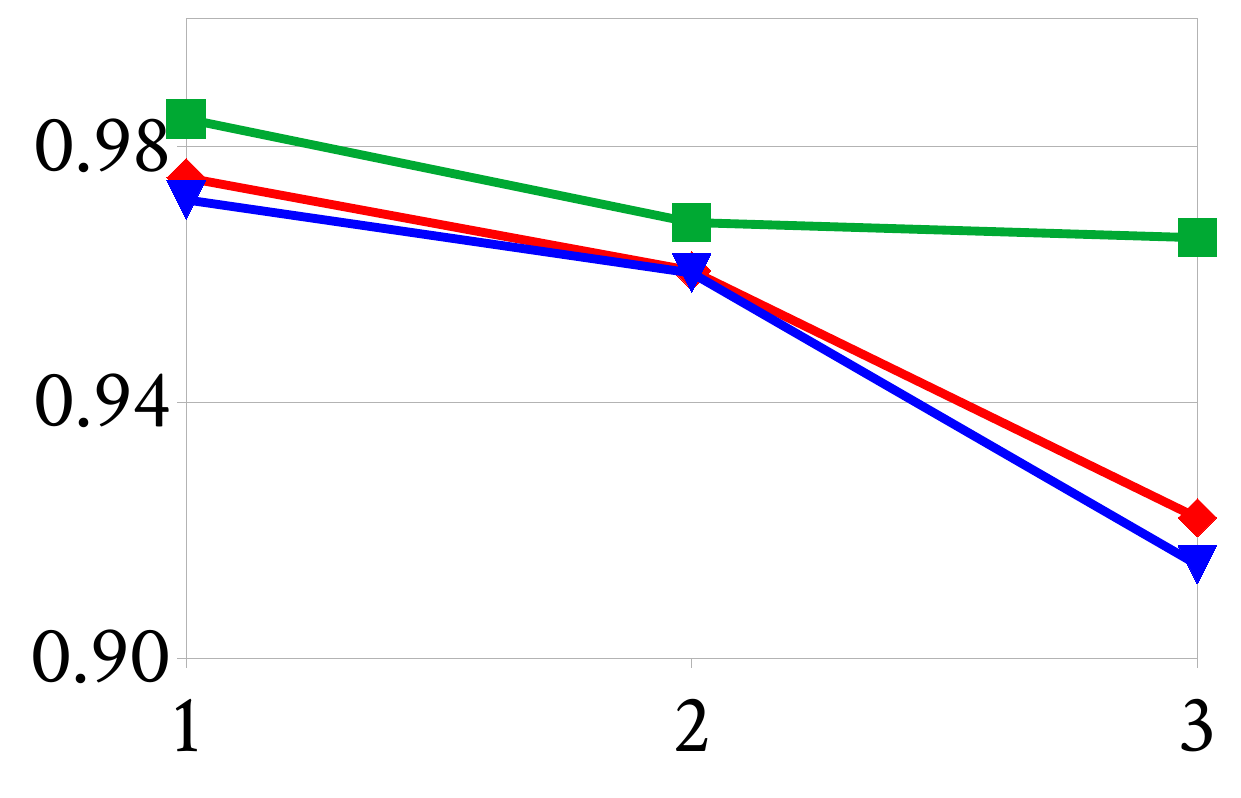}
		\subcaption{Deriv--Min}
	\end{subfigure}
	\begin{subfigure}[t]{0.22\textwidth}
		\includegraphics[width=\textwidth]{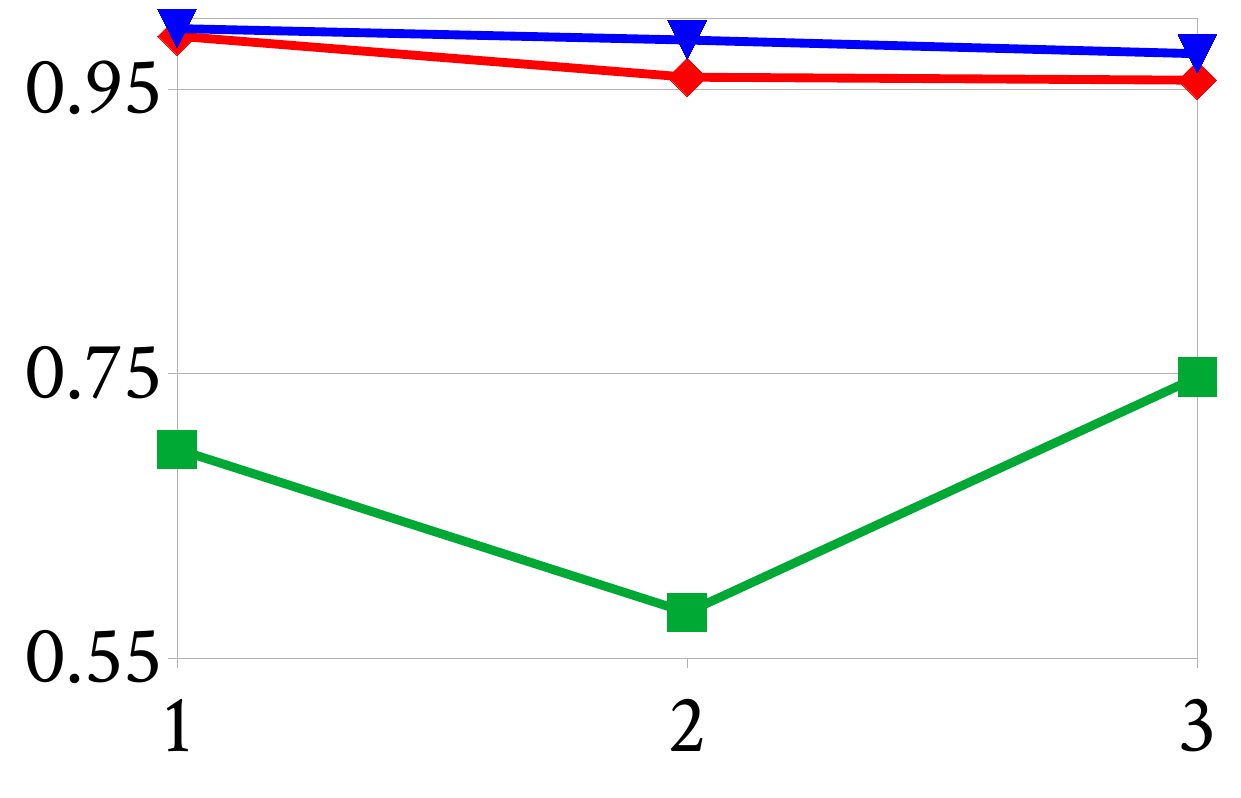}
		\subcaption{Infl--Avg}
	\end{subfigure}
	\begin{subfigure}[t]{0.22\textwidth}
		\includegraphics[width=\textwidth]{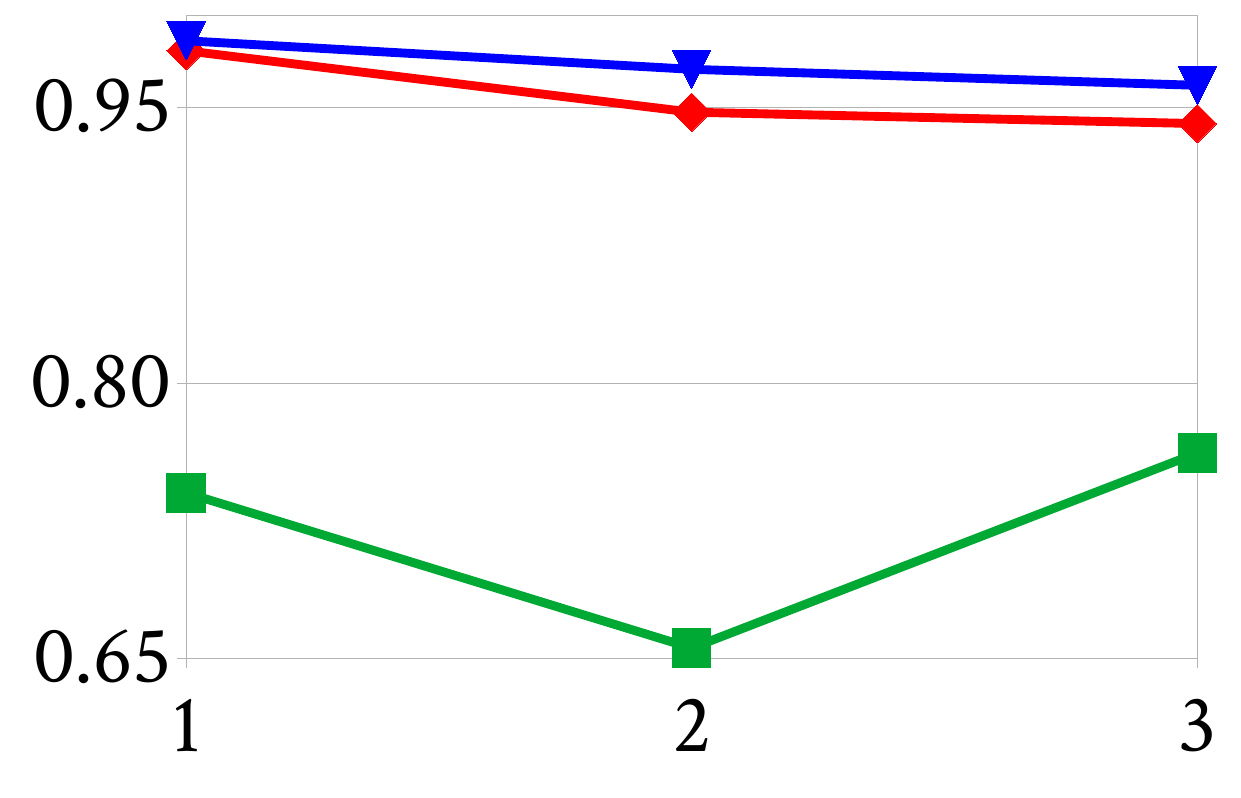}
		\subcaption{Infl--Min}
	\end{subfigure}
	\hfill
	\begin{subfigure}[t]{0.07\textwidth}
		\includegraphics[width=\textwidth]{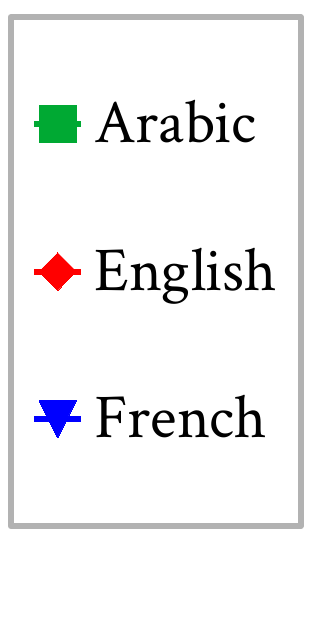}
		%		\subcaption{Legend}
	\end{subfigure}
	
	\caption{Effect of Transformer depth ($N$) on morphological consistency in Arabic, French, and English.  
		Plots show derivational (Deriv) and inflectional (Infl) morphology on the ``Avg'' and ``Min'' datasets, using average cosine similarity between words and their root.}
	
	\label{fig:ablation-N-morph-consist}
\end{figure}

\paragraph{Morphological consistency.}
Figure~\ref{fig:ablation-N-morph-consist} shows that Transformer depth $N$ affects derivational and inflectional morphology in distinct ways. 
For derivational relations, performance declines with depth in English and French, which may reflect that deeper layers capture broader regularities at the expense of fine-grained root--form patterns. 
This decline might also be influenced by the fixed embedding size ($d=16$), suggesting that increasing $N$ without scaling $d$ could restrict representational capacity. 
Arabic shows less decline, partly because derivational clusters combine both templatic processes and semantic categories. 
For example, different agentive forms arise depending on verb class, making them less tightly tied to form alone. 
Inflectional morphology behaves differently: English and French remain mostly stable, while Arabic exhibits a sharp drop at $N=2$ followed by recovery at $N=3$.
This pattern reflects Arabic's complex combination of templatic and affixal morphological processes.
Overall, derivational structure is more vulnerable to erosion with depth, whereas inflectional patterns may benefit from added layers in language-specific ways. 
These results suggest that, under a fixed embedding size, shallow networks better preserve morphological clustering, while deeper networks risk losing consistency unless representational capacity is increased.

%\begin{tabular}{|c|c|c|c|c|c|c|c|c|c|c|c|c|c|c|c|}
%	\hline
%	& deriv & deriv & deriv &  & deriv & deriv & deriv &  & infl & infl & infl &  & infl & infl & infl \\
%	\hline
%	& avg & avg & avg &  & min & min & min &  & avg & avg & avg &  & min & min & min \\
%	\hline
%	N & Arabic & English & French &  & Arabic & English & French &  & Arabic & English & French &  & Arabic & English & French \\
%	\hline
%	1 & 0.9829 & 0.9737 & 0.9709 &  & 0.9844 & 0.9752 & 0.9717 &  & 0.6971 & 0.9873 & 0.9925 &  & 0.7403 & 0.9810 & 0.9864 \\
%	\hline
%	2 & 0.9634 & 0.9592 & 0.9585 &  & 0.9682 & 0.9606 & 0.9603 &  & 0.5825 & 0.9584 & 0.9845 &  & 0.6557 & 0.9475 & 0.9709 \\
%	\hline
%	3 & 0.9657 & 0.9345 & 0.9255 &  & 0.9658 & 0.9220 & 0.9147 &  & 0.7478 & 0.9563 & 0.9749 &  & 0.7620 & 0.9413 & 0.9622 \\
%	\hline
%\end{tabular}

\begin{figure}[!htp]
	\centering\small
	
	\begin{subfigure}[t]{0.27\textwidth}
		\includegraphics[width=\textwidth]{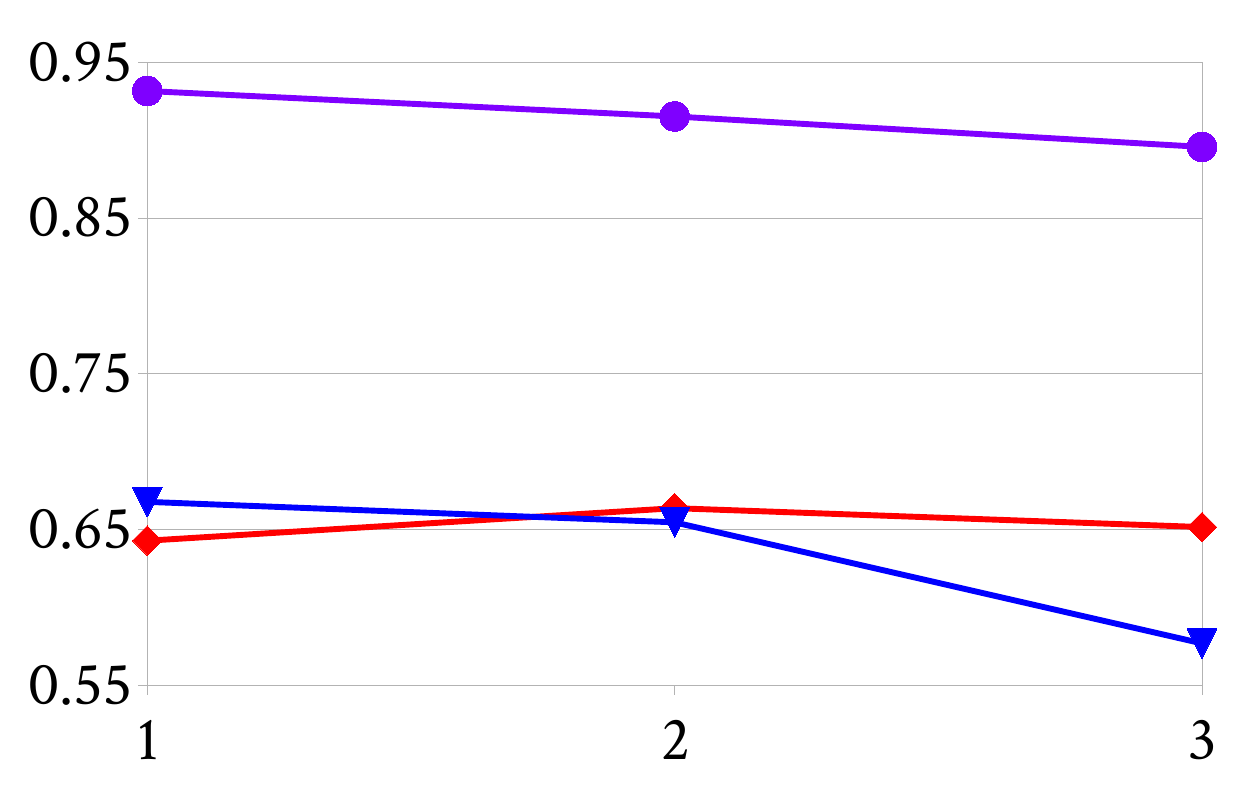}
		\subcaption{Obfus. clusters}
	\end{subfigure}
	\begin{subfigure}[t]{0.27\textwidth}
		\includegraphics[width=\textwidth]{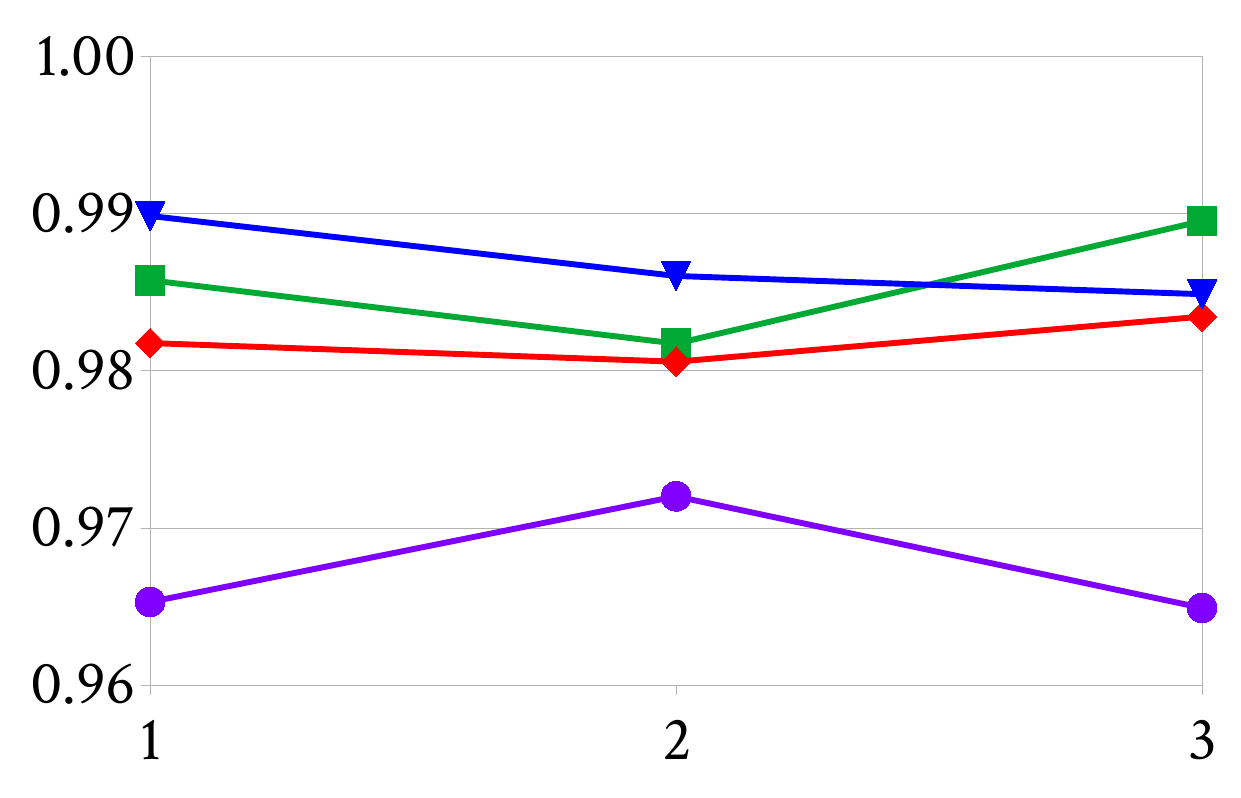}
		\subcaption{Obfus. tuples-*}
	\end{subfigure}
	\begin{subfigure}[t]{0.27\textwidth}
		\includegraphics[width=\textwidth]{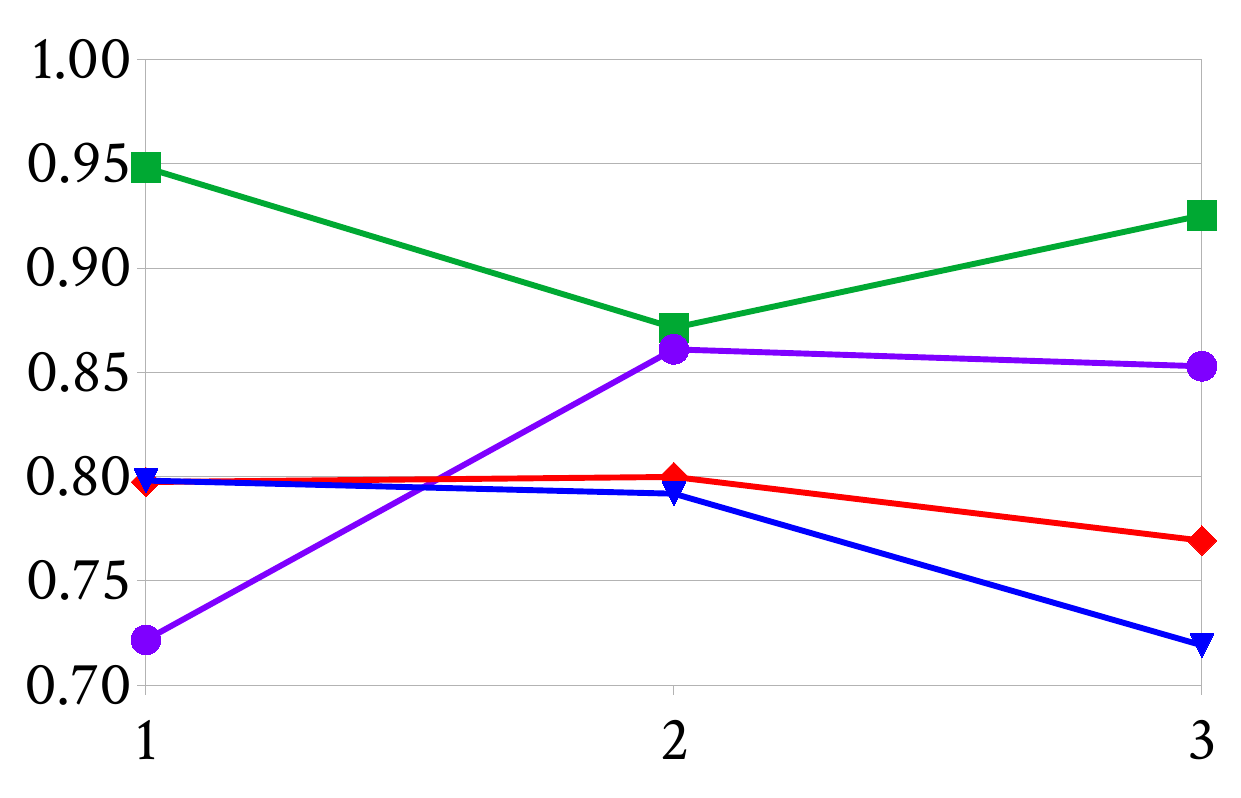}
		\subcaption{Obfus. tuples-c}
	\end{subfigure}
	\hfill
	\begin{subfigure}[t]{0.07\textwidth}
		\includegraphics[width=\textwidth]{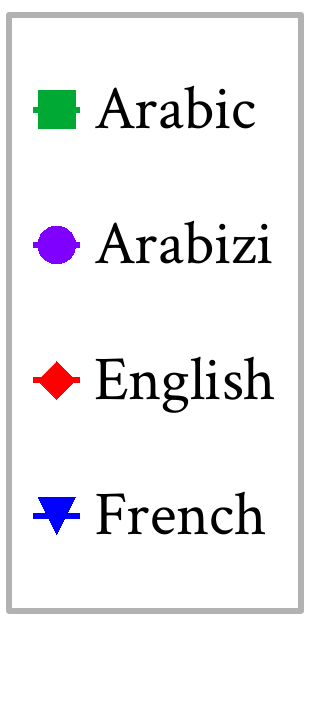}
		%		\subcaption{Legend}
	\end{subfigure}
	
	\caption{Effect of Transformer depth ($N$) on robustness to orthographic noise in Arabic, Arabizi, French, and English. 
		Plots show: (a) cluster-based obfuscations, (b) tuple obfuscations with \texttt{*} substitutions, and 
		(c) tuple obfuscations with visually or phonetically similar substitutions. 
		Scores are reported as Average Cosine Similarity (ACS). 
		Note that Arabic is absent from (a), as it is predominantly written in standard form and does not exhibit such obfuscations.}
	
	\label{fig:ablation-N-morph-noise}
\end{figure}

\paragraph{Robustness to orthographic noise.}
Figure~\ref{fig:ablation-N-morph-noise} shows that Transformer depth ($N$) affects resilience to orthographic variation differently across languages. 
Under cluster-based obfuscations, performance declines with increasing depth. 
Arabizi remains robust (ACS $>0.89$ at $N=3$), while English and French deteriorate more sharply. 
This likely reflects the more standardized nature of their pre-training data. 
In contrast, tuple obfuscations with simple ``*'' substitutions yield uniformly high similarity ($>0.96$) across all languages, indicating that minimal masking is easily handled. 
The most challenging scenario is tuple obfuscations with phonetically or visually similar substitutions.
Arabic shows partial recovery at greater depths, Arabizi maintains strong robustness, and English and French decline steadily.
This pattern reflects reliance on rigid orthography. 
Overall, robustness depends on both the type of perturbation and the language system. 
Shallow models preserve similarity better under cluster-level variation, while deeper models can sometimes aid recovery from fine-grained substitutions in morphologically rich languages like Arabic.

%\begin{tabular}{|c|c|c|c|c|c|c|c|c|c|c|c|c|c|}
%	\hline
%	&  &  & ACS & ACS & ACS & ACS1* & ACS1* & ACS1* & ACS1* & ACS1c & ACS1c & ACS1c & ACS1c \\
%	\hline
%	N &  &  & Arabizi & English & French & Arabic & Arabizi & English & French & Arabic & Arabizi & English & French \\
%	\hline
%	1 &  &  & 0.9317 & 0.6428 & 0.6678 & 0.9858 & 0.9653 & 0.9818 & 0.9899 & 0.9481 & 0.7217 & 0.7974 & 0.7981 \\
%	\hline
%	2 &  &  & 0.9153 & 0.6637 & 0.6547 & 0.9817 & 0.9720 & 0.9806 & 0.9860 & 0.8714 & 0.8610 & 0.7999 & 0.7918 \\
%	\hline
%	3 &  &  & 0.8957 & 0.6516 & 0.5769 & 0.9895 & 0.9649 & 0.9835 & 0.9849 & 0.9253 & 0.8529 & 0.7693 & 0.7191 \\
%	\hline
%\end{tabular}

\begin{figure}[!htp]
	\centering\small
	
	\begin{subfigure}[t]{0.24\textwidth}
		\includegraphics[width=\textwidth]{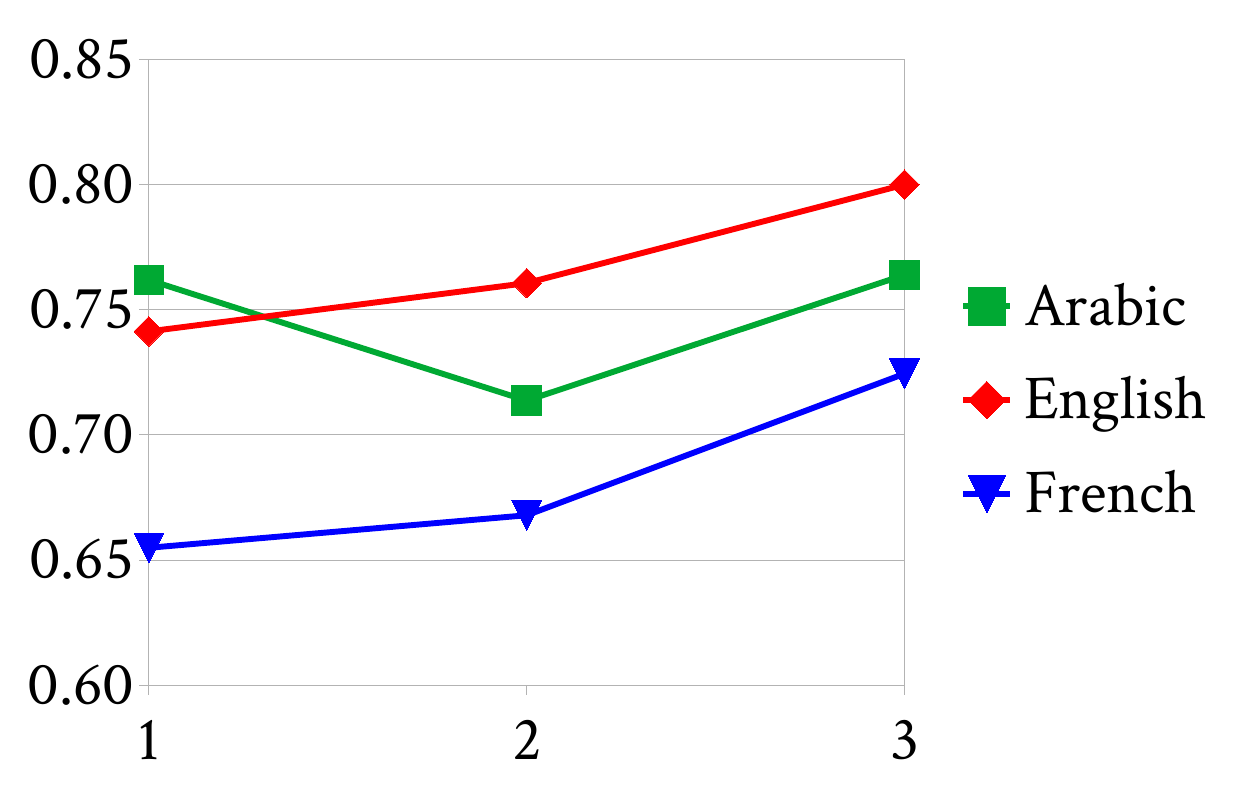}
		\subcaption{Overall Accuracy}
	\end{subfigure}
	\begin{subfigure}[t]{0.24\textwidth}
		\includegraphics[width=\textwidth]{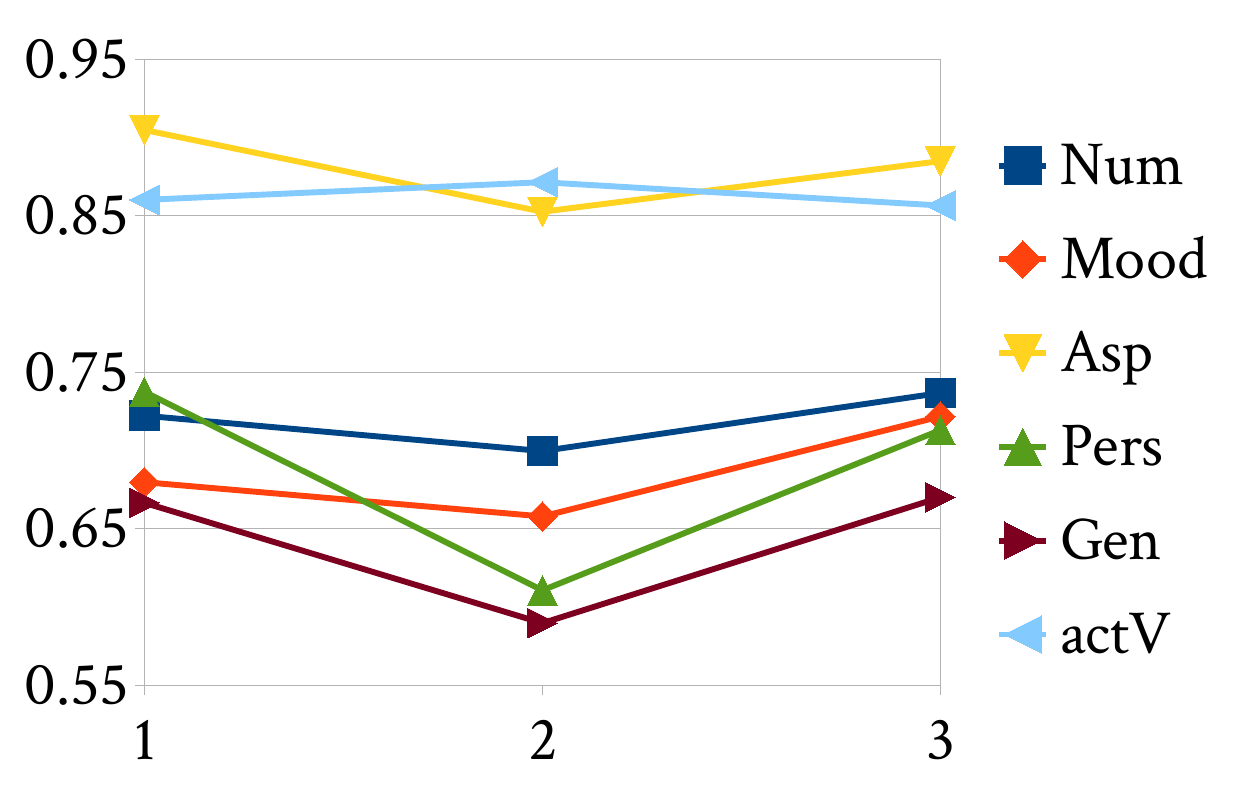}
		\subcaption{Accuracy of Arabic tags}
	\end{subfigure}
	\begin{subfigure}[t]{0.24\textwidth}
		\includegraphics[width=\textwidth]{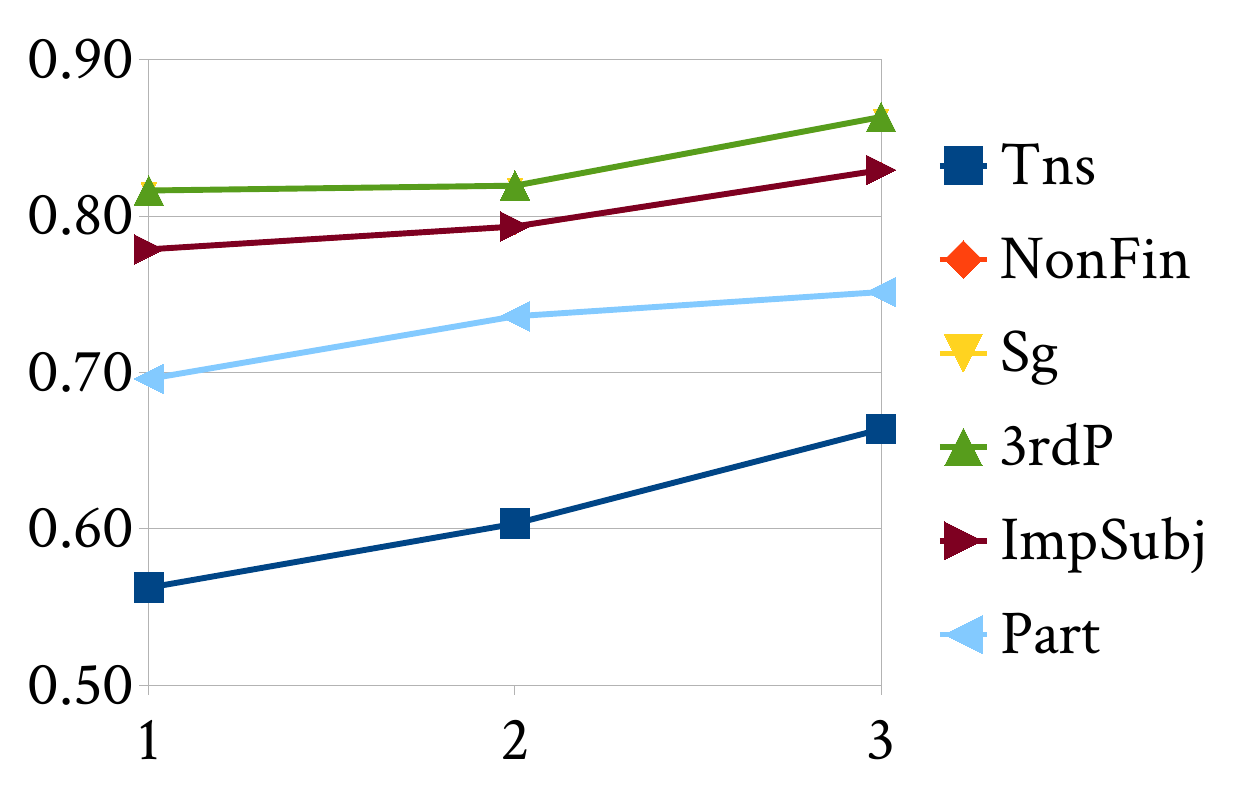}
		\subcaption{Accuracy of English tags}
	\end{subfigure}
	\begin{subfigure}[t]{0.24\textwidth}
		\includegraphics[width=\textwidth]{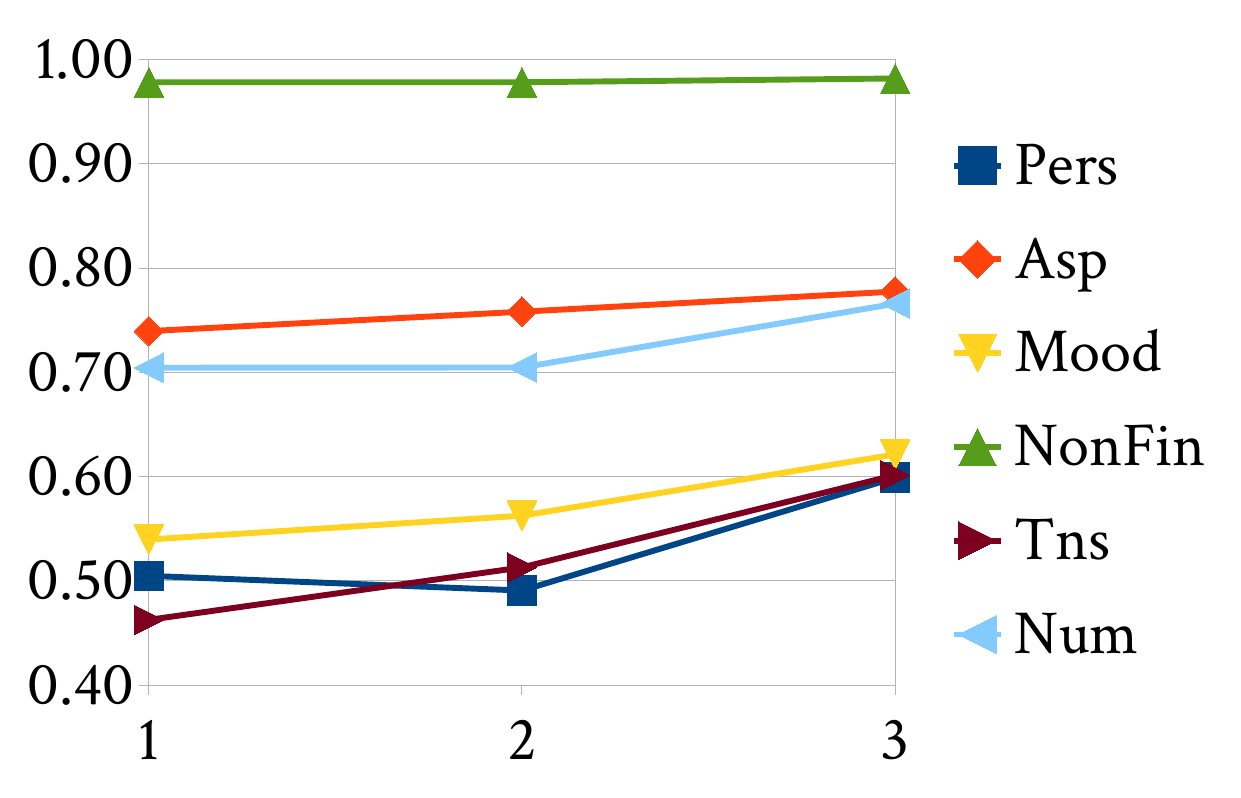}
		\subcaption{Accuracy of French tags}
	\end{subfigure}
	
	\caption{Effect of Transformer depth ($N$) on morphological tagging in Arabic, English, and French. 
		(a) Overall accuracy across the three languages; (b) Arabic tags; (c) English tags; (d) French tags. 
		In English, the curves for \textit{Sg} and \textit{3rdP} overlap, as do those for \textit{NonFin} and \textit{ImpSubj}.}

	\label{fig:ablation-N-morph-tag}
\end{figure}

\paragraph{Morphological tagging.}
Figure~\ref{fig:ablation-N-morph-tag} shows that morphological tagging accuracy is sensitive to Transformer depth. 
Overall performance generally improves with additional blocks, although Arabic exhibits a dip at $N=2$ before recovering at $N=3$. 
English consistently achieves the highest scores, reflecting the relative simplicity of its binary features, whereas French lags behind due to its larger and more complex feature inventory. 
At the level of individual categories, classification gains with depth are modest, indicating that the models converge quickly. 
The irregular trajectory of Arabic, with accuracies dropping at $N=2$ and rebounding at $N=3$, likely reflects optimization instability. 
In English, coinciding curves for features such as \textit{Sg}/\textit{3rdP} and \textit{NonFin}/\textit{ImpSubj} confirm strong correlations already noted. 
Overall, the results suggest that shallow models capture broad morphological patterns. 
Moderate depth is required for finer distinctions, particularly in morphologically simpler languages, while excessive stacking does not guarantee consistent improvement.

%\begin{tabular}{|c|c|c|c|c|c|c|c|c|c|c|c|c|c|c|c|c|c|c|c|c|c|c|c|}
%	\hline
%	N &  &  & Overall &  &  & Arabic &  &  &  &  &  & English &  &  &  &  &  & French &  &  &  &  &  \\
%	\hline
%	N &  &  & Arabic & English & French & Num & Mood & Asp & Pers & Gen & actV & Tns & NonFin & Sg & 3rdP & ImpSubj & Part & Pers & Asp & Mood & NonFin & Tns & Num \\
%	\hline
%	1 &  &  & 0.7619 & 0.7414 & 0.6549 & 0.7223 & 0.6798 & 0.9050 & 0.7376 & 0.6666 & 0.8603 & 0.5624 & 0.7786 & 0.8164 & 0.8163 & 0.7787 & 0.6959 & 0.5049 & 0.7396 & 0.5397 & 0.9782 & 0.4627 & 0.7045 \\
%	\hline
%	2 &  &  & 0.7137 & 0.7607 & 0.6679 & 0.6997 & 0.6579 & 0.8527 & 0.6107 & 0.5897 & 0.8716 & 0.6033 & 0.7931 & 0.8191 & 0.8194 & 0.7933 & 0.7359 & 0.4908 & 0.7582 & 0.5626 & 0.9782 & 0.5130 & 0.7047 \\
%	\hline
%	3 &  &  & 0.7640 & 0.8000 & 0.7245 & 0.7369 & 0.7218 & 0.8851 & 0.7131 & 0.6701 & 0.8567 & 0.6637 & 0.8292 & 0.8631 & 0.8633 & 0.8294 & 0.7514 & 0.5990 & 0.7775 & 0.6213 & 0.9818 & 0.6015 & 0.7659 \\
%	\hline
%\end{tabular}

\begin{figure}[!htp]
	\centering\small
	
	\begin{subfigure}[t]{0.22\textwidth}
		\includegraphics[width=\textwidth]{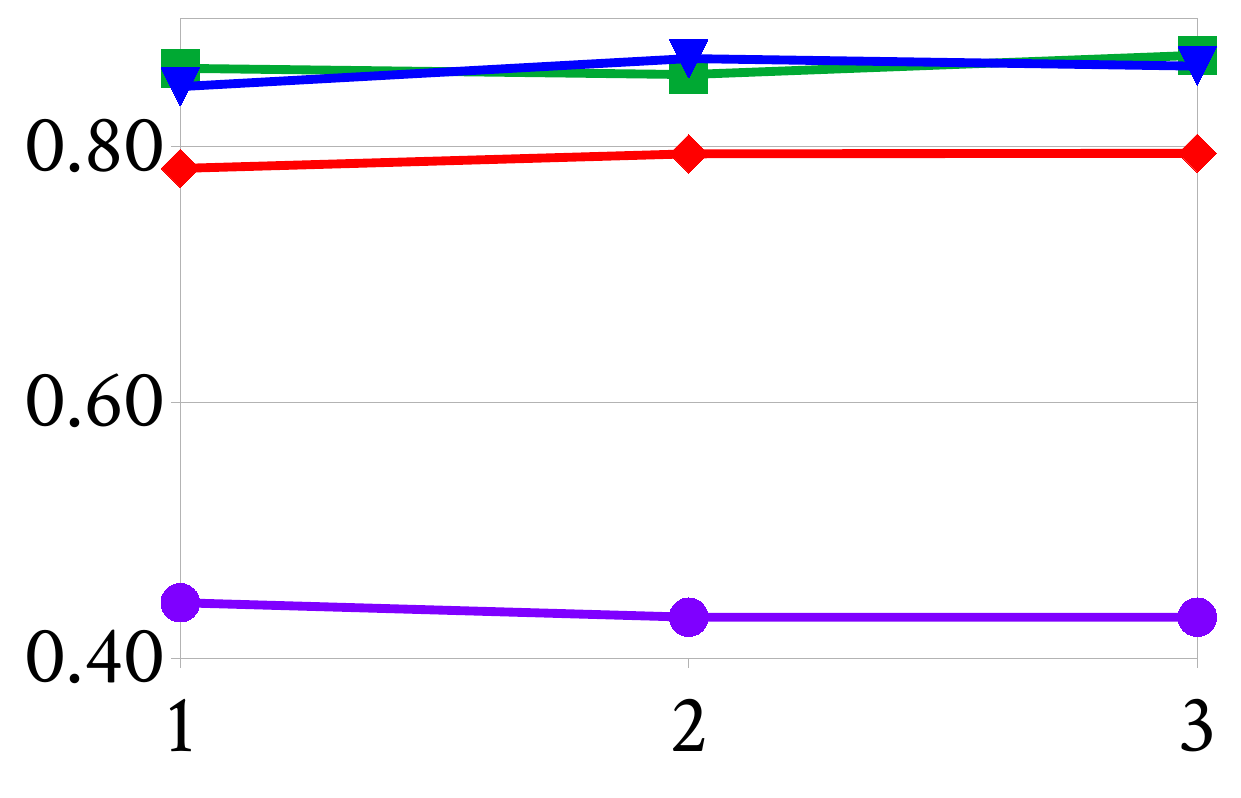}
		\subcaption{Overall accuracy}
	\end{subfigure}
	\begin{subfigure}[t]{0.22\textwidth}
		\includegraphics[width=\textwidth]{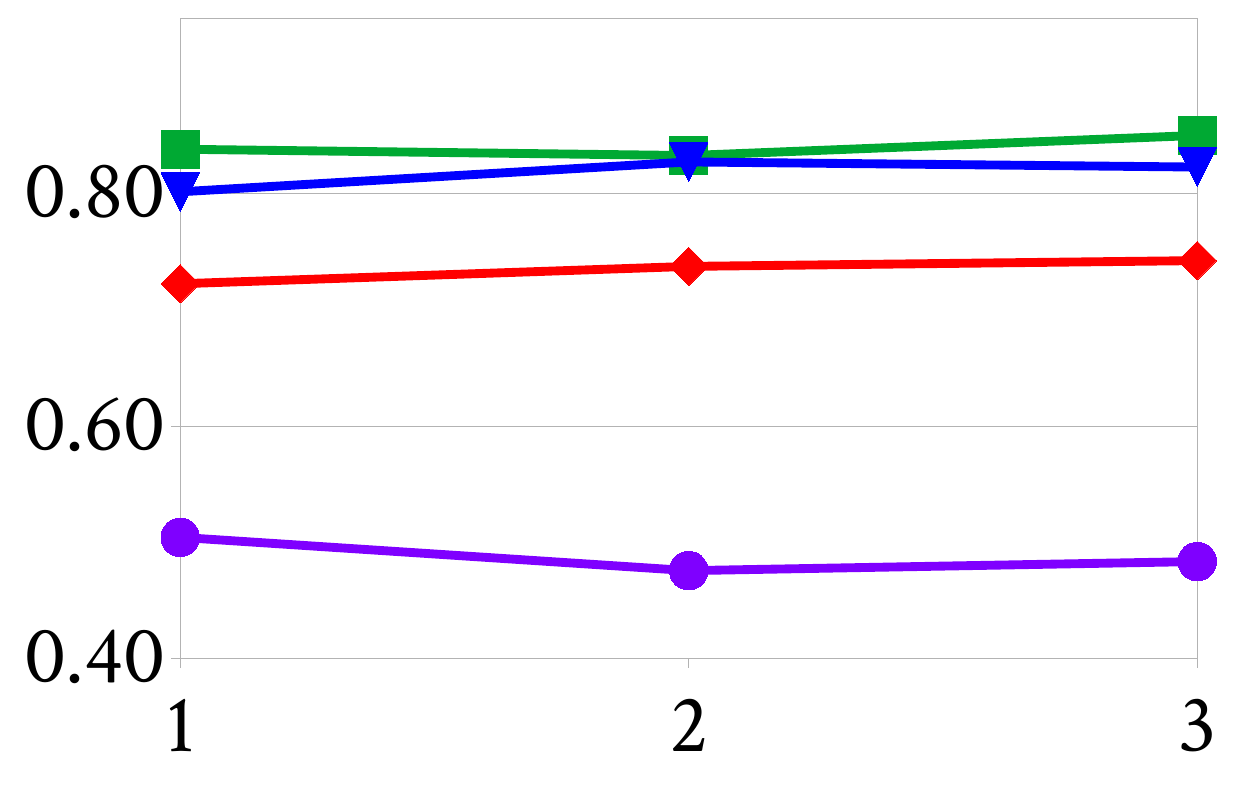}
		\subcaption{Noun F1-score}
	\end{subfigure}
	\begin{subfigure}[t]{0.22\textwidth}
		\includegraphics[width=\textwidth]{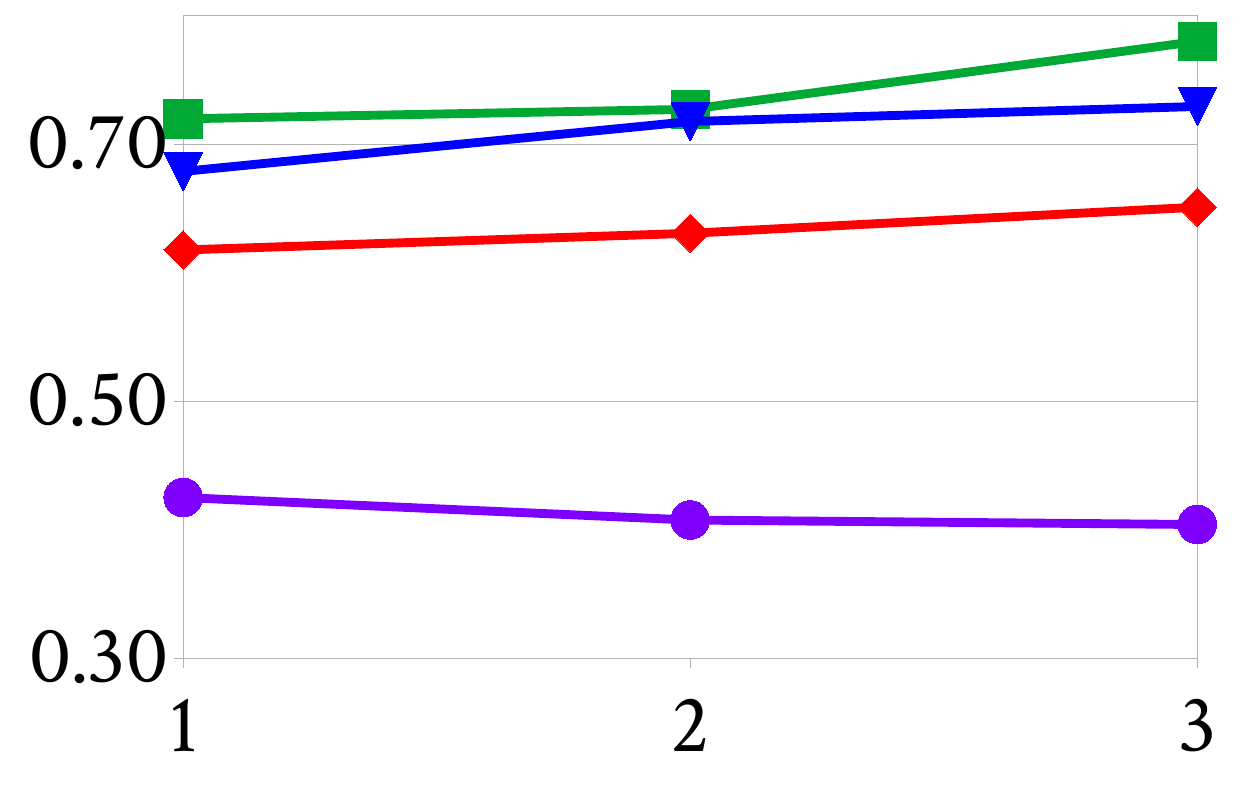}
		\subcaption{Verb F1-score}
	\end{subfigure}
	\begin{subfigure}[t]{0.22\textwidth}
		\includegraphics[width=\textwidth]{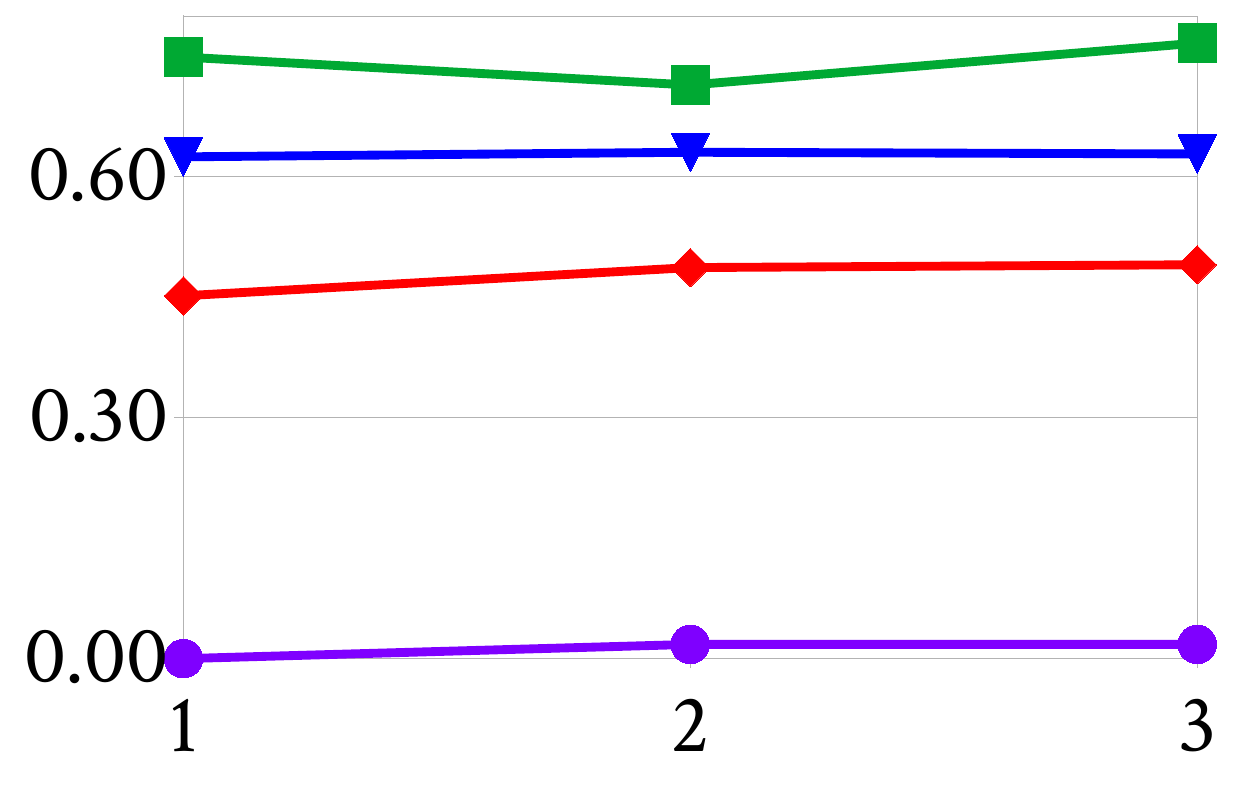}
		\subcaption{Adjective F1-score}
	\end{subfigure}
	\hfill
	\begin{subfigure}[t]{0.07\textwidth}
		\includegraphics[width=\textwidth]{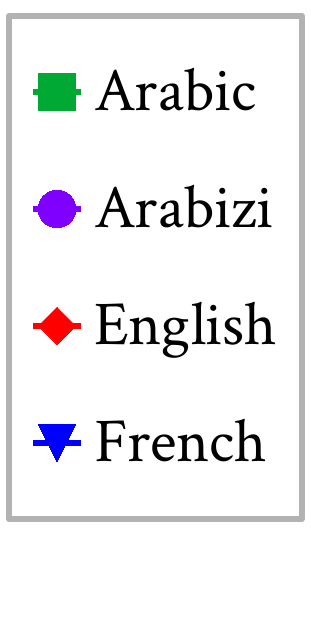}
		%		\subcaption{Legend}
	\end{subfigure}
	
	\caption{Effect of Transformer depth ($N$) on PoS tagging in Arabic, Arabizi, English, and French. 
		Results were obtained by training a BiGRU decoder on frozen chDzDT embeddings.}
		
	\label{fig:ablation-N-pos-tag}
\end{figure}

\paragraph{Part-of-speech tagging.}
Figure~\ref{fig:ablation-N-pos-tag} shows that increasing Transformer depth has only marginal effects on PoS tagging performance. 
Overall accuracy remains nearly constant across depths for all four languages, with differences rarely exceeding one to two points. 
At the category level, nouns and verbs show slight improvements at $N=3$, particularly in Arabic and French, but the effect size is modest. 
Adjectives are the least stable, though variations remain small in absolute terms. 
Arabizi consistently lags behind the other languages, and depth does not meaningfully change this trend. 
These results suggest that PoS tagging benefits little from deeper embeddings, as shallow configurations already provide sufficient input for the downstream classifier.

%\begin{tabular}{|c|c|c|c|c|c|c|c|c|c|c|c|c|c|c|c|c|c|c|}
%	\hline
%	&  &  & Overall &  &  &  & Noun &  &  &  & Verb &  &  &  & Adjective &  &  &  \\
%	\hline
%	N &  &  & Arabic & Arabizi & English & French & Arabic & Arabizi & English & French & Arabic & Arabizi & English & French & Arabic & Arabizi & English & French \\
%	\hline
%	1 &  &  & 0.8613 & 0.4437 & 0.7831 & 0.8470 & 0.8379 & 0.5043 & 0.7223 & 0.8013 & 0.7199 & 0.4253 & 0.6179 & 0.6790 & 0.7488 & 0.0000 & 0.4516 & 0.6246 \\
%	\hline
%	2 &  &  & 0.8564 & 0.4324 & 0.7945 & 0.8689 & 0.8325 & 0.4757 & 0.7372 & 0.8270 & 0.7273 & 0.4078 & 0.6309 & 0.7178 & 0.7140 & 0.0177 & 0.4868 & 0.6303 \\
%	\hline
%	3 &  &  & 0.8715 & 0.4324 & 0.7949 & 0.8630 & 0.8498 & 0.4835 & 0.7422 & 0.8226 & 0.7802 & 0.4043 & 0.6512 & 0.7295 & 0.7662 & 0.0177 & 0.4903 & 0.6282 \\
%	\hline
%\end{tabular}

\begin{figure}[!htp]
	\centering\small
	
	\begin{subfigure}[t]{0.22\textwidth}
		\includegraphics[width=\textwidth]{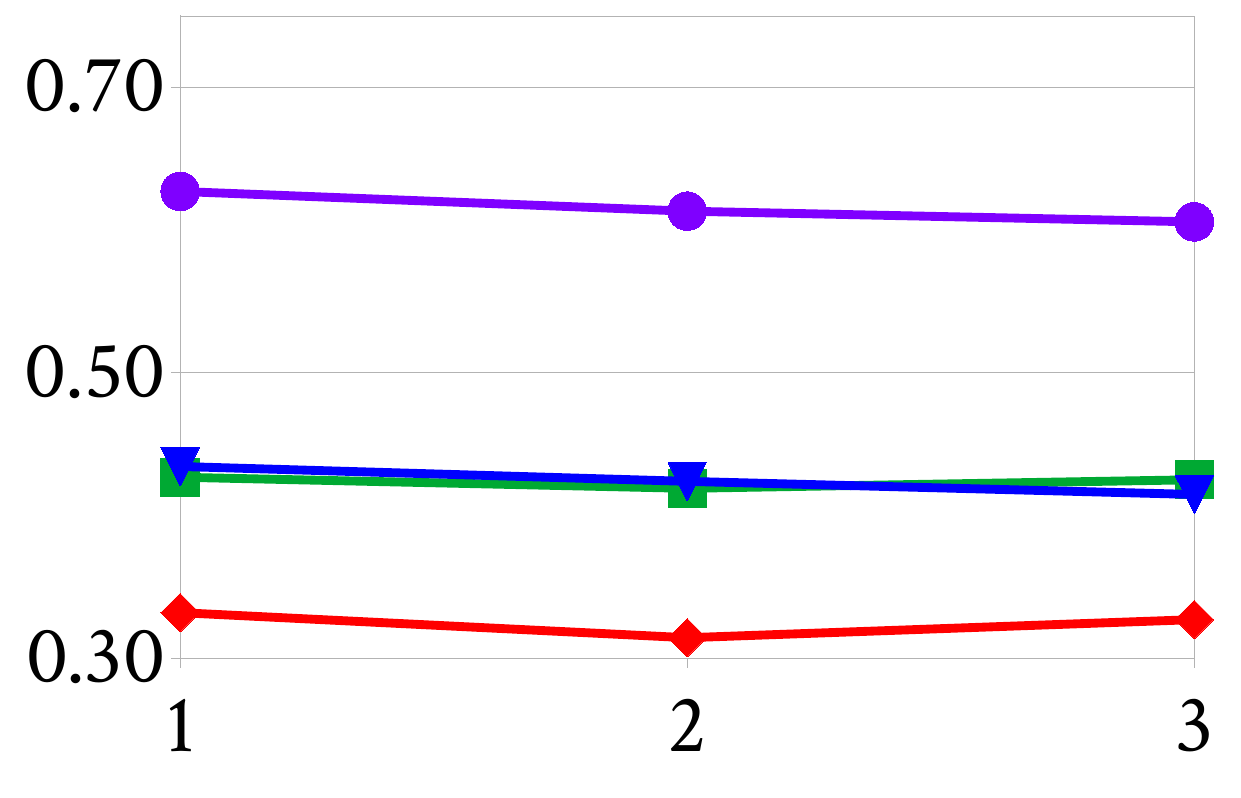}
		\subcaption{Overall Accuracy}
	\end{subfigure}
	\begin{subfigure}[t]{0.22\textwidth}
		\includegraphics[width=\textwidth]{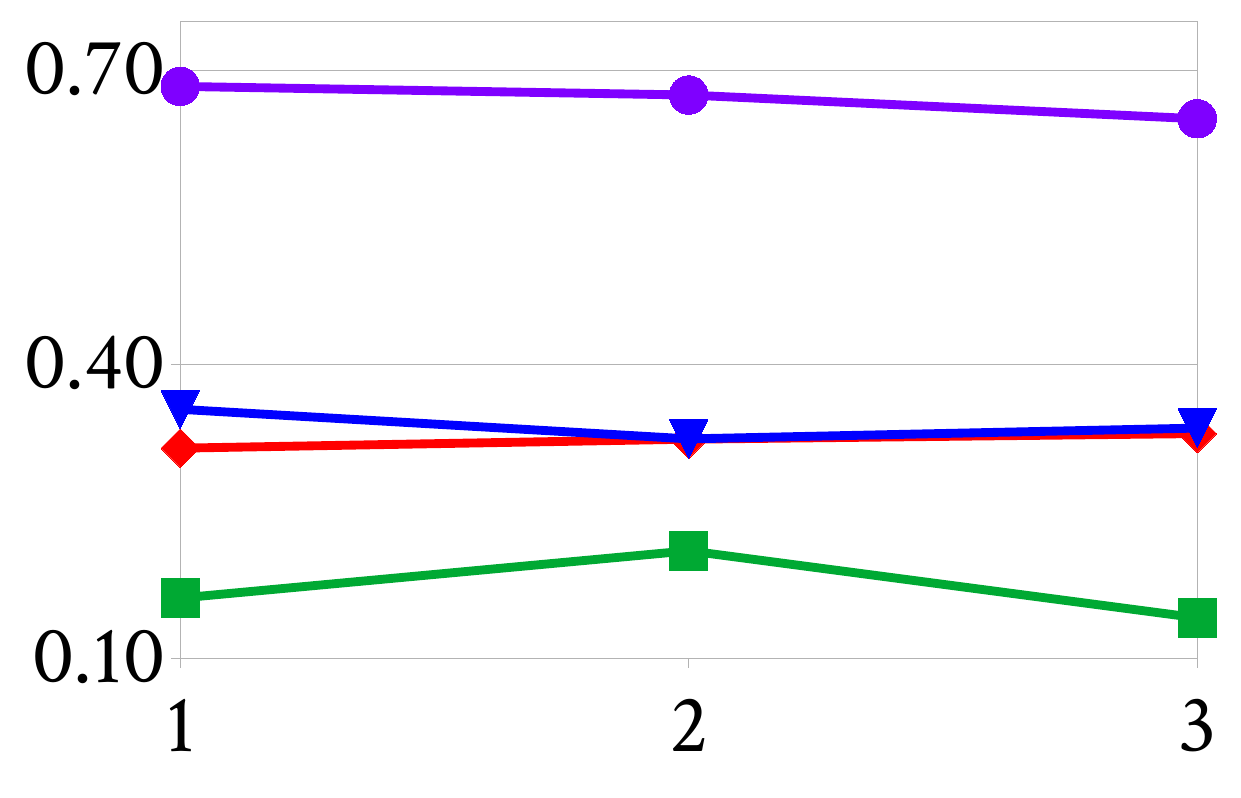}
		\subcaption{Positive F1-score}
	\end{subfigure}
	\begin{subfigure}[t]{0.22\textwidth}
		\includegraphics[width=\textwidth]{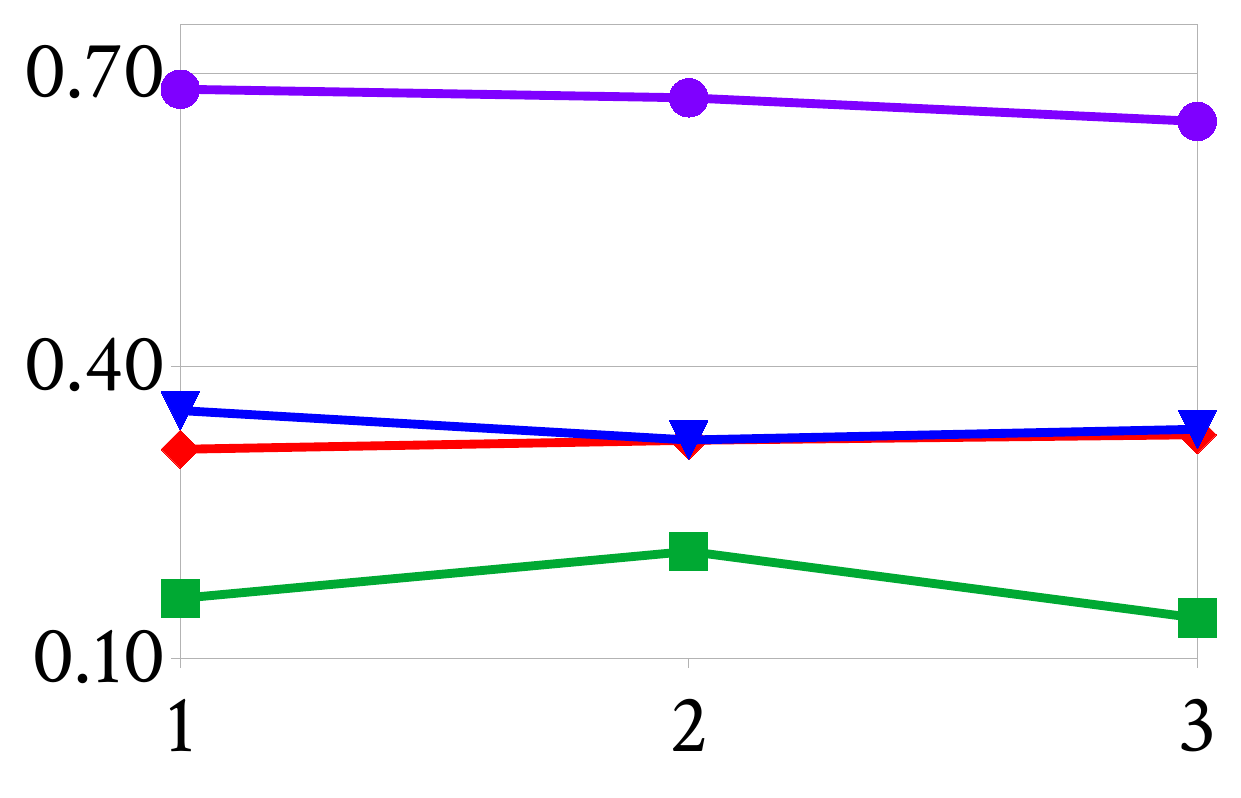}
		\subcaption{Negative F1-score}
	\end{subfigure}
	\begin{subfigure}[t]{0.22\textwidth}
		\includegraphics[width=\textwidth]{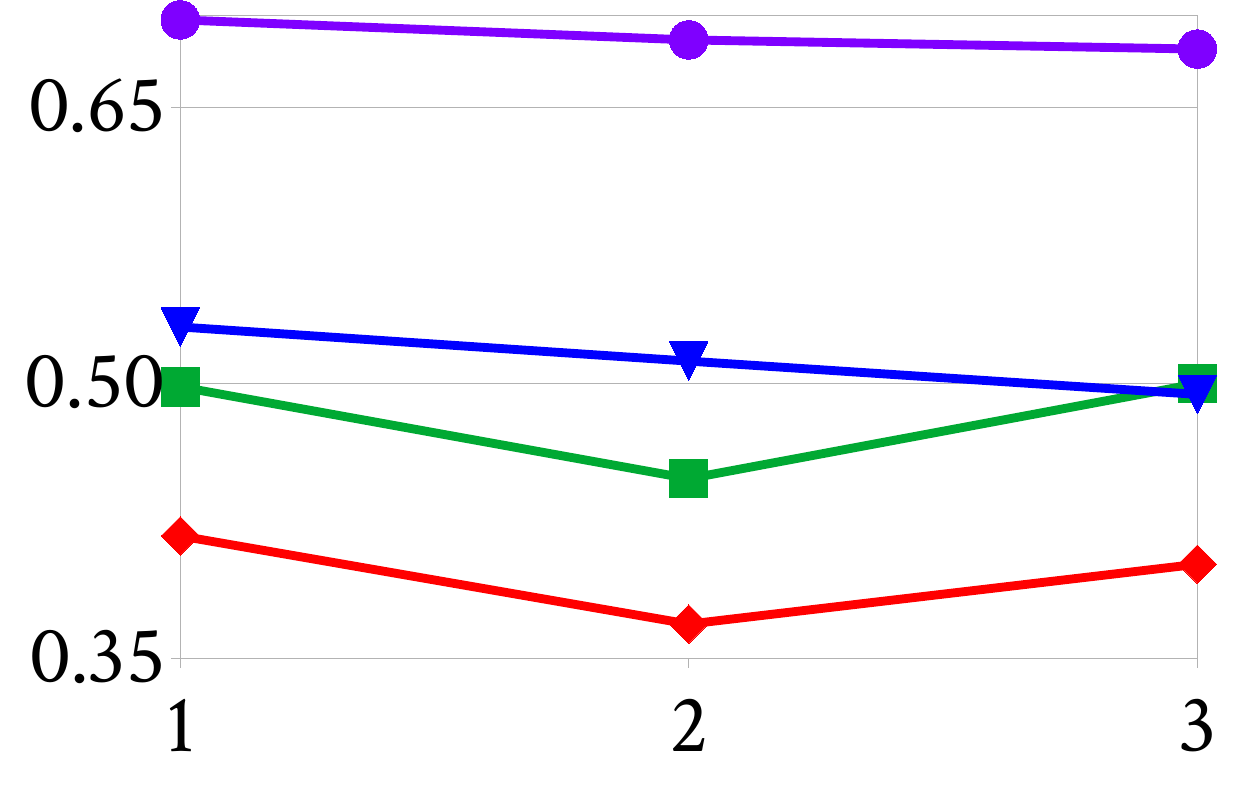}
		\subcaption{Neutral F1-score}
	\end{subfigure}
	\hfill
	\begin{subfigure}[t]{0.07\textwidth}
		\includegraphics[width=\textwidth]{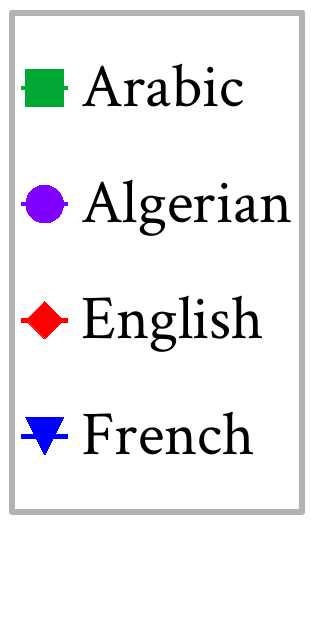}
		%		\subcaption{Legend}
	\end{subfigure}
	
	\caption{Effect of Transformer depth ($N$) on sentiment analysis in Arabic, Algerian dialect, English, and French. 
		Results were obtained by training a BiGRU--dense layer decoder on frozen chDzDT embeddings.}
	
	\label{fig:ablation-N-sa}
\end{figure}

\paragraph{Sentiment analysis.}
Figure~\ref{fig:ablation-N-sa} shows the effect of Transformer depth ($N$) on sentiment analysis. 
Similar to PoS tagging, performance exhibits only minor and inconsistent variations across $N=1$ to $N=3$.
Overall accuracy remains broadly stable, with fluctuations likely due to optimization noise. 
The Algerian dialect achieves the highest scores across classes, suggesting that its training data provides relatively clear sentiment cues.
English performs weaker overall, with only slight gains from $N=1$ to $N=3$. 
Arabic and French occupy an intermediate range but show no consistent trend with increasing depth. 
At the class level, positive and negative F1 scores fluctuate irregularly, while neutral classification is more stable but does not improve meaningfully. 
As DzDT embeddings are frozen and morphology-based, sentiment classification relies more on distributional and lexical patterns captured by the BiGRU sentence encoder than on deeper character-level abstraction
Overall, increasing Transformer depth in the embedding model yields little additional benefit for sentiment, as the downstream BiGRU--dense layer decoder can already leverage the morphological embeddings effectively.

%\begin{tabular}{|c|c|c|c|c|c|c|c|c|c|c|c|c|c|c|c|c|c|c|}
%	\hline
%	&  &  & Overall &  &  &  & Positive &  &  &  & Negative &  &  &  & Neutral &  &  &  \\
%	\hline
%	N &  &  & Arabic & Algerian & English & French & Arabic & Algerian & English & French & Arabic & Algerian & English & French & Arabic & Algerian & English & French \\
%	\hline
%	1 &  &  & 0.4270 & 0.6271 & 0.3322 & 0.4345 & 0.1618 & 0.6839 & 0.3147 & 0.3544 & 0.4583 & 0.4586 & 0.1908 & 0.3912 & 0.4980 & 0.6978 & 0.4168 & 0.5306 \\
%	\hline
%	2 &  &  & 0.4192 & 0.6133 & 0.3146 & 0.4241 & 0.2100 & 0.6751 & 0.3239 & 0.3242 & 0.4894 & 0.4165 & 0.1935 & 0.4175 & 0.4480 & 0.6869 & 0.3688 & 0.5120 \\
%	\hline
%	3 &  &  & 0.4252 & 0.6058 & 0.3273 & 0.4149 & 0.1416 & 0.6509 & 0.3296 & 0.3352 & 0.4615 & 0.4714 & 0.1841 & 0.3985 & 0.4997 & 0.6819 & 0.4014 & 0.4939 \\
%	\hline
%\end{tabular}

\subsubsection{Number of attention heads}

Multi-head attention allows the model to capture multiple relational patterns in parallel. 
\textit{Does increasing the number of attention heads ($H$) improve embedding quality and task performance, or mainly add computational overhead?} 
While additional heads may broaden the range of captured dependencies, they also fragment the embedding space into smaller subspaces.  
To test this factor, we fixed $N=2$ and $d=16$ and varied $H \in \{1, 2, 4\}$ (since $d$ is not divisible by 3, we used 4 heads).

\paragraph{Model efficiency.}
Unlike Transformer depth, the number of attention heads does not change the total parameter count, since the projection matrices always map from $d$ to $d$ regardless of the number of heads. 
Training times across different $H$ values were generally similar, with observed variations attributable to system-level factors rather than architectural differences. 
Accordingly, we do not include plots for this setting.

\begin{figure}[!htp]
	\centering\small
	
	\begin{subfigure}[t]{0.22\textwidth}
		\includegraphics[width=\textwidth]{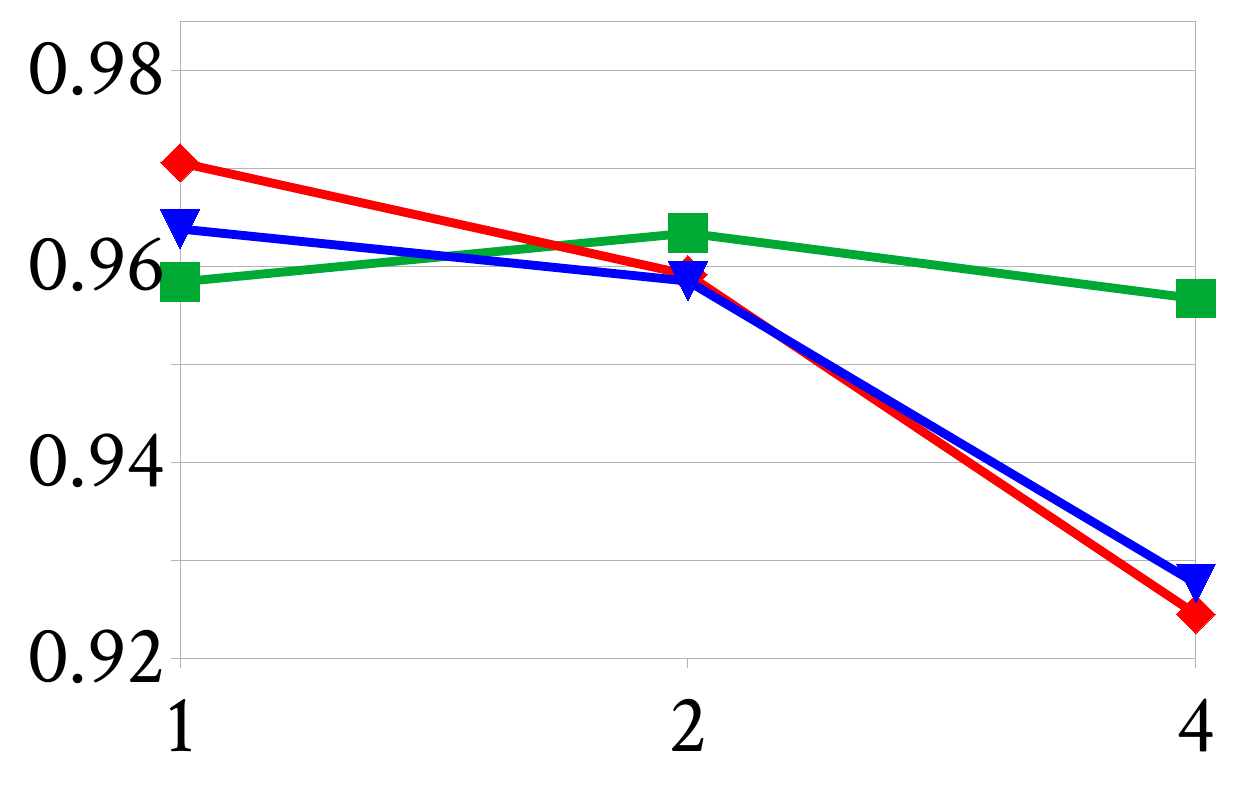}
		\subcaption{Deriv--Avg}
	\end{subfigure}
	\begin{subfigure}[t]{0.22\textwidth}
		\includegraphics[width=\textwidth]{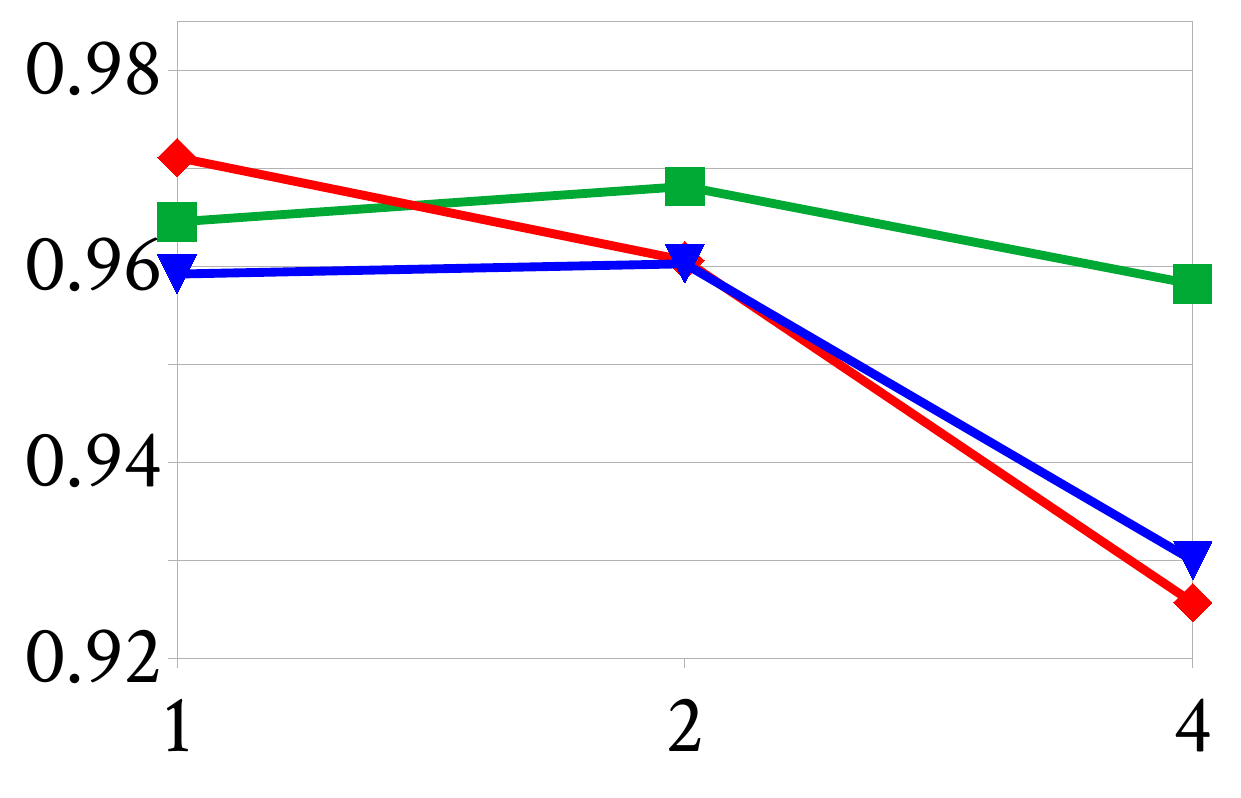}
		\subcaption{Deriv--Min}
	\end{subfigure}
	\begin{subfigure}[t]{0.22\textwidth}
		\includegraphics[width=\textwidth]{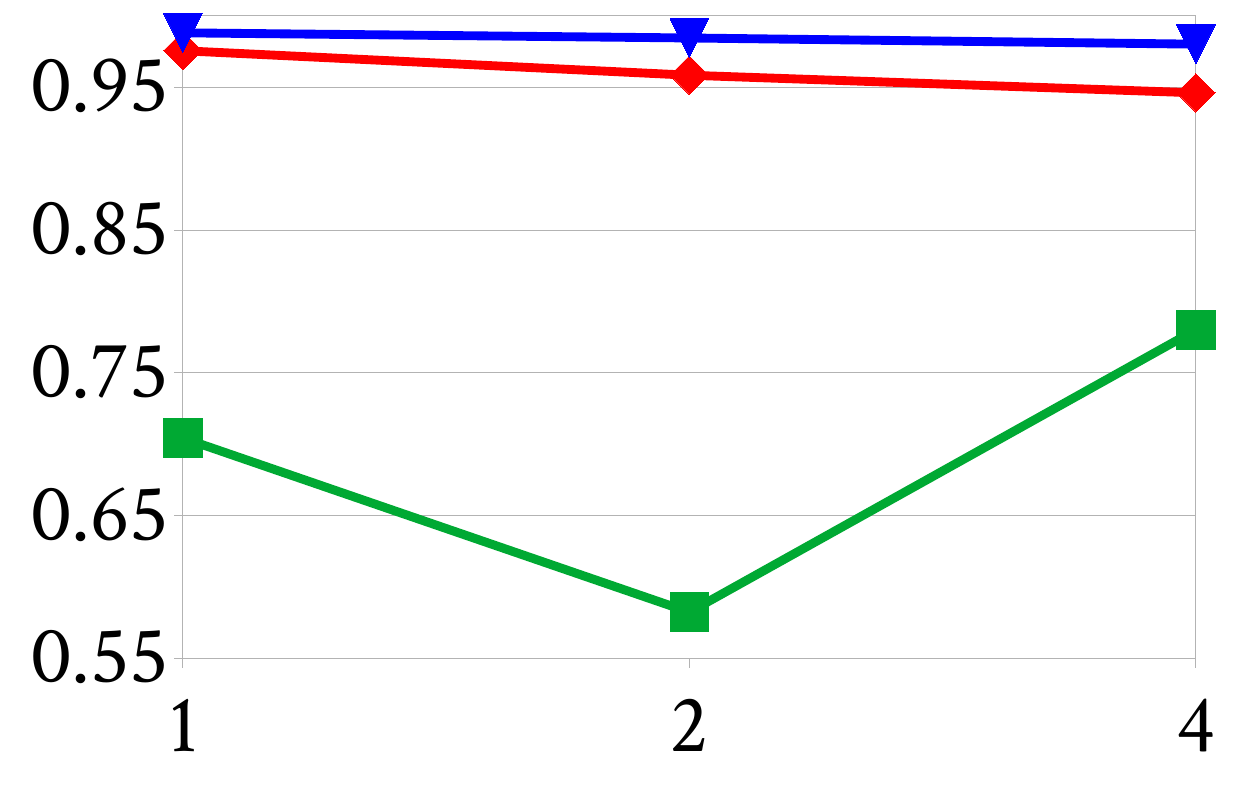}
		\subcaption{Infl--Avg}
	\end{subfigure}
	\begin{subfigure}[t]{0.22\textwidth}
		\includegraphics[width=\textwidth]{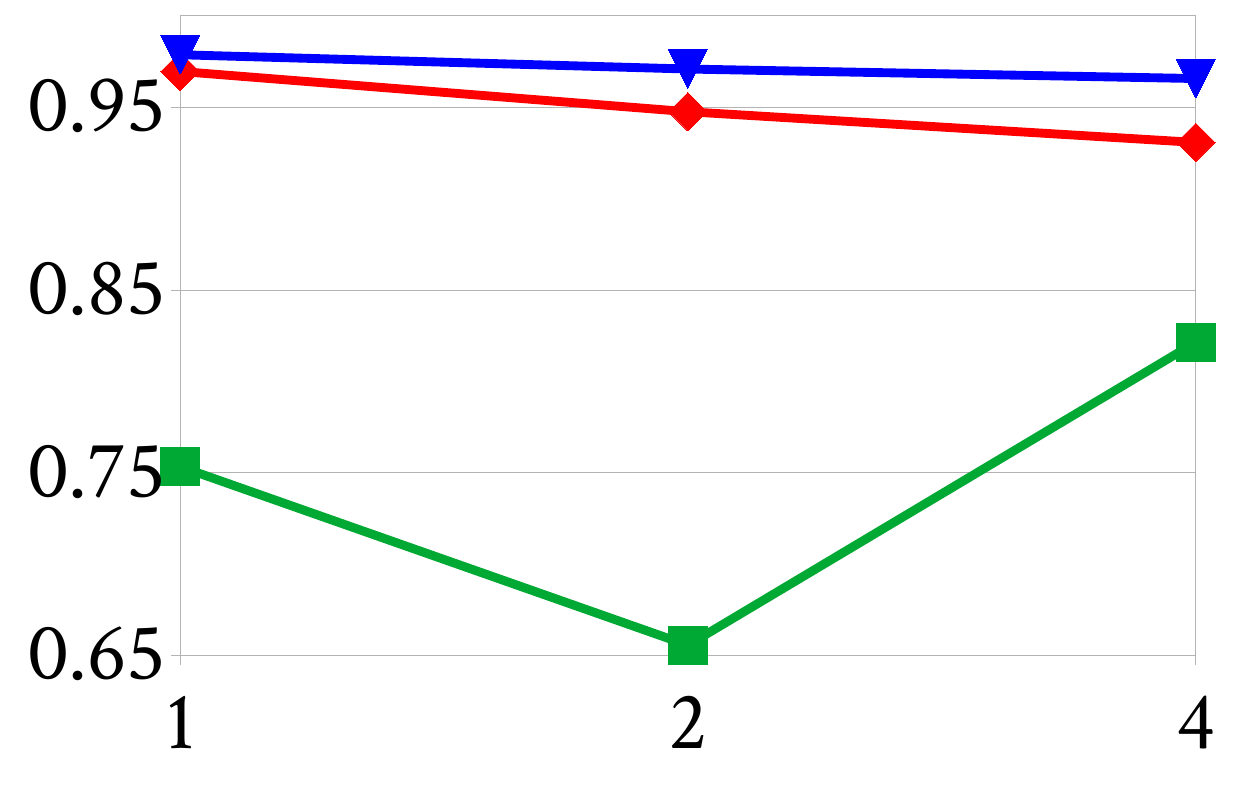}
		\subcaption{Infl--Min}
	\end{subfigure}
	\hfill
	\begin{subfigure}[t]{0.07\textwidth}
		\includegraphics[width=\textwidth]{IMG/abl_morph_legend.pdf}
		%		\subcaption{Legend}
	\end{subfigure}
	
	\caption{Effect of the number of attention heads ($H$) on morphological consistency in Arabic, French, and English.  
		Plots show derivational (Deriv) and inflectional (Infl) morphology on the ``Avg'' and ``Min'' datasets, using average cosine similarity between words and their root.}

	\label{fig:ablation-H-morph-consist}
\end{figure}

\paragraph{Morphological consistency.}
Figure~\ref{fig:ablation-H-morph-consist} shows that increasing the number of attention heads ($H$) affects derivational and inflectional morphology in distinct ways. 
For derivation, performance generally deteriorates with larger $H$, especially in English and French, where clustering quality declines sharply. 
Arabic appears more stable, but this is partly due to dataset design: derivational clusters are defined by both templatic processes and semantic categories, making the measure less sensitive to changes in $H$. 
This illustrates the difficulty of evaluating derivation in languages where semantics and morphology are intertwined. 
In contrast, inflectional morphology shows modest declines in English and French, while Arabic exhibits a non-monotonic U-shaped pattern, with a sharp drop at $H=2$ followed by a strong recovery at $H=4$. 
This behavior may stem from the interaction of templatic and affixal processes, which require multiple attention heads to be captured effectively. 
Overall, the findings suggest that, for the tested configurations, higher $H$ tends to reduce derivational consistency, while inflectional patterns depend on language-specific structures. 
Multi-head attention thus enhances representational diversity but does not uniformly support morphological regularities.

%\begin{tabular}{|c|c|c|c|c|c|c|c|c|c|c|c|c|c|c|c|}
%	\hline
%	& deriv & deriv & deriv &  & deriv & deriv & deriv &  & infl & infl & infl &  & infl & infl & infl \\
%	\hline
%	& avg & avg & avg &  & min & min & min &  & avg & avg & avg &  & min & min & min \\
%	\hline
%	H & Arabic & English & French &  & Arabic & English & French &  & Arabic & English & French &  & Arabic & English & French \\
%	\hline
%	1 & 0.9584 & 0.9706 & 0.9638 &  & 0.9645 & 0.9711 & 0.9592 &  & 0.7044 & 0.9756 & 0.9881 &  & 0.7534 & 0.9696 & 0.9787 \\
%	\hline
%	2 & 0.9634 & 0.9592 & 0.9585 &  & 0.9682 & 0.9606 & 0.9603 &  & 0.5825 & 0.9584 & 0.9845 &  & 0.6557 & 0.9475 & 0.9709 \\
%	\hline
%	4 & 0.9567 & 0.9245 & 0.9277 &  & 0.9582 & 0.9257 & 0.9300 &  & 0.7802 & 0.9461 & 0.9802 &  & 0.8214 & 0.9307 & 0.9657 \\
%	\hline
%\end{tabular}

\begin{figure}[!htp]
	\centering\small
	
	\begin{subfigure}[t]{0.27\textwidth}
		\includegraphics[width=\textwidth]{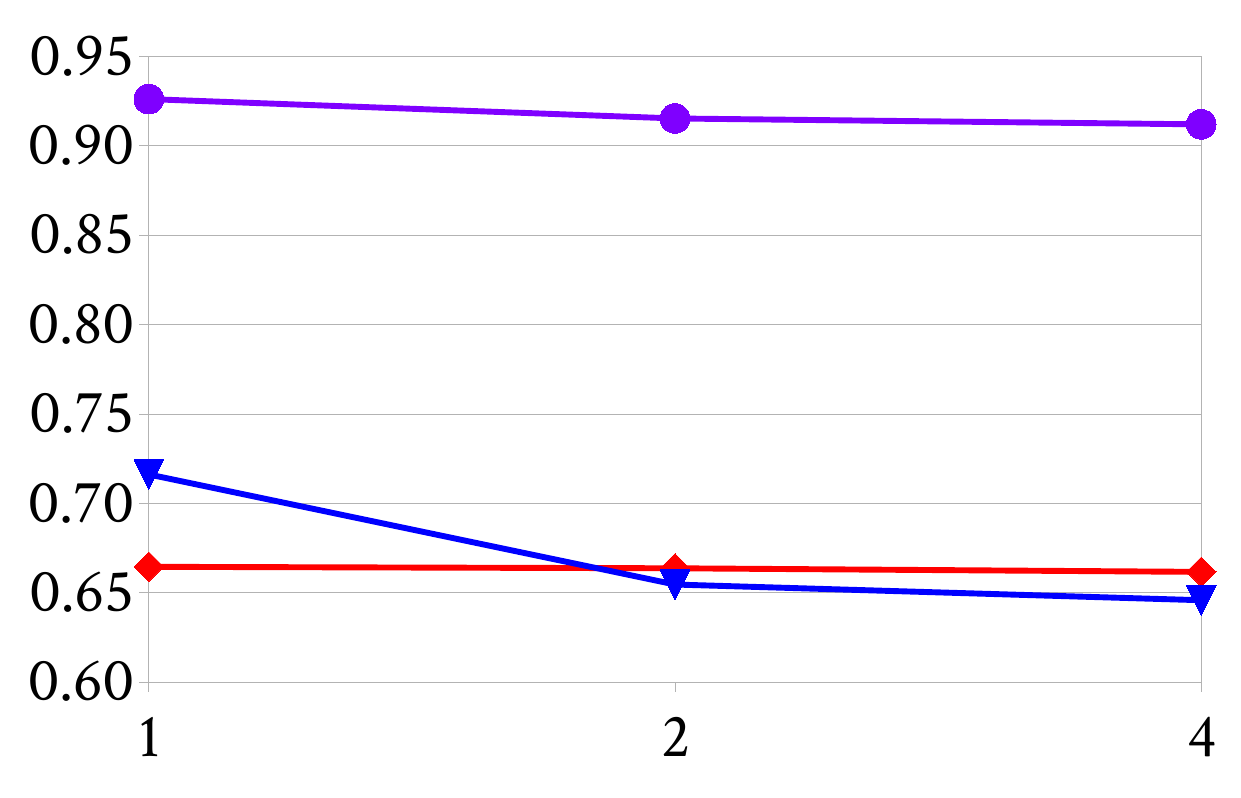}
		\subcaption{Obfus. clusters}
	\end{subfigure}
	\begin{subfigure}[t]{0.27\textwidth}
		\includegraphics[width=\textwidth]{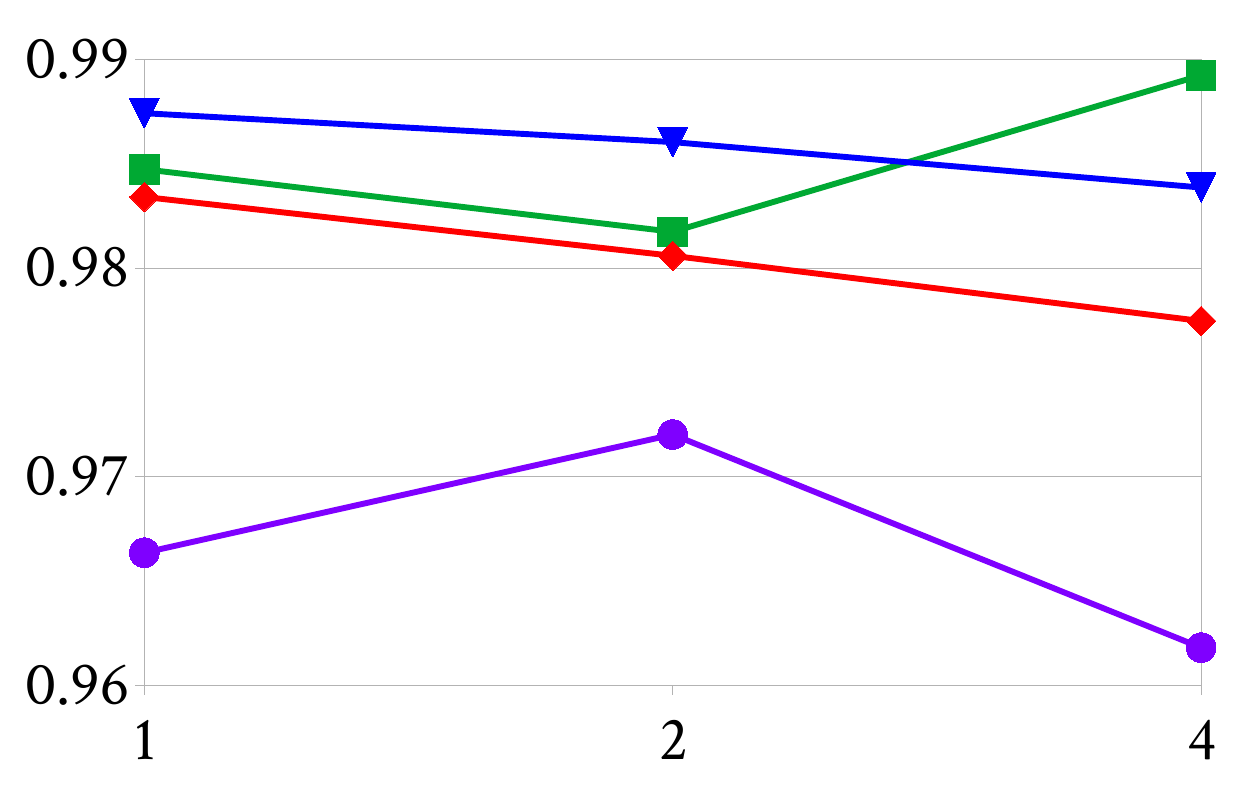}
		\subcaption{Obfus. tuples-*}
	\end{subfigure}
	\begin{subfigure}[t]{0.27\textwidth}
		\includegraphics[width=\textwidth]{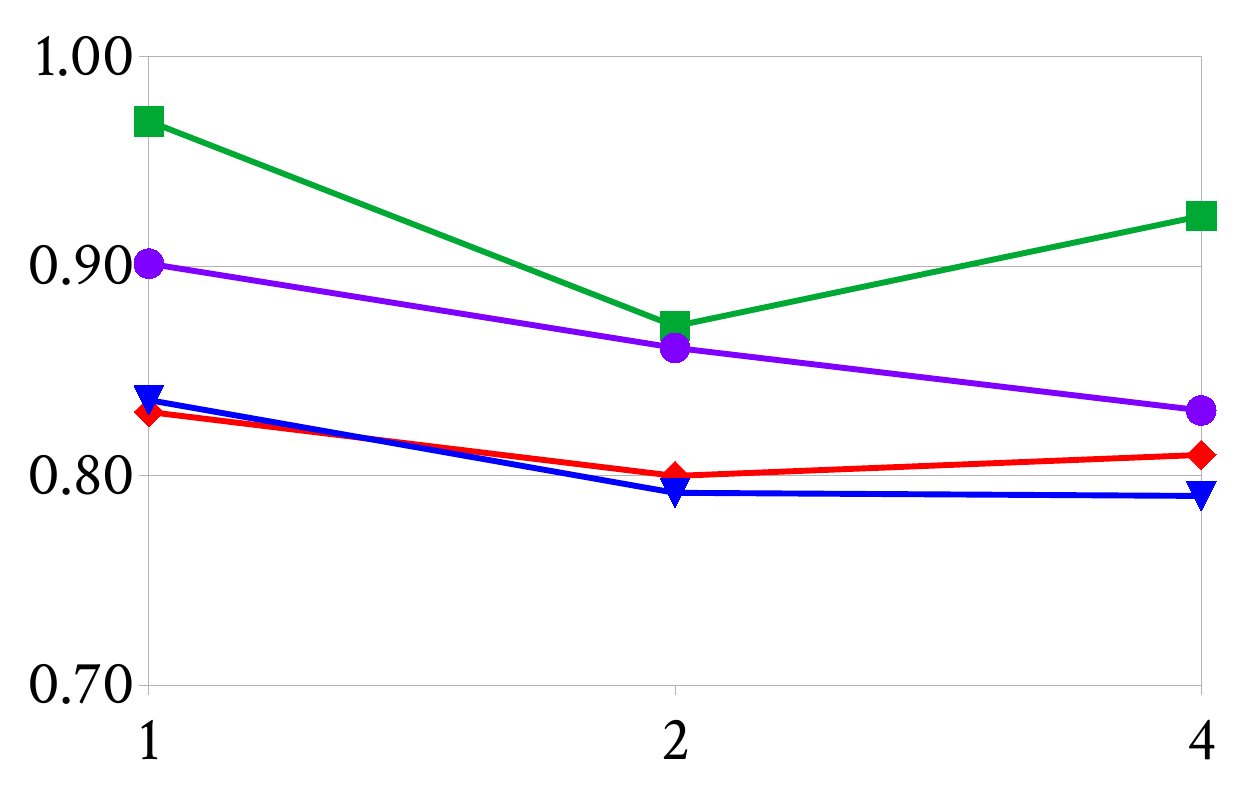}
		\subcaption{Obfus. tuples-c}
	\end{subfigure}
	\hfill
	\begin{subfigure}[t]{0.07\textwidth}
		\includegraphics[width=\textwidth]{IMG/abl_noise_legend.pdf}
%		\subcaption{Legend}
	\end{subfigure}
	
	\caption{Effect of the number of attention heads ($H$) on robustness to orthographic noise in Arabic, Arabizi, French, and English. 
		Plots show: (a) cluster-based obfuscations, (b) tuple obfuscations with \texttt{*} substitutions, and 
		(c) tuple obfuscations with visually or phonetically similar substitutions. 
		Scores are reported as Average Cosine Similarity (ACS). 
		Note that Arabic is absent from (a), as it is predominantly written in standard form and does not exhibit such obfuscations.}
	
	\label{fig:ablation-H-morph-noise}
\end{figure}

\paragraph{Robustness to orthographic noise.}
Figure~\ref{fig:ablation-H-morph-noise} shows the effect of the number of attention heads ($H$) on robustness to orthographic noise across Arabic, Arabizi, French, and English. 
Overall, $H$ has a weaker influence than Transformer depth, but consistent patterns emerge. 
Cluster-based obfuscations remain stable in Arabizi, with ACS values above 0.91.
English and French show lower and more variable scores, reflecting their standardized pre-training corpora compared to Arabizi's noisier data. 
Tuple obfuscations with simple ``*'' substitutions are robust across all settings, with ACS above 0.96, indicating that minimal masking is largely unaffected by $H$. 
More complex tuple obfuscations using phonetically or visually similar substitutions are more challenging.
Arabic shows resilience, especially at higher $H$, likely because its templatic morphology provides redundant cues that buffer against character-level perturbations. 
Arabizi also remains stable, whereas English and French degrade progressively, reflecting their stronger dependence on fixed orthography. 
In sum, although attention head count has less impact than depth, higher $H$ can modestly improve robustness in morphologically rich or orthographically flexible languages. 
Rigid systems, by contrast, remain more sensitive to obfuscation.

%\begin{tabular}{|c|c|c|c|c|c|c|c|c|c|c|c|c|}
%	\hline
%	&  & ACS & ACS & ACS & ACS1* & ACS1* & ACS1* & ACS1* & ACS1c & ACS1c & ACS1c & ACS1c \\
%	\hline
%	H &  & Arabizi & English & French & Arabic & Arabizi & English & French & Arabic & Arabizi & English & French \\
%	\hline
%	1 &  & 0.9262 & 0.6646 & 0.7162 & 0.9847 & 0.9664 & 0.9834 & 0.9874 & 0.9691 & 0.9012 & 0.8304 & 0.8361 \\
%	\hline
%	2 &  & 0.9153 & 0.6637 & 0.6547 & 0.9817 & 0.9720 & 0.9806 & 0.9860 & 0.8714 & 0.8610 & 0.7999 & 0.7918 \\
%	\hline
%	4 &  & 0.9120 & 0.6617 & 0.6459 & 0.9892 & 0.9618 & 0.9775 & 0.9839 & 0.9239 & 0.8311 & 0.8099 & 0.7903 \\
%	\hline
%\end{tabular}

\begin{figure}[!htp]
	\centering\small
	
	\begin{subfigure}[t]{0.24\textwidth}
		\includegraphics[width=\textwidth]{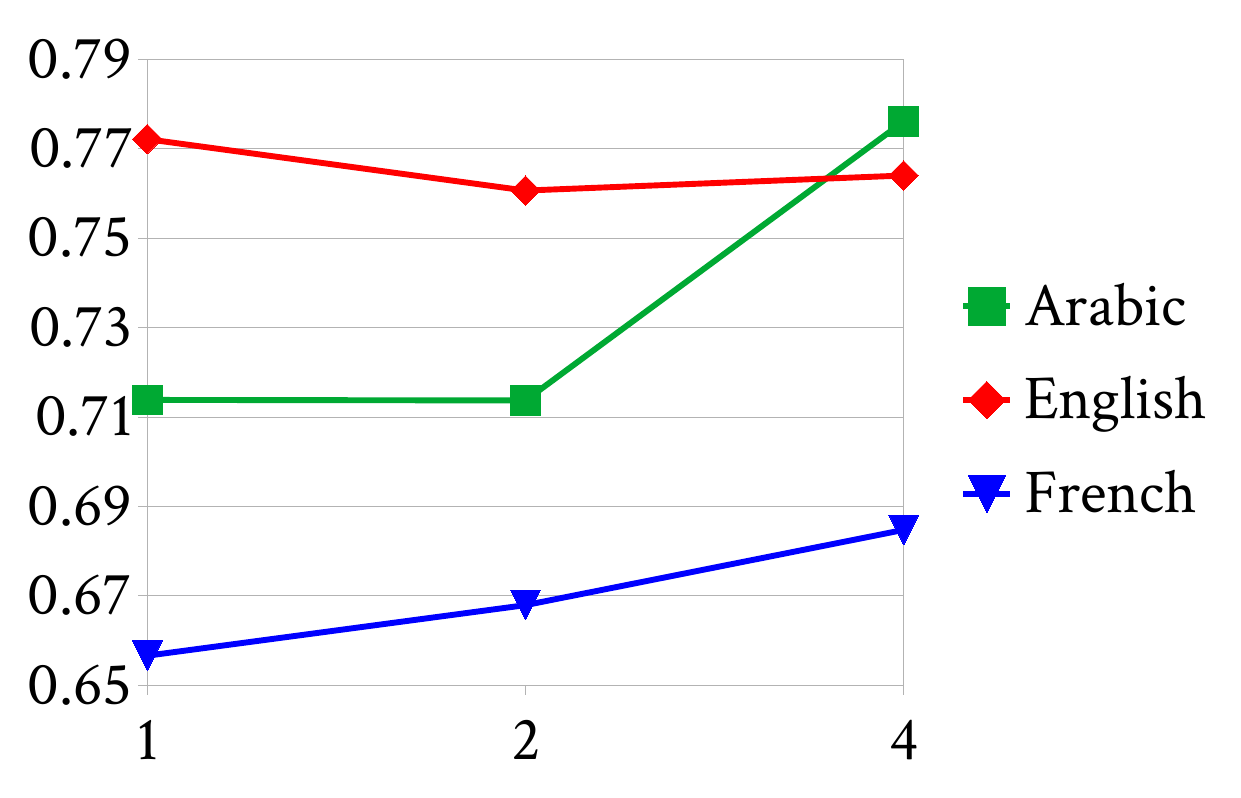}
		\subcaption{Overall Accuracy}
	\end{subfigure}
	\begin{subfigure}[t]{0.24\textwidth}
		\includegraphics[width=\textwidth]{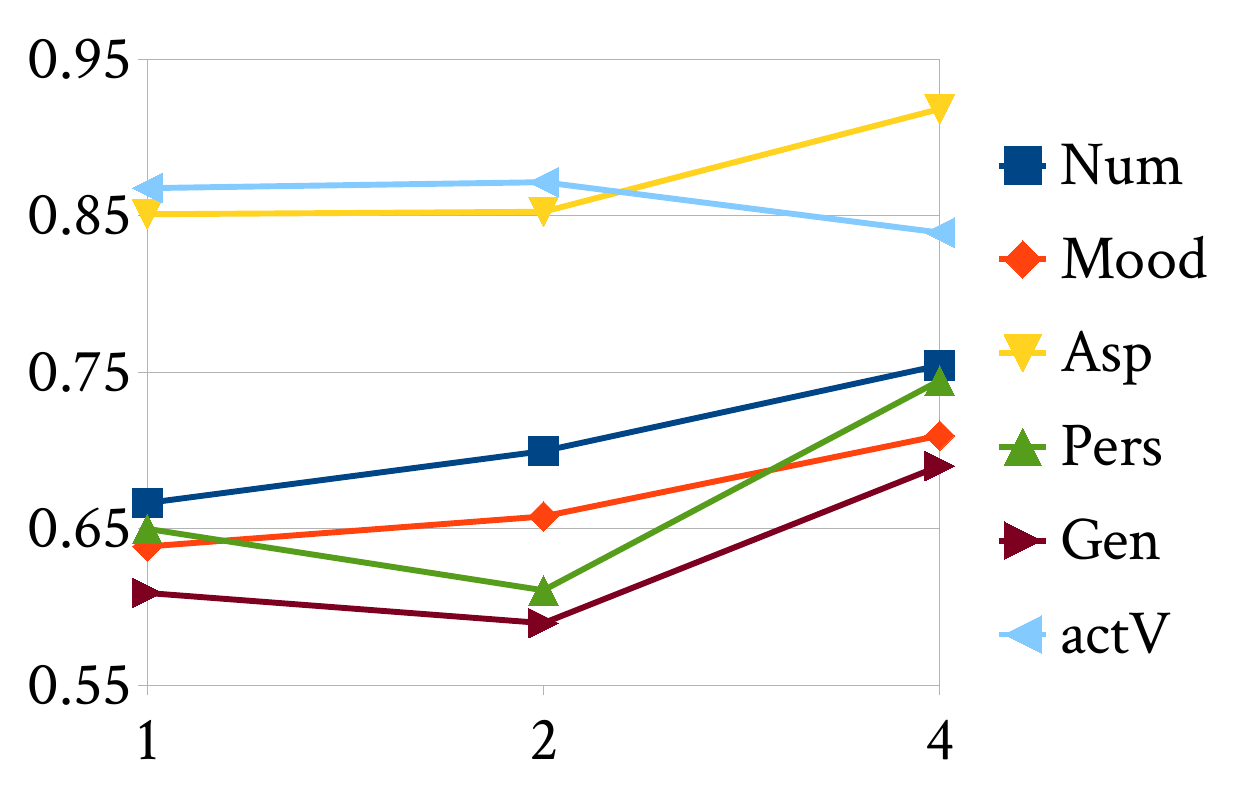}
		\subcaption{Accuracy of Arabic tags}
	\end{subfigure}
	\begin{subfigure}[t]{0.24\textwidth}
		\includegraphics[width=\textwidth]{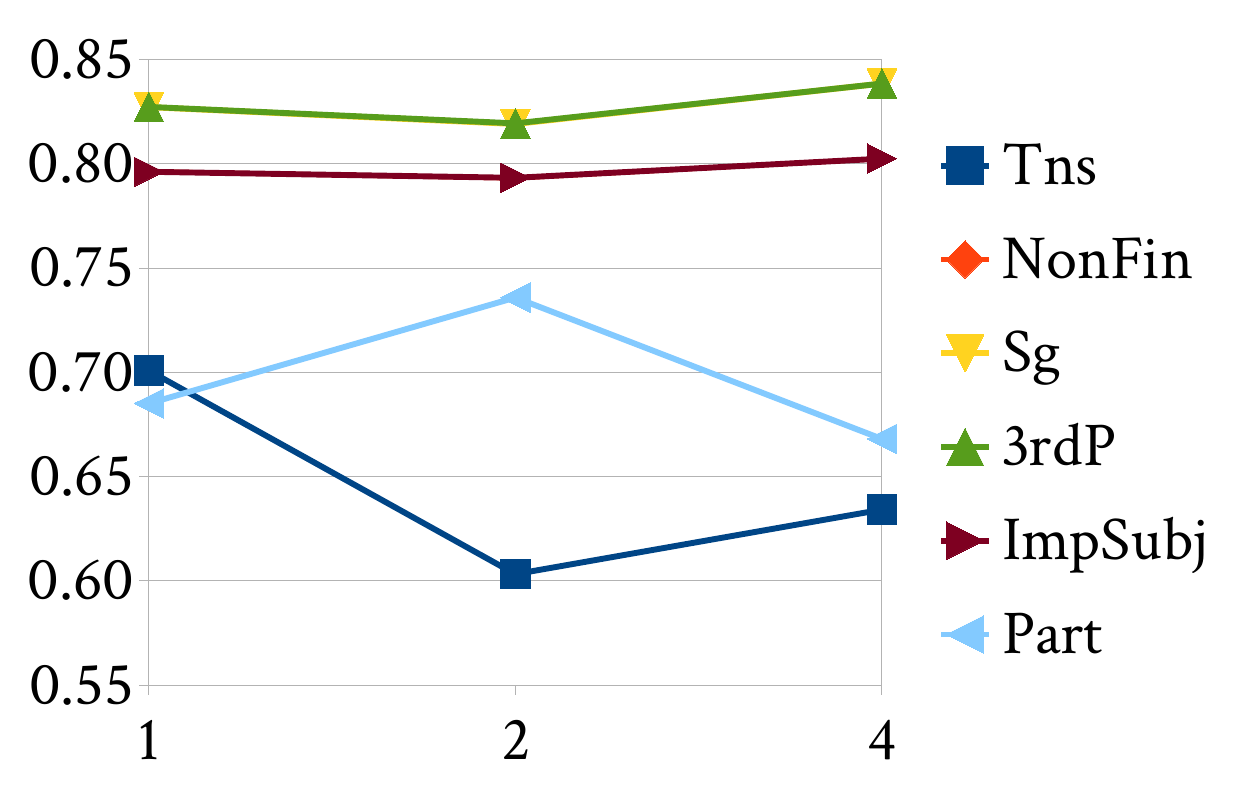}
		\subcaption{Accuracy of English tags}
	\end{subfigure}
	\begin{subfigure}[t]{0.24\textwidth}
		\includegraphics[width=\textwidth]{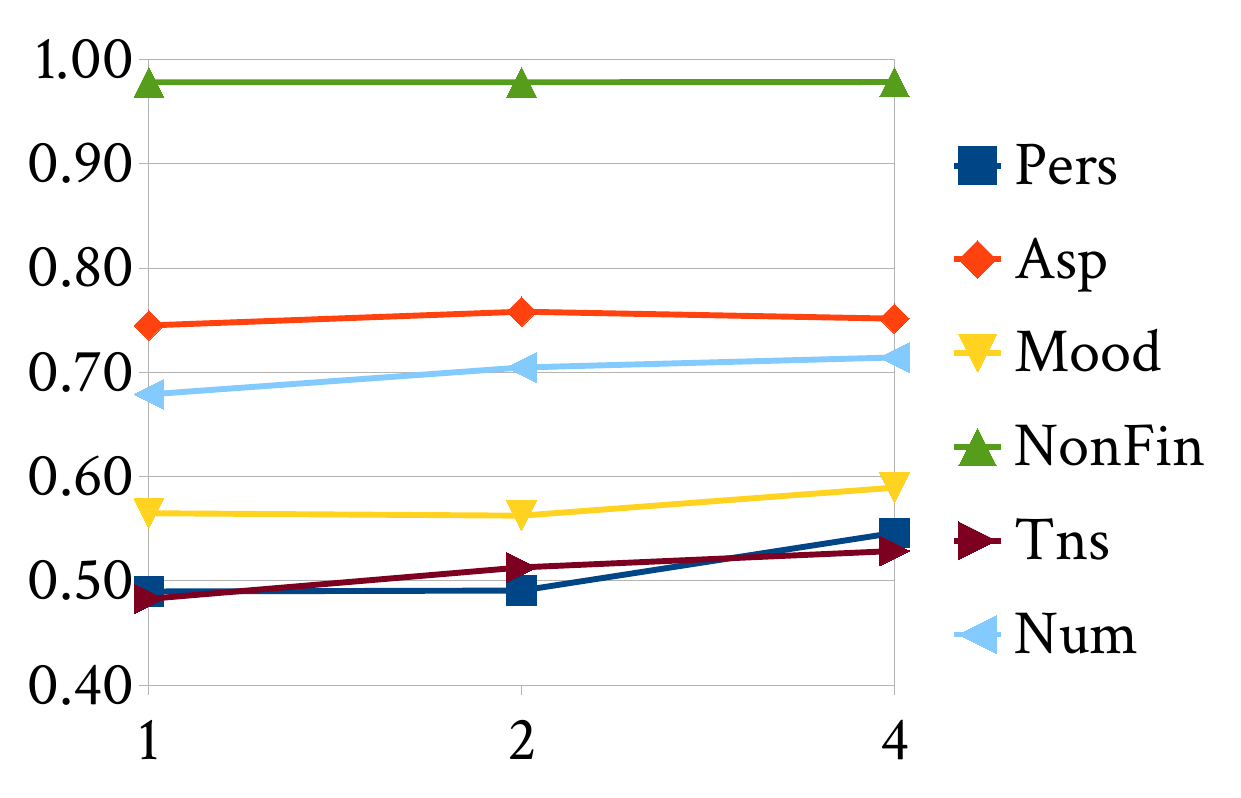}
		\subcaption{Accuracy of French tags}
	\end{subfigure}
		
	\caption{Effect of the number of attention heads ($H$) on morphological tagging in Arabic, English, and French. 
		(a) Overall accuracy across the three languages; (b) Arabic tags; (c) English tags; (d) French tags. 
		In English, the curves for \textit{Sg} and \textit{3rdP} overlap, as do those for \textit{NonFin} and \textit{ImpSubj}.}
	
	\label{fig:ablation-H-morph-tag}
\end{figure}

\paragraph{Morphological tagging.}
Figure~\ref{fig:ablation-H-morph-tag} shows that the number of attention heads affects morphological tagging unevenly across languages. 
With a single head, the model achieves a reasonable baseline, although accuracy is lower for French, reflecting its more complex conjugation system. 
Increasing to two heads yields only marginal improvements, with English showing a slight decline. 
By contrast, four heads provide the best overall accuracy, with notable gains for Arabic, where features such as aspect and mood benefit from richer attention patterns. 
English remains relatively stable, and overlapping curves for \textit{Sg}/\textit{3rdP} and \textit{NonFin}/\textit{ImpSubj} again indicate strong feature correlations. 
Overall, these results suggest that while additional heads enhance the model's ability to capture diverse morphological cues, the improvements are not uniform across languages.

%\begin{tabular}{|c|c|c|c|c|c|c|c|c|c|c|c|c|c|c|c|c|c|c|c|c|c|c|}
%	\hline
%	H &  & Overall &  &  & Arabic &  &  &  &  &  & English &  &  &  &  &  & French &  &  &  &  &  \\
%	\hline
%	H &  & Arabic & English & French & Num & Mood & Asp & Pers & Gen & actV & Tns & NonFin & Sg & 3rdP & ImpSubj & Part & Pers & Asp & Mood & NonFin & Tns & Num \\
%	\hline
%	1 &  & 0.7139 & 0.7722 & 0.6566 & 0.6665 & 0.6387 & 0.8511 & 0.6500 & 0.6090 & 0.8678 & 0.7009 & 0.7963 & 0.8272 & 0.8273 & 0.7962 & 0.6851 & 0.4899 & 0.7450 & 0.5649 & 0.9782 & 0.4828 & 0.6789 \\
%	\hline
%	2 &  & 0.7137 & 0.7607 & 0.6679 & 0.6997 & 0.6579 & 0.8527 & 0.6107 & 0.5897 & 0.8716 & 0.6033 & 0.7931 & 0.8191 & 0.8194 & 0.7933 & 0.7359 & 0.4908 & 0.7582 & 0.5626 & 0.9782 & 0.5130 & 0.7047 \\
%	\hline
%	4 &  & 0.7761 & 0.7641 & 0.6848 & 0.7545 & 0.7096 & 0.9184 & 0.7448 & 0.6901 & 0.8394 & 0.6342 & 0.8025 & 0.8385 & 0.8385 & 0.8025 & 0.6681 & 0.5461 & 0.7515 & 0.5895 & 0.9784 & 0.5287 & 0.7143 \\
%	\hline
%\end{tabular}

\begin{figure}[!htp]
	\centering\small
	
	\begin{subfigure}[t]{0.22\textwidth}
		\includegraphics[width=\textwidth]{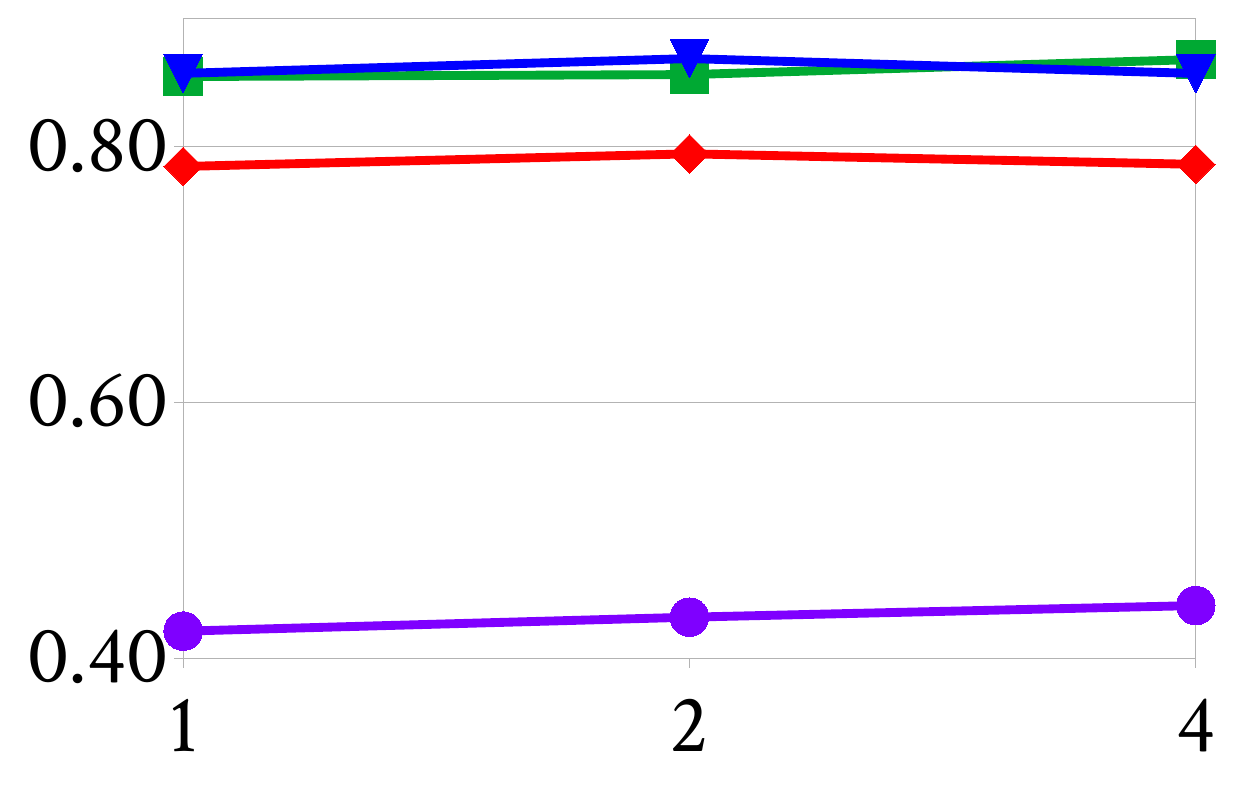}
		\subcaption{Overall accuracy}
	\end{subfigure}
	\begin{subfigure}[t]{0.22\textwidth}
		\includegraphics[width=\textwidth]{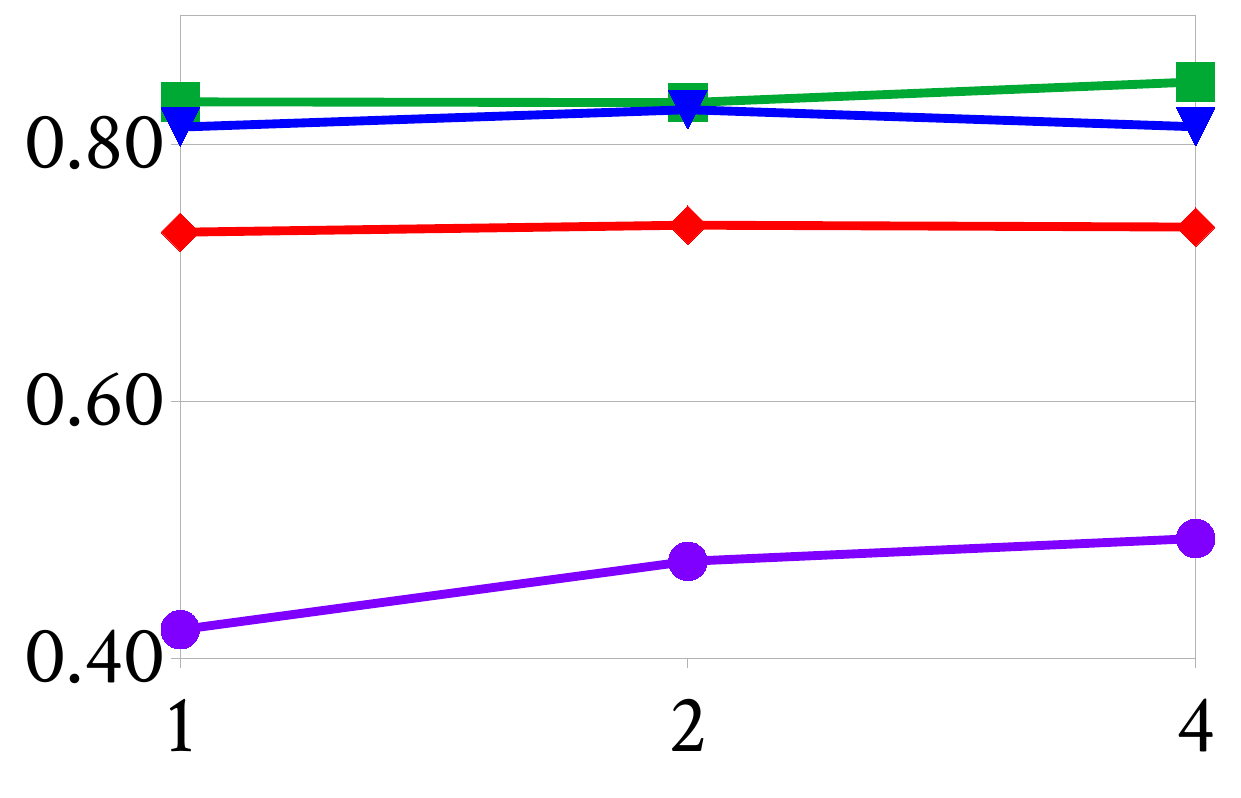}
		\subcaption{Noun tag F1-score}
	\end{subfigure}
	\begin{subfigure}[t]{0.22\textwidth}
		\includegraphics[width=\textwidth]{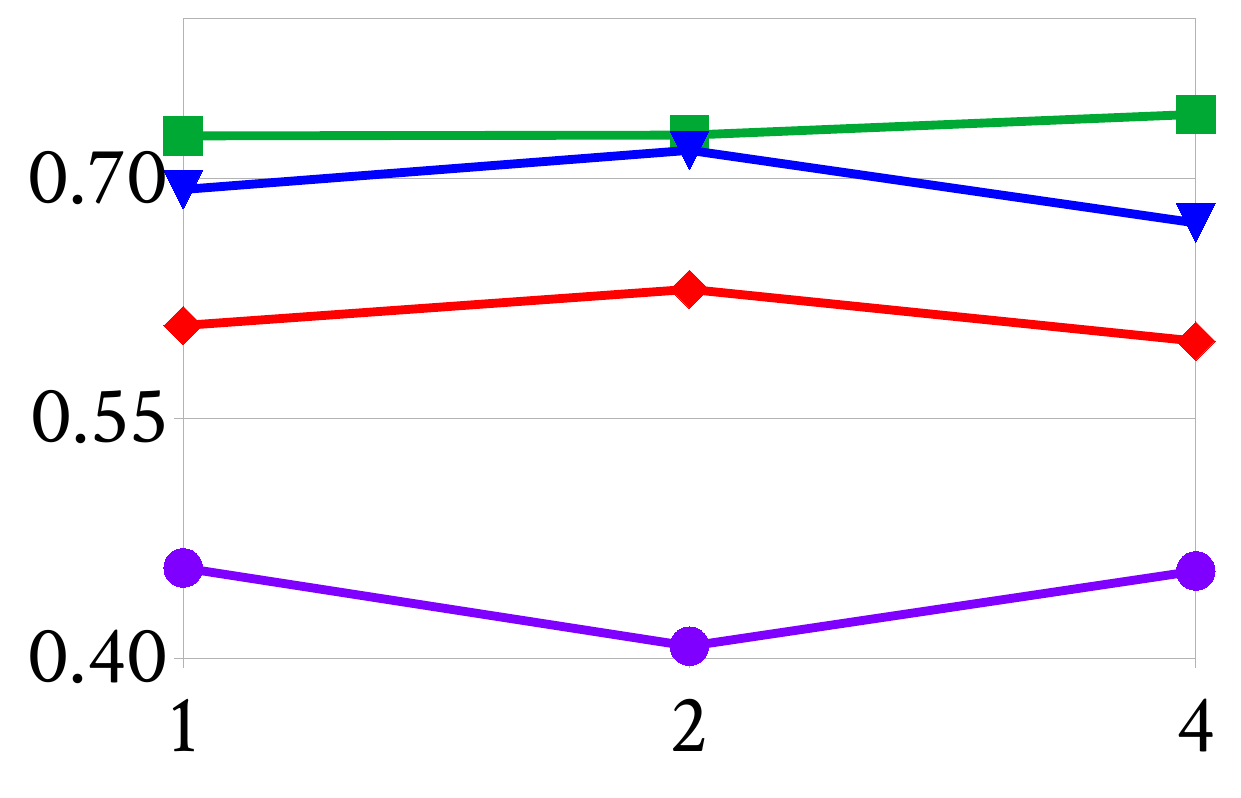}
		\subcaption{Verb tag F1-score}
	\end{subfigure}
	\begin{subfigure}[t]{0.22\textwidth}
		\includegraphics[width=\textwidth]{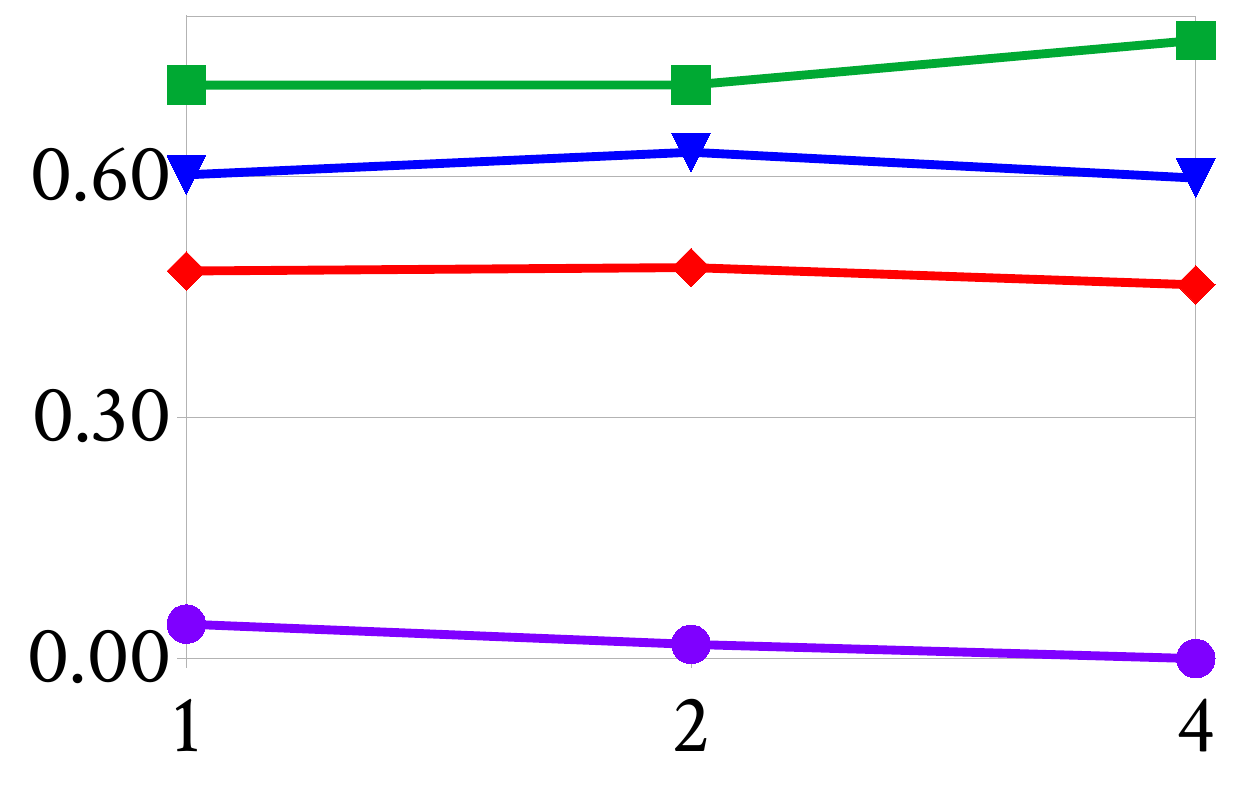}
		\subcaption{Adjective tag F1-score}
	\end{subfigure}
	\hfill
	\begin{subfigure}[t]{0.07\textwidth}
		\includegraphics[width=\textwidth]{IMG/abl_pos_legend.pdf}
		%		\subcaption{Legend}
	\end{subfigure}
	
	\caption{Effect of the number of attention heads ($H$) on PoS tagging in Arabic, Arabizi, English, and French. Results were obtained by training a BiGRU decoder on frozen chDzDT embeddings.}
	
	\label{fig:ablation-H-pos-tag}
\end{figure}

\paragraph{Part-of-speech tagging.}
Figure~\ref{fig:ablation-H-pos-tag} shows the effect of varying the number of attention heads on PoS tagging performance. 
Overall accuracy remains largely stable across $H=1$, $H=2$, and $H=4$, with differences generally within one percentage point. 
At the feature level, nouns and verbs show modest gains with additional heads, particularly in Arabic and French, while adjectives remain unstable across all settings. 
Arabizi consistently underperforms, with adjective tagging performing poorly regardless of $H$. 
English appears almost insensitive to the number of heads, showing only minor fluctuations.
Taken together, similar to the effect of $N$, these results indicate that PoS tagging is comparatively insensitive to both architectural dimensions: increasing $H$ yields only marginal improvements. 
This contrasts with morphologically richer tasks, where depth produced clearer gains. 
In practice, a small number of heads is sufficient to provide embeddings that, when processed by the BiGRU decoder, capture the distinctions needed for PoS tagging.

%\begin{tabular}{|c|c|c|c|c|c|c|c|c|c|c|c|c|c|c|c|c|c|}
%	\hline
%	&  & Overall &  &  &  & Noun &  &  &  & Verb &  &  &  & Adjective &  &  &  \\
%	\hline
%	H &  & Arabic & Arabizi & English & French & Arabic & Arabizi & English & French & Arabic & Arabizi & English & French & Arabic & Arabizi & English & French \\
%	\hline
%	1 &  & 0.8549 & 0.4215 & 0.7846 & 0.8574 & 0.8331 & 0.4225 & 0.7317 & 0.8135 & 0.7268 & 0.4567 & 0.6083 & 0.6933 & 0.7139 & 0.0429 & 0.4825 & 0.6022 \\
%	\hline
%	2 &  & 0.8564 & 0.4324 & 0.7945 & 0.8689 & 0.8325 & 0.4757 & 0.7372 & 0.8270 & 0.7273 & 0.4078 & 0.6309 & 0.7178 & 0.7140 & 0.0177 & 0.4868 & 0.6303 \\
%	\hline
%	4 &  & 0.8682 & 0.4415 & 0.7863 & 0.8573 & 0.8486 & 0.4934 & 0.7357 & 0.8137 & 0.7403 & 0.4548 & 0.5985 & 0.6724 & 0.7696 & 0.0000 & 0.4652 & 0.5981 \\
%	\hline
%\end{tabular}

\begin{figure}[!htp]
	\centering\small
	
	\begin{subfigure}[t]{0.22\textwidth}
		\includegraphics[width=\textwidth]{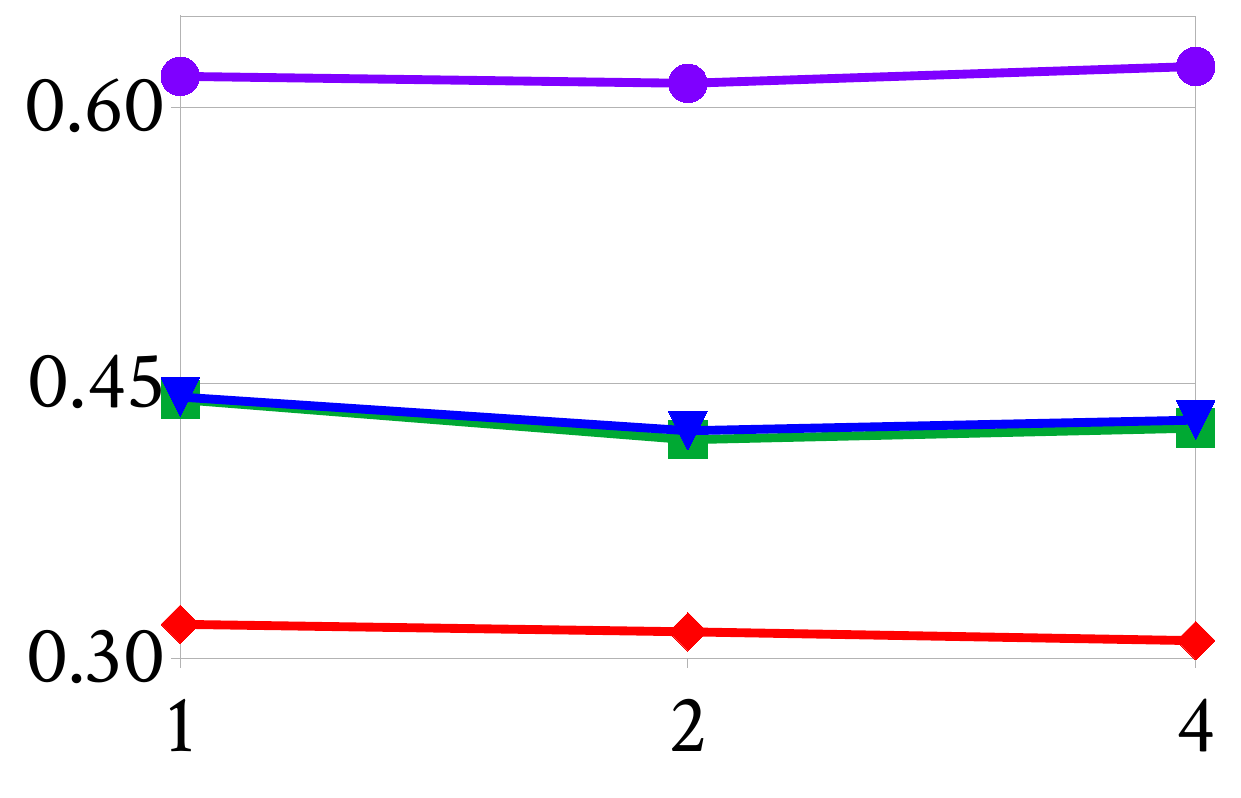}
		\subcaption{Overall Accuracy}
	\end{subfigure}
	\begin{subfigure}[t]{0.22\textwidth}
		\includegraphics[width=\textwidth]{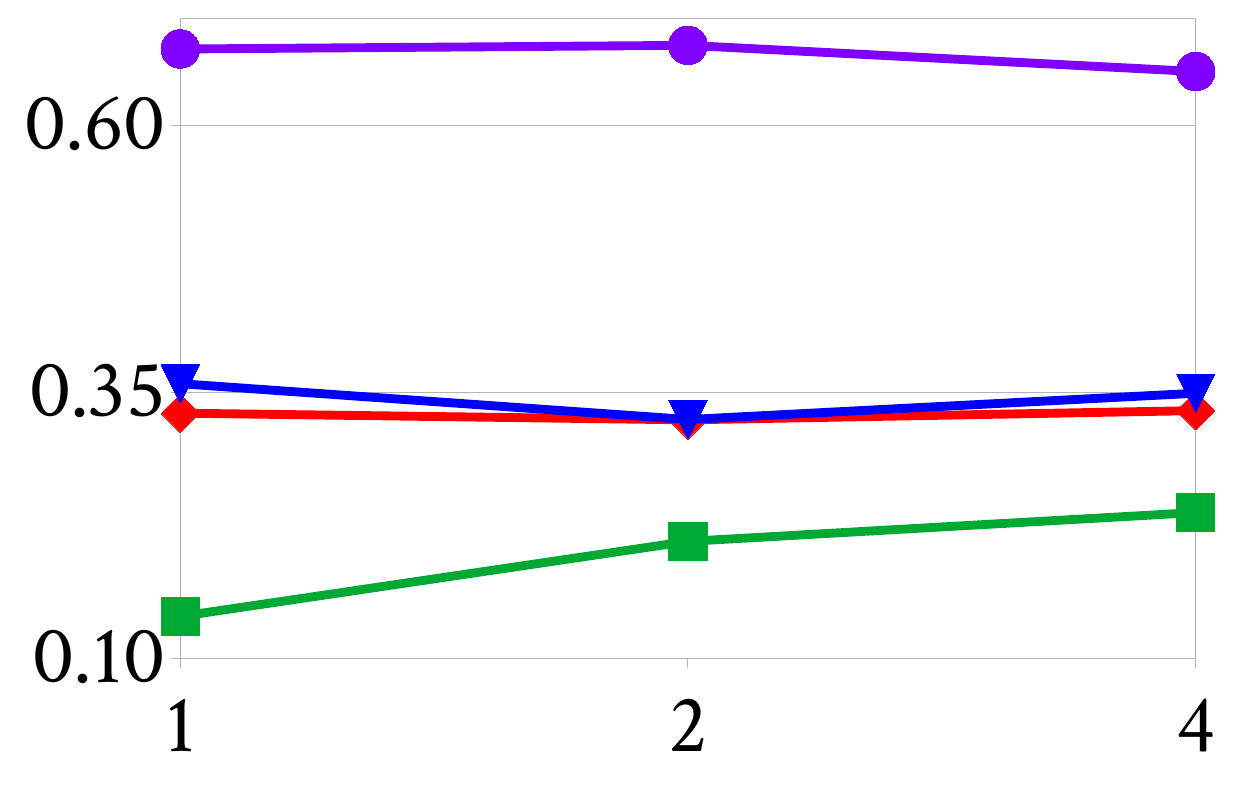}
		\subcaption{Positive F1-score}
	\end{subfigure}
	\begin{subfigure}[t]{0.22\textwidth}
		\includegraphics[width=\textwidth]{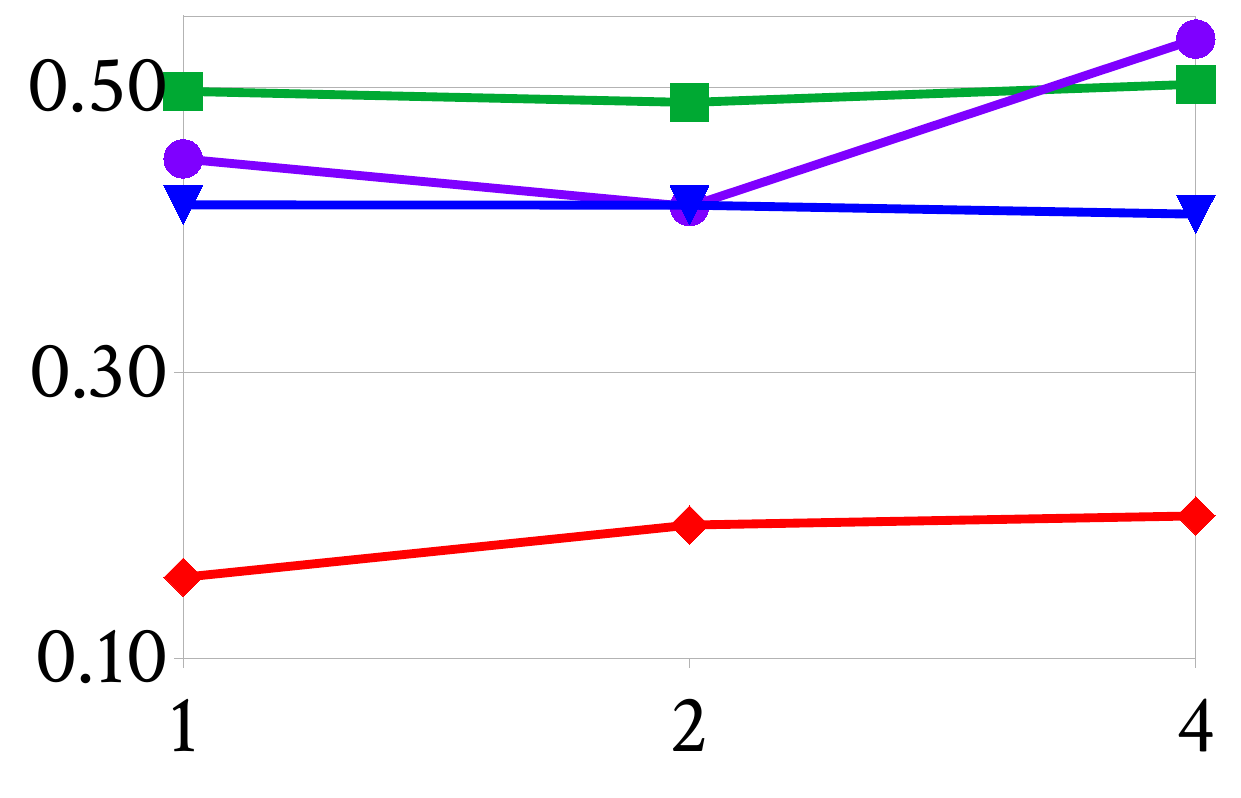}
		\subcaption{Negative F1-score}
	\end{subfigure}
	\begin{subfigure}[t]{0.22\textwidth}
		\includegraphics[width=\textwidth]{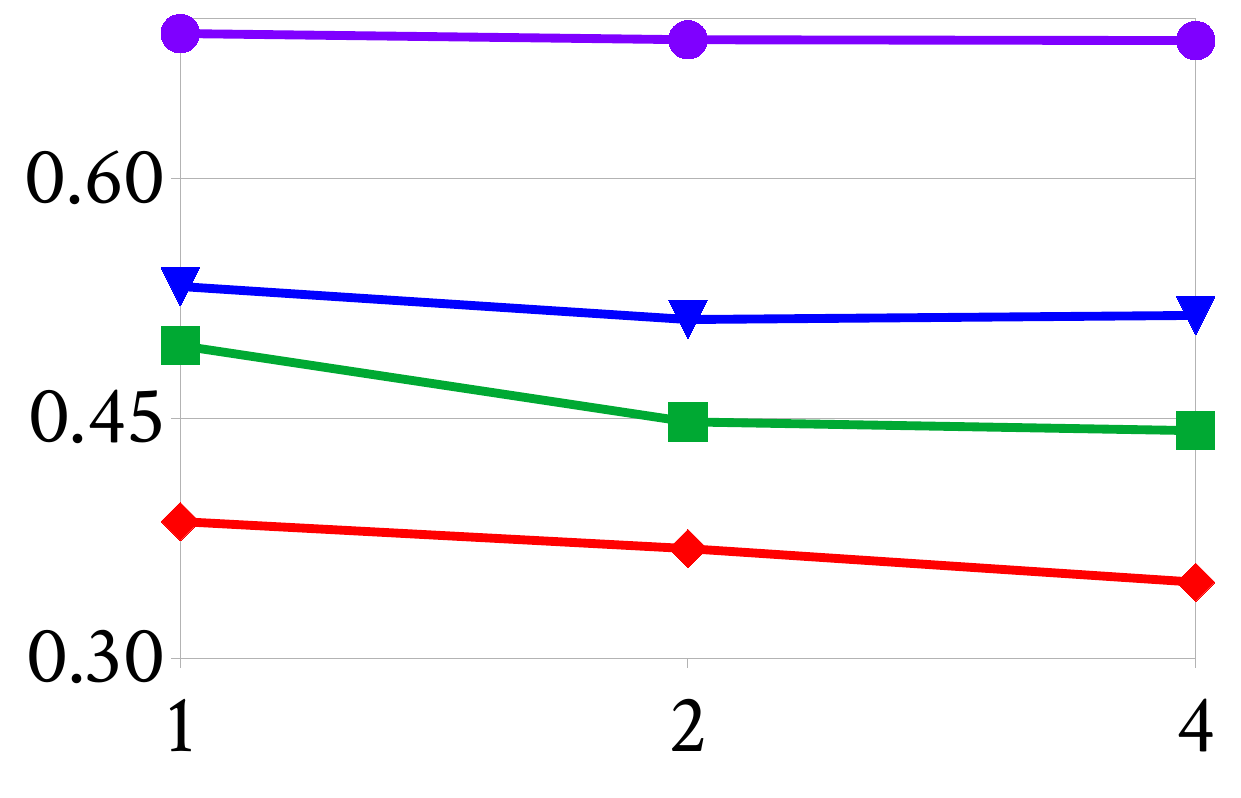}
		\subcaption{Neutral tag F1-score}
	\end{subfigure}
	\hfill
	\begin{subfigure}[t]{0.07\textwidth}
		\includegraphics[width=\textwidth]{IMG/abl_sa_legend.pdf}
		%		\subcaption{Legend}
	\end{subfigure}
	
	\caption{Effect of the number of attention heads ($H$) on sentiment analysis in Arabic, Algerian dialect, English, and French. 
		Results were obtained by training a BiGRU--dense layer decoder on frozen chDzDT embeddings.}

	\label{fig:ablation-H-sa}
\end{figure}

\paragraph{Sentiment analysis.}
Figure~\ref{fig:ablation-H-sa} shows the impact of varying the number of attention heads on sentiment classification. 
Similar to the results observed for depth ($N$), changes in $H$ produce only marginal and inconsistent effects. 
Overall accuracy remains stable across $H=1$, $H=2$, and $H=4$, with variations generally within a narrow range.
At the class level, the behavior is also irregular. 
Positive F1-scores improve modestly with higher $H$ in Arabic and English, but decline in the Algerian dialect. 
Negative and neutral scores show no systematic trend, with fluctuations often attributable to optimization variance rather than consistent gains. 
French results remain largely unchanged across all values of $H$.
Taken together, these findings suggest that the number of attention heads is not a decisive factor for sentiment analysis, echoing the patterns observed in PoS tagging. 
Unlike morphologically oriented tasks, where increasing capacity can sometimes yield benefits, sentence-level classification appears largely insensitive to this parameter. 
From a practical standpoint, the choice of head count has little effect on computational cost in our setup, so smaller values may be preferred for simplicity without compromising accuracy.

%\begin{tabular}{|c|c|c|c|c|c|c|c|c|c|c|c|c|c|c|c|c|c|}
%	\hline
%	&  & Overall &  &  &  & Positive &  &  &  & Negative &  &  &  & Neutral &  &  &  \\
%	\hline
%	H &  & Arabic & Algerian & English & French & Arabic & Algerian & English & French & Arabic & Algerian & English & French & Arabic & Algerian & English & French \\
%	\hline
%	1 &  & 0.4411 & 0.6171 & 0.3187 & 0.4425 & 0.1397 & 0.6717 & 0.3302 & 0.3579 & 0.4972 & 0.4499 & 0.1569 & 0.4177 & 0.4957 & 0.6908 & 0.3857 & 0.5327 \\
%	\hline
%	2 &  & 0.4192 & 0.6133 & 0.3146 & 0.4241 & 0.2100 & 0.6751 & 0.3239 & 0.3242 & 0.4894 & 0.4165 & 0.1935 & 0.4175 & 0.4480 & 0.6869 & 0.3688 & 0.5120 \\
%	\hline
%	4 &  & 0.4255 & 0.6225 & 0.3098 & 0.4299 & 0.2370 & 0.6506 & 0.3326 & 0.3489 & 0.5022 & 0.5338 & 0.1999 & 0.4112 & 0.4425 & 0.6864 & 0.3478 & 0.5146 \\
%	\hline
%\end{tabular}

\subsubsection{Embedding size}

The dimensionality of embeddings ($d$) determines the model's representational capacity. 
\textit{What is the trade-off between smaller embeddings, which reduce parameter count and training time, and larger embeddings, which may capture richer features but increase computational demands?} 
To investigate this, we fixed $N=2$ and $H=2$, and varied $d \in \{8, 16, 32\}$.

\begin{figure}[!htp]
	\centering
	
	\includegraphics[width=0.4\textwidth]{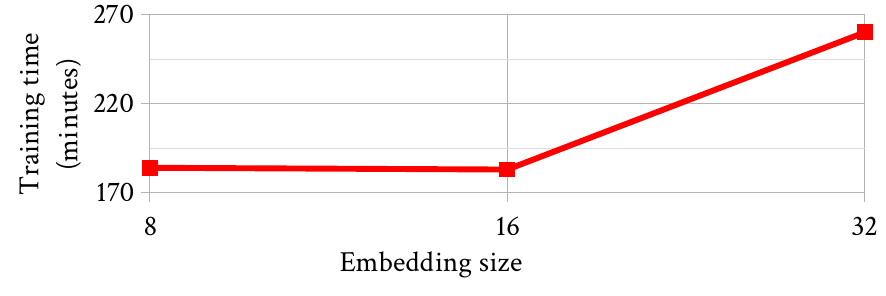}
	\hfill
	\includegraphics[width=0.4\textwidth]{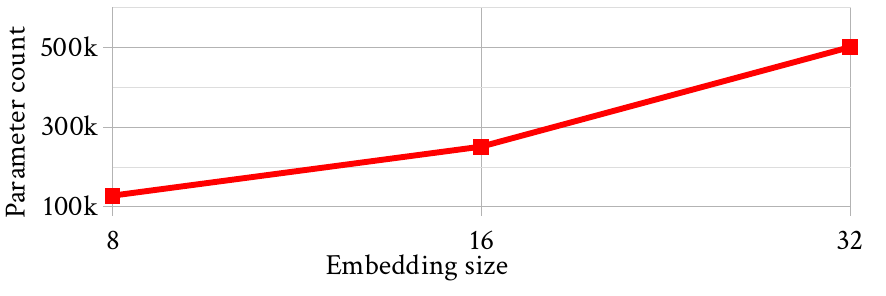}
	
	\caption{Effect of embedding size ($d$) on training time and parameter count.}
	
	\label{fig:ablation-d-t_theta}
\end{figure}

\paragraph{Model efficiency.}  
Figure~\ref{fig:ablation-d-t_theta} shows that the parameter count grows roughly linearly with embedding size $d$. 
This follows from the dependence of layer weights on hidden dimensionality.
By contrast, training time does not increase strictly monotonically: $d=8$ and $d=16$ yield similar times, whereas $d=32$ shows a substantial increase.

\begin{figure}[!htp]
	\centering\small
	
	\begin{subfigure}[t]{0.22\textwidth}
		\includegraphics[width=\textwidth]{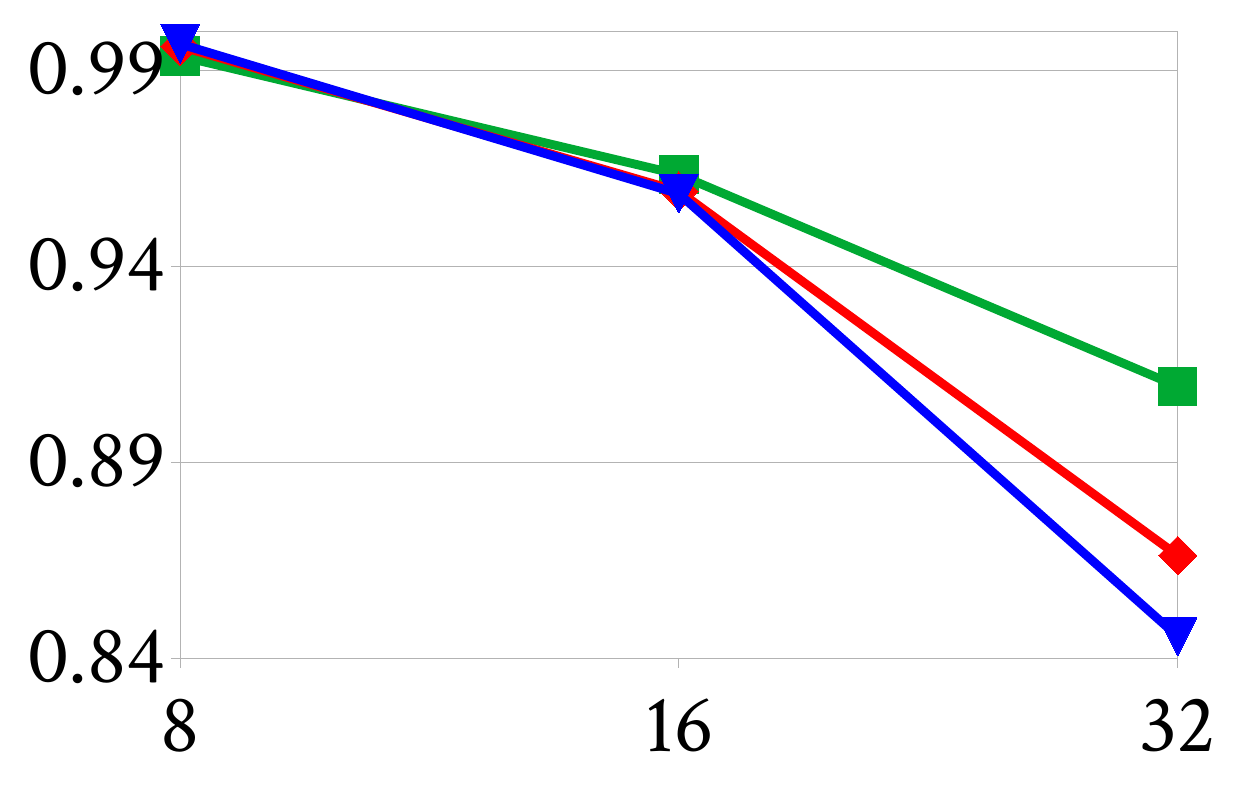}
		\subcaption{Deriv--Avg}
	\end{subfigure}
	\begin{subfigure}[t]{0.22\textwidth}
		\includegraphics[width=\textwidth]{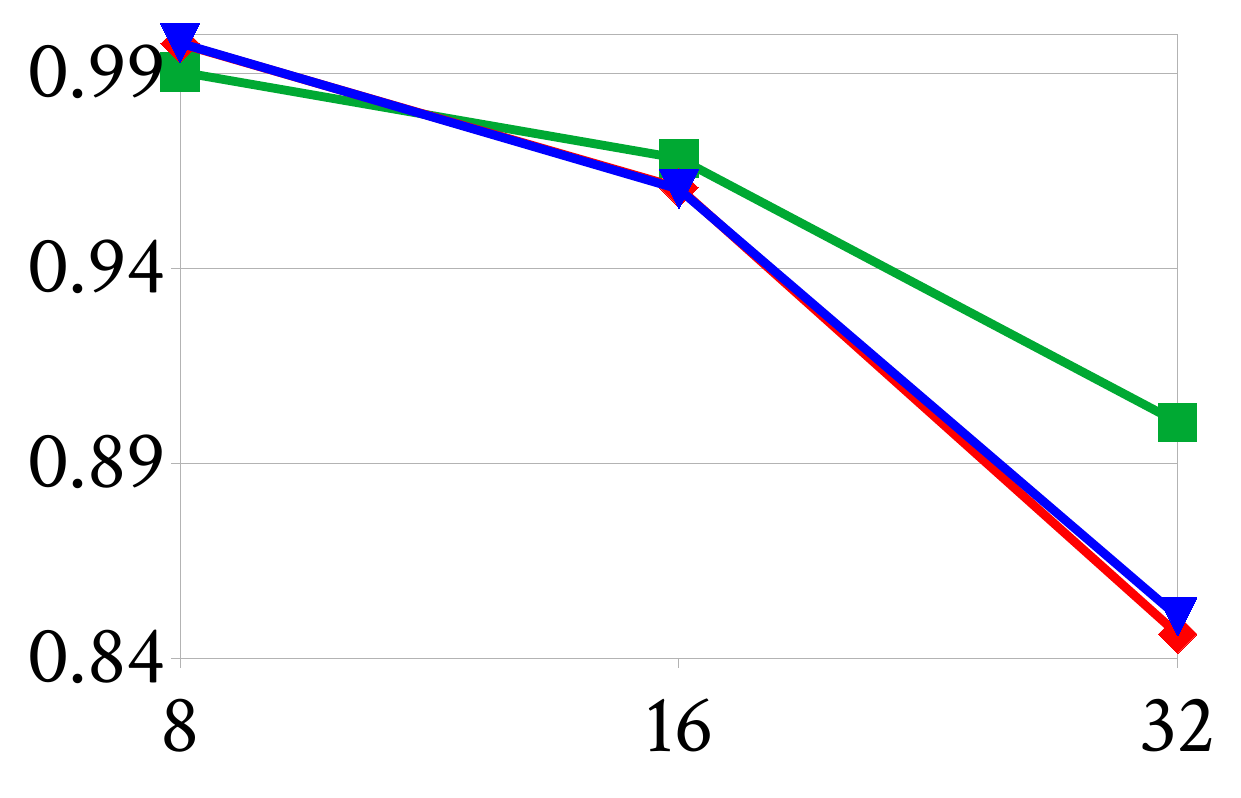}
		\subcaption{Deriv--Min}
	\end{subfigure}
	\begin{subfigure}[t]{0.22\textwidth}
		\includegraphics[width=\textwidth]{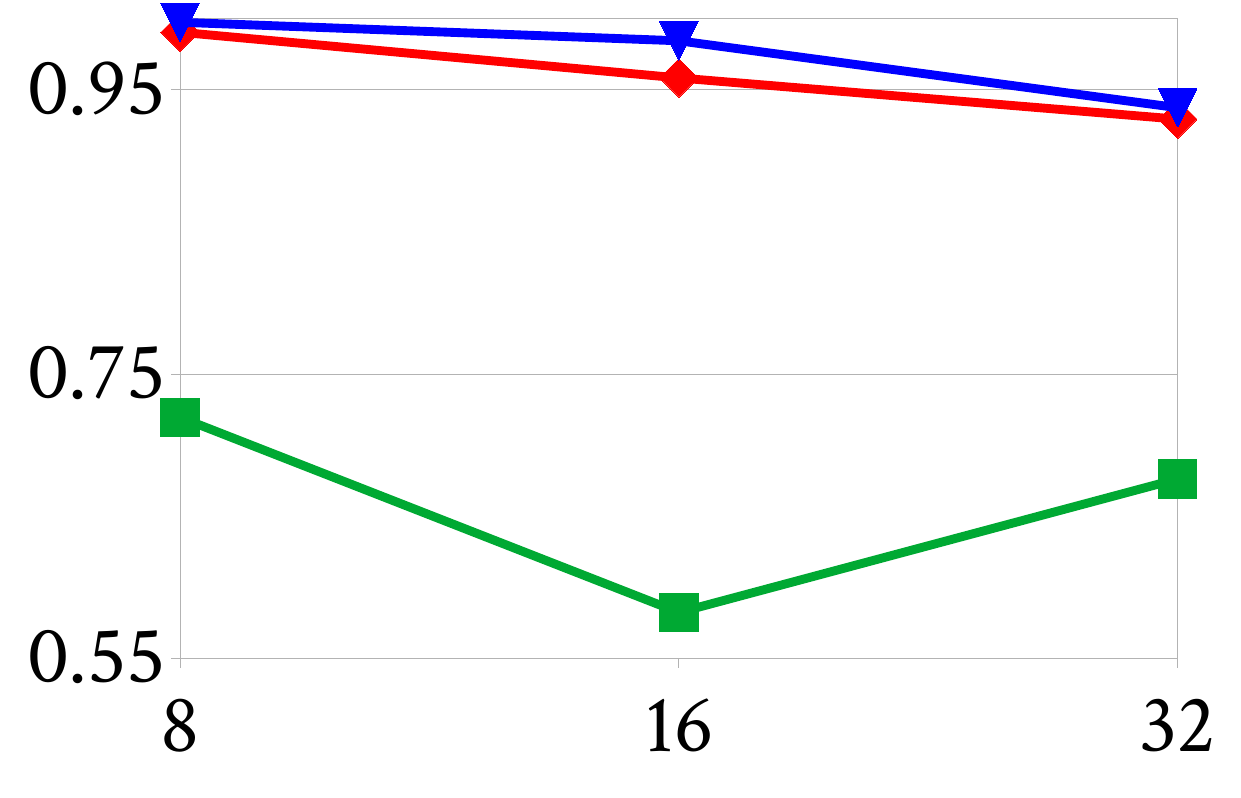}
		\subcaption{Infl--Avg}
	\end{subfigure}
	\begin{subfigure}[t]{0.22\textwidth}
		\includegraphics[width=\textwidth]{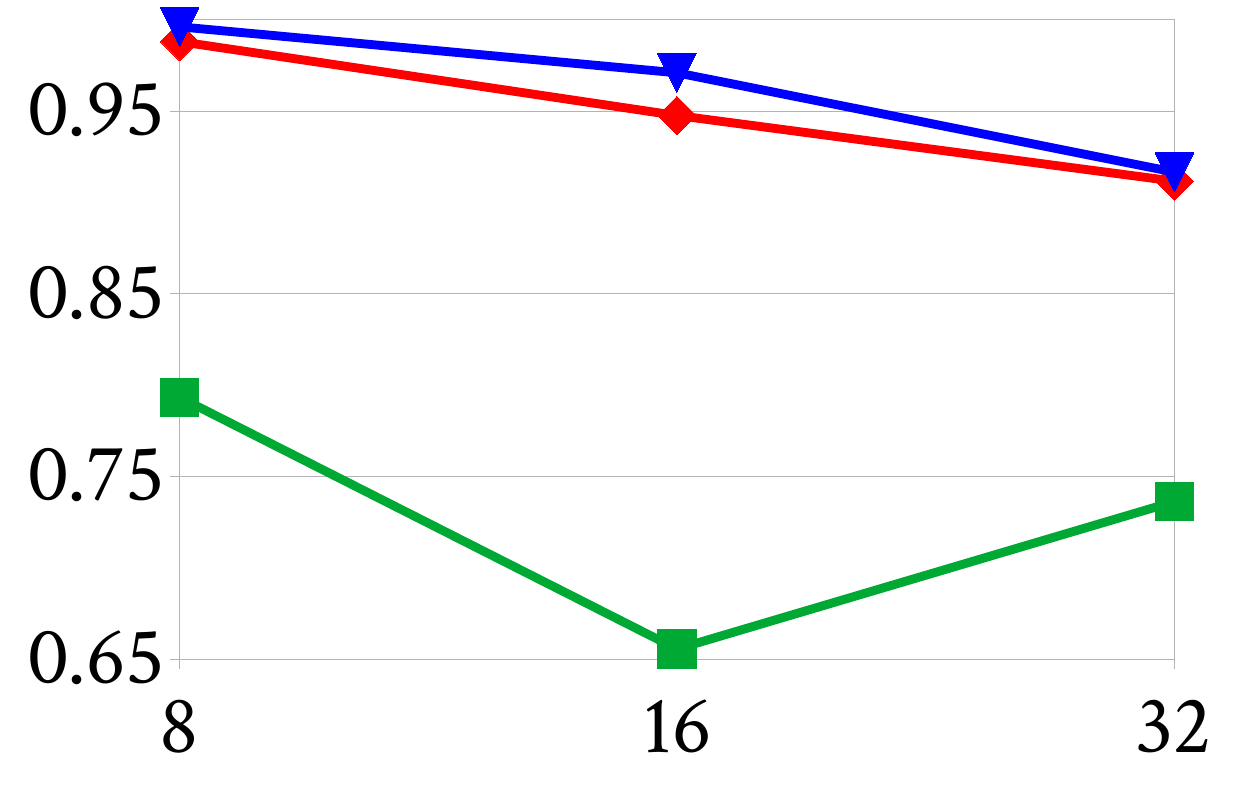}
		\subcaption{Infl--Min}
	\end{subfigure}
	\hfill
	\begin{subfigure}[t]{0.07\textwidth}
		\includegraphics[width=\textwidth]{IMG/abl_morph_legend.pdf}
		%		\subcaption{Legend}
	\end{subfigure}
	
	\caption{Effect of the embedding size ($d$) on morphological consistency in Arabic, French, and English.  
		Plots show derivational (Deriv) and inflectional (Infl) morphology on the ``Avg'' and ``Min'' datasets, using average cosine similarity between words and their root.}
	
	\label{fig:ablation-d-morph-consist}
\end{figure}

\paragraph{Morphological consistency.}
Figure~\ref{fig:ablation-d-morph-consist} shows that embedding size $d$ strongly affects morphological consistency. 
For derivational morphology, performance peaks at $d=8$ but declines as $d$ increases.
English and French deteriorate sharply at $d=32$, while Arabic degrades more gradually due to its dataset design, which groups derivations by both form and meaning. 
Inflectional morphology exhibits a different pattern. 
English and French remain relatively stable with modest decreases, whereas Arabic shows a sharp drop at $d=16$ followed by partial recovery at $d=32$. 
This pattern in Arabic echoes the earlier non-monotonic behaviors observed with attention heads.
It likely reflects the structural complexity of Arabic inflection, where templatic and affixal processes respond unevenly to increased dimensionality.
Overall, smaller embeddings ($d=8$) best preserve derivational consistency.
Compact encodings make words more similar to their roots, enhancing morphological alignment, although unrelated words may sometimes appear erroneously similar. 
Inflectional patterns are less sensitive, but the Arabic results highlight that optimal dimensionality can be language-specific.

%\begin{tabular}{|c|c|c|c|c|c|c|c|c|c|c|c|c|c|c|c|}
%	\hline
%	& deriv & deriv & deriv &  & deriv & deriv & deriv &  & infl & infl & infl &  & infl & infl & infl \\
%	\hline
%	& avg & avg & avg &  & min & min & min &  & avg & avg & avg &  & min & min & min \\
%	\hline
%	d & Arabic & English & French &  & Arabic & English & French &  & Arabic & English & French &  & Arabic & English & French \\
%	\hline
%	8 & 0.9936 & 0.9961 & 0.9967 &  & 0.9903 & 0.9976 & 0.9977 &  & 0.7198 & 0.9906 & 0.9976 &  & 0.7935 & 0.9880 & 0.9961 \\
%	\hline
%	16 & 0.9634 & 0.9592 & 0.9585 &  & 0.9682 & 0.9606 & 0.9603 &  & 0.5825 & 0.9584 & 0.9845 &  & 0.6557 & 0.9475 & 0.9709 \\
%	\hline
%	32 & 0.9094 & 0.8663 & 0.8457 &  & 0.9005 & 0.8463 & 0.8508 &  & 0.6764 & 0.9294 & 0.9375 &  & 0.7364 & 0.9116 & 0.9166 \\
%	\hline
%\end{tabular}

\begin{figure}[!htp]
	\centering\small
	
	\begin{subfigure}[t]{0.27\textwidth}
		\includegraphics[width=\textwidth]{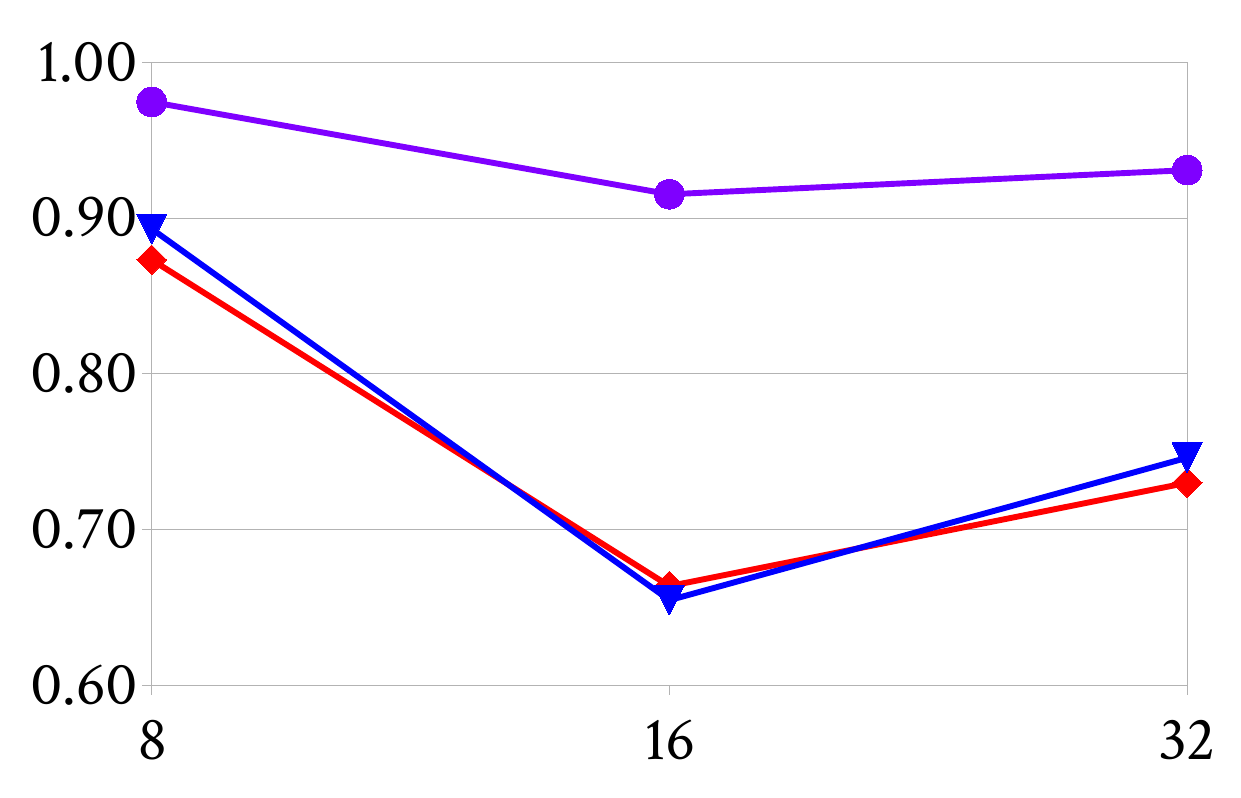}
		\subcaption{Obfus. clusters}
	\end{subfigure}
	\begin{subfigure}[t]{0.27\textwidth}
		\includegraphics[width=\textwidth]{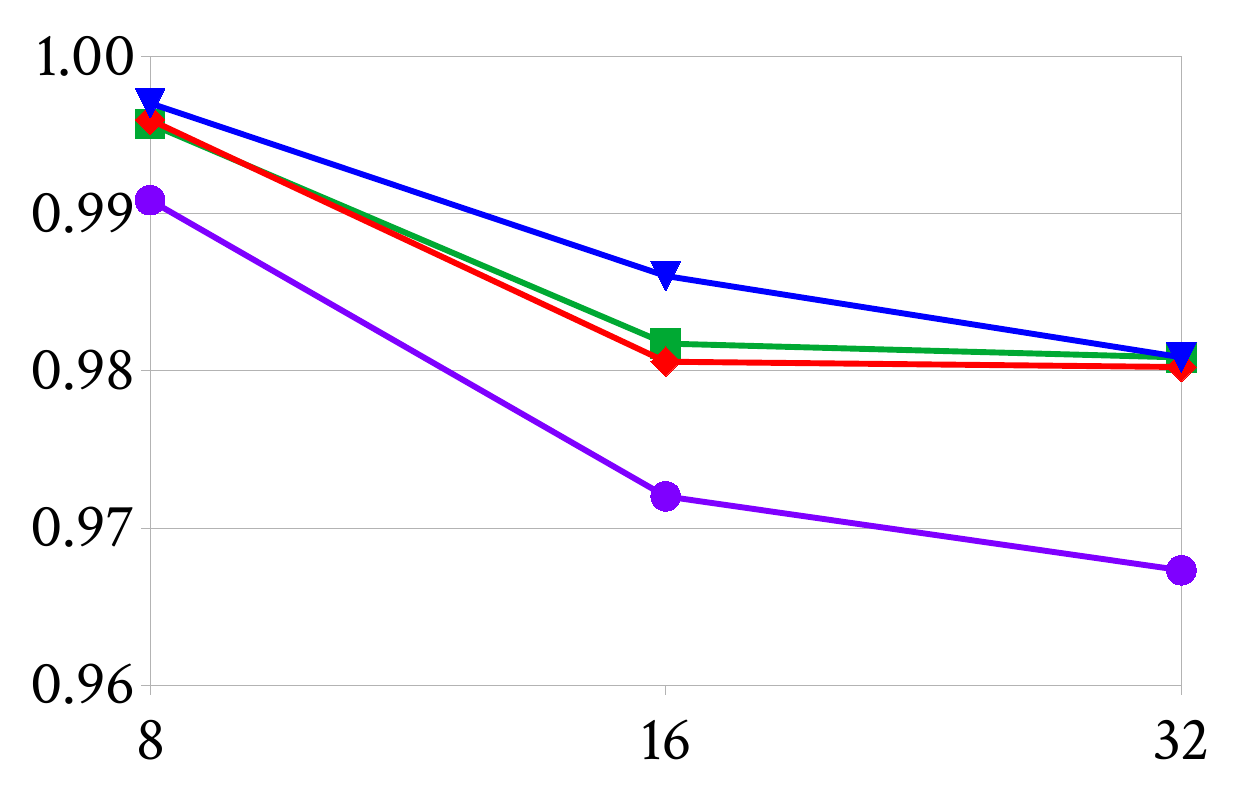}
		\subcaption{Obfus. tuples-*}
	\end{subfigure}
	\begin{subfigure}[t]{0.27\textwidth}
		\includegraphics[width=\textwidth]{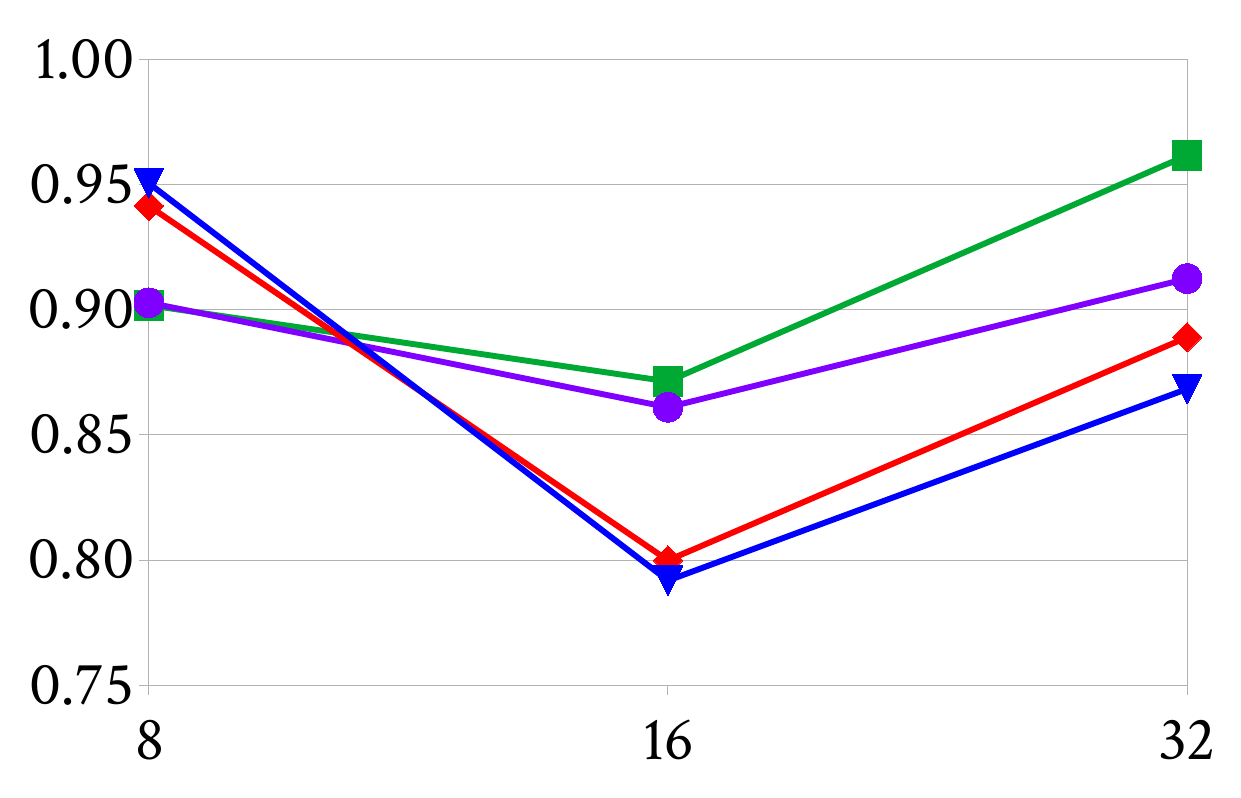}
		\subcaption{Obfus. tuples-c}
	\end{subfigure}
	\hfill
	\begin{subfigure}[t]{0.07\textwidth}
		\includegraphics[width=\textwidth]{IMG/abl_noise_legend.pdf}
		%		\subcaption{Legend}
	\end{subfigure}
	
	\caption{Effect of the embedding size ($d$) on robustness to orthographic noise in Arabic, Arabizi, French, and English. 
		Plots show: (a) cluster-based obfuscations, (b) tuple obfuscations with \texttt{*} substitutions, and 
		(c) tuple obfuscations with visually or phonetically similar substitutions. 
		Scores are reported as Average Cosine Similarity (ACS). 
		Note that Arabic is absent from (a), as it is predominantly written in standard form and does not exhibit such obfuscations.}
	
	\label{fig:ablation-d-morph-noise}
\end{figure}

\paragraph{Robustness to orthographic noise.}
Figure~\ref{fig:ablation-d-morph-noise} shows the effect of embedding size ($d$) on robustness to orthographic noise, revealing a stronger influence than depth or attention heads. 
ACS for cluster-based obfuscations declines gradually with larger $d$, especially in English and French.
Arabizi remains relatively stable ($\approx0.92$), likely due to pre-training exposure to noisy input. 
Tuple obfuscations with ``*'' substitutions are highly robust across all settings ($>0.96$), confirming that minimal masking is largely unaffected by embedding dimensionality. 
More challenging tuple obfuscations with phonetically or visually similar substitutions show sharper contrasts.
Arabic recovers at larger $d$, reflecting redundancy in its templatic morphology; Arabizi maintains strong stability, while English and French deteriorate steadily, underscoring their reliance on rigid orthography. 
Overall, embedding size significantly affects robustness.
Larger embeddings benefit morphologically rich or flexible languages such as Arabic and Arabizi, but may increase susceptibility to noise in alphabetic systems like English and French.

%\begin{tabular}{|c|c|c|c|c|c|c|c|c|c|c|c|}
%	\hline
%	& ACS & ACS & ACS & ACS1* & ACS1* & ACS1* & ACS1* & ACS1c & ACS1c & ACS1c & ACS1c \\
%	\hline
%	d & Arabizi & English & French & Arabic & Arabizi & English & French & Arabic & Arabizi & English & French \\
%	\hline
%	8 & 0.9746 & 0.8733 & 0.8931 & 0.9957 & 0.9909 & 0.9960 & 0.9970 & 0.9018 & 0.9028 & 0.9415 & 0.9506 \\
%	\hline
%	16 & 0.9153 & 0.6637 & 0.6547 & 0.9817 & 0.9720 & 0.9806 & 0.9860 & 0.8714 & 0.8610 & 0.7999 & 0.7918 \\
%	\hline
%	32 & 0.9308 & 0.7300 & 0.7461 & 0.9809 & 0.9673 & 0.9802 & 0.9809 & 0.9617 & 0.9125 & 0.8891 & 0.8684 \\
%	\hline
%\end{tabular}

\begin{figure}[!htp]
	\centering\small
	
	\begin{subfigure}[t]{0.24\textwidth}
		\includegraphics[width=\textwidth]{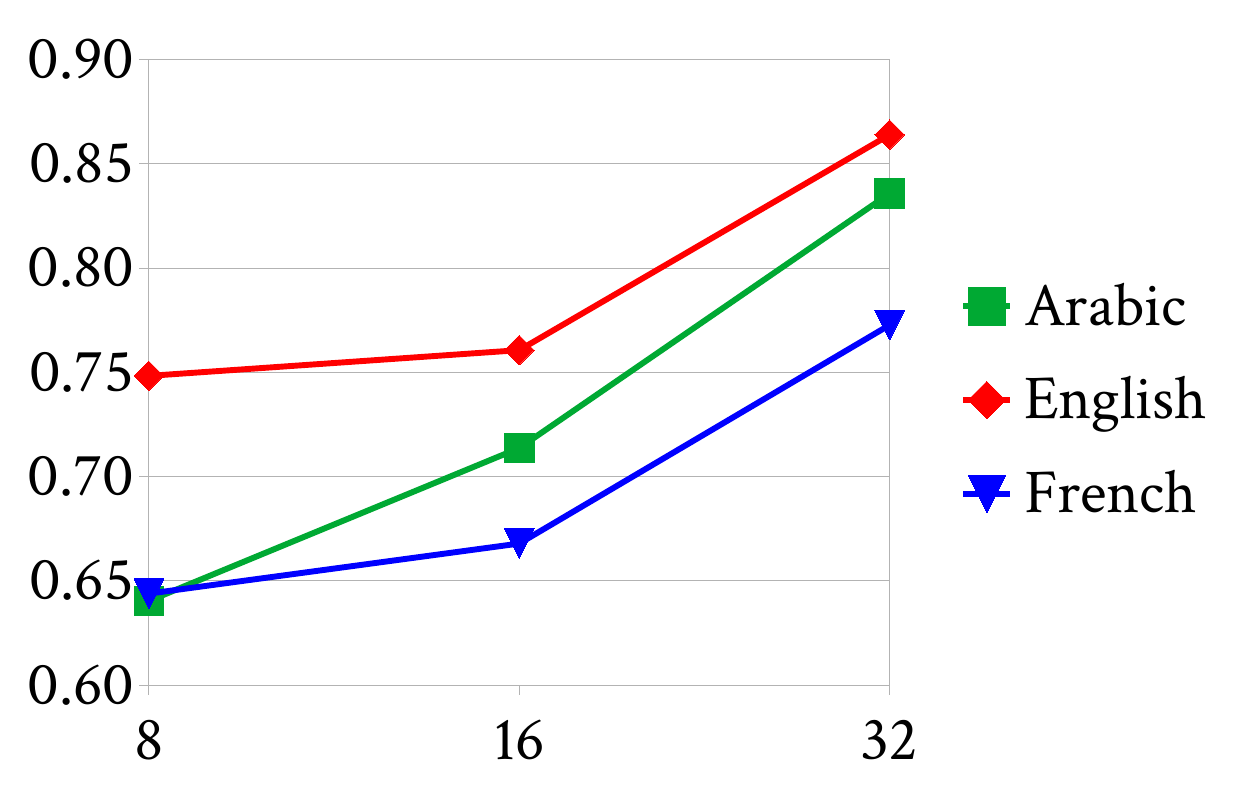}
		\subcaption{Overall Accuracy}
	\end{subfigure}
	\begin{subfigure}[t]{0.24\textwidth}
		\includegraphics[width=\textwidth]{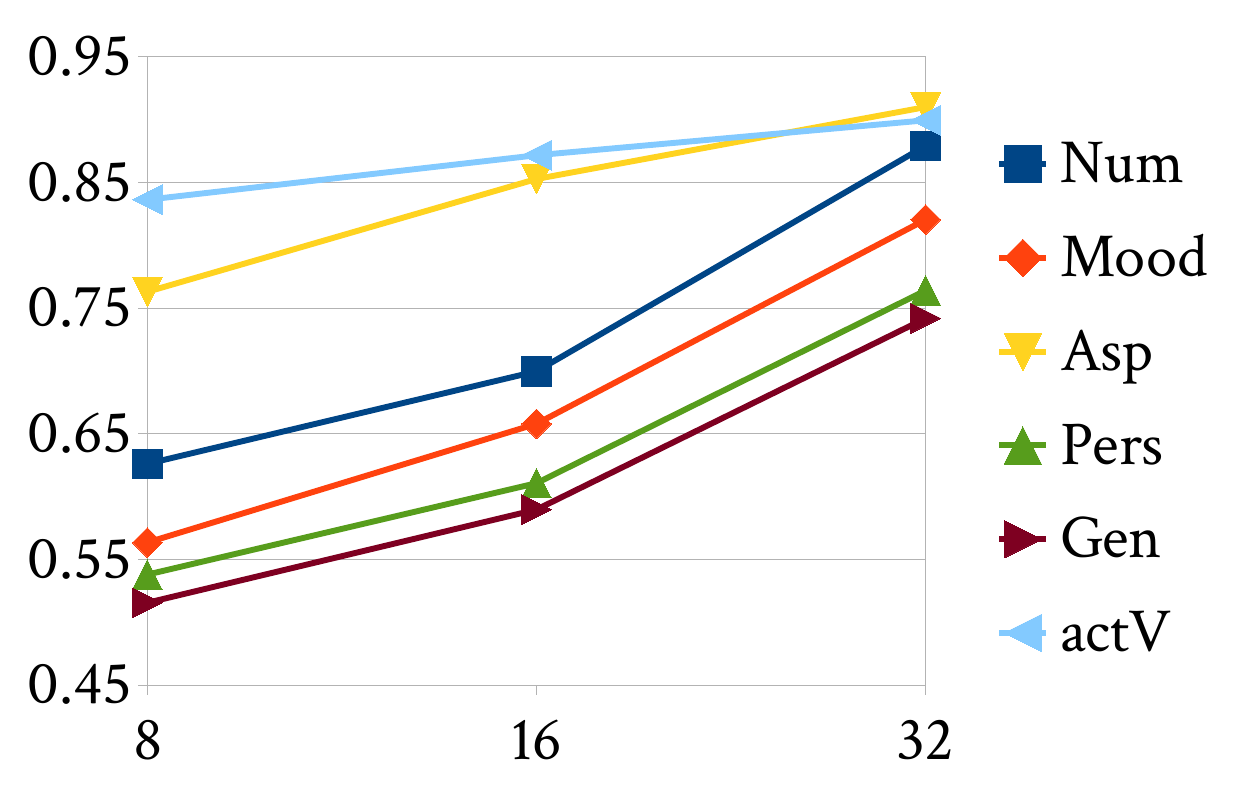}
		\subcaption{Accuracy of Arabic tags}
	\end{subfigure}
	\begin{subfigure}[t]{0.24\textwidth}
		\includegraphics[width=\textwidth]{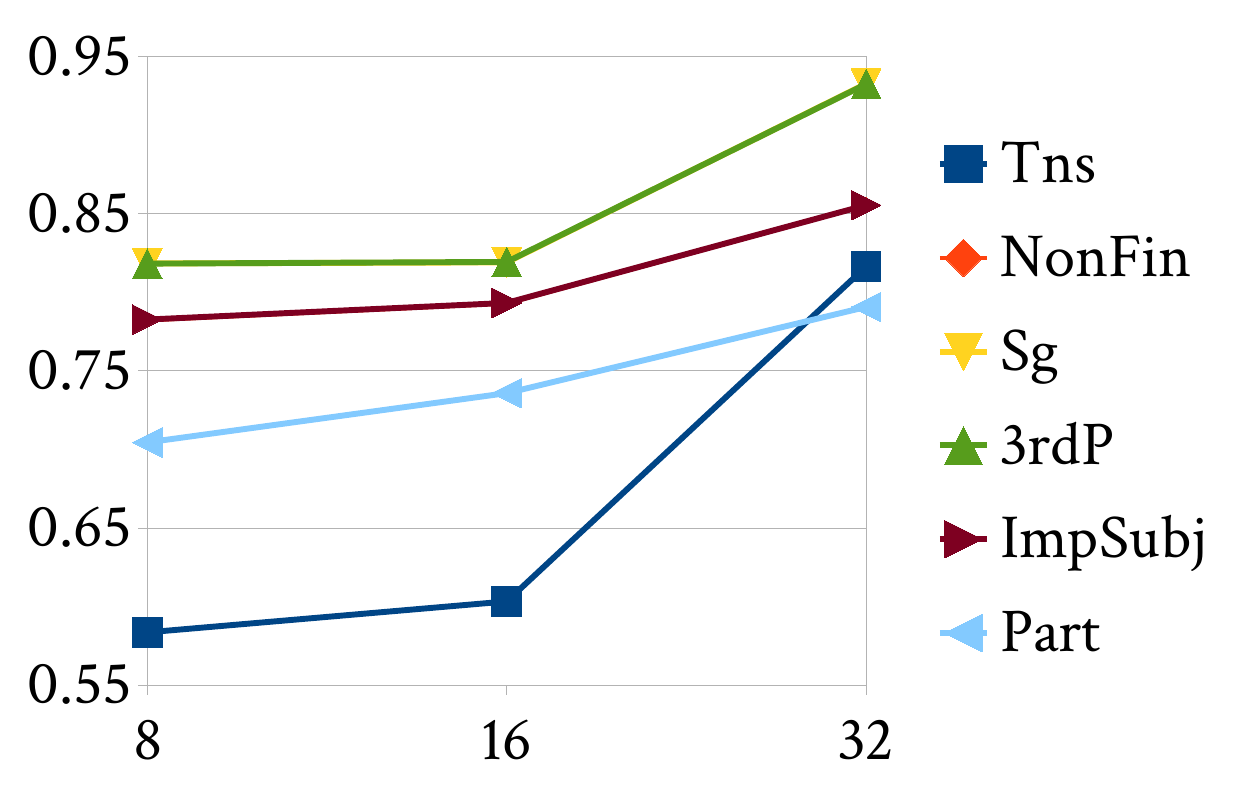}
		\subcaption{Accuracy of English tags}
	\end{subfigure}
	\begin{subfigure}[t]{0.24\textwidth}
		\includegraphics[width=\textwidth]{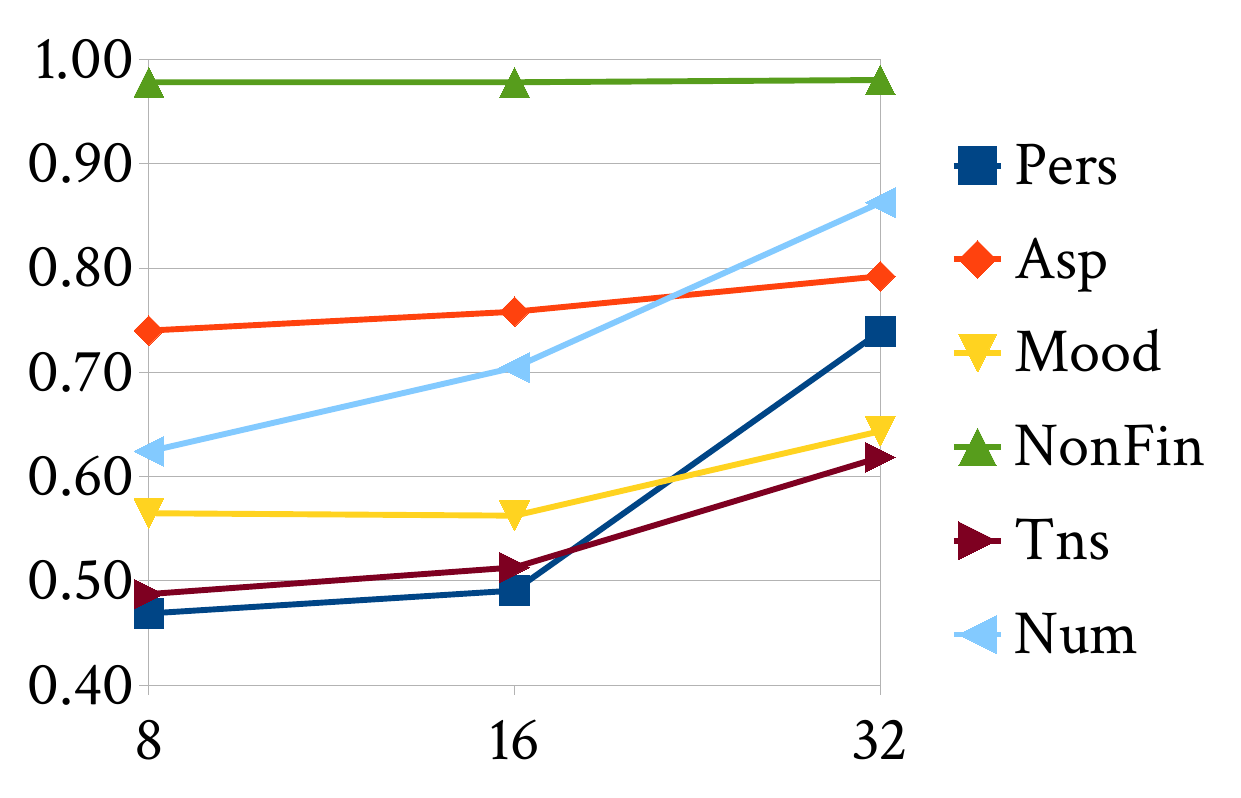}
		\subcaption{Accuracy of French tags}
	\end{subfigure}
	
	\caption{Effect of the embedding size ($d$) on morphological tagging in Arabic, English, and French. 
		(a) Overall accuracy across the three languages; (b) Arabic tags; (c) English tags; (d) French tags. 
		In English, the curves for \textit{Sg} and \textit{3rdP} overlap, as do those for \textit{NonFin} and \textit{ImpSubj}.}
	
	\label{fig:ablation-d-morph-tag}
\end{figure}

\paragraph{Morphological tagging.}
Figure~\ref{fig:ablation-d-morph-tag} shows that embedding dimensionality strongly affects morphological tagging performance. 
With a very small size ($d=8$), accuracy is lowest across all three languages, reflecting limited representational capacity. 
Increasing to $d=16$ yields moderate improvements but remains constrained, especially for French. 
By contrast, $d=32$ produces substantial gains in overall accuracy, with Arabic and English exceeding 0.80 and French reaching 0.77. 
The sharper rise for Arabic suggests that richer embeddings better capture templatic and feature-rich morphology, while English benefits from clearer binary distinctions. 
French continues to lag, highlighting the challenge of modeling its more elaborate verb system. 
Overall, the results indicate that too small an embedding size restricts the model, whereas moderate dimensionality provides sufficient representational space to capture morphological detail.

%\begin{tabular}{|c|c|c|c|c|c|c|c|c|c|c|c|c|c|c|c|c|c|c|c|c|c|}
%	\hline
%	d & Overall &  &  & Arabic &  &  &  &  &  & English &  &  &  &  &  & French &  &  &  &  &  \\
%	\hline
%	d & Arabic & English & French & Num & Mood & Asp & Pers & Gen & actV & Tns & NonFin & Sg & 3rdP & ImpSubj & Part & Pers & Asp & Mood & NonFin & Tns & Num \\
%	\hline
%	8 & 0.6404 & 0.7483 & 0.6439 & 0.6261 & 0.5635 & 0.7630 & 0.5380 & 0.5156 & 0.8362 & 0.5836 & 0.7824 & 0.8184 & 0.8183 & 0.7827 & 0.7045 & 0.4688 & 0.7401 & 0.5649 & 0.9782 & 0.4874 & 0.6242 \\
%	\hline
%	16 & 0.7137 & 0.7607 & 0.6679 & 0.6997 & 0.6579 & 0.8527 & 0.6107 & 0.5897 & 0.8716 & 0.6033 & 0.7931 & 0.8191 & 0.8194 & 0.7933 & 0.7359 & 0.4908 & 0.7582 & 0.5626 & 0.9782 & 0.5130 & 0.7047 \\
%	\hline
%	32 & 0.8358 & 0.8639 & 0.7727 & 0.8791 & 0.8204 & 0.9099 & 0.7640 & 0.7419 & 0.8994 & 0.8165 & 0.8555 & 0.9328 & 0.9326 & 0.8555 & 0.7907 & 0.7391 & 0.7922 & 0.6436 & 0.9803 & 0.6185 & 0.8627 \\
%	\hline
%\end{tabular}

\begin{figure}[!htp]
	\centering\small
	
	\begin{subfigure}[t]{0.22\textwidth}
		\includegraphics[width=\textwidth]{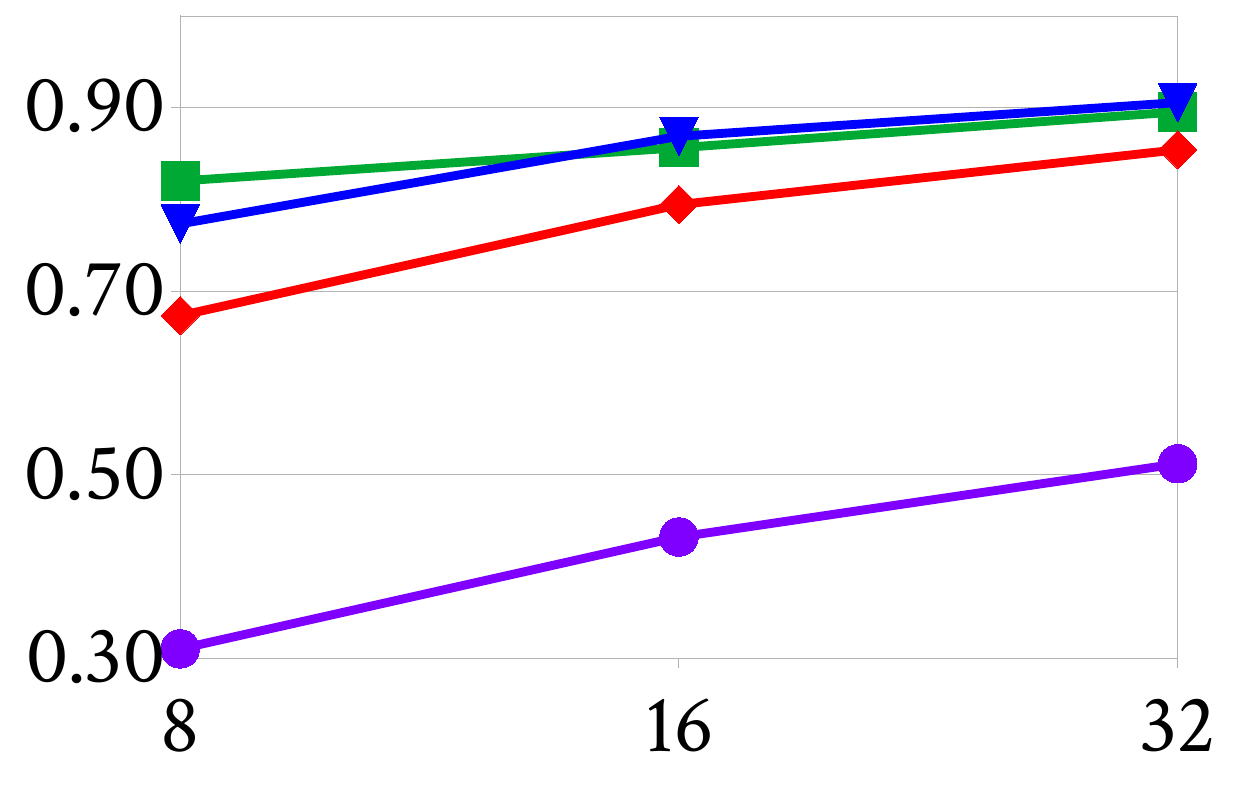}
		\subcaption{Overall accuracy}
	\end{subfigure}
	\begin{subfigure}[t]{0.22\textwidth}
		\includegraphics[width=\textwidth]{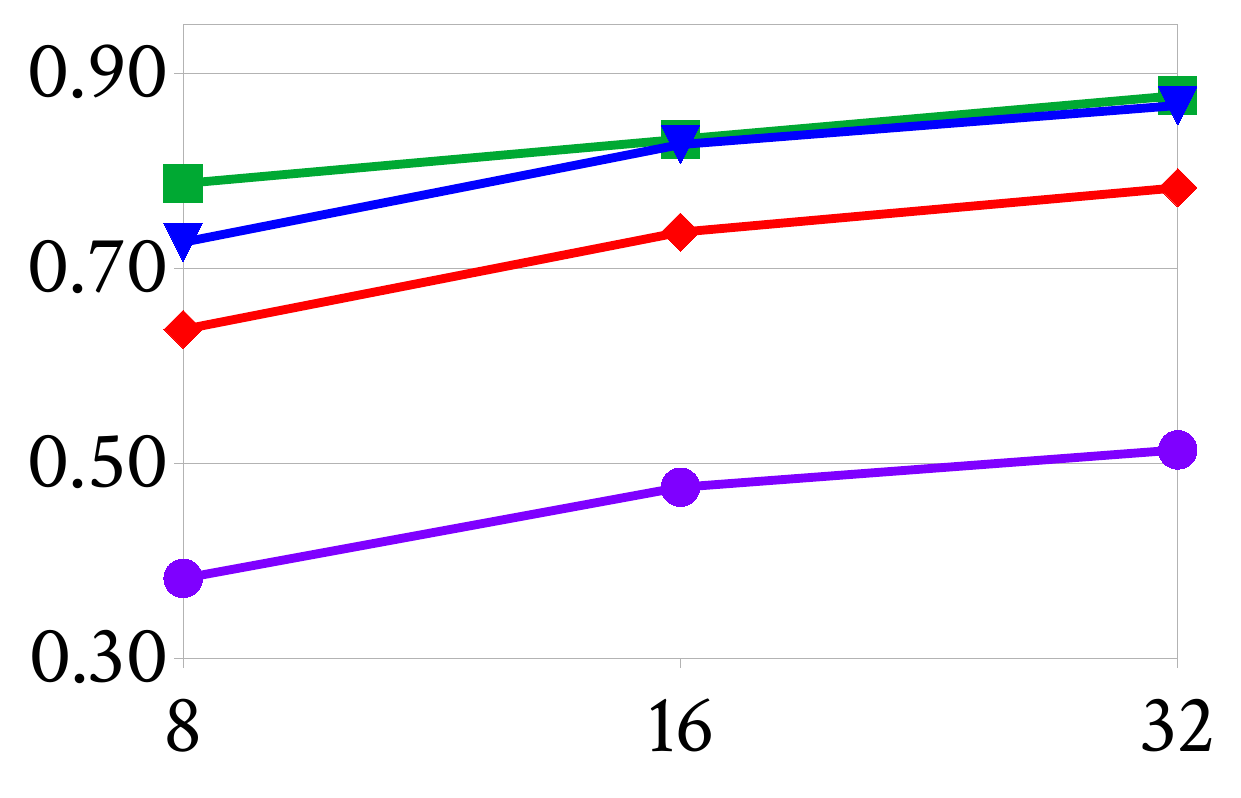}
		\subcaption{Noun tag F1-score}
	\end{subfigure}
	\begin{subfigure}[t]{0.22\textwidth}
		\includegraphics[width=\textwidth]{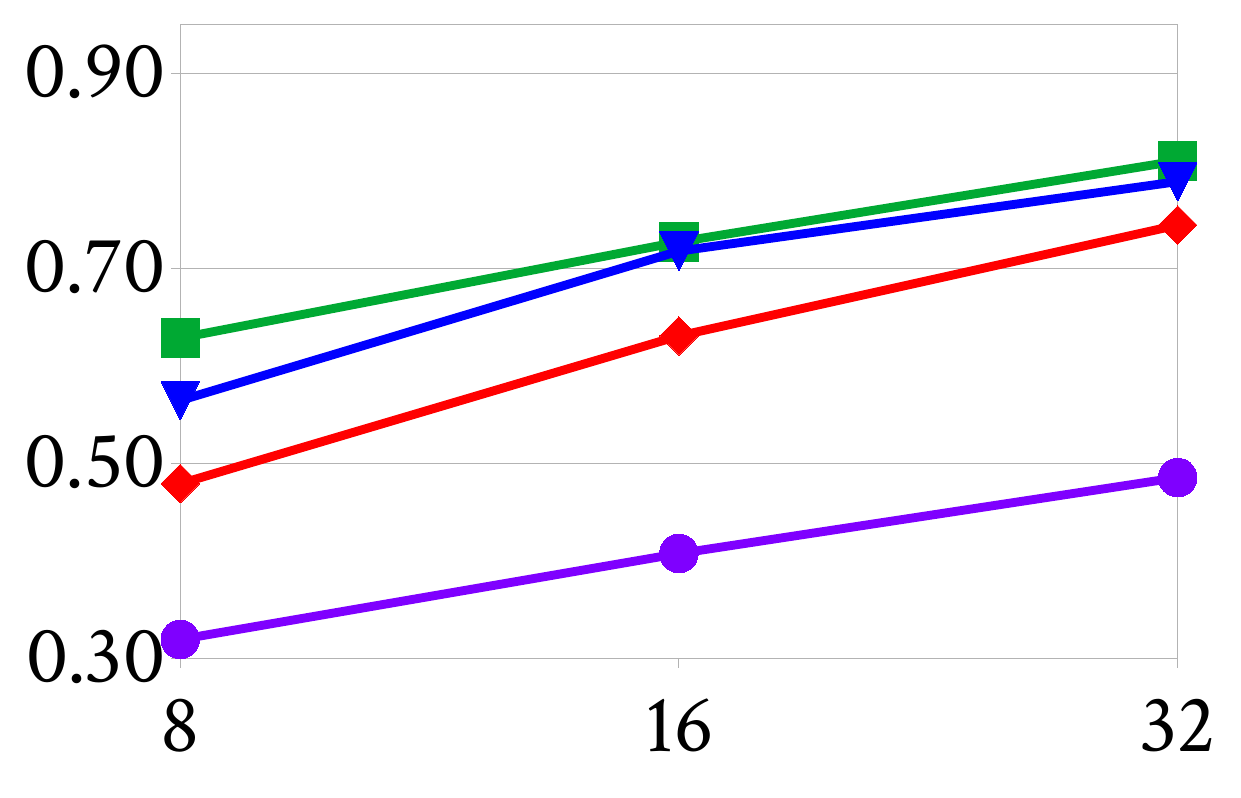}
		\subcaption{Verb tag F1-score}
	\end{subfigure}
	\begin{subfigure}[t]{0.22\textwidth}
		\includegraphics[width=\textwidth]{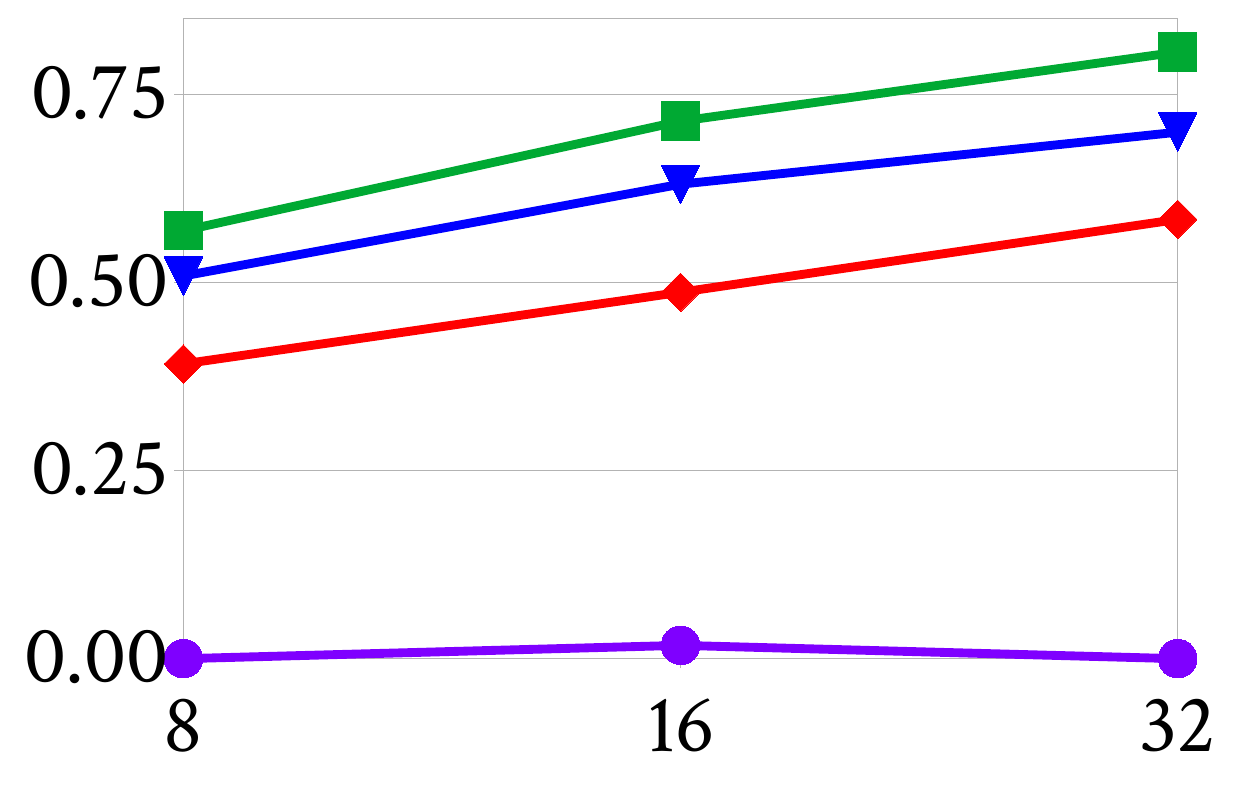}
		\subcaption{Adjective tag F1-score}
	\end{subfigure}
	\hfill
	\begin{subfigure}[t]{0.07\textwidth}
		\includegraphics[width=\textwidth]{IMG/abl_pos_legend.pdf}
		%		\subcaption{Legend}
	\end{subfigure}
	
	\caption{Effect of the embedding size ($d$) on PoS tagging in Arabic, Arabizi, English, and French. 
		Results were obtained by training a BiGRU decoder on frozen chDzDT embeddings.}
	
	\label{fig:ablation-d-pos-tag}
\end{figure}

\paragraph{Part-of-speech tagging.}
Figure~\ref{fig:ablation-d-pos-tag} shows that embedding size ($d$) has a clear and consistent impact on PoS tagging performance across all languages. 
Increasing $d$ from $8$ to $16$ leads to substantial improvements in overall accuracy, with further gains observed at $d=32$. 
Noun and verb F1-scores benefit particularly from larger embeddings, reflecting that more expressive character-based representations facilitate downstream decoding for lexical categorization. 
However, the patterns are not uniform. 
Arabizi performance increases with $d$, but absolute scores remain lower than in other languages, and adjective tagging is unstable, sometimes failing regardless of embedding size. 
English and French show diminishing returns at $d=32$, where improvements are modest compared to the jump from $8$ to $16$. 
Compared to depth ($N$) and attention head count ($H$), embedding size has the strongest effect on PoS tagging.
This highlights that token-level representation capacity is more critical for capturing lexical distinctions than architectural depth or head multiplicity. 
In practical terms, allocating parameters to larger embeddings may offer a better efficiency--accuracy trade-off for PoS tagging than increasing $N$ or $H$.

%\begin{tabular}{|c|c|c|c|c|c|c|c|c|c|c|c|c|c|c|c|c|}
%	\hline
%	& Overall &  &  &  & Noun &  &  &  & Verb &  &  &  & Adjective &  &  &  \\
%	\hline
%	d & Arabic & Arabizi & English & French & Arabic & Arabizi & English & French & Arabic & Arabizi & English & French & Arabic & Arabizi & English & French \\
%	\hline
%	8 & 0.8202 & 0.3106 & 0.6736 & 0.7735 & 0.7871 & 0.3823 & 0.6375 & 0.7269 & 0.6287 & 0.3197 & 0.4794 & 0.5647 & 0.5686 & 0.0000 & 0.3916 & 0.5084 \\
%	\hline
%	16 & 0.8564 & 0.4324 & 0.7945 & 0.8689 & 0.8325 & 0.4757 & 0.7372 & 0.8270 & 0.7273 & 0.4078 & 0.6309 & 0.7178 & 0.7140 & 0.0177 & 0.4868 & 0.6303 \\
%	\hline
%	32 & 0.8955 & 0.5122 & 0.8542 & 0.9056 & 0.8773 & 0.5140 & 0.7828 & 0.8671 & 0.8103 & 0.4856 & 0.7445 & 0.7891 & 0.8059 & 0.0000 & 0.5836 & 0.6994 \\
%	\hline
%\end{tabular}

\begin{figure}[!htp]
	\centering\small
	
	\begin{subfigure}[t]{0.22\textwidth}
		\includegraphics[width=\textwidth]{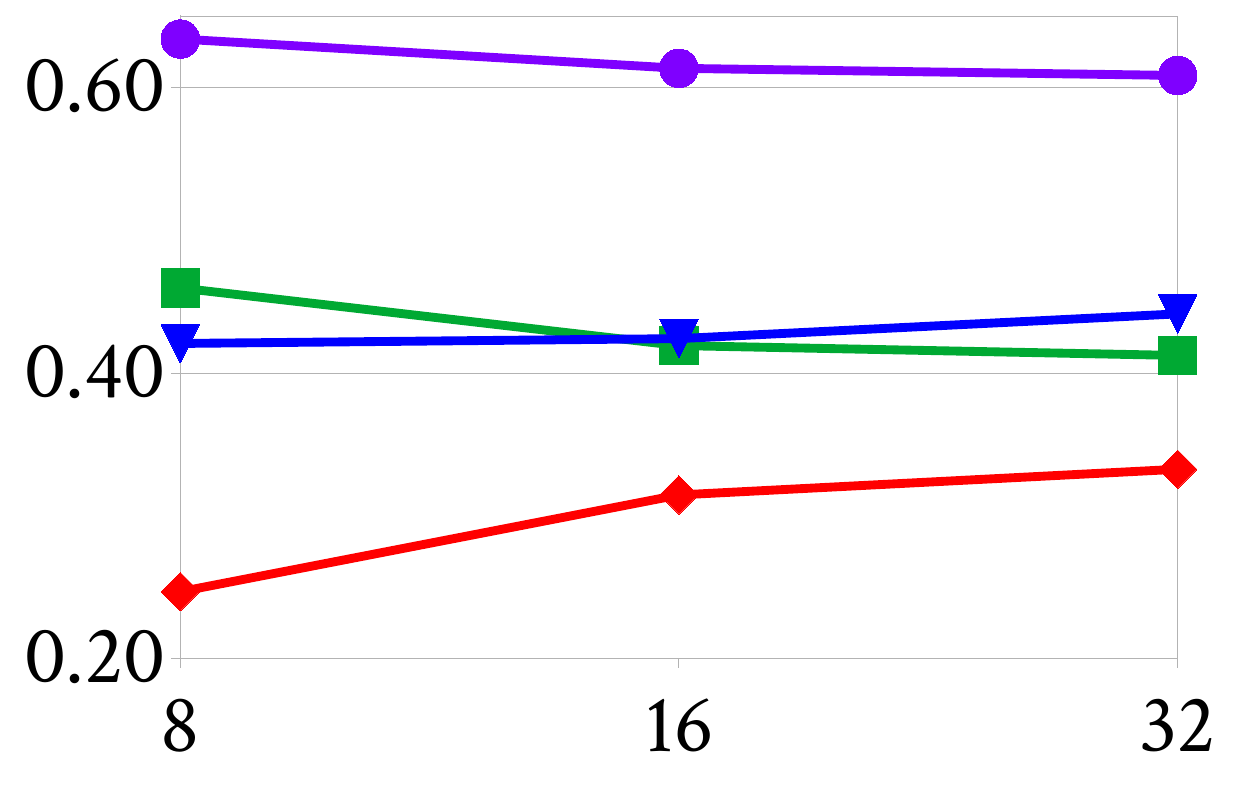}
		\subcaption{Overall Accuracy}
	\end{subfigure}
	\begin{subfigure}[t]{0.22\textwidth}
		\includegraphics[width=\textwidth]{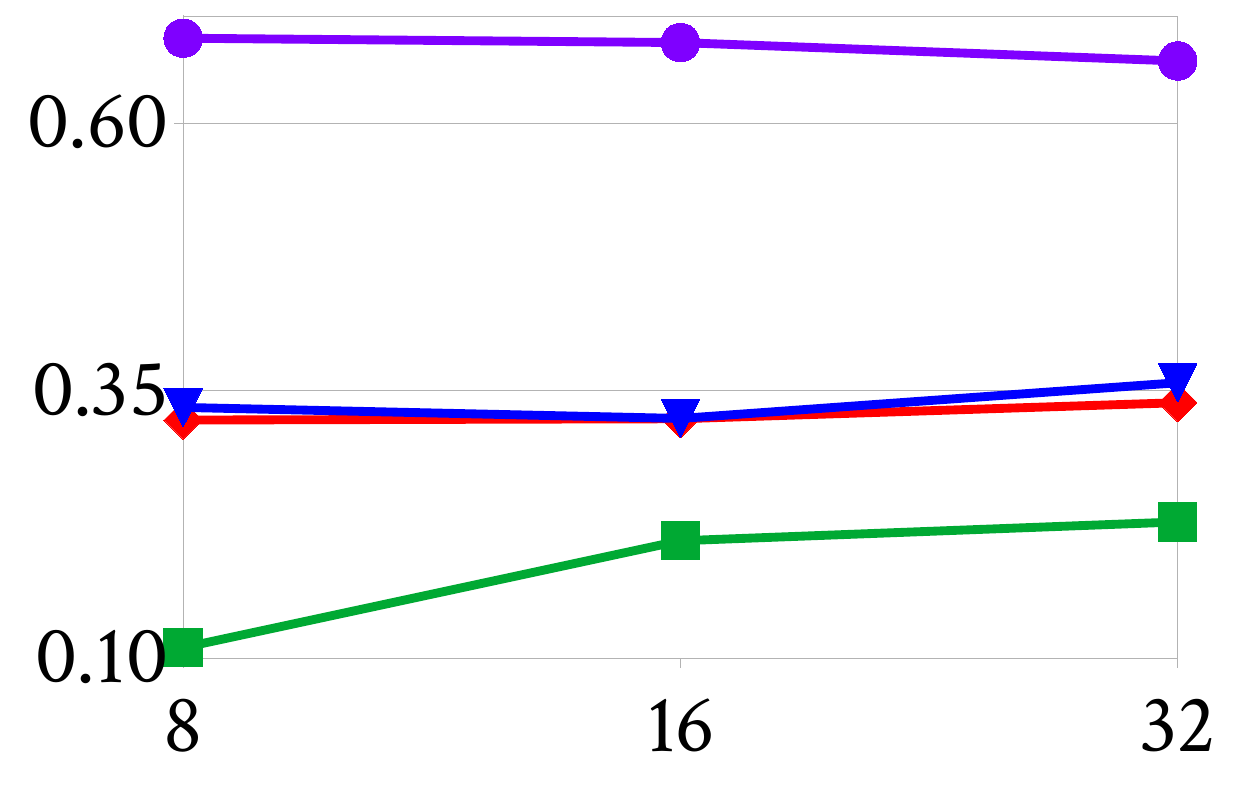}
		\subcaption{Positive F1-score}
	\end{subfigure}
	\begin{subfigure}[t]{0.22\textwidth}
		\includegraphics[width=\textwidth]{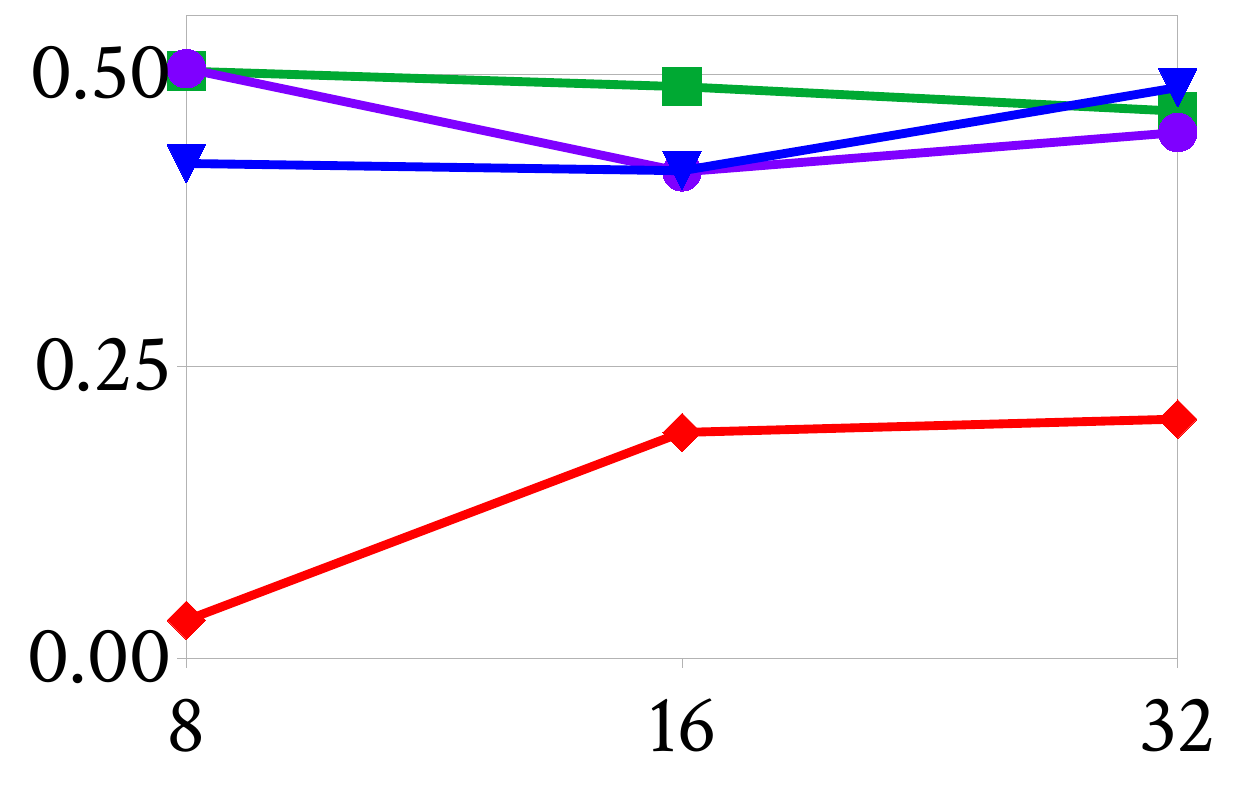}
		\subcaption{Negative F1-score}
	\end{subfigure}
	\begin{subfigure}[t]{0.22\textwidth}
		\includegraphics[width=\textwidth]{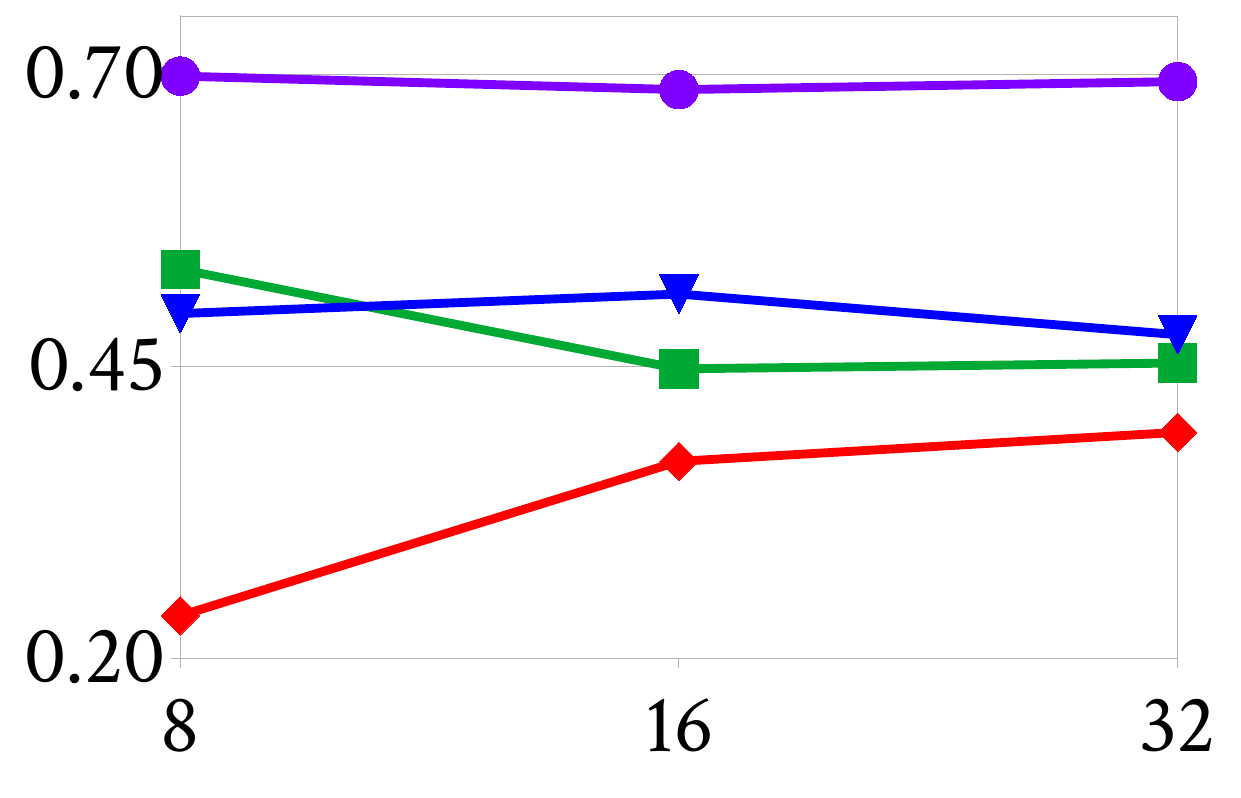}
		\subcaption{Neutral tag F1-score}
	\end{subfigure}
	\hfill
	\begin{subfigure}[t]{0.07\textwidth}
		\includegraphics[width=\textwidth]{IMG/abl_sa_legend.pdf}
		%		\subcaption{Legend}
	\end{subfigure}
	
	\caption{Effect of the embedding size ($d$) on sentiment analysis in Arabic, Algerian dialect, English, and French. 
		Results were obtained by training a BiGRU--dense layer decoder on frozen chDzDT embeddings.}

	\label{fig:ablation-d-sa}
\end{figure}

\paragraph{Sentiment analysis.}
Figure~\ref{fig:ablation-d-sa} shows the effect of embedding size on sentiment classification. 
In contrast to morphologically sensitive tasks, increasing $d$ from 8 to 32 yields no consistent gains. 
Overall accuracy remains broadly stable across languages, with variations within the expected range of optimization variability. 
%At the class level, performance exhibits similarly mixed trends: positive F1-scores improve slightly with larger embeddings in English and French but decline in Algerian dialect; negative scores fluctuate without a clear pattern; neutral performance shows inconsistent improvements across languages.
At the class level, performance exhibits mixed trends. 
Positive F1-scores improve slightly with larger embeddings in English and French but decline in the Algerian dialect. 
Negative scores fluctuate without a clear pattern, while neutral performance shows inconsistent improvements across languages. 
Arabic results are particularly unstable, suggesting that larger embeddings do not systematically improve sentiment discrimination. 
Taken together, these findings reinforce the observation that sentiment analysis, unlike morpho-syntactic tagging, is relatively insensitive to increases in representational capacity. 
From a practical perspective, smaller embeddings may suffice, reducing computational cost without sacrificing performance.

%\begin{tabular}{|c|c|c|c|c|c|c|c|c|c|c|c|c|c|c|c|c|}
%	\hline
%	& Overall &  &  &  & Positive &  &  &  & Negative &  &  &  & Neutral &  &  &  \\
%	\hline
%	d & Arabic & Algerian & English & French & Arabic & Algerian & English & French & Arabic & Algerian & English & French & Arabic & Algerian & English & French \\
%	\hline
%	8 & 0.4595 & 0.6338 & 0.2469 & 0.4207 & 0.1103 & 0.6792 & 0.3227 & 0.3345 & 0.5027 & 0.5048 & 0.0328 & 0.4237 & 0.5332 & 0.6985 & 0.2367 & 0.4953 \\
%	\hline
%	16 & 0.4192 & 0.6133 & 0.3146 & 0.4241 & 0.2100 & 0.6751 & 0.3239 & 0.3242 & 0.4894 & 0.4165 & 0.1935 & 0.4175 & 0.4480 & 0.6869 & 0.3688 & 0.5120 \\
%	\hline
%	32 & 0.4123 & 0.6083 & 0.3326 & 0.4414 & 0.2276 & 0.6579 & 0.3389 & 0.3576 & 0.4684 & 0.4502 & 0.2049 & 0.4887 & 0.4530 & 0.6938 & 0.3936 & 0.4772 \\
%	\hline
%\end{tabular}

\section{Conclusion}
\label{sec:concl}

This work addresses the limitations of existing PLMs in handling the Algerian dialect, a variety characterized by complex morphology, frequent code-switching, and non-standard orthography. 
Traditional word- or subword-level approaches often struggle with high variability and fragmented forms, limiting effectiveness on downstream tasks.

To overcome these challenges, we proposed \textbf{chDzDT}, a character-level pre-trained language model designed specifically for Algerian dialectal text. 
By constructing word representations from characters, chDzDT reduces reliance on fixed vocabularies and improves robustness to spelling inconsistencies, mixed scripts, and informal user-generated content. 
Experimental results show that the model produces stable and expressive embeddings that enhance performance on morphological and syntactic tasks. 
While semantic--pragmatic tasks did not see systematic improvement, we hypothesize that combining character-level embeddings with word- or subword-level representations could benefit meaning-oriented applications.

These findings highlight the potential of character-level modeling for resource-scarce, morphologically rich dialects, providing a foundation for more inclusive and adaptable NLP systems. 
Future directions include integrating word-level semantic signals and leveraging cross-lingual and cross-dialectal transfer.
Additionally, extending coverage to Tifinagh script for Berber-origin words and addressing word segmentation challenges could further enhance token-level and semantic processing.

\section*{Declarations}

%\subsection*{Acknowledgements}
%We thank ...

\subsection*{Ethical disclaimer}
This study uses publicly available datasets, including YouTube comments, Wikipedia, and Tatoeba, in accordance with their terms of use and applicable licenses.  
All data were handled under fair use for research purposes.  
Some datasets contain sensitive or offensive language, which was included solely to evaluate model robustness.  
The presence of such content does not reflect the authors' views, and all analyses were conducted following established ethical research guidelines.

\subsection*{Data and software availability}
The code used in this research is available on GitHub\footnote{Code and evaluation datasets: \url{https://github.com/kariminf/chdzdt}} under the Apache 2 license.  
Some datasets prepared for this study's evaluation, or those with less restrictive licenses, are included in the same repository under Creative Commons licenses.  
For datasets requiring external access, we provide scripts to preprocess them after downloading from their original sources.  
The pre-training dataset includes words extracted from YouTube comments and therefore cannot be redistributed.  
The trained models are published in the Hugging Face repository\footnote{chDzDT models: \url{https://huggingface.co/kariminf/chdzdt}} under the Apache 2 license.

\subsection*{Competing interests}
The authors declare that they have no known competing financial interests or personal relationships that could have influenced the work reported in this paper.

\subsection*{Generative AI and AI-assisted technologies}
During the preparation of this work, the authors used ChatGPT (OpenAI) to improve readability and refine the language of the manuscript. 
Additionally, ChatGPT was employed to assist in searching for related works, identifying potential datasets, and exploring suitable evaluation tests. 
After using this tool, the authors reviewed and edited the content as needed and take full responsibility for all aspects of the publication.

%\printbibliography
{
	\small \bibliographystyle{apalike-ejor}
	\bibliography{citation}
}

\end{document}